\documentclass[10pt,twocolumn,letterpaper]{article}

\usepackage[pagenumbers]{cvpr} %

\usepackage{graphicx}
\usepackage{amsmath}
\usepackage{amssymb}
\usepackage{booktabs}
\usepackage{subcaption}

\usepackage{multirow}
\usepackage{pifont}
\usepackage{mathrsfs}
\usepackage{multirow,overpic}

\usepackage{verbatim}
\usepackage{color}
\usepackage{float}
\usepackage{enumitem}
\usepackage{booktabs}
\usepackage{tabulary,multirow,overpic,xcolor}

\usepackage[pagebackref,breaklinks,colorlinks]{hyperref}

\usepackage[capitalize]{cleveref}
\crefname{section}{Sec.}{Secs.}
\Crefname{section}{Section}{Sections}
\Crefname{table}{Table}{Tables}
\crefname{table}{Tab.}{Tabs.}

\usepackage{xcolor}
\usepackage[font=footnotesize,labelfont=bf,aboveskip=1pt,belowskip=-10pt]{caption}  %
\usepackage{makecell}
\usepackage{tabulary}
\definecolor{demphcolor}{RGB}{144,144,144}
\newcommand{\demph}[1]{\textcolor{demphcolor}{#1}}
\definecolor{mygray}{gray}{0.4}
\usepackage{pifont}%
\newlength\savewidth\newcommand\shline{\noalign{\global\savewidth\arrayrulewidth
  \global\arrayrulewidth 1pt}\hline\noalign{\global\arrayrulewidth\savewidth}}

\newcommand{\tablestyle}[2]{\setlength{\tabcolsep}{#1}\renewcommand{\arraystretch}{#2}\centering\footnotesize}
\makeatletter\renewcommand\paragraph{\@startsection{paragraph}{4}{\z@}
  {.5em \@plus1ex \@minus.2ex}{-.5em}{\normalfont\normalsize\bfseries}}\makeatother
\newdimen\abovecrulesep
\newdimen\belowcrulesep
\makeatletter
\patchcmd{\@@@cmidrule}{\aboverulesep}{\abovecrulesep}{}{}
\patchcmd{\@xcmidrule}{\belowrulesep}{\belowcrulesep}{}{}
\makeatother
\newcolumntype{C}[1]{>{\centering\arraybackslash}p{#1}}
\newcolumntype{R}[1]{>{\raggedleft\arraybackslash}p{#1}}
\newcolumntype{L}[1]{>{\raggedright\arraybackslash}p{#1}}

\usepackage{etoolbox}
\preto\align{\small}
\expandafter\preto\csname align*\endcsname{\small}
\preto\equation{\small}
\usepackage{algorithm}
\usepackage{algorithmic}

\def\cvprPaperID{2178} %
\def\confName{CVPR}
\def\confYear{2023}

\begin{document}

\title{Adaptive Human Matting for Dynamic Videos}

\author{{Chung-Ching Lin, Jiang Wang, Kun Luo, Kevin Lin, Linjie Li, Lijuan Wang, Zicheng Liu}\\
Microsoft\\
{\tt\small \{chungching.lin,jiangwang,kun.luo,keli,linjli,lijuanw,zliu\}@microsoft.com}}

\makeatletter
\g@addto@macro\@maketitle{
\vspace{-3em}
  \begin{figure}[H]
  \setlength{\linewidth}{\textwidth}
  \setlength{\hsize}{\textwidth}
  \centering

  \includegraphics[trim=0 0 0 0, clip,width=0.22\textwidth]{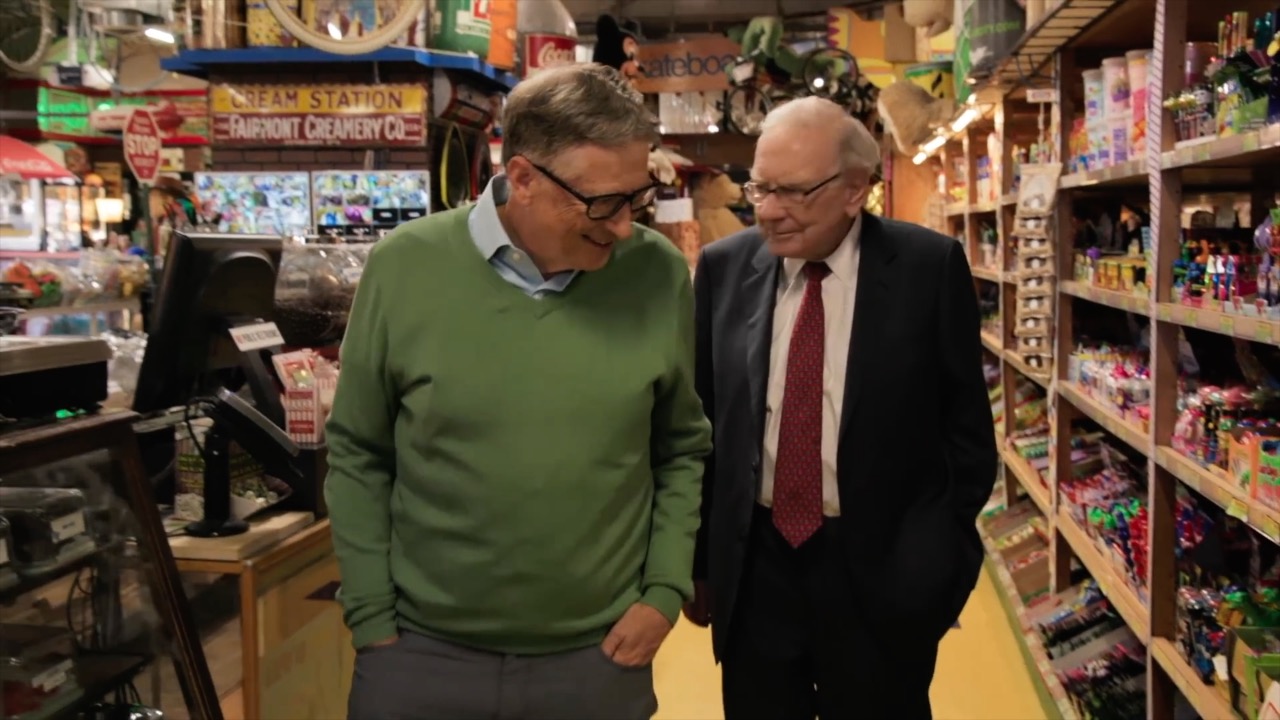}
  \includegraphics[trim=0 0 0 0, clip,width=0.22\textwidth]{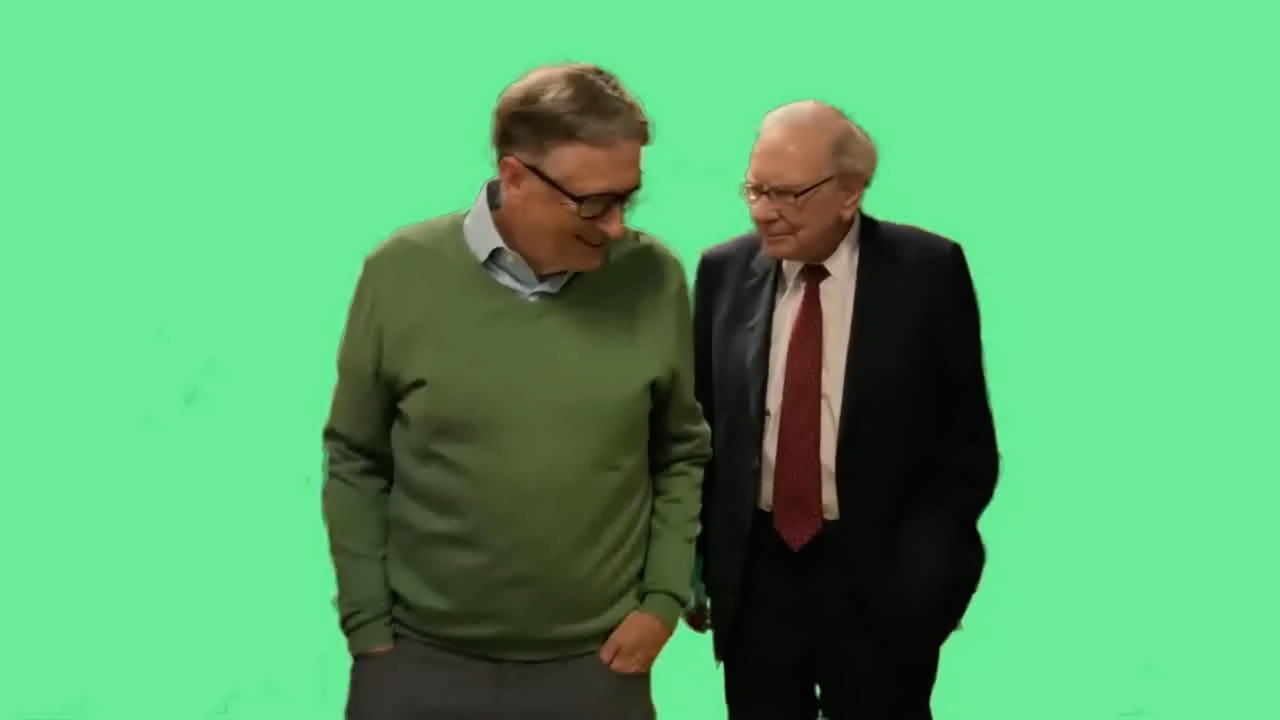}  
  \includegraphics[trim=0 0 0 0, clip,width=0.22\textwidth]{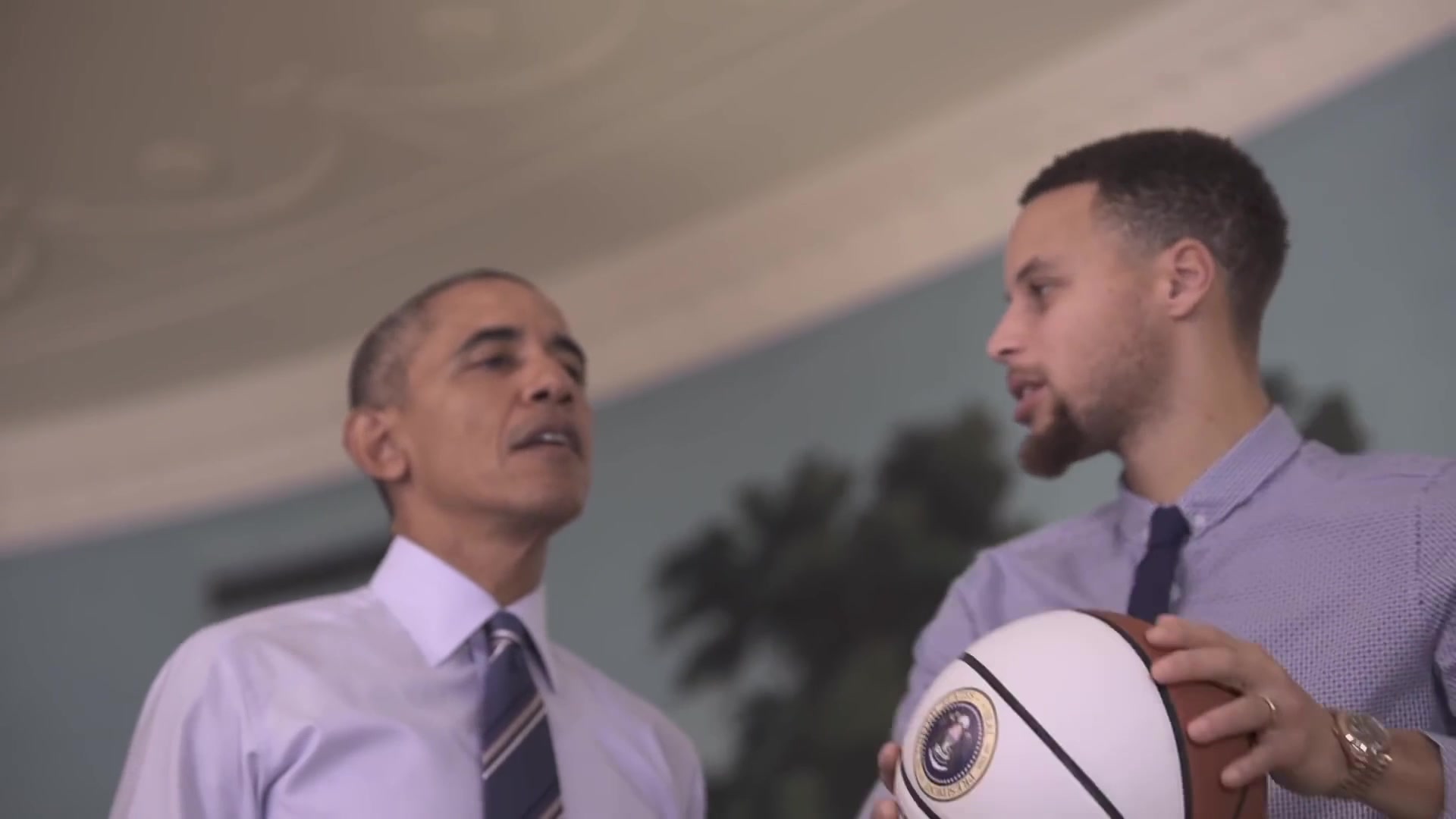}        %
  \includegraphics[trim=0 0 0 0, clip,width=0.22\textwidth]{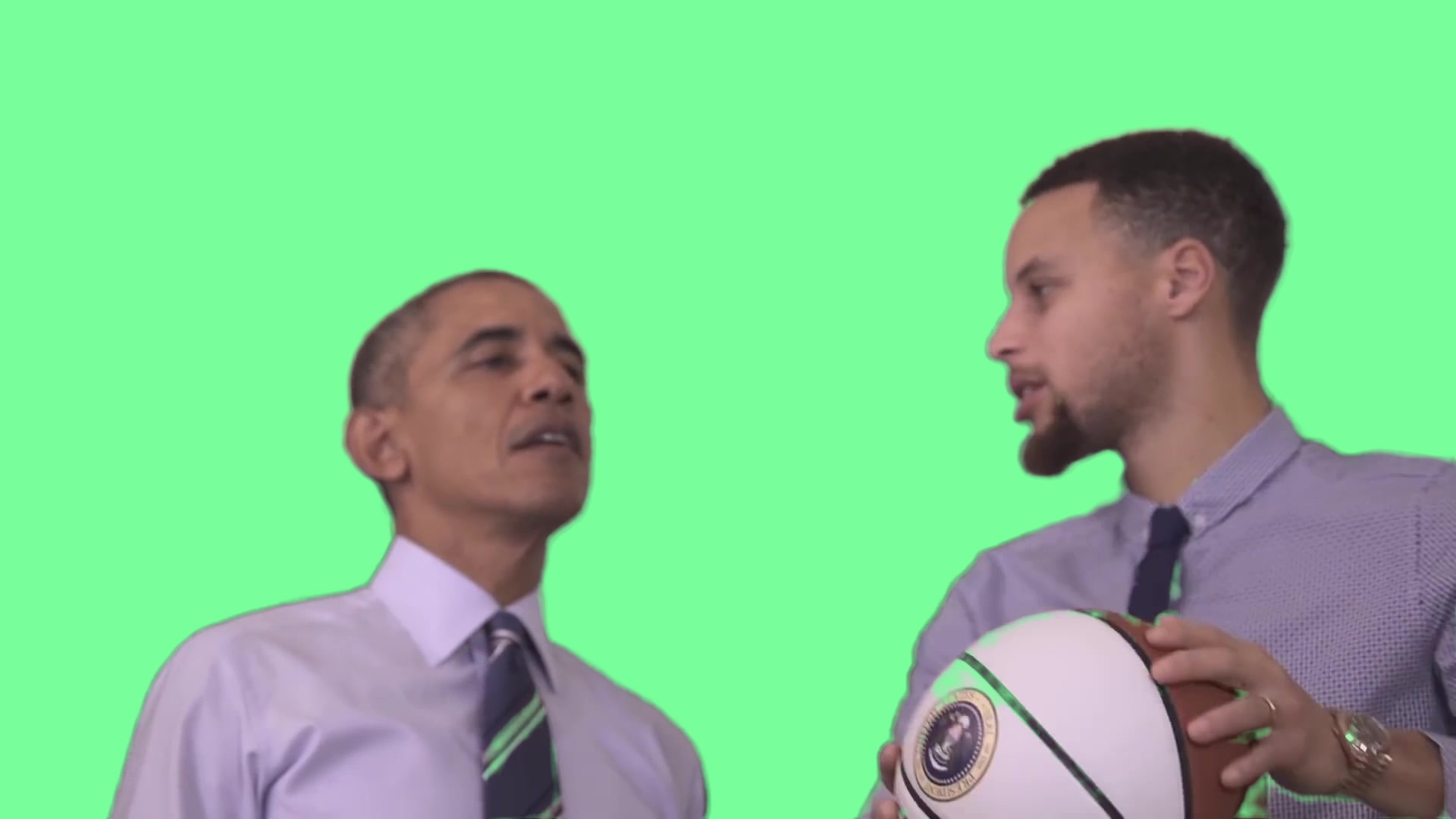}\\
  \vspace{1pt}
  \includegraphics[trim=0 0 0 0, clip,width=0.22\textwidth]{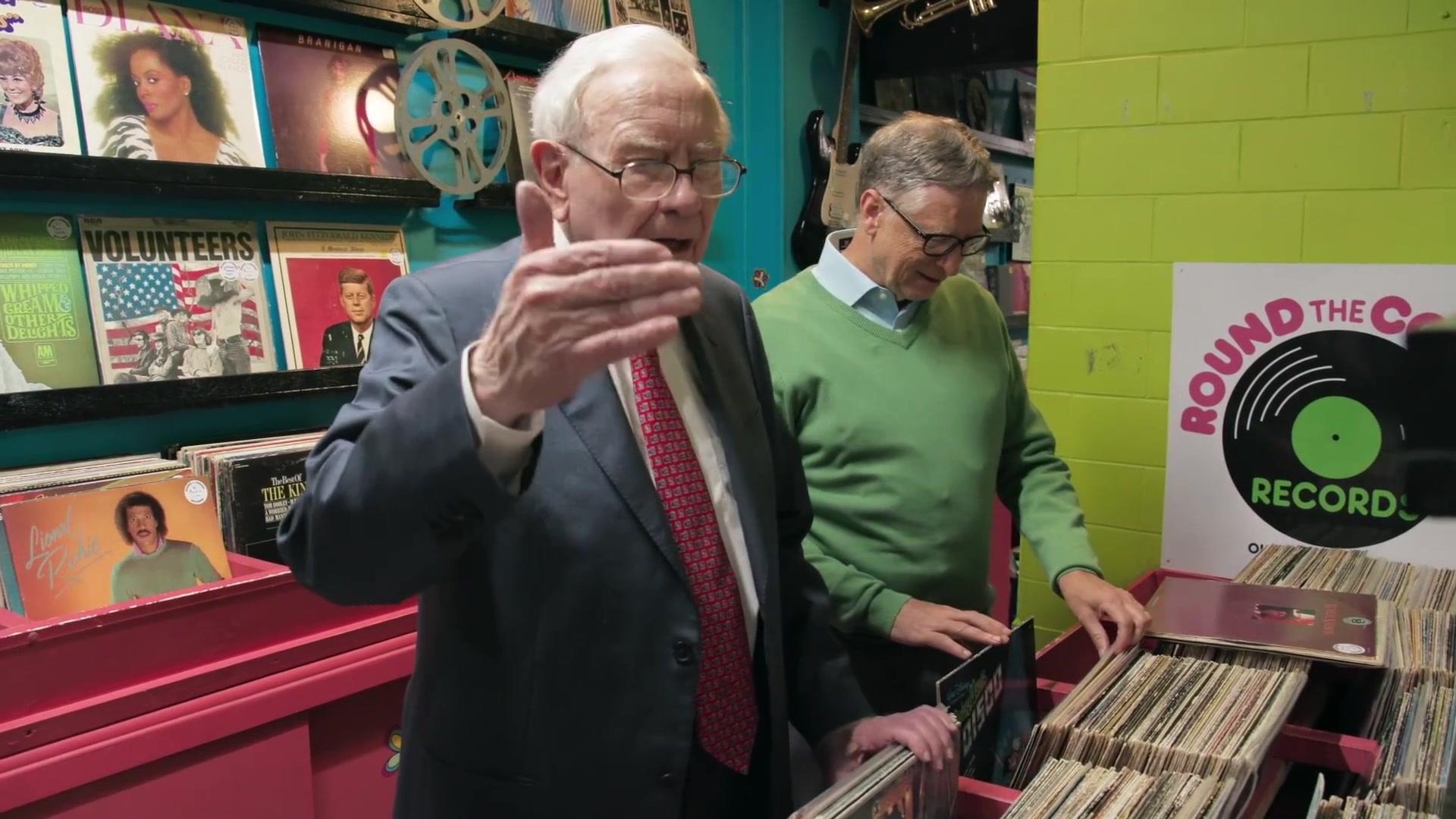}
  \includegraphics[trim=0 0 0 0, clip,width=0.22\textwidth]{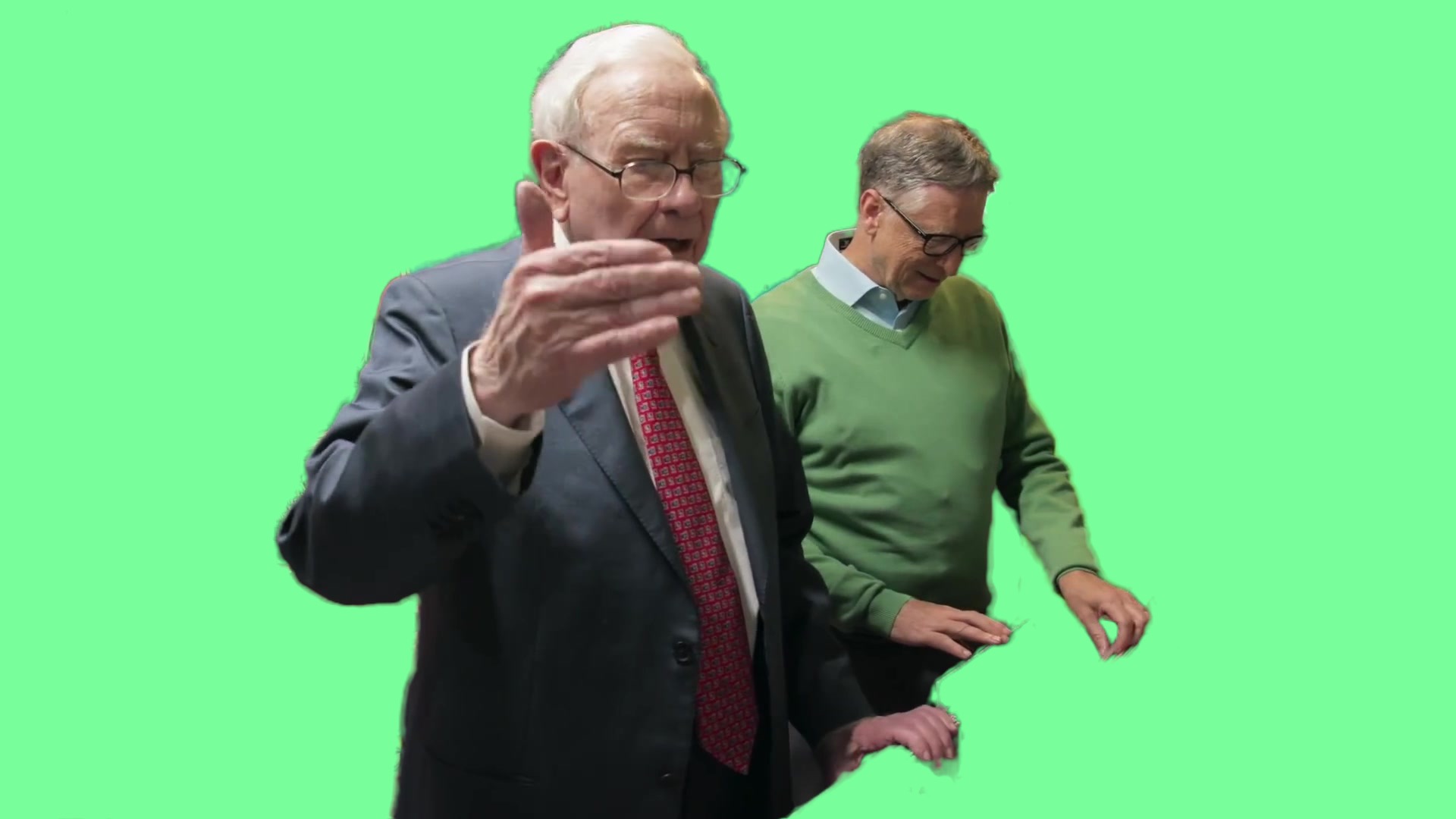}  
  \includegraphics[trim=0 0 0 0, clip,width=0.22\textwidth]{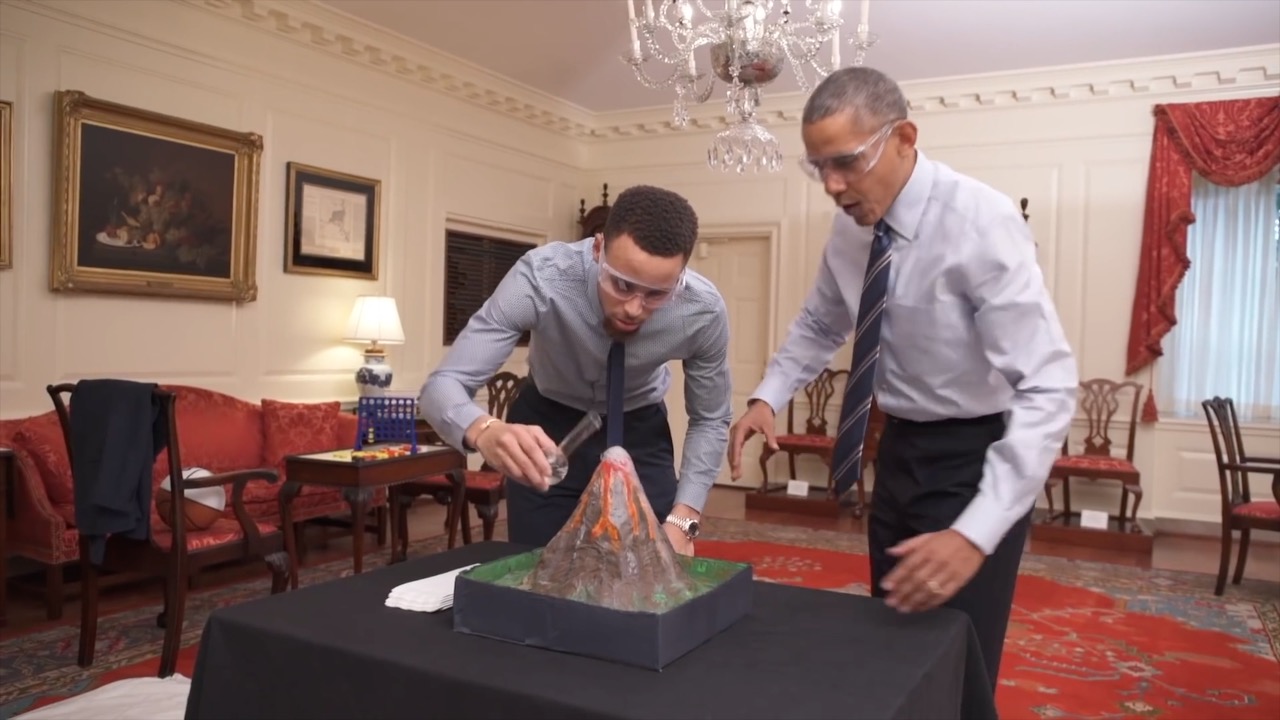}        %
  \includegraphics[trim=0 0 0 0, clip,width=0.22\textwidth]{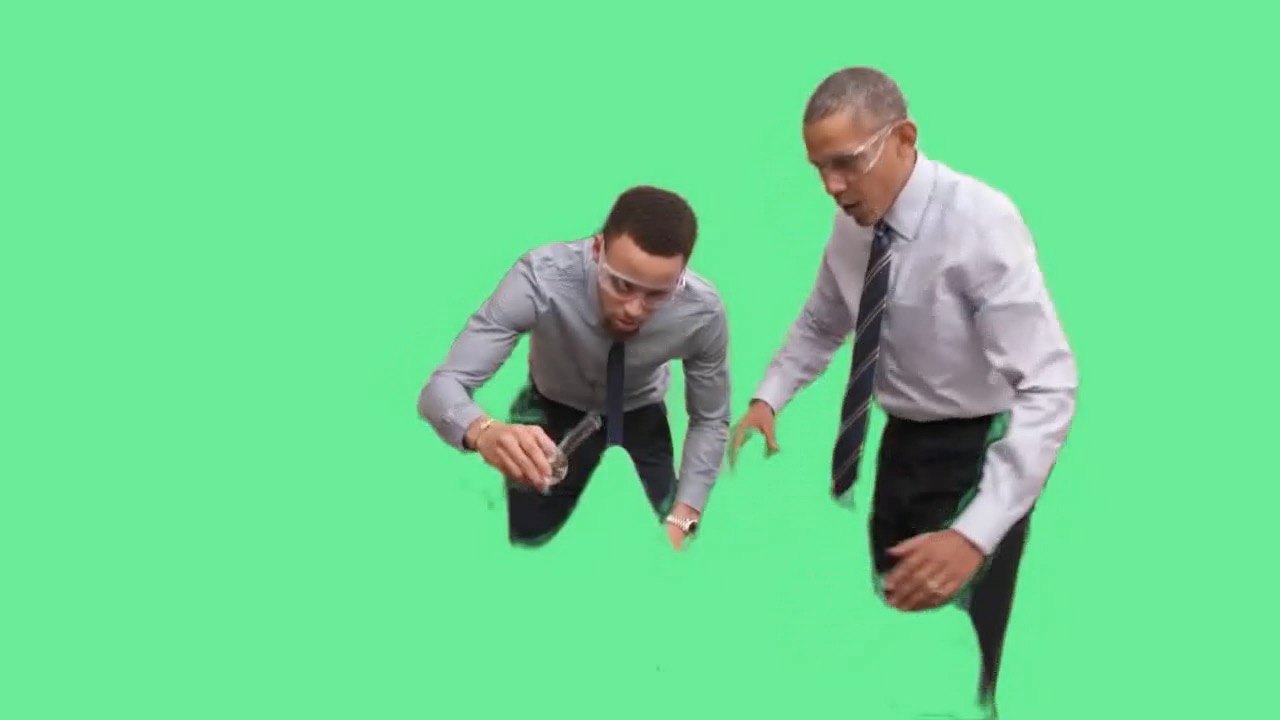}

  \caption{Our method predicts reliable alpha mattes on dynamic videos without requiring trimaps or pre-captured backgrounds.}
  
  \label{fig:first}
   \vspace{1.8em}
  \end{figure}
}
\makeatother

\maketitle

\begin{abstract}
  \vspace{-6pt}
The most recent efforts in video matting have focused on eliminating trimap dependency since trimap annotations are expensive and trimap-based methods are less adaptable for real-time applications. Despite the latest tripmap-free methods showing promising results, their performance often degrades when dealing with highly diverse and unstructured videos. We address this limitation by introducing \textbf{Ada}ptive \textbf{M}atting for Dynamic Videos, termed \textbf{AdaM}, which is a framework designed for simultaneously differentiating foregrounds from backgrounds and capturing alpha matte details of human subjects in the foreground. Two interconnected network designs are employed to achieve this goal: (1) an encoder-decoder network that produces alpha mattes and intermediate masks which are used to guide the transformer in adaptively decoding foregrounds and backgrounds, and (2) a transformer network in which long- and short-term attention combine to retain spatial and temporal contexts, facilitating the decoding of foreground details. We benchmark and study our methods on recently introduced datasets, showing that our model notably improves matting realism and temporal coherence in complex real-world videos and achieves new best-in-class generalizability. Further details and examples are available at \url{https://github.com/microsoft/AdaM}.

\end{abstract}
  \vspace{-10pt}

\section{Introduction}

Video human matting aims to estimate a precise alpha matte to extract the human foreground from each frame of an input video.
In comparison with image matting \cite{xu2017deep, hou2019context, forte2020f, qiao2020attention, li2020natural, sun2021semantic, yu2021mask, dai2022boosting}, video matting \cite{bai2011towards, choi2012video, gong2010real, li2013motion, BMSengupta20, sun2021modnet, shahrian2014temporally, lee2010temporally} presents additional challenges, such as preserving spatial and temporal coherence.

Many different solutions have been put forward for the video matting problem. A straightforward approach is to build on top of image matting models \cite{xu2017deep}, which is to implement an image matting approach frame by frame. It may, however, result in inconsistencies in alpha matte predictions across frames, which will inevitably lead to flickering artifacts \cite{wang2021video}. 
On the other hand, top performers leverage dense trimaps to predict alpha mattes, which is expensive and difficult to generalize across large video datasets. 
To alleviate the substantial trimap limitation, OTVM \cite{seong2022one} proposed a one-trimap solution recently. BGM \cite{BMSengupta20, lin2021real} proposes a trimap-free solution, which needs to take an additional background picture without the subject at the time of capture. While the setup is less time-consuming than creating trimaps, it may not work well if used in a dynamic background environment. The manual prior required by these methods limits their use in some real-time applications, such as video conferencing. 
Lately, more general solutions, e.g., MODNet \cite{ke2022modnet} and RVM \cite{lin2022robust}, have been proposed which involve manual-free matting without auxiliary inputs. However, in challenging real-world videos, backgrounds are inherently non-differentiable at some points, causing these solutions to produce blurry alpha mattes.

\begin{figure}[t!]
\begin{center}
\includegraphics[trim=0 0 0 0, clip,width=0.86\linewidth]{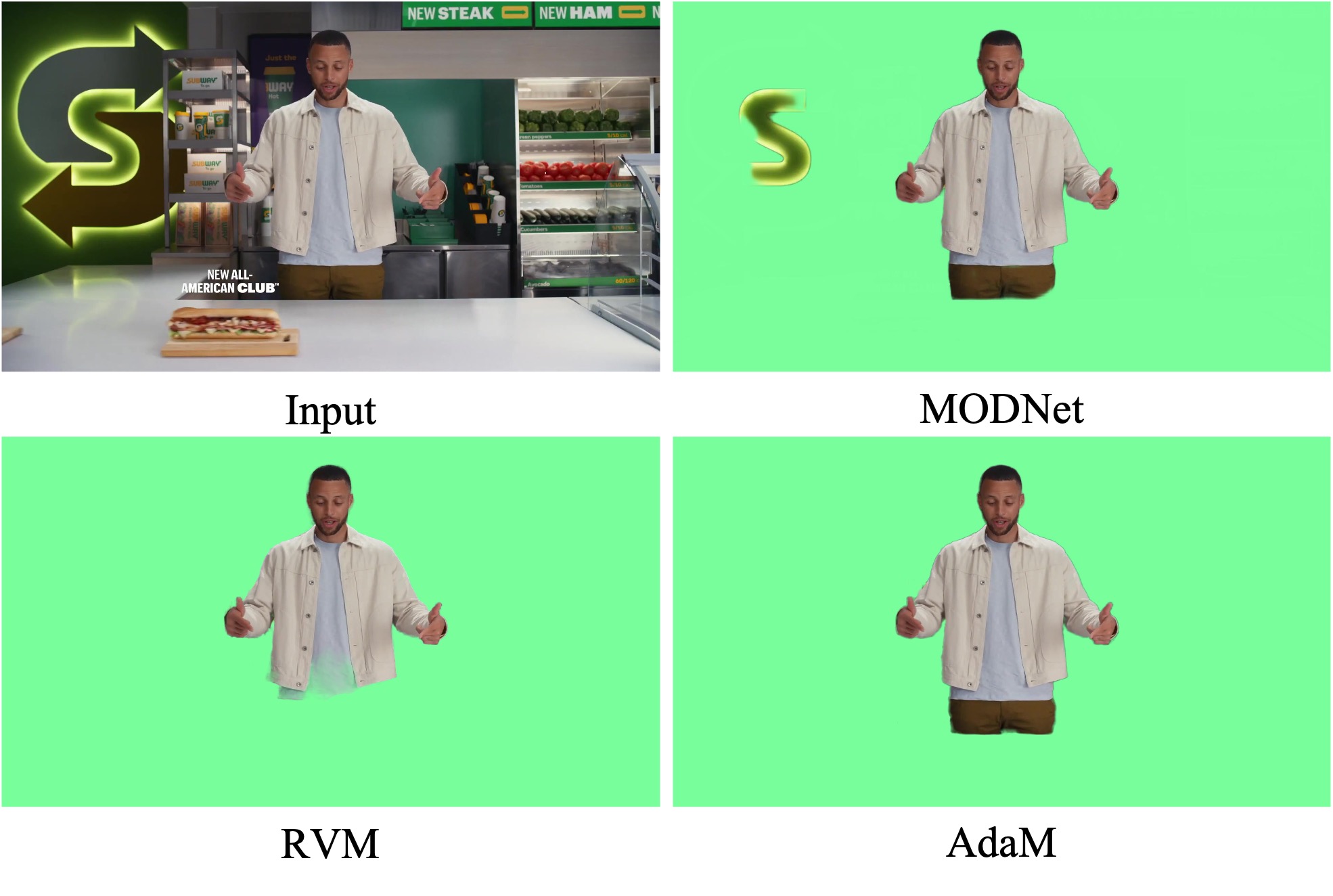}
\caption{
Qualitative sample results of MODNet \cite{ke2022modnet}, RVM \cite{lin2022robust} and the proposed AdaM on real video scenes.}
\vspace{-15pt}
\label{fig:teaser}
\end{center}
\end{figure}

It is quite challenging to bring together the benefits of both worlds, i.e., a manual-free model that produces accurate alpha mattes in realistic videos. 
In our observation, the significant challenges can mostly be explained by the inherent unconstrained and diverse nature of real-world videos. 
As a camera moves in unstructured and complex scenes, foreground colors can resemble passing objects in the background, which makes it hard to separate foreground subjects from cluttered backgrounds.
This might result in blurred foreground boundaries or revealing backgrounds behind the subjects, as shown in Fig. \ref{fig:teaser}.
MODNet \cite{ke2022modnet} and RVM \cite{lin2022robust} are both nicely designed models with auxiliary-free architectures to implicitly model backgrounds.
In complex scenes, however, models without guidance may experience foreground-background confusion, thereby degrading the alpha matte accuracy.

In this paper, we aim to expand the applicability of the matting architecture such that it can serve as a reliable framework for human matting in real-world videos. Our method does not require manual efforts (e.g., manually annotated trimaps or pre-captured backgrounds). The main idea is straightforward: understanding and structuring background appearance can make the underlying matting network easier to render high-fidelity foreground alpha mattes in dynamic video scenes. 
Toward this goal, an interconnected two-network framework is employed: (1) an encoder-decoder network with skip connections produces alpha mattes and intermediate masks that guide the transformer network in adaptively enhancing foregrounds and backgrounds, and (2) a transformer network with long- and short-term attention that retain both spatial and temporal contexts, enabling foreground details to be decoded.
Based on a minimal-auxiliary strategy, the transformer network obtains an initial mask from an off-the-shelf segmenter for coarse foreground/background (Fg/Bg) guidance, but the decoder network predicts subsequent masks automatically in a data-driven manner. 
The proposed method seeks to produce accurate alpha mattes in challenging real-world environments while eliminating the sensitivities associated with handling an ill-initialized mask. 
Compared to the recently published successes in video matting study, our main contributions are as follows:
\vspace{-5pt}
\begin{itemize}
\setlength\itemsep{0pt}
    \item We propose a framework for human matting with unified handling of complex unconstrained videos without requiring manual efforts. The proposed method provides a data-driven estimation of the foreground masks to guide the network to distinguish foregrounds and backgrounds adaptively.
    \item Our network architecture and training scheme have been carefully designed to take advantage of both long- and short-range spatial and motion cues. It reaches top-tier performance on the VM \cite{lin2022robust} and CRGNN \cite{wang2021video} benchmarks.
\end{itemize}
\vspace{-5pt}

\section{Related Work}
The recent video matting studies can be summarized into two main categories: 1) trimap-based matting \cite{wang2021video, sun2021deep, seong2022one}; 2) trimap-free matting \cite{BMSengupta20, lin2021real, ke2022modnet, lin2022robust, li2022vmformer}. 

\vspace{2pt}
\noindent\textbf{Trimap-based matting.}
Previous solutions mainly focus on estimating alpha mattes from image feature maps with auxiliary trimap supervisions, and extending the propagation temporally. %
The trimap serves to reduce the difficulty of the matting problem.
Several methods \cite{wang2021video, sun2021deep}, have recently been proposed, which require multiple human-annotated trimaps for network guidance.
Yet, the manual labor involved in creating well-defined trimap annotations is very costly.
Recently, Seong et al. \cite{seong2022one} propose one-trimap video matting network. The method requires only a single trimap input and performs trimap propagation and alpha prediction as a joint task.
The model, however, is still limited by its requirement for a trimap in real-time applications. In addition, with an ill-initialized trimap, errors can be propagated over time.

\begin{figure*}[t!]
\begin{center}
\includegraphics[trim=0 0 0 0, clip,width=0.8\textwidth]{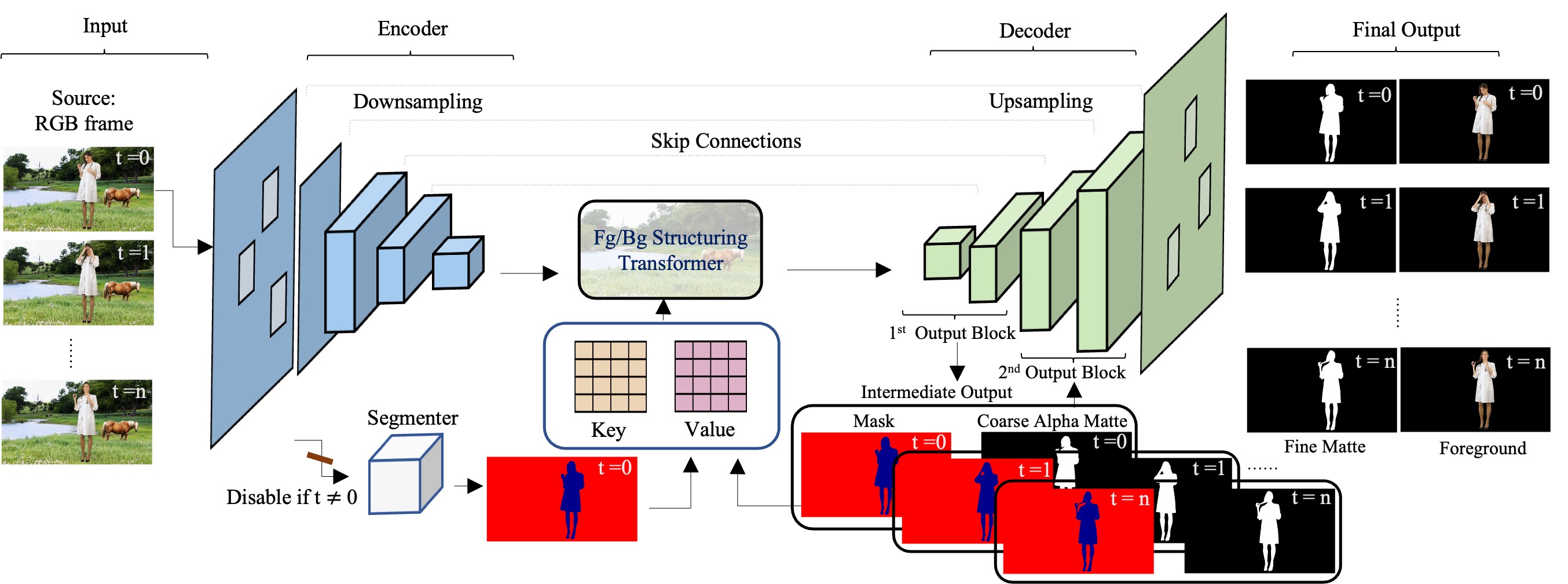}
\vspace{2mm}
\caption{\small
\textbf{Overview of AdaM.} \footnotesize Our framework includes an off-the-shelf segmenter \cite{he2017mask}, an encoder-decoder network, and a Fg/Bg structuring transformer network.
Given an RGB frame, the encoder (Sec. \ref{sec:encoder}) extracts low- and high-resolution features. At frame $t=0$ only, the segmenter obtains an initial coarse mask for embedding into the initial Key and Value feature map. The transformer network (Sec. \ref{sec:transformer}) employs long- and short-term attention to explicitly model Fg/Bg appearance. The decoder (Sec. \ref{sec:decoder}) concatenates the Fg/Bg features encoded in the transformer and the high-resolution fine features from the encoder to produce an intermediate Fg/Bg mask and a final foreground alpha matte. The newly predicted Fg/Bg mask functions as a new reference mask used to update Value adaptively.
}
\vspace{-8pt}
\label{fig:overview}
\end{center}

\end{figure*}

\vspace{2pt}
\noindent\textbf{Trimap-free matting.}
In order to relax the strong requirement of trimaps, studies from recent years have explored the idea of using background data \cite{BMSengupta20, lin2021real}, annotated or predicted segmentation masks \cite{lee2010temporally, tang2010temporally, gong2010real} as the guidance for matting networks, or designing fully automated auxiliary-free \cite{sun2021modnet, lin2022robust, sun2021modnet} methods.

\vspace{2pt}
\noindent\textbf{Background-based matting.}
As an alternative to trimap input, background matting methods \cite{BMSengupta20, lin2021real} require users to take an additional picture of the background without the subject. This step is much less time-consuming than creating a trimap. However, the methods are constrained to videos with relatively static backgrounds. It is also less feasible to use the methods in real-time because background inputs must be captured beforehand.

\vspace{2pt}
\noindent\textbf{Mask-guided matting.}
Some works in image matting use predicted or annotated segmentation masks  to provide a rough guide to the foreground region \cite{yu2021mask}, thereby relaxing the model dependence on the trimap. 
However, the direct application of these image matting methods to individual video frames neglects the temporal information in videos. The frame-by-frame mask generation could also largely restrict its use in real-time settings.

\vspace{2pt}
\noindent\textbf{Auxiliary-free matting.}
Recently, some methods \cite{ke2022modnet, lin2022robust, li2022vmformer} explore auxiliary-free frameworks for video matting. For example, MODNet \cite{ke2022modnet} considers frames as independent images and proposes a technique for suppressing flicker by making comparisons between adjacent frames. %
Concurrently, RVM \cite{lin2022robust} proposes a recurrent architecture to exploit temporal information in videos and achieves significant improvements in temporal coherence and matting quality. 
However, due to the lack of guidance, these methods can be largely compromised in challenging scenes where the cluttered background is present or the color distribution of the background is similar to that of the foreground. 

Unlike previous auxiliary-free methods, our model employs a minimal-auxiliary approach. Except for the initial mask generated by an off-the-shelf segmenter, our network predicts the subsequent masks on its own and uses them to guide the network to render reliable alpha mattes in real-world videos.

\section{Method}
An overview of AdaM architecture is illustrated in Figure \ref{fig:overview}.
Our framework includes an off-the-shelf segmenter \cite{he2017mask}, a fully-convolutional encoder-decoder network, and a Fg/Bg structuring transformer network. The pipeline begins with an RGB video frame being input into the encoder and the segmenter.  
The encoder (Section \ref{sec:encoder}) takes the RGB input and extracts low-resolution coarse features as well as high-resolution fine features. The low-resolution coarse features are fed into the transformer network for efficient Fg/Bg modeling, while the high-resolution fine features are supplied to the decoder module for fine-grained alpha matte decoding. 
Given the first video frame, the segmenter estimates an initial coarse mask. It is noted that the segmenter only serves as an adjunct for obtaining the introductory mask and it could be replaced with a more sophisticated segmentation network \cite{cheng2021maskformer, cheng2021mask2former, ke2022mask}. The initial mask and the low-resolution features are embedded into the initial Key and Value and stored in the transformer. Each of the subsequent input frames acts as a query.
The transformer network (Section \ref{sec:transformer}) employs long- and short-term attention to explicitly model Fg/Bg appearance. Specifically, a per-pixel comparison is made between the coarse low-resolution features and the Key/Value to estimate the features of the foreground or background.
Short-term attention is applied to capture local-to-local spatial interdependency and motion continuity, while long-term attention helps to model local-to-global context reasoning. With the Fg/Bg structuring network, more meaningful semantic information is encoded and a cleaner foreground region is supplied to the decoder, making it easier for the decoder to recover finer details in the foreground. Following that, the Fg/Bg features encoded in the transformer and the high-resolution fine features from the encoder are concatenated together in the decoder (Section \ref{sec:decoder}), producing an intermediate Fg/Bg mask and a final foreground alpha matte. The newly predicted Fg/Bg mask functions as a new reference mask used to update Value adaptively.

\subsection{Encoder} \label{sec:encoder}
We adopt MobileNetV2 \cite{sandler2018mobilenetv2} as our backbone. In a given video, frames are sequentially processed. For each video frame, the encoder network extracts features at 1/2, 1/4, 1/8, and 1/16 scales with different levels of abstraction ($F^{1/2}, F^{1/4}, F^{1/8}, F^{1/16}$). $F^{1/16}$ is fed into the transformer to model the foreground and background more efficiently. The rest of the features are fed into the decoder through skip connections to provide more fine-grained details for alpha matte predictions.

\subsection{Fg/Bg Structuring Network} \label{sec:transformer}
Our method is designed to explicitly model Fg/Bg appearance.
The core design is inspired by recent successes of memory mechanisms applied to natural language processing \cite{sukhbaatar2015end, miller2016key, kumar2016ask} and vision models \cite{lu2020video, oh2019video, seong2020kernelized, yang2021associating}.
As the transformer architecture consists of Query, Key and Value in the multi-head attention, it is suitable to dynamically determine which input regions we want to attend more than others. We thus adopt the transformer architecture and extend the concept of memory networks in \cite{oh2019video, yang2021associating} to make it appropriate to model and propagate spatio-temporal Fg/Bg information for the video matting task.

In our pipeline, the Fg/Bg information of previous frames is stored in Key and Value, and a current frame to be processed acts as Query. At each frame, the transformer network takes the low-resolution feature map from the encoder and the previously estimated mask from the decoder as inputs. 
Through visual semantics and the mask information, the network learns to predict whether a given feature belongs in the foreground or the background.
Key provides invariant visual semantic information and serves as an addressing function. The feature map of the current frame is used as a Query. 
Comparing the similarity of Query and Key, the network retrieves relevant Fg/Bg features in Value, and propagates them over space and time through long- and short-term attention. The Fg/Bg features of the current frame are then fed to the decoder for fine-grained alpha estimation.

We first review and define the essential  elements of the transformer.
An attention operation is the core component of a transformer layer, which linearly projects input features $Z$ into queries $Q$, keys $K$, and values $V$:
\begin{equation}
    Q = Z W_Q, \quad K = Z W_K, \quad V = Z W_V,
\end{equation}
where $W_Q$, $W_K$, $W_V$ are learnable linear projection layers. 
To obtain an output feature $Z'$, a self-attention operation can be expressed as:
\begin{equation}
Z' = \mathrm {Attn}(Q, K, V) = \mathrm {Softmax}(QK^\top / \sqrt {d})V.
\end{equation}

We extend the basic attention operation to include the previously computed features stored in Key and Value.
This is to encourage the network to learn spatial interdependence and motion continuity embedded in videos.
In specific, after processing a video frame, the computed features are retained for subsequent iterations.  
When processing the frame $I^t$ at time $t$ ($t>0$), our model refers to Keys and Values computed from earlier iterations to learn long-term and short-term contextual dependencies. 
The attention operation of our Fg/Bg structuring transformer $\Xi$ can be represented as: 
\begin{align}
    &Z'^t = \boldsymbol{\Xi}(Q^t, \boldsymbol{\hat{K}}^{t}, \boldsymbol{\hat{V}}^t), \\
    &\boldsymbol{\hat{K}}^t = concat(K^{t-n},\ldots, K^{t-1}),  \\
    &\boldsymbol{\hat{V}}^t = concat(V^{t-n},\ldots, V^{t-1}),
\end{align}
where $concat()$ denotes concatenation along the feature dimension, $K^{t-1}$ and $V^{t-1}$ are Key and Value at time $t-1$, respectively.  
In this formulation, the query $Q^t$ attends to $\boldsymbol{\hat{K}}^t$ and $\boldsymbol{\hat{V}}^t$ that contain previously computed features up to $n$ steps in the past.

\vspace{3pt}
\noindent\textbf{Fg/Bg update scheme.}
As outlined in the method overview, given the previous frame $I^{t-1}$, our decoder outputs an intermediate mask, $m^{t-1}$, which is used to embed estimated Fg/Bg features of $I^{t-1}$ into Value.
The Fg/Bg embedded feature map, $V^{t-1}_{f/b}$, can be expressed as: 
\begin{equation} \label{eq:fgbg_emb}
    {V}^{t-1}_{f/b} = V^{t-1} + (m^{t-1}_d  E_f + (1-m^{t-1}_d) E_b), 
\end{equation}
where $m^{t-1}_d$ is the downsized Fg/Bg mask of $m^{t-1}$, and $E_f$ and $E_b$ are learnable embeddings for foreground $f$ and background $b$, respectively. 
To process frame $I^t$, we concatenate Value feature map $\boldsymbol{\hat{V}}_{f/b}^t$ to include $V^{t-1}_{f/b}$:
\begin{align} 
    &\boldsymbol{\hat{V}}_{f/b}^t = concat(V^{t-n}_{f/b},\ldots, V^{t-1}_{f/b}),
\end{align}
As both $\boldsymbol{\hat{K}}^t$ and $\boldsymbol{\hat{V}}^{t}_{f/b}$ contain Fg/Bg information accumulated from prior iterations, the network performs attention operations along the temporal domain for each input query. 

\vspace{3pt}
\noindent\textbf{Long-term local-to-global attention.} In highly dynamic scenes, the foreground and background variations over a sequence of frames could be non-local. In light of this, we employ long-term local-to-global attention to capture non-local variations. A conventional transformer operation is used to perform non-local attention on $\boldsymbol{\hat{K}}^t$ and  $\boldsymbol{\hat{V}}^{t}_{f/b}$:
\begin{equation}
Z'^t_L = \mathrm{Attn}(Q^t, \boldsymbol{\hat{K}}^t, \boldsymbol{\hat{V}}^t_{f/b}),
\end{equation}
In this formulation, every pixel in the query feature map can attend to all the stored Key and Value features. The network can thus capture long-term temporal variations. As non-local propagation involves high computational complexity and many features are redundant, this operation is only performed every $l$ frames.

\vspace{3pt}
\noindent\textbf{Short-term local-to-local attention.} Generally, scene changes are smooth and continuous across frames. In the context of video matting, capturing fine-grained local temporal variations would be beneficial for ensuring temporal consistency. Thus, in addition to long-term local-to-global attention, our network also performs short-term local-to-local attention:
\begin{equation}
Z'^t_S = \mathrm{S\text{-}Attn}_{\omega, s}(Q^t, \boldsymbol{\hat{K}}^t, \boldsymbol{\hat{V}}^t_{f/b}).
\end{equation}
In this short-term attention module, $\mathrm{S\text{-}Attn}_{\omega, s}$ restricts the attention to a spatial ($\omega$) and temporal ($s$) tube around the pixel position with a $\omega\times\omega\times s$ tube size.

Finally, the outputs of long-term global attention and short-term local attention module are combined to obtain the final output of $\Xi$: $Z'^t = Z'^t_L + Z'^t_S$.

\subsection{Decoder} \label{sec:decoder}

We use FPN \cite{lin2017feature} as our decoder, which contains four upscaling blocks and two output blocks. The decoder takes the output of the transformer and multi-scale features at different stages to estimate the alpha mattes.  
At each upscaling block, the module concatenates the upsampled output from the previous block and a feature map from the encoder at the corresponding scale through skip-connections. The upscaling process repeats at 1/16, 1/8, 1/4 and 1/2 scale.

We employ a two-stage refinement to progressively refine alpha mattes. 
The model first produces a Fg/Bg mask and a coarse alpha matte at 1/4 scale of the original resolution, and then predicts a finer alpha matte at full resolution. 
The Fg/Bg mask prediction is used to update Value.
The coarse alpha matte prediction and the feature map at 1/4 scale of the original resolution are concatenated and fed to the subsequent upscaling block for further alpha matte refinement. 
The second output block produces the final alpha matte at the original resolution.

\subsection{Losses}

Our model predicts a Fg/Bg mask $m_p$, a coarse alpha matte $\alpha^c_p$, and a fine-grained alpha matte $\alpha^f_p$.
We represent the ground truth alpha matte $\alpha_{GT}$. 
To learn Fg/Bg masks, we use the ground truth alpha matte $\alpha_{GT}$ to generate pseudo labels $m^*_{GT}$ and compute binary cross entropy loss as:
\begin{equation}
    \mathcal{L}^{m}(m_p) = m^{*}_{GT}(-log(m_p)) + (1-m^*_{GT})(-log(1-m_p)),
\end{equation}
where $m^*_{GT}= \alpha_{GT} < \tau$ and $\tau$ is a threshold. 

To learn alpha matte predictions, we compute the L1 loss $\mathcal{L}^{\alpha}_{l1}$ and Laplacian pyramid loss ($\mathcal{L}^{\alpha}_{lap}$) \cite{hou2019context, forte2020f} as: 
\begin{equation}
    \mathcal{L}^{\alpha}_{l1}(\alpha_p) = ||\alpha_{p} - \alpha_{GT}||_1 
\end{equation}
\begin{equation}
    \mathcal{L}^{\alpha}_{lap}(\alpha_p) = \sum^{5}_{s=1}\frac{2^{s-1}}{5}||L^s_{pyr}(\alpha_p) - L^s_{pyr}(\alpha_{GT})||_1
\end{equation}

The overall loss function $\mathcal{L}$ is: 
\begin{align}
    \mathcal{L} =  \omega_{m} \mathcal{L}^{m}_{}(m_p) + & \omega^c_{\alpha}(\mathcal{L}^{\alpha}_{l1}(\alpha^c_p) + \mathcal{L}^{\alpha}_{lap} (\alpha^c_p)) \nonumber \\
    + &\omega^f_{\alpha}(\mathcal{L}^{\alpha}_{l1}(\alpha^f_p) + \mathcal{L}^{\alpha}_{lap} (\alpha^f_p)), 
\end{align}
where $\omega_{m}$, $\omega^c_{\alpha}$, and $\omega^f_{\alpha}$ are loss weights. For our experiments, we empirically determined these weights to be $\omega_{m}=0.5$, $\omega^c_{\alpha}=0.5$, and $\omega^f_{\alpha}=1$.

\section{Experiments}

\begin{table*}[t]
\tablestyle{4.5pt}{1.1} 
\def\w{20pt} 
\centering
    
    \begin{tabular}{cccccccccccccc}
        \shline
        ~ & \multicolumn{2}{c}{Features} & ~ & \multicolumn{5}{c}{VM 512$\times$288} & ~ & \multicolumn{4}{c}{VM 1920$\times$1080}\\
        \cmidrule{2-3} \cmidrule{5-9} \cmidrule{11-14} Method & Backbone & HD Training & ~ & MAD $\downarrow$  & MSE $\downarrow$  & Grad $\downarrow$ & Conn $\downarrow$ & dtSSD $\downarrow$ & ~ & MAD $\downarrow$ & MSE $\downarrow$ & Grad $\downarrow$ & dtSSD $\downarrow$ \\
        \hline
        FBA~\cite{forte2020f}  & ResNet50 & \checkmark & ~ & 8.36 & 3.37 & 2.09 & 0.75 & 2.09 & ~ & - & - & - & - \\
        BGMv2~\cite{lin2021real}  & MobileNetV2 & \checkmark & ~ & 25.19 & 19.63 & 2.28 & 3.26 & 2.74 & ~ & - & - & - & - \\
        MODNet~\cite{ke2022modnet} & MobileNetV2 & \checkmark & ~ & 9.41 & 4.30 & 1.89 & 0.81 & 2.23 & ~ & 11.13 & 5.54 & 15.3 & 3.08 \\
        RVM~\cite{lin2022robust} & MobileNetV3 & \checkmark & ~ & 6.08 & 1.47 & 0.88 & 0.41 & {1.36} & ~ & 6.57 & 1.93 & 10.55 & 1.90 \\
        \demph{RVM-Large~\cite{lin2022robust}} & \demph{ResNet50} & \demph{\checkmark} & ~ & \demph{-} & \demph{-} & \demph{-} & \demph{-} & \demph{-} & ~ & \demph{5.81} & \demph{0.97} & \demph{9.65} & \demph{1.78} \\
        \hline
AdaM  & MobileNetV2  & - & ~ & \textbf{5.55} & \textbf{0.84} & \textbf{0.80} & \textbf{0.33} & 1.38 & ~ & \textbf{4.61} & \textbf{0.46} & \textbf{6.06} & \textbf{1.47} \\
AdaM  & MobileNetV2 & \checkmark & ~ & \textbf{5.30} & \textbf{0.78} & \textbf{0.72} & \textbf{0.30} & \textbf{1.33} & ~ & \textbf{4.42} & \textbf{0.39} & \textbf{5.12} & \textbf{1.39} \\ 
        \shline
    \end{tabular}
    \caption{Comparison with state-of-the-art methods on {the test split of VM \cite{lin2021real} 512$\times$288 and VM 1920$\times$1080 datasets.} }
    \label{table:vm}
\end{table*}

\begin{table}[t]
\tablestyle{4pt}{1.1} 
\centering
\vspace{5pt}
    \begin{tabular}{ccccccc}
        \shline
        Method & HD Traning & MAD $\downarrow$  & MSE $\downarrow$  & Grad $\downarrow$ & dtSSD $\downarrow$  \\
        \hline
        MODNet~\cite{ke2022modnet} & \checkmark  & 9.50 & 4.33 & 16.94 & 6.22 \\
        RVM~\cite{lin2022robust} & \checkmark  & 15.42 & 9.22 & 18.34 & 6.95 \\
        \hline
AdaM  & - &  \textbf{7.13} & \textbf{3.05} & 17.96 & \textbf{5.76} \\
AdaM  & \checkmark & \textbf{5.94} & \textbf{2.79} & \textbf{16.71} & \textbf{5.45} \\
        \shline
    \end{tabular}
    \caption{Evaluation of the transfer capability of MODNet, RVM and AdaM on the whole CRGNN Real dataset.}
    \label{table:crgnn}
\end{table}
We evaluate our framework on VideoMatte240K \cite{lin2021real} and CRGNN \cite{wang2021video} benchmark datasets, and compared it to leading trimiap-free video matting baselines, including FBA~\cite{forte2020f}, BGMv2~\cite{lin2021real} MODNet~\cite{ke2022modnet} and RVM~\cite{lin2022robust}.
We then study in greater depth the properties of AdaM by an ablation study. All design decisions and hyperparameter tuning were performed on the validation set.

\subsection{\noindent\textbf{Datasets and Metrics}}
The VideoMatte240K (VM) dataset \cite{lin2021real} consists of 484 videos. %
RVM \cite{lin2022robust} split VM dataset into 475/4/5 video clips for training, validating and testing.
The training set is converted into SD and HD sets for different training stages. In evaluation, the testing set is converted and split into VM 512$\times$288 and VM 1920$\times$1080 sets.
To ensure a fair comparison, we use the same training, validation, and test sets created by RVM. \cite{lin2022robust}.
VM  is the largest public video matting dataset containing continuous frame sequences for evaluating motion coherence. 
However, due to the high cost of collecting video matting annotations, most existing datasets, including VM, only provide ground-truth alpha and foreground, which must be composited with background images. 
As far as we are aware, there are only limited annotated real video datasets that can be used to evaluate video matting. 
To study the generalization capability of the proposed method, we evaluate our method on the real video dataset provided by CRGNN. \cite{wang2021video}. 
The CRGNN real dataset (CRGNN-R) consists of 19 real-world videos. Annotations are made every 10 frames at a 30 fps frame rate, which adds up to 711 labeled frames.

In accordance with previous methods, the video matting accuracy is evaluated by the Mean of Absolute Difference (MAD), Mean Squared Error (MSE), spatial Gradient (Grad) \cite{rhemann2009perceptually} and Connectivity (Conn) \cite{rhemann2009perceptually} errors. 
Conn metric is not analyzed at high resolution because it is too expansive to compute.
We also evaluate the temporal consistency of alpha matte predictions using dtSSD \cite{erofeev2015perceptually}.

\subsection{\noindent\textbf{Implementation Details}}

The video matting dataset we use to train our network is VideoMatte240K \cite{lin2021real} (VM).
Since this is a composited dataset, synthesized videos are usually not realistic and lack the context of real-world footage. 
Similar to the reasons explained in RVM \cite{lin2022robust}, we train our network on both matting and semantic segmentation datasets (PASCAL \cite{pascal-voc-2012}, COCO \cite{lin2014microsoft}, YouTubeVOS \cite{xu2018youtube}) to reduce the gap between synthetic and real domains.
Since the high-definition image matting dataset (Adobe Image Matting \cite{xu2017deep}, AIM) is not available for licensing reasons, we do not train our model using AIM and Distinctions-646 \cite{qiao2020attention}.
It is expected that our model could be further enhanced when including HD image matting data in the training pipeline.

We divide the training process into three stages. 
First, we initialize the MobileNetV2 \cite{sandler2018mobilenetv2} encoder with the weights pre-trained on ImageNet \cite{deng2009imagenet} and train the whole model on image segmentation datasets (PASCAL \cite{pascal-voc-2012}, COCO \cite{lin2014microsoft}). We apply multiple random motion augmentations \cite{lin2022robust} on static image segmentation datasets to synthesize video sequences. The model is trained for 100K iterations using AdamW with a learning rate of 1e-4 and weight decay of 0.03. %
Second, we train our model alternatively on low-resolution VM data (odd stages) and video segmentation data (even stages). 
The model is trained for another 100K iterations using AdamW with a learning rate of 2e-4 and weight decay of 0.07. 
Third, we train our model only on VM HD data to learn fine-grained alpha matte predictions. The model is trained for 20K iterations using AdamW with a learning rate of 1e-5 and weight decay of 0.07. 
The linear warmup and cosine learning rate scheduler is used at all three training stages. 

The initial segmentation mask ($t=0$) is estimated by an off-the-shelf Mask R-CNN detector \cite{he2017mask, wu2019detectron2} with ResNet50 \cite{he2016deep} backbone. 
The transformer in AdaM consists of three layers with a hidden size of 256D. The step $l$ in long-term attention is 10. To reduce computational complexity, the network stores up to 10 sets of Key and Value features for long-term attention. The window size, $\omega$ and $s$, for the short-term attention are 7 and 1, respectively. 
All experiments are performed on the Nvidia Tesla V100 GPUs. The mini-batch size is 2 clips per GPU.
\subsection{\noindent\textbf{Results on the VM Benchmark}}

The experimental results on VM dataset are summarized in Table~\ref{table:vm}.
Without training on high-definition data, AdaM outperforms previous state-of-the-art methods in most metrics by a considerable margin in both VM sets. %
In addition, AdaM with MobileNetV2 backbone outperforms RVM-Large with ResNet-50 in VM 1920$\times$1080 set. 
We attribute the notable improvement in performance to our integrated framework, which is designed to simultaneously differentiate foreground from background and capture alpha matte details in dynamic environments. The last set of results demonstrates that performance could be further enhanced with training on high-definition video data. 

\vspace{-3pt}
\subsection{\noindent\textbf{Results on the CRGNN Benchmark}}

To evaluate the robustness and generalization of our method trained on the composited dataset to a real-video dataset, we compare the matting quality by transferring MODNet \cite{ke2022modnet}, RVM \cite{lin2022robust}, and AdaM to the real dataset CRGNN-R.
We do not fine-tune the pre-trained models on CRGNN dataset. %
Table \ref{table:crgnn} shows AdaM is competitive among these leading methods, suggesting the proposed method can overcome the domain shift problem and generalize to real-world data more effectively.
We believe the proposed framework is a core factor for its general success.

\subsection{\noindent\textbf{Qualitative Analysis}}

\begin{figure}[]
\begin{center}
\includegraphics[trim=0 0 0 0, clip,width=1\linewidth]{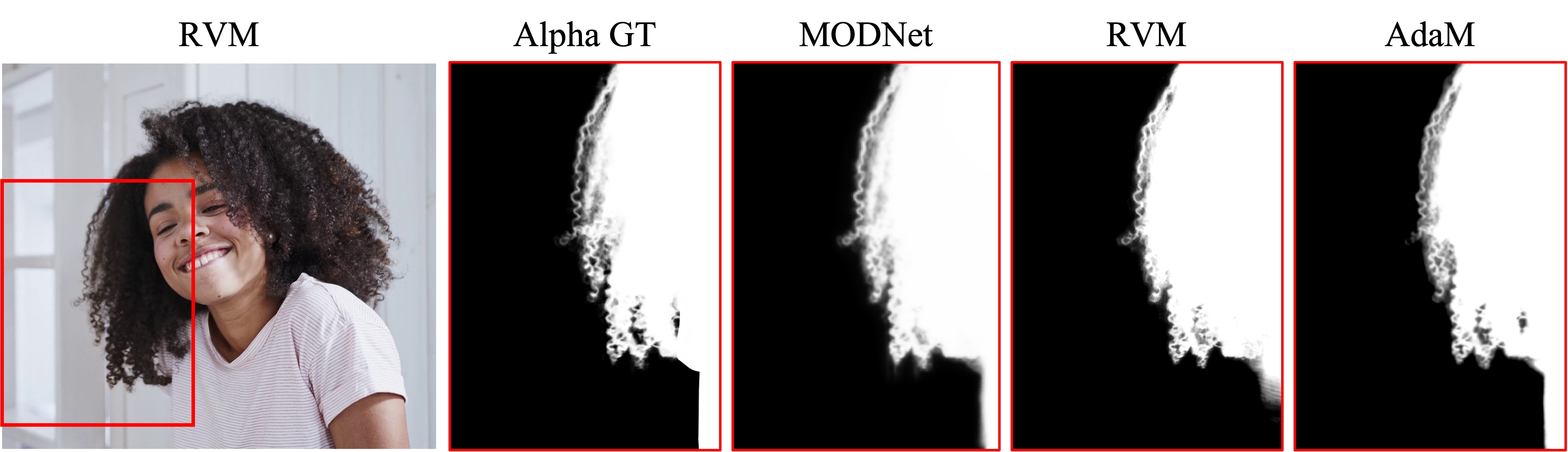}
\caption{\footnotesize
Comparison of alpha matte details.
}
\vspace{-1pt}
\label{fig:details}
\end{center}
\end{figure}

\begin{figure*}[]
\captionsetup[subfigure]{labelformat=empty}
\begin{center}
    \scriptsize
\begin{minipage}[]{.99\textwidth}
    \centering
    \footnotesize
    \begin{subfigure}[b]{0.21\textwidth}
        \caption{Input}
        \includegraphics[width=\textwidth]{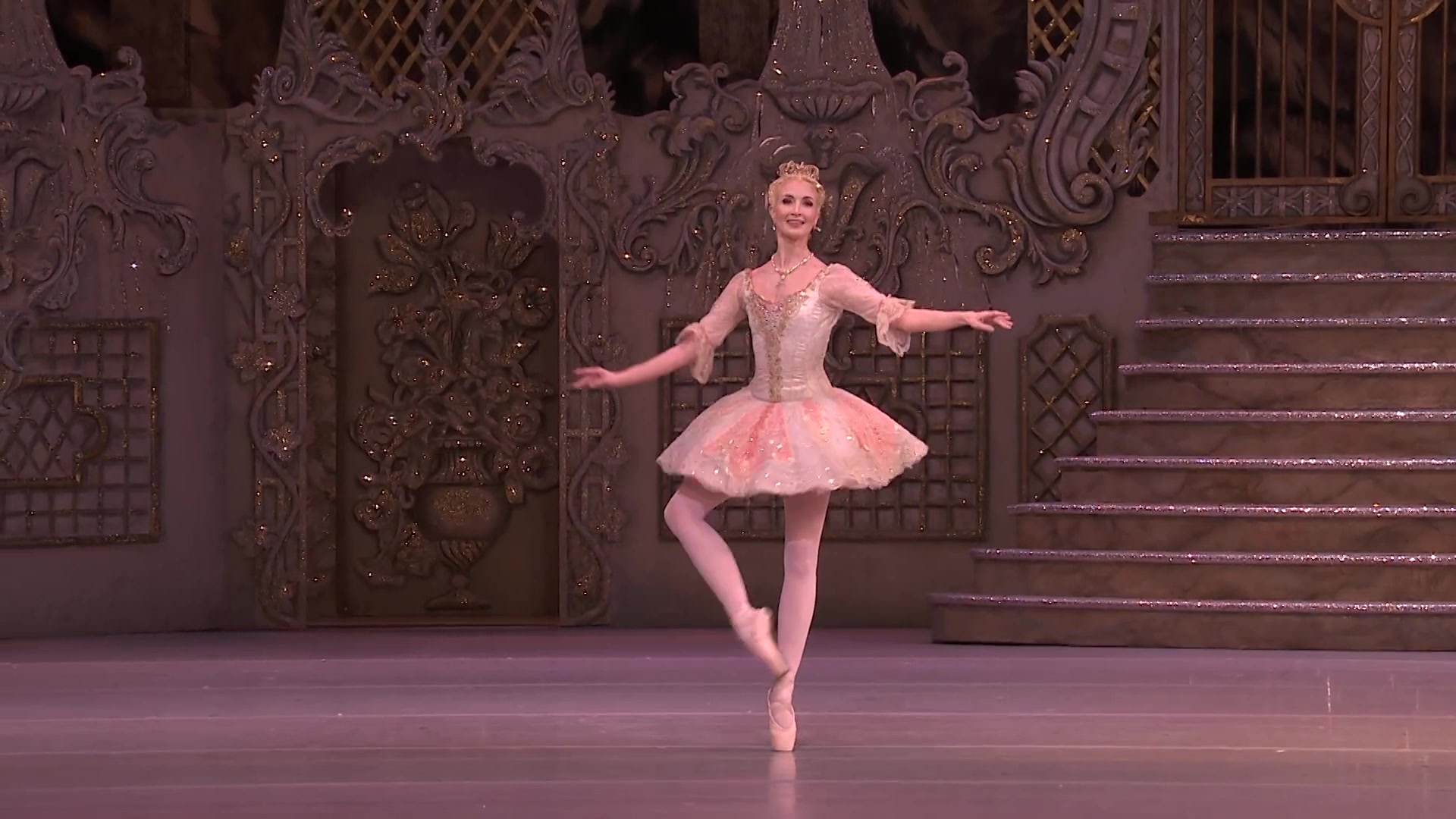}
    \end{subfigure}\hspace{0em}
    \begin{subfigure}[b]{0.21\textwidth}
        \caption{MODNet}
        \includegraphics[width=\textwidth]{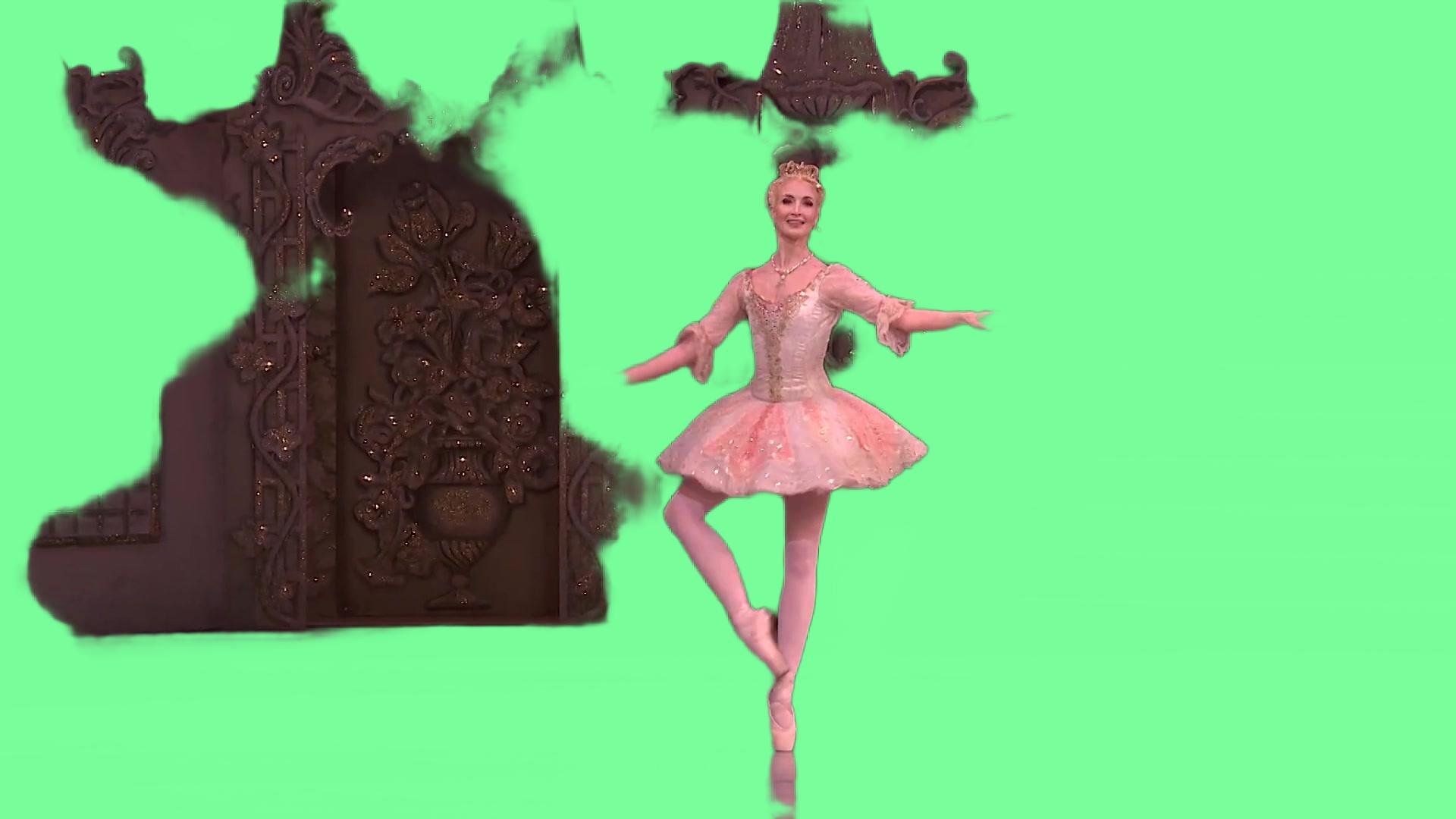}
    \end{subfigure}\hspace{0em}
    \begin{subfigure}[b]{0.21\textwidth}
        \caption{RVM}
        \includegraphics[width=\textwidth]{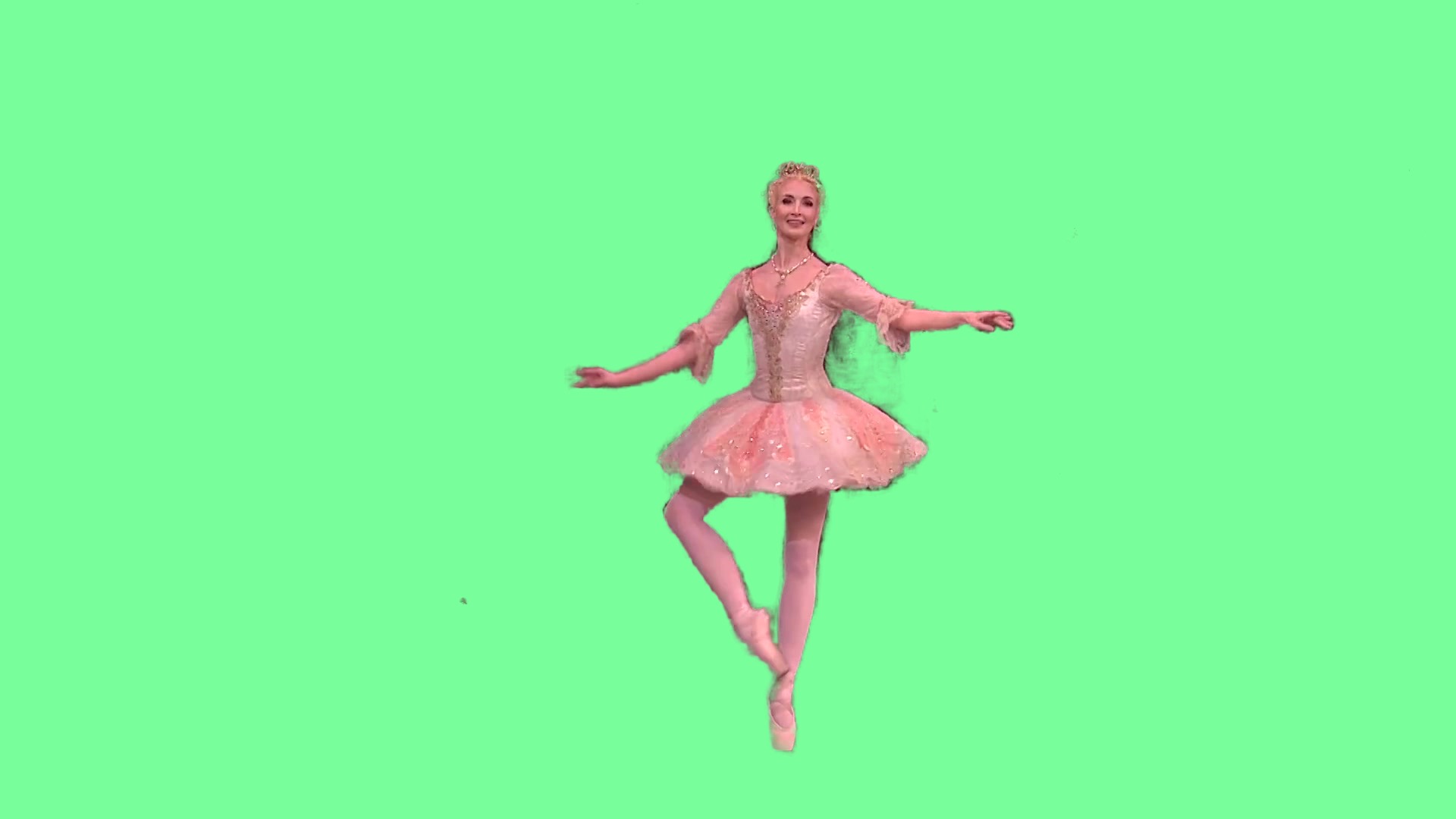}
    \end{subfigure}\hspace{0em}
    \begin{subfigure}[b]{0.21\textwidth}
        \caption{AdaM}
        \includegraphics[width=\textwidth]{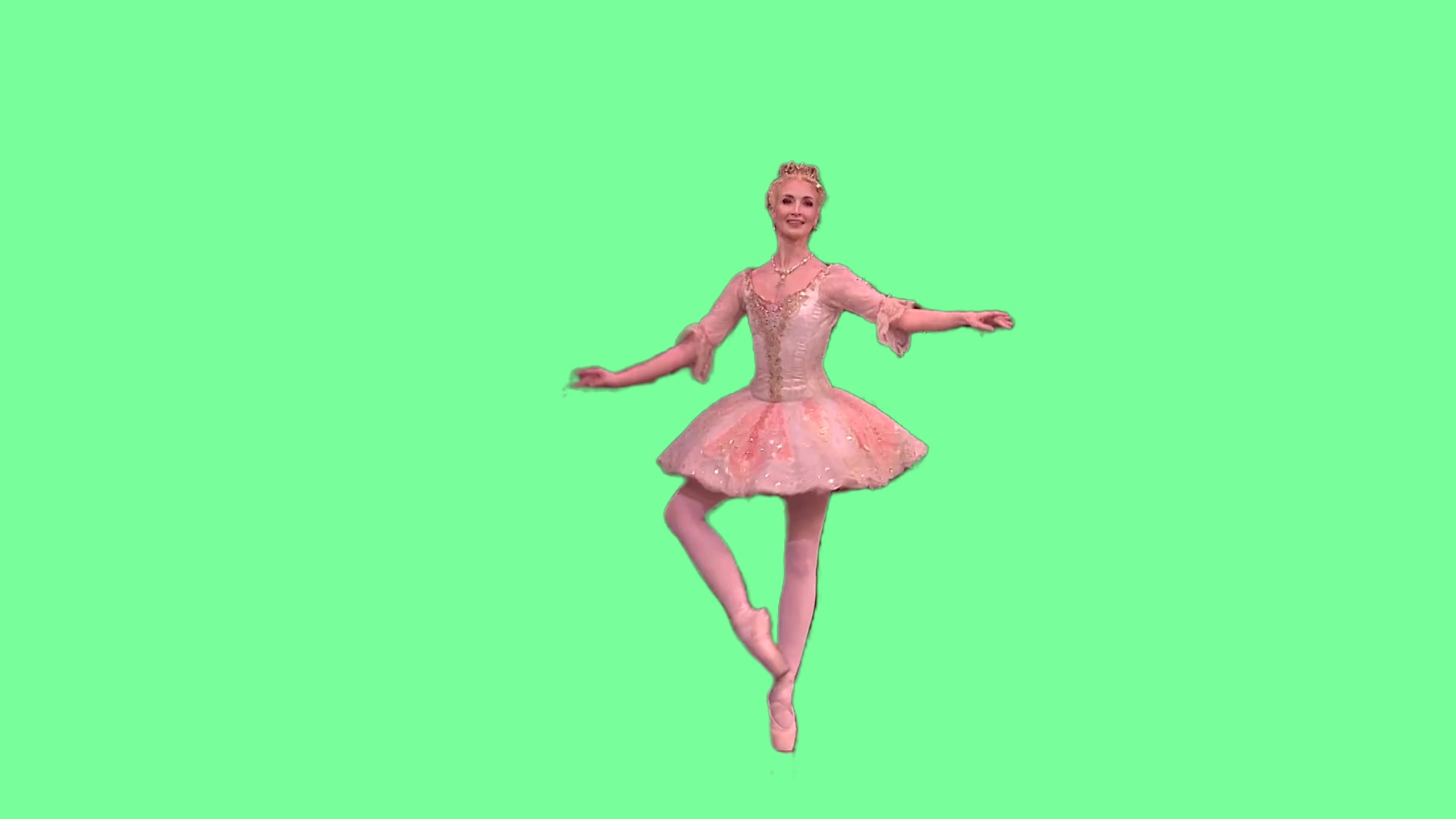}
    \end{subfigure}\hspace{0em}
    \subcaption{\footnotesize (a) Video scenes: The ballet dancer moves at a moderate tempo, but the camera moves quickly.}
    \vspace{10pt}
    \end{minipage}
   
\begin{minipage}[]{.99\textwidth}
        \centering
        \footnotesize
    \includegraphics[trim=0 0 0 0, clip,width=0.21\textwidth]{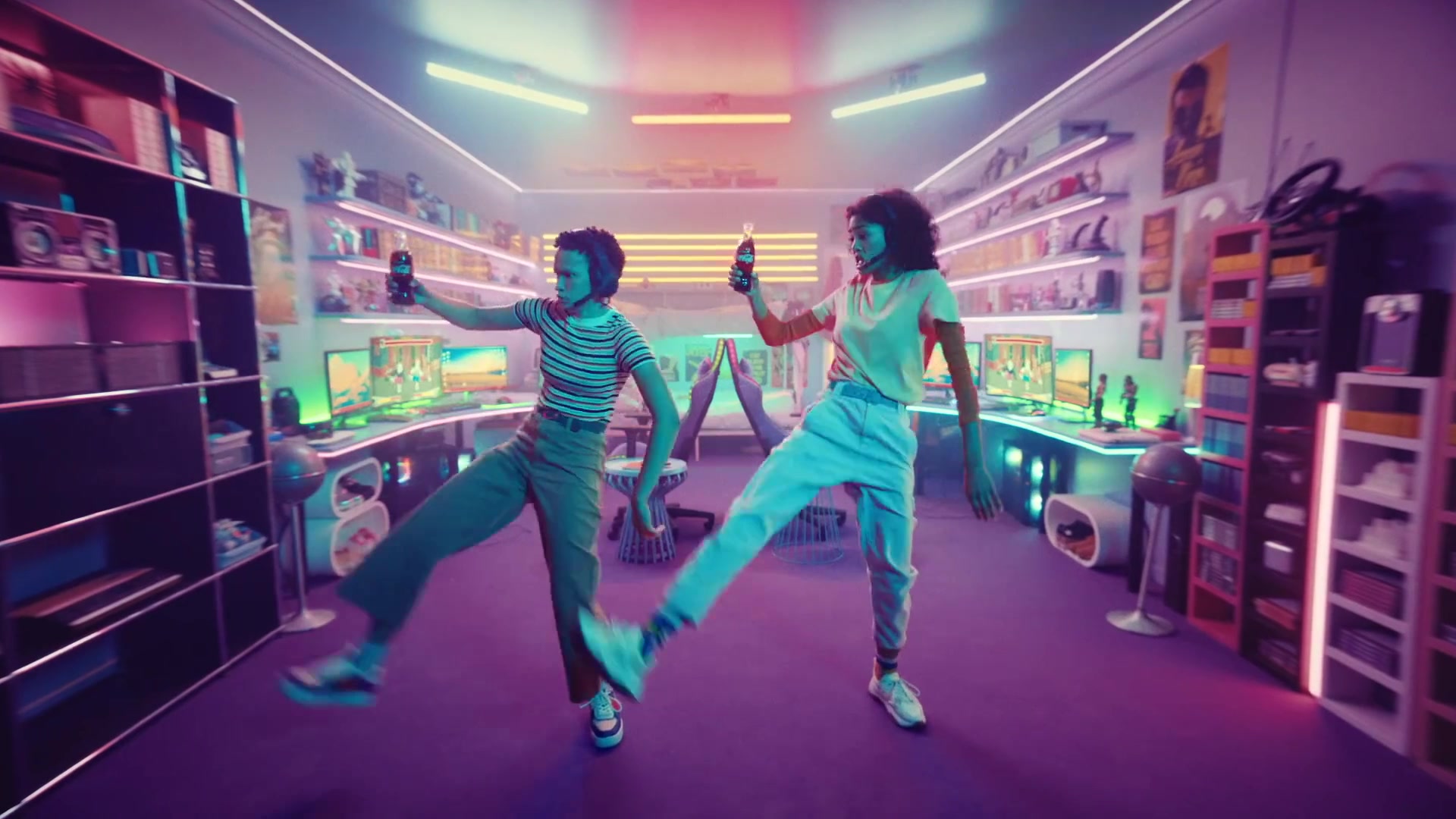}\hspace{0em}
    \includegraphics[trim=0 0 0 0, clip,width=0.21\textwidth]{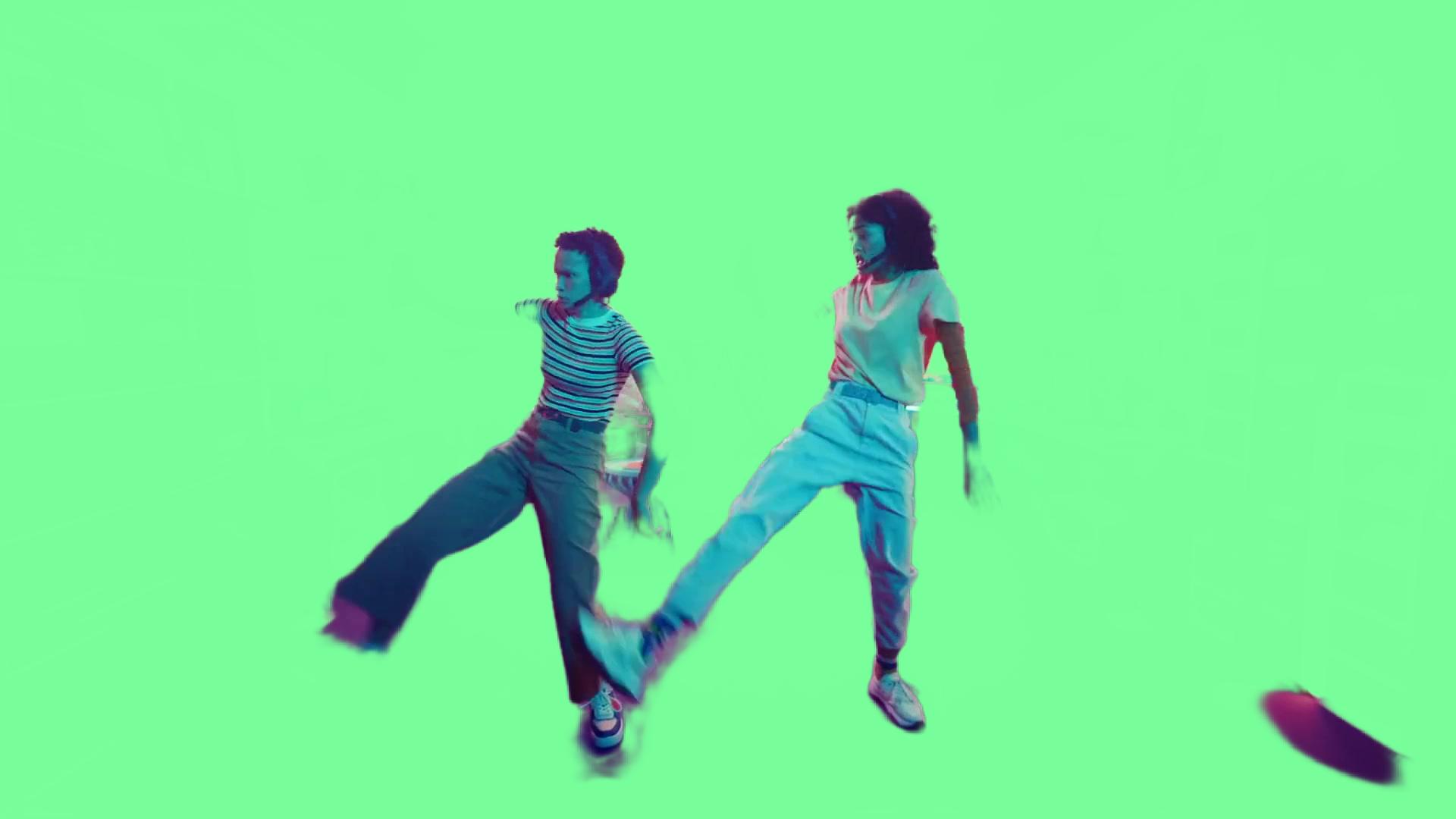}\hspace{0em}
    \includegraphics[trim=0 0 0 0, clip,width=0.21\textwidth]{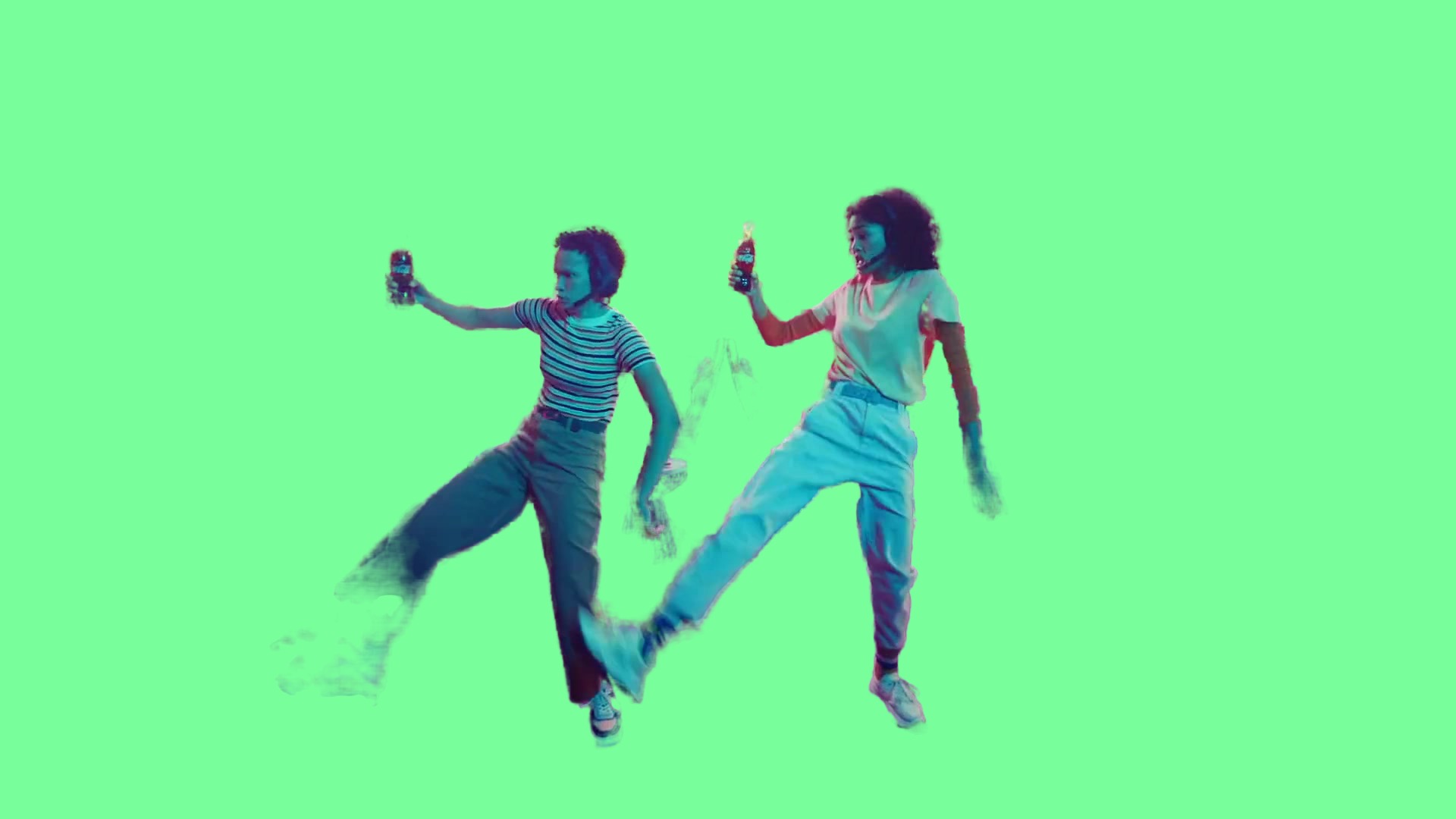}\hspace{0em}
    \includegraphics[trim=0 0 0 0, clip,width=0.21\textwidth]{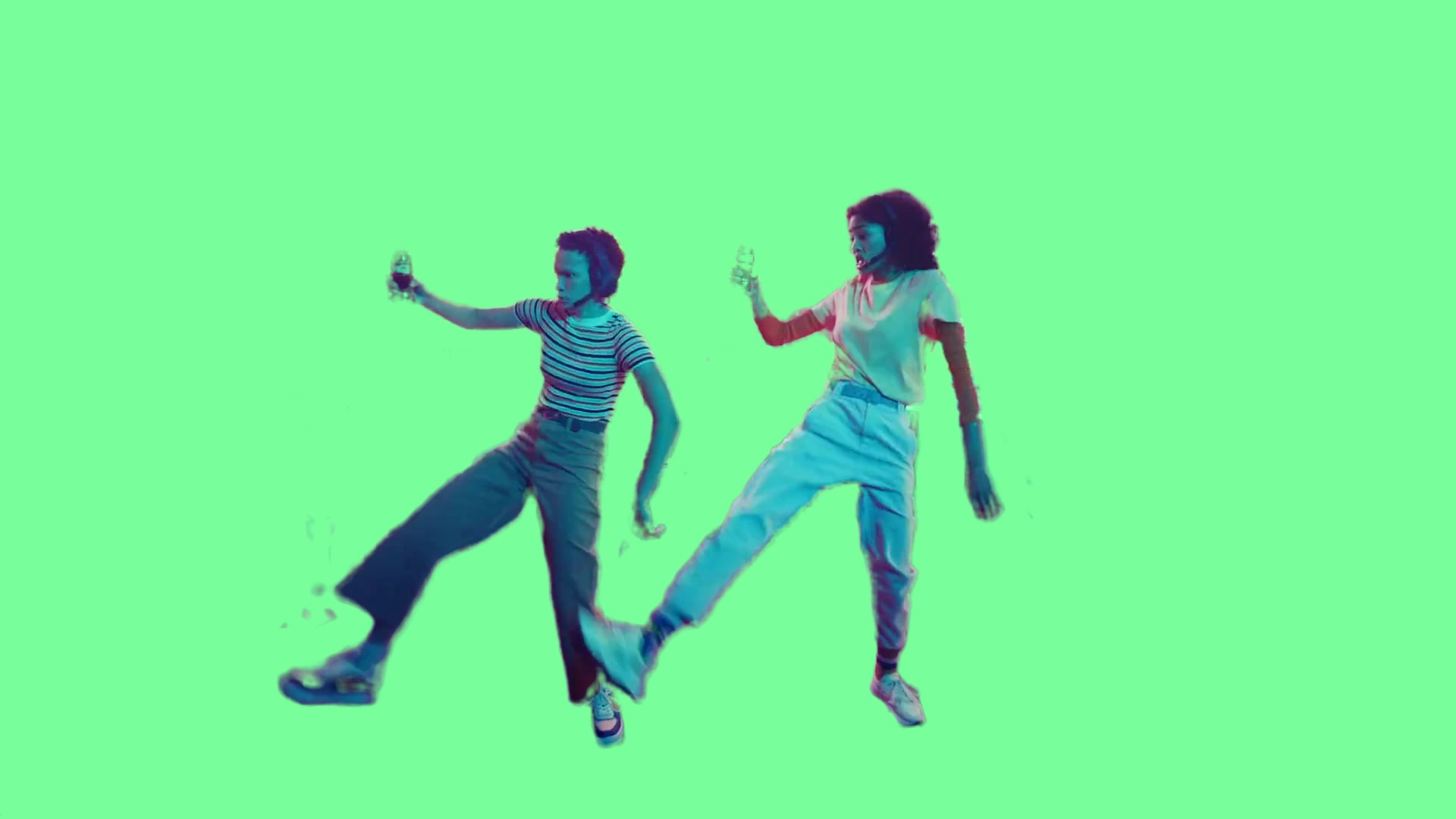}\hspace{0em}
    \subcaption{\footnotesize (b) Video scenes: The camera slowly zooms in, but the subjects are moving at a very rapid pace.}
    \vspace{10pt}
    \end{minipage}

\begin{minipage}[]{.99\textwidth}
        \centering
        \footnotesize
    \includegraphics[trim=0 0 0 0, clip,width=0.21\textwidth]{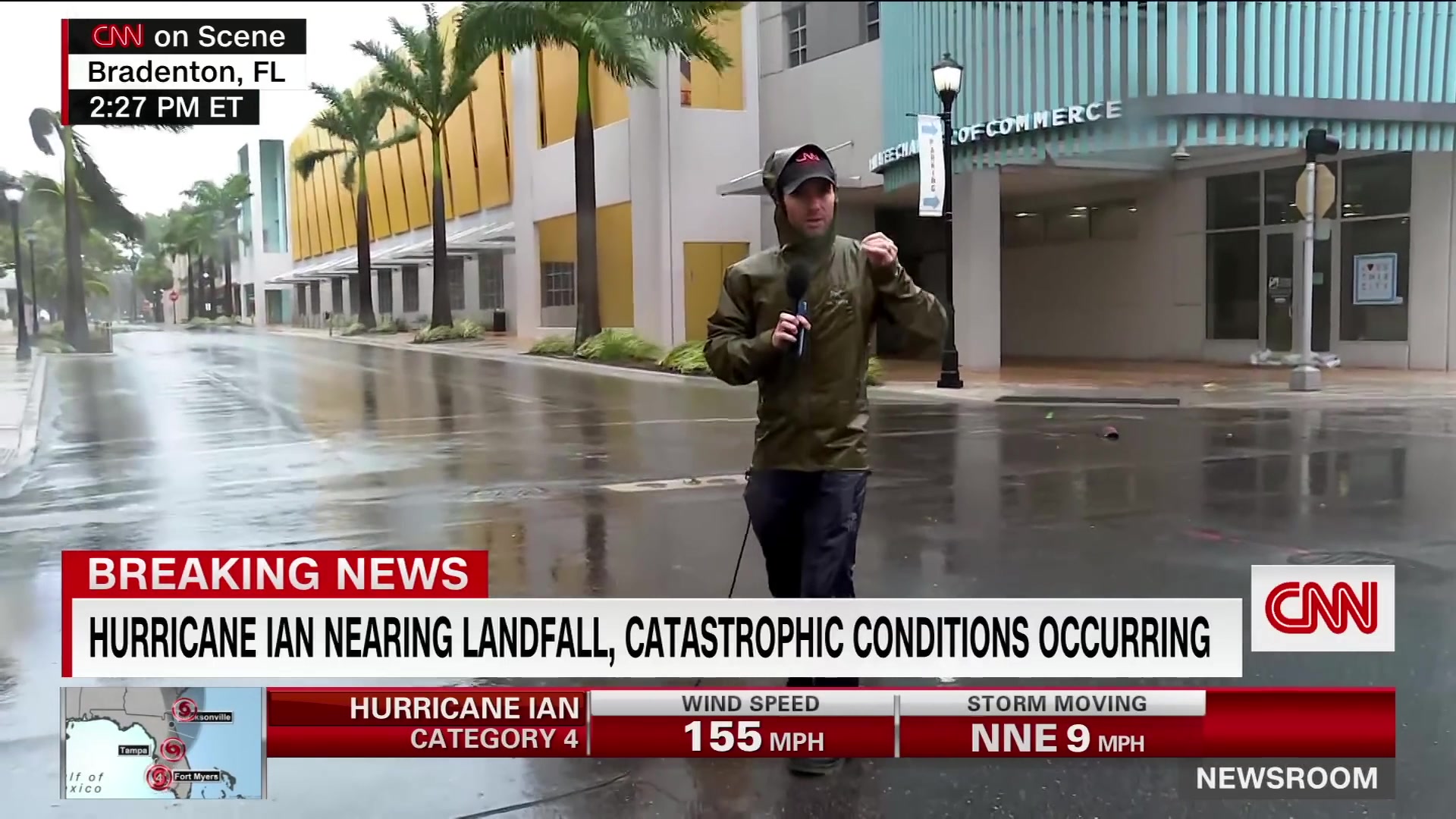}\hspace{0em}
    \includegraphics[trim=0 0 0 0, clip,width=0.21\textwidth]{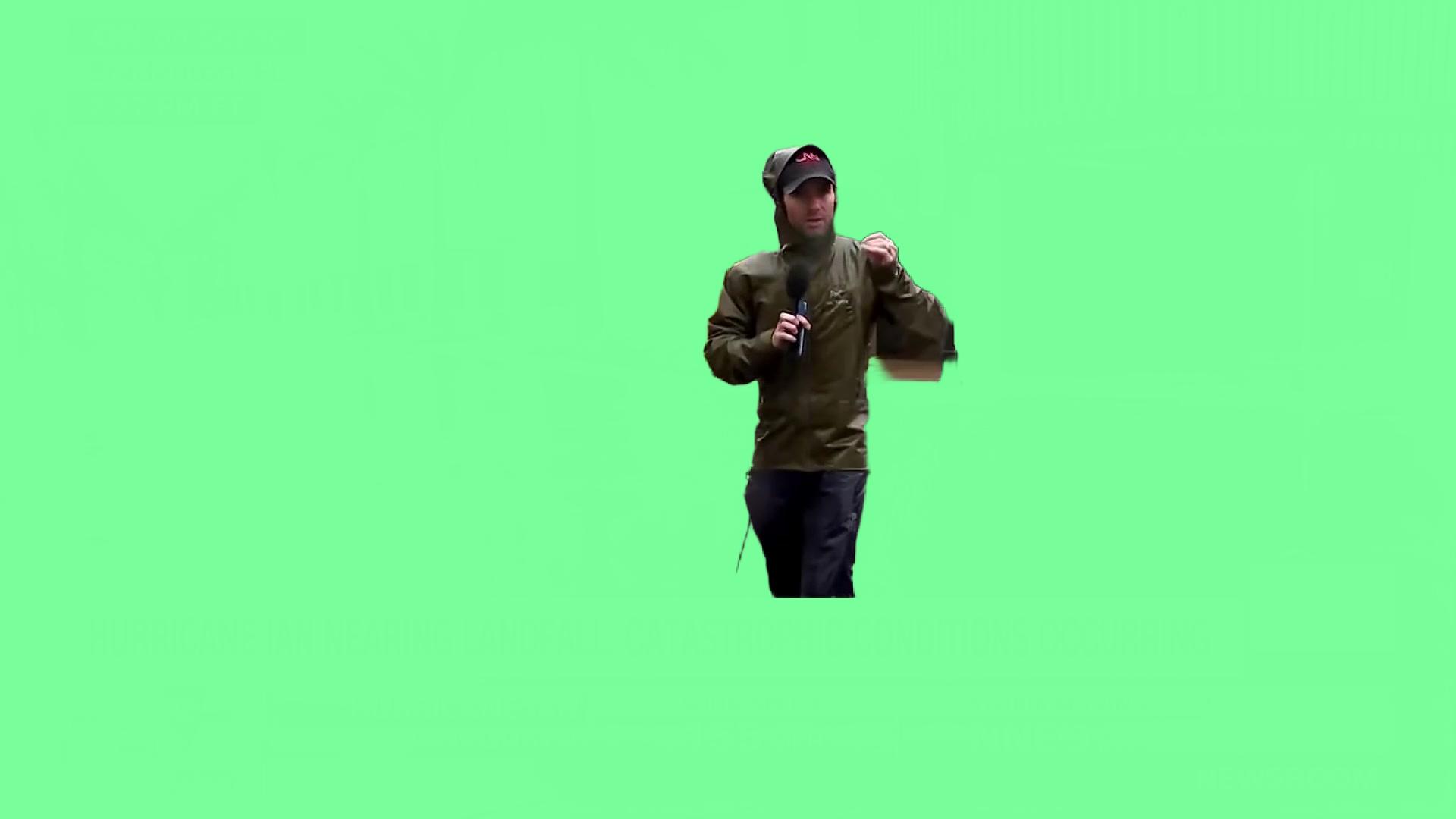}\hspace{0em}
    \includegraphics[trim=0 0 0 0, clip,width=0.21\textwidth]{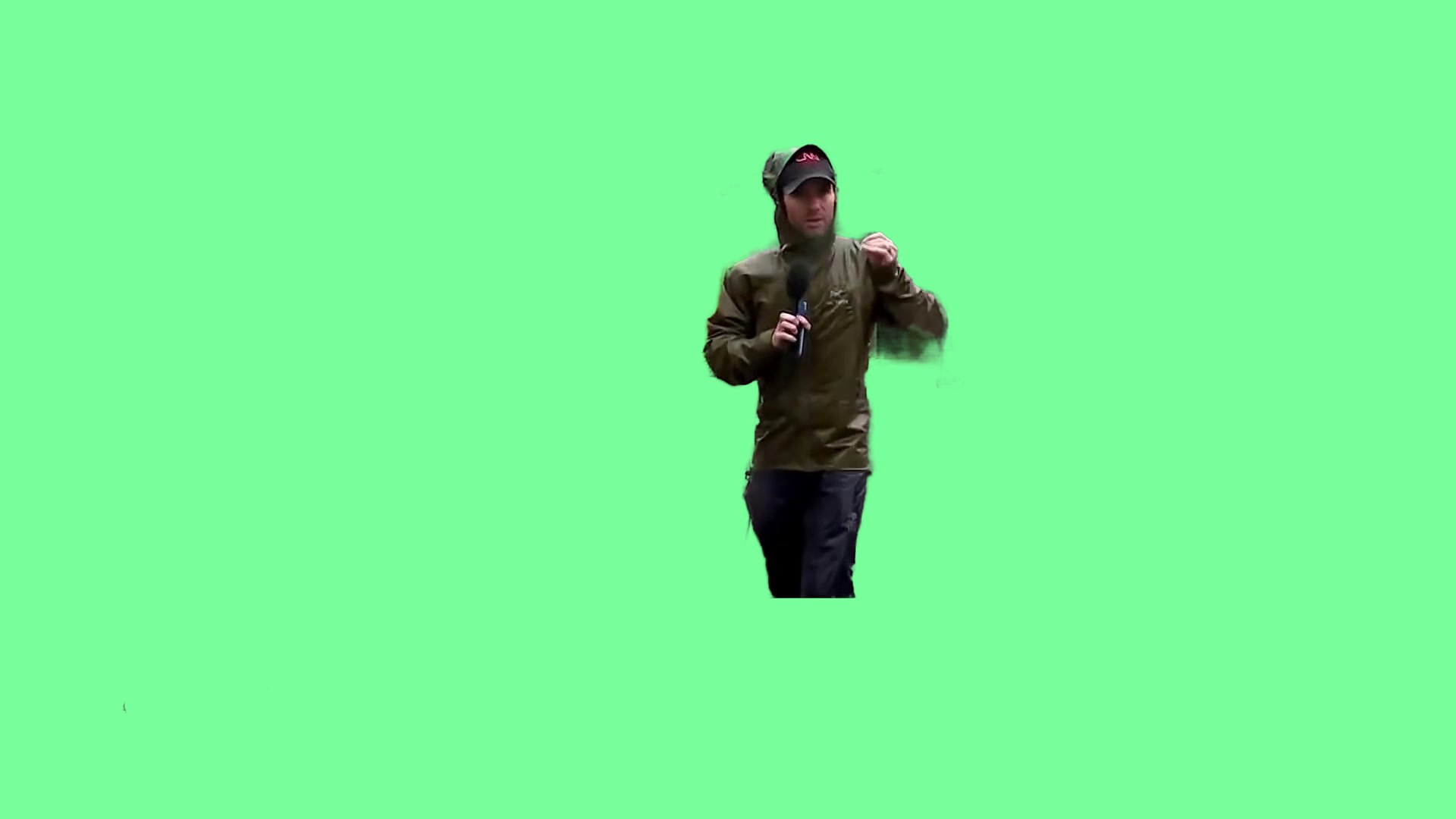}\hspace{0em}
    \includegraphics[trim=0 0 0 0, clip,width=0.21\textwidth]{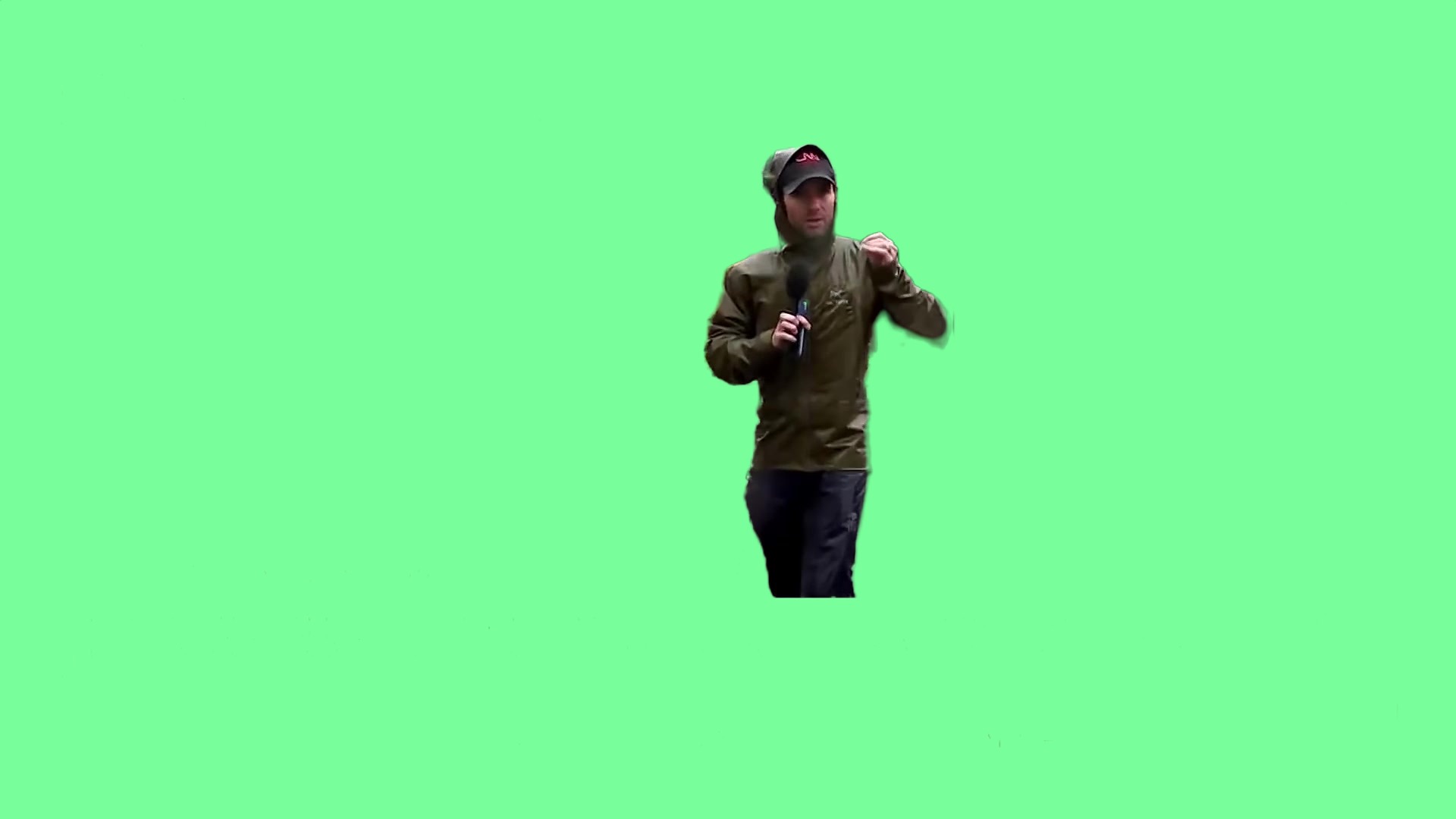}\hspace{0em}
    \subcaption{\footnotesize (c) Video scenes: A reporter is standing in a hurricane area. A strong wind blows his clothes. }
    \vspace{10pt}
    \end{minipage}

\begin{minipage}[]{.99\textwidth}
        \centering
        \footnotesize
    \includegraphics[trim=0 0 0 0, clip,width=0.21\textwidth]{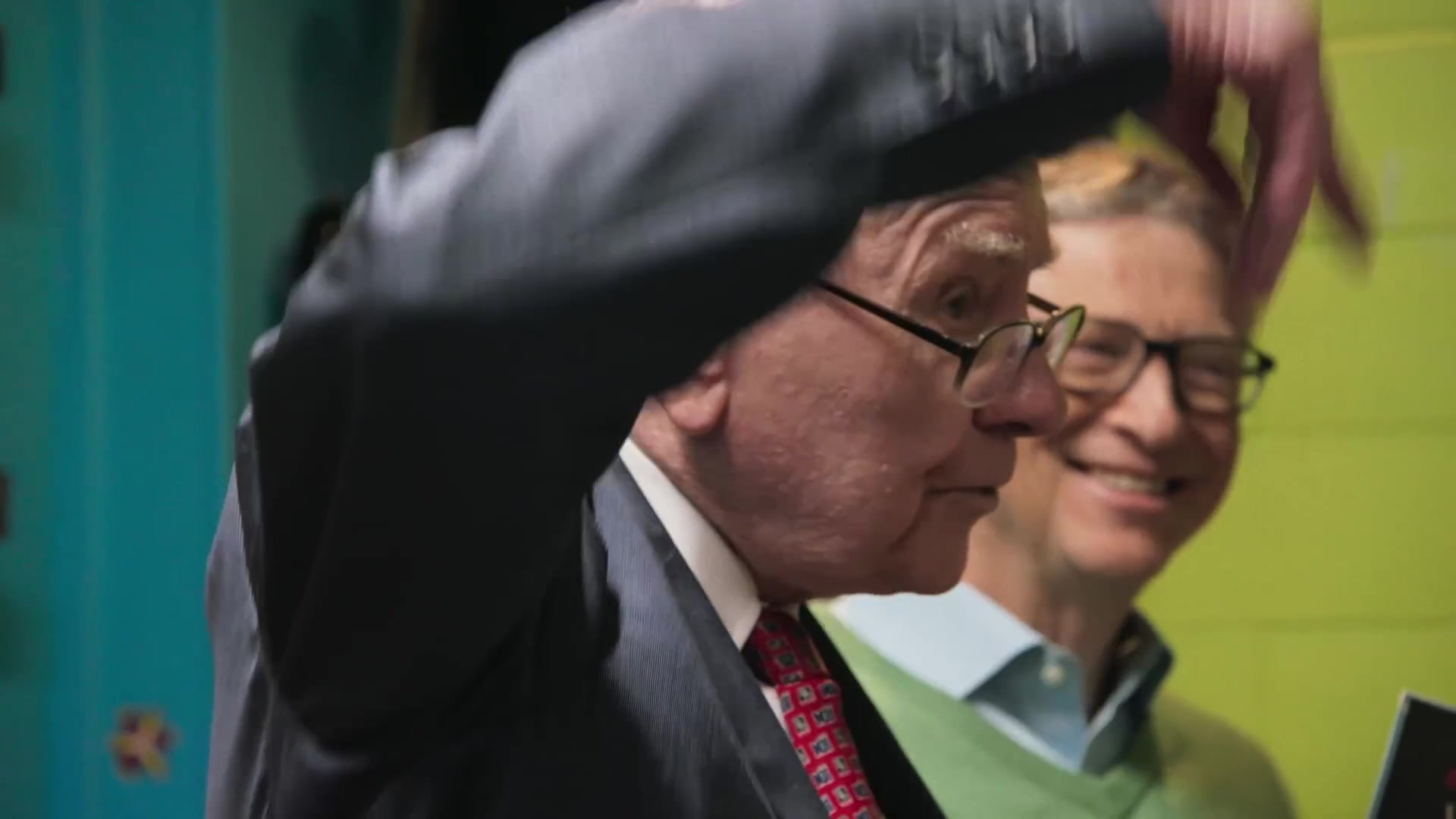}\hspace{0em}
    \includegraphics[trim=0 0 0 0, clip,width=0.21\textwidth]{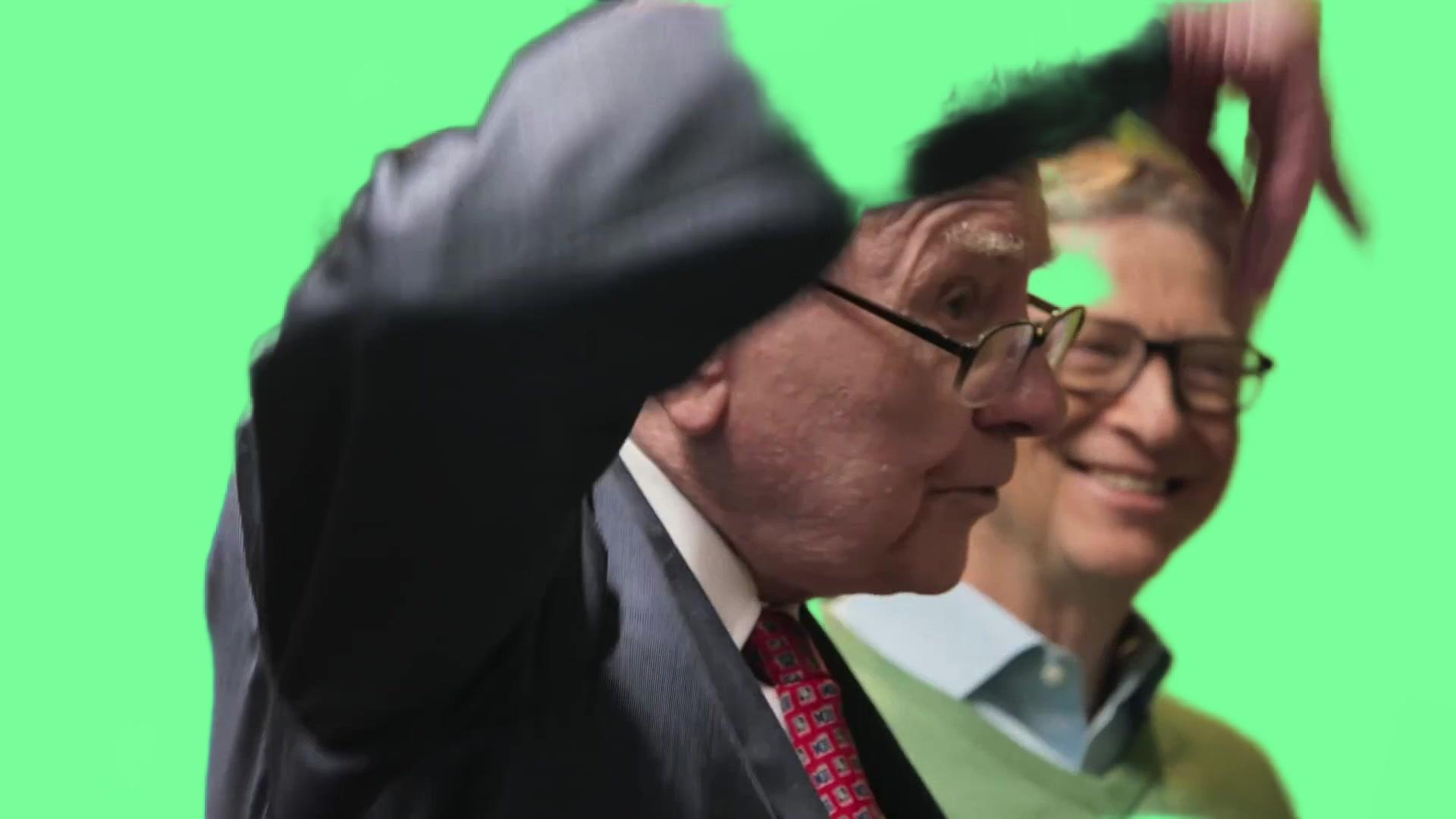}\hspace{0em}
    \includegraphics[trim=0 0 0 0, clip,width=0.21\textwidth]{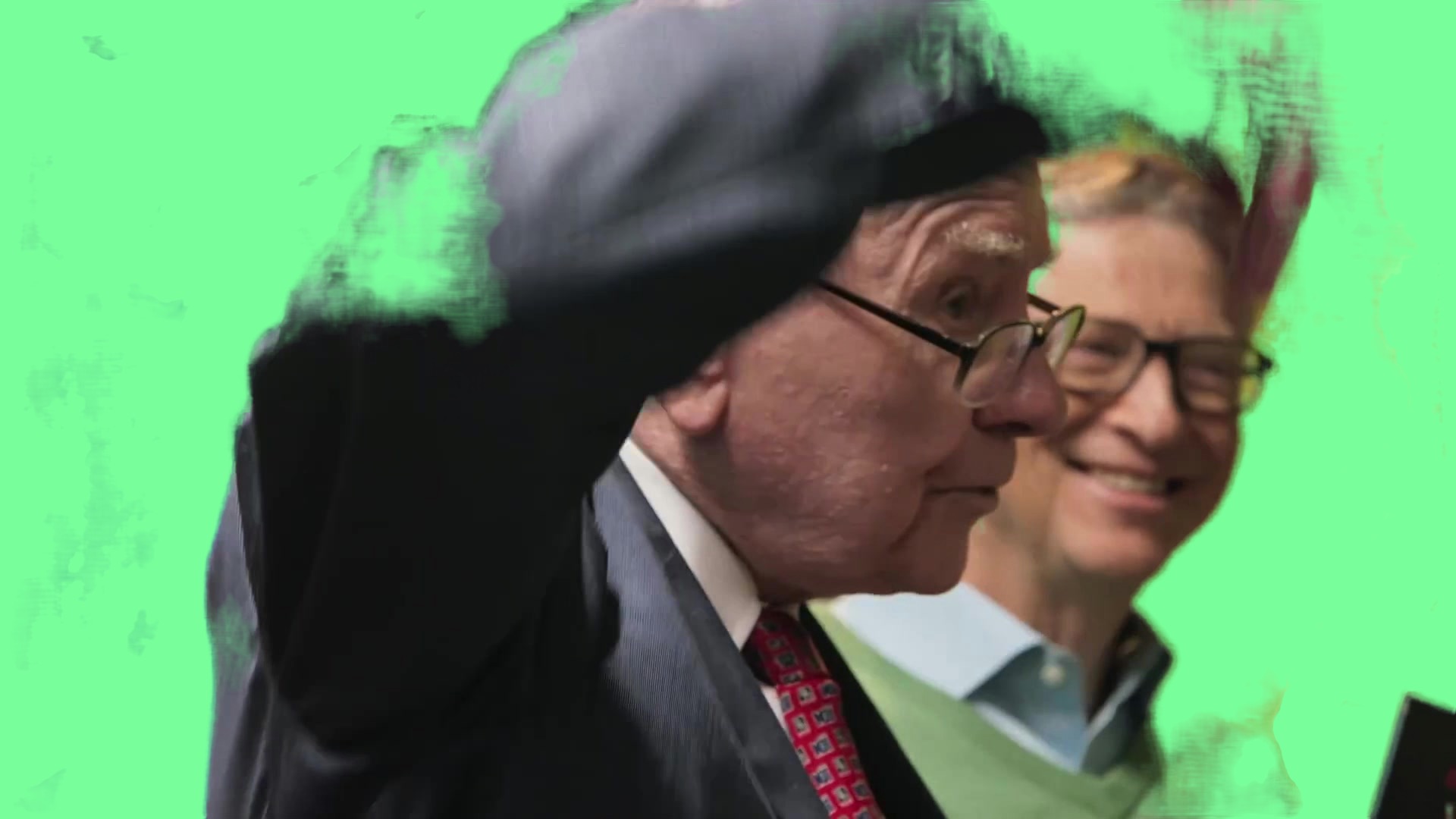}\hspace{0em}
    \includegraphics[trim=0 0 0 0, clip,width=0.21\textwidth]{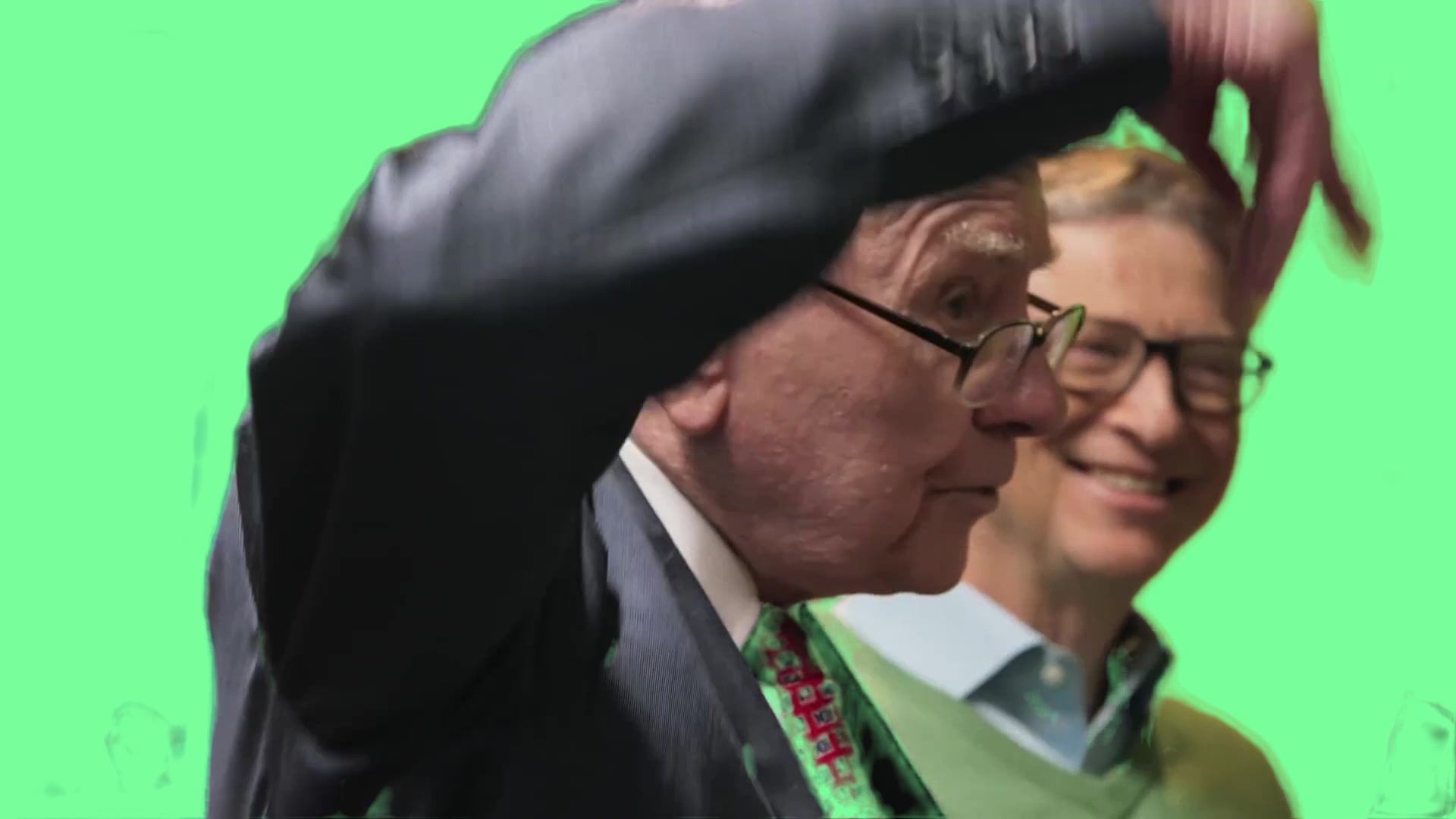}\hspace{0em}
    \subcaption{\footnotesize (d) Video scenes: The camera zooms in following the scenes in left panel of Figure \ref{fig:first}. }
    \vspace{10pt}
    \end{minipage}

\begin{minipage}[]{.99\textwidth}
        \centering
        \footnotesize
    \includegraphics[trim=0 0 0 0, clip,width=0.21\textwidth]{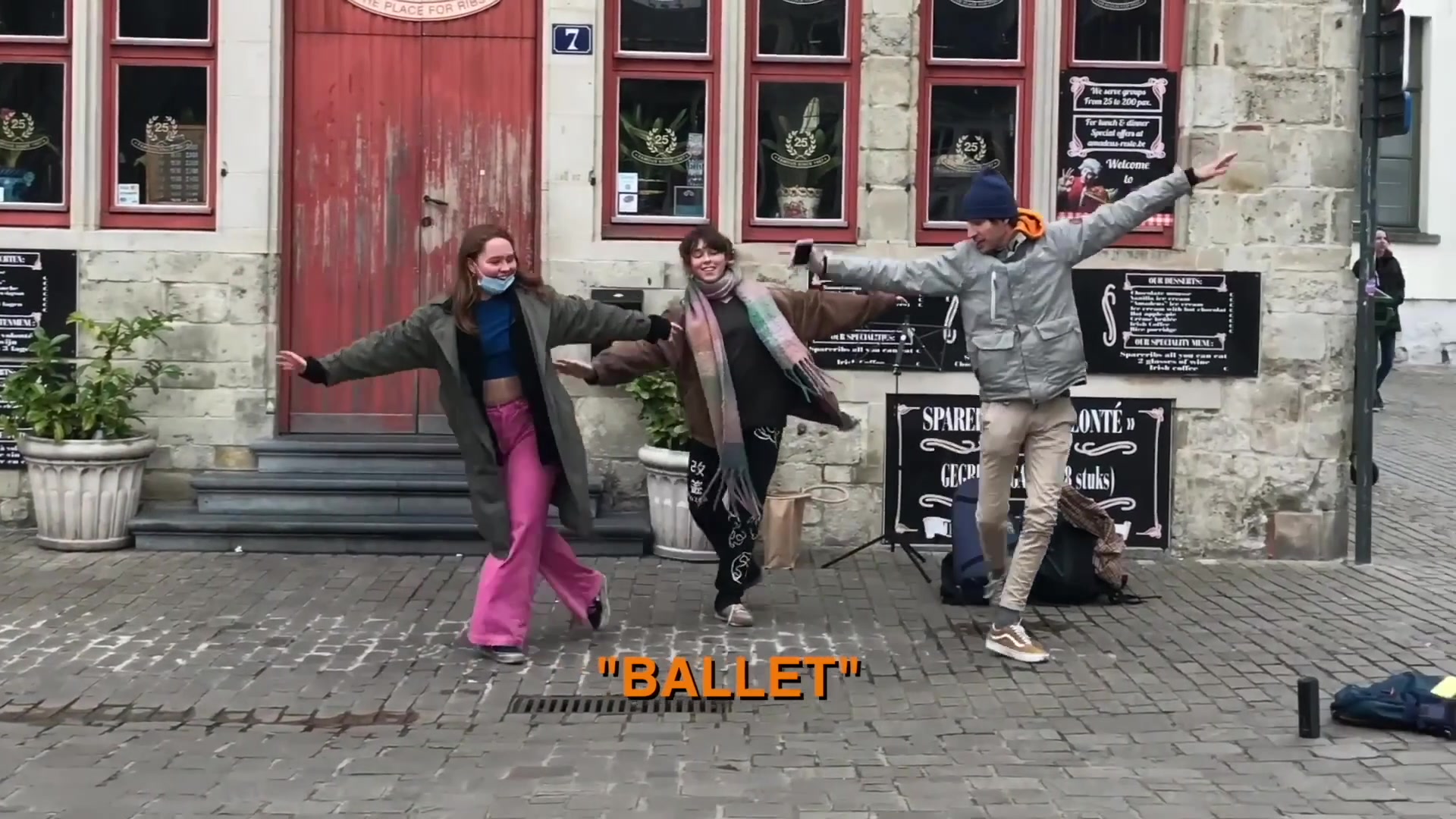}\hspace{0em}
    \includegraphics[trim=0 0 0 0, clip,width=0.21\textwidth]{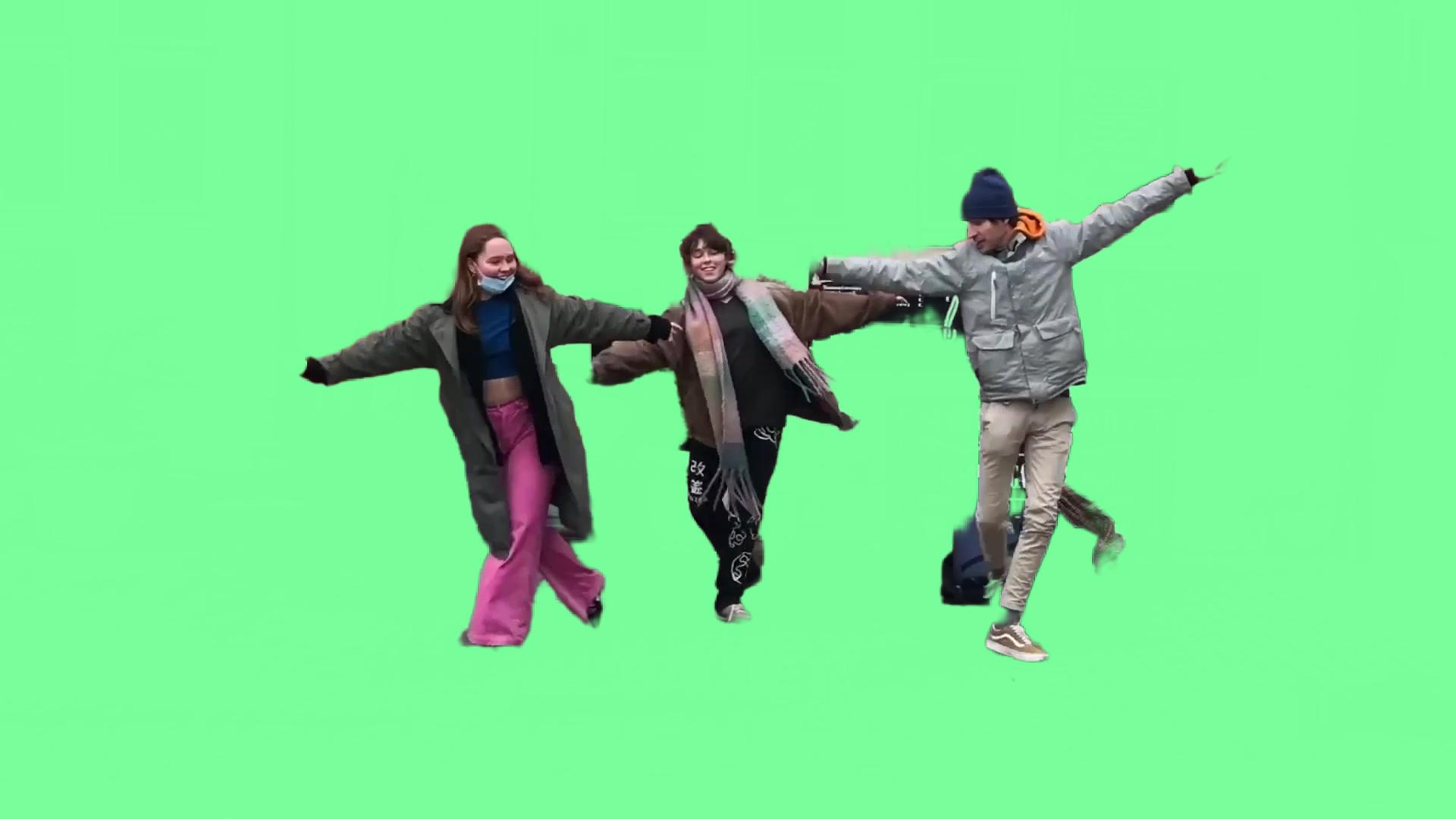}\hspace{0em}
    \includegraphics[trim=0 0 0 0, clip,width=0.21\textwidth]{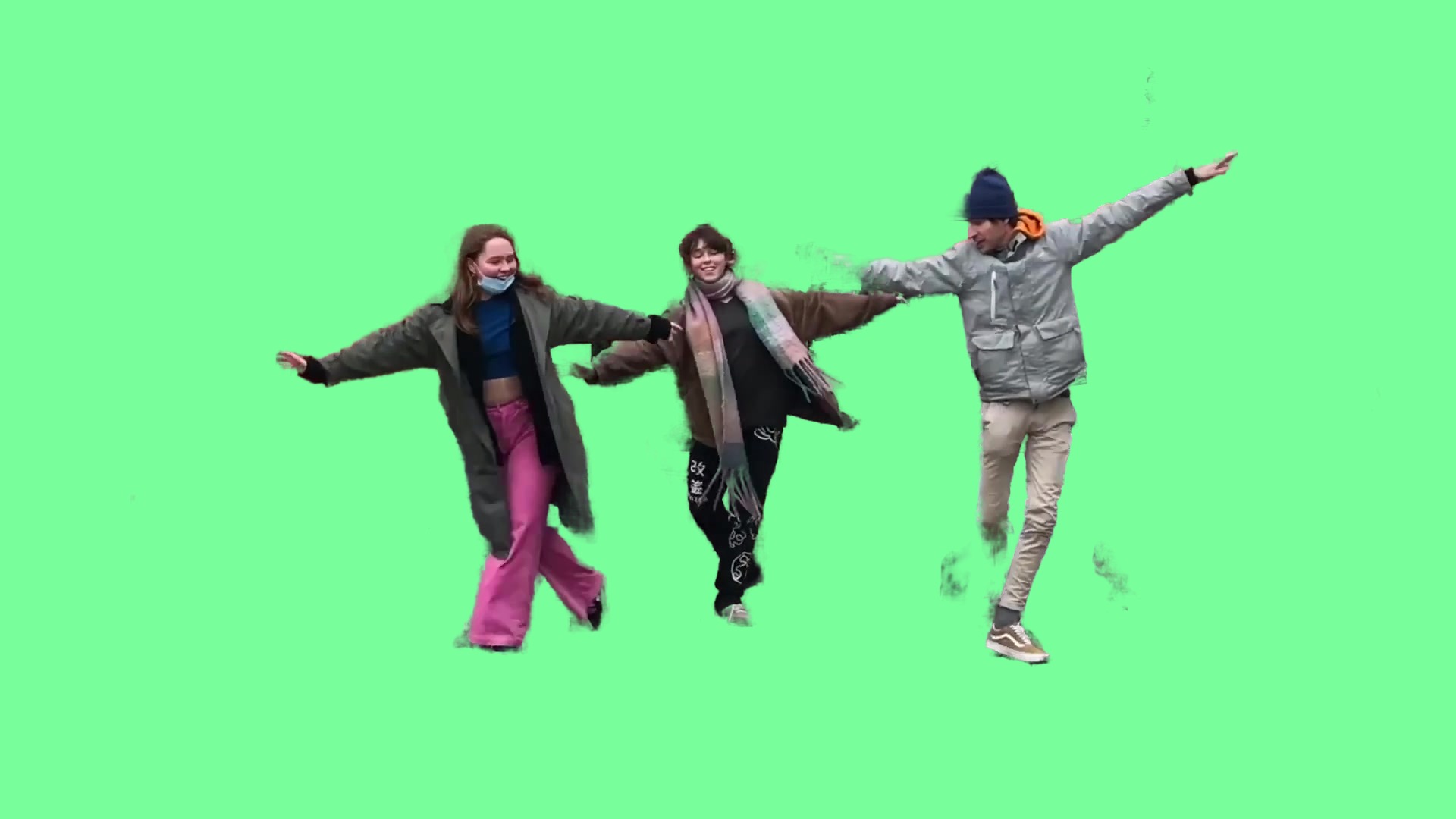}\hspace{0em}
    \includegraphics[trim=0 0 0 0, clip,width=0.21\textwidth]{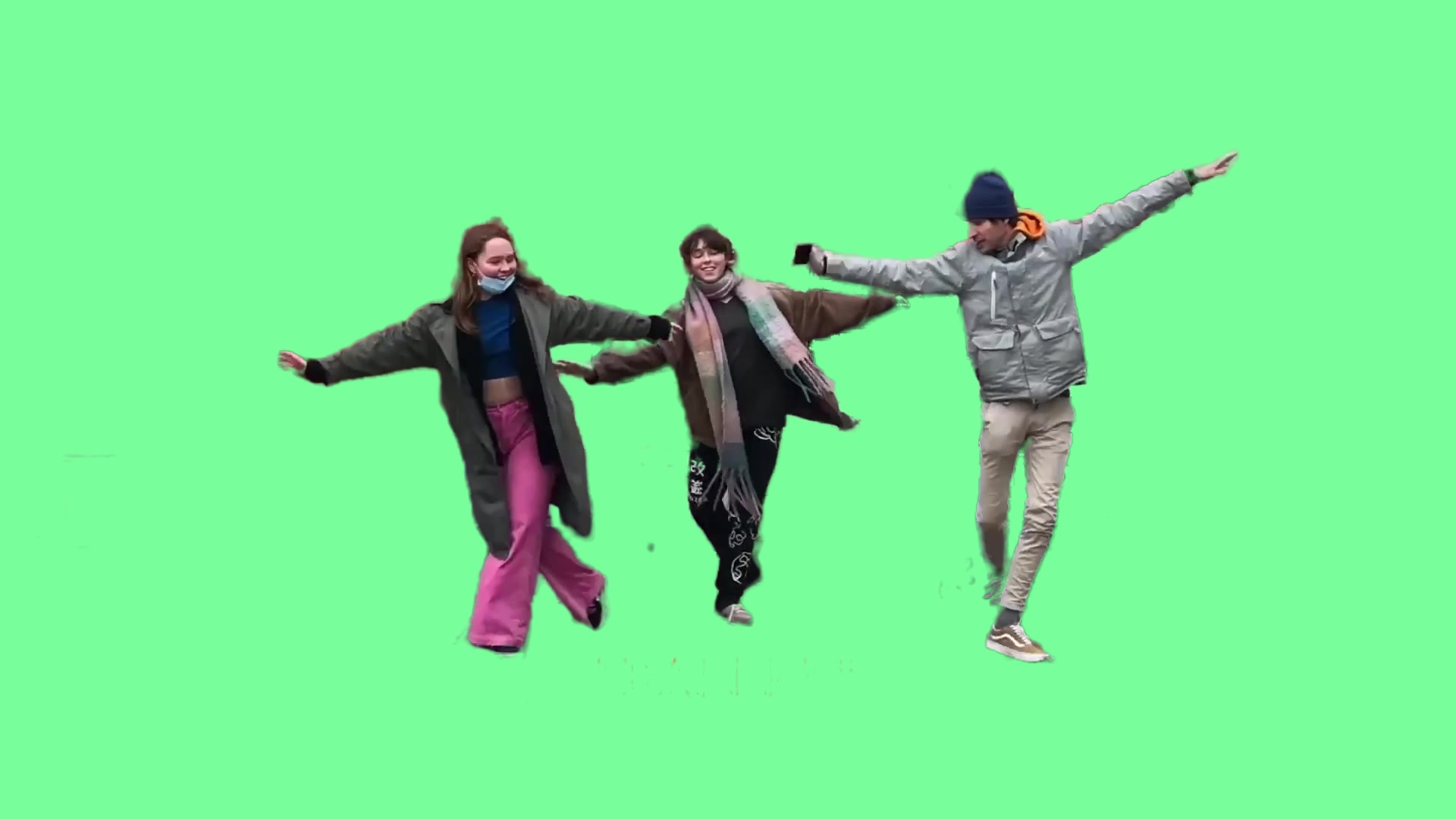}\hspace{0em}
    \subcaption{\footnotesize (e) Video scenes: Street scenes are captured by a camera following the subjects' movements.}
    \vspace{3pt}
    \end{minipage}
   
  \label{fig:visual}

  \end{center}

    \centering

   \caption{\small Comparisons of our model to existing methods on complex real-world videos.}
   
\label{fig:visual2}
\end{figure*}

As MODNet, RVM and the proposed method do not require manual inputs (e.g., trimaps or pre-captured backgrounds), we are able to conduct further tests on real-world videos. 
In Figure \ref{fig:details}, we compare the alpha matte predictions. Our method is able to predict fine-grained details more accurately. 
Figure \ref{fig:visual2} presents qualitative comparisons on real-world videos from YouTube.\footnote{\scriptsize The examples are drawn from test experiments on videos retrieved from YouTube for research purposes to evaluate AdaM's robustness in complex scenarios.} Each sub-figure illustrates a challenging case. 
Despite being top-performing methods, MODNet and RVM produce blurry foreground boundaries in videos with dynamic backgrounds or reveal some backgrounds behind the subjects, indicating that real-world video matting presents a fundamental challenge for existing techniques. 
Our visualization depicts comparably reliable results in highly challenging scenes (e.g., fast camera movement in Figure \ref{fig:visual2} (a), fast human motion in in Figure \ref{fig:visual2} (b), and cluttered background in Figure \ref{fig:visual2} (e)).

\subsection{\noindent\textbf{Ablation Study}}

For a controlled evaluation, the ablations are performed using the same training recipe on the VM 1920$\times$1080 set.

\begin{figure}[h]
\begin{center}
\includegraphics[trim=0 0 0 0, clip,width=.85\linewidth]{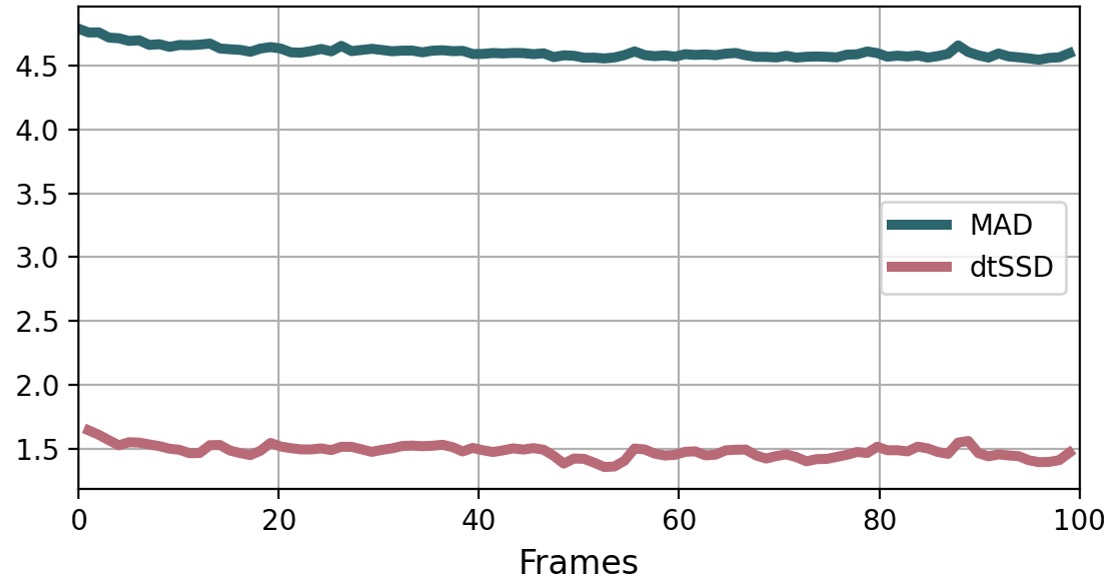}
\caption{
MAD and dtSSD performance of AdaM across frames.}
\vspace{-6pt}
\label{fig:temporal}
\end{center}
\end{figure}

\vspace{-6pt}
\begin{figure}[h]
\begin{center}
\includegraphics[trim=0 0 0 0, clip,width=1\linewidth]{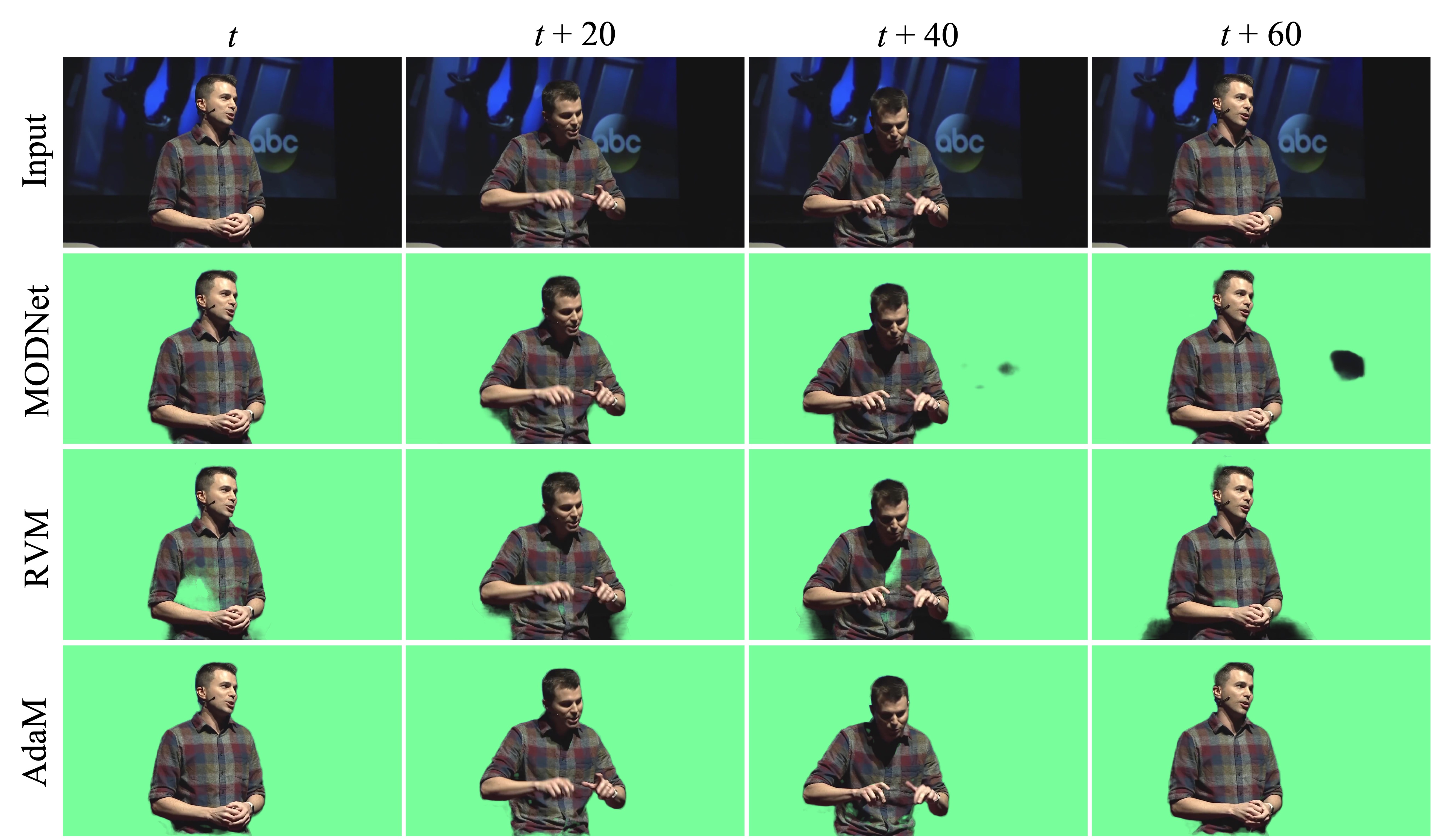}
\caption{\footnotesize
Evaluation of temporal coherence. Video scenes: A Ted talk speaker talks with very fast pose gesture movements.
}
\vspace{-12pt}
\label{fig:motion}
\end{center}
\end{figure}

\vspace{-6pt}
\noindent\textbf{Performance over temporal dimension.} Figure~\ref{fig:temporal} shows the average alpha MAD and dtSSD of AdaM along time steps in VM test set.
Overall, we observe the error (MAD) and temporal consistency (dtSSD) measures improve in the first 10 frames and then stay stable over time.
Further, we examine the temporal coherence of AdaM. In Figure \ref{fig:motion}, a Ted talk speaker gives a speech with rapid gesture movements. In comparison with MODNet \cite{ke2022modnet} and RVM \cite{lin2022robust}, our model produces less undesirable artifacts and more consistent alpha matte predictions.

\vspace{3pt}
\noindent\textbf{Performance in static and dynamic background.}
Table \ref{tab:static} compares our model performance on static and dynamic backgrounds in VM. In dynamic videos, both foreground subjects and background move, while in static videos, only foreground subjects move. In static videos, BGMv2 produces the best results as it uses ground-truth background as an additional input. However, its performance drops significantly in dynamic videos. Overall, our model performs robustly on both static and dynamic backgrounds and presents little performance difference in two sets. 
\begin{table}[hbt!]
\centering
\tablestyle{3.5pt}{1.1} 
   \begin{tabular}{cc|cccc} 
   \shline
   BG  &  Method & MAD $\downarrow$  & MSE $\downarrow$  & Grad $\downarrow$ & dtSSD $\downarrow$ \\ 
   \hline
    \multirow{4}{*}{Static} & BGMv2 \cite{lin2021real} & 4.33 & 0.32 & 4.19 & 1.33 \\
     & MODNet \cite{ke2022modnet} & 11.04 & 5.42 & 15.80 & 3.10 \\
     & RVM \cite{lin2022robust} & 5.64 & 1.07 & 9.80 & 1.84 \\
     & AdaM  & 4.56 & 0.43 & 5.91 & 1.41 \\
    \hline
    \multirow{4}{*}{Dynamic} & BGMv2 \cite{lin2021real}  & 42.45 & 37.05 & 17.30 & 4.61 \\
     & MODNet \cite{ke2022modnet}  & 11.23 & 5.65 & 14.79 & 3.06 \\
     & RVM \cite{lin2022robust}  & 7.50 & 2.80 & 11.30 & 1.96 \\
     & AdaM  & 4.65 & 0.48 & 6.20 & 1.51 \\
   \shline
   \end{tabular}\\
\caption{ Performance in static and dynamic backgrounds.} 
   \label{tab:static}
\vspace{-3pt}

\end{table}

\vspace{3pt}
\noindent\textbf{Shrot- and long-term attention.}
To evaluate the effect of short- and long-term attention in transformer, we train several models, turning them on and off. Table \ref{tab:attention} shows that the models with single attention produce inferior results. Comparing these two models, the model with short-term attention performs better in dtSSD. As short-term local attention propagates contextual information among neighboring pixels in nearby frames, it helps enhance temporal consistency in alpha matte predictions. In contrast, long-term global attention addresses recurrence and dependencies of Fg/Bg, contributing to MAD and MSE. With both short- and long-range attention, spatial interdependency and motion consistency can be captured.

\begin{table}[!h]
\centering
\tablestyle{7pt}{1.1} 
   \begin{tabular}{cc|cccc} 
   \shline
   Short- & Long- & MAD $\downarrow$  & MSE $\downarrow$  & Grad $\downarrow$ & dtSSD $\downarrow$ \\ 
   \hline
   \checkmark &            & 5.32 & 0.61 & 6.76 & 1.55  \\ 
              & \checkmark & 4.97 & 0.53 & 6.70 & 1.63  \\ 
   \checkmark & \checkmark & 4.61 & 0.46 & 6.06 & 1.47  \\ 
   \shline
   \end{tabular}\\
\caption{Short-term and long-term attention.}
   \label{tab:attention}
\end{table}

\vspace{6pt}
\noindent\textbf{Fg/Bg update mechanism.}
We study the effect of the Fg/Bg update scheme described in Sec. \ref{sec:transformer}. In this experiment, we test two variants. First, we disable the update mechanism. This is equivalent to removing Equation \ref{eq:fgbg_emb}. Second, we directly convert the final alpha matte prediction into a Fg/Bg mask and use it to update Value in transformer. The network without an update does not directly absorb Fg/Bg information. Although the previous Key and Value still hold information stored from prior time-steps, the network processes Fg/Bg information in an implicit way. As Table \ref{tab:update} shows, the performance decreases without the update mechanism. When the network is updated using alpha matte prediction, the performance also declines. As an alpha matte is represented by a float value in the range [0,1], in a direct conversion, an incorrect alpha matte prediction will be more likely to result in Fg/Bg mask errors, which will be propagated across frames. The mask update scheme, on the other hand, is intended to provide effective control over the integrated network for utilizing meaningful semantics to a greater extent, thus yielding better performance.
\vspace{3pt}
\begin{table}[!h]
\centering
\tablestyle{5pt}{1.1} 
   \begin{tabular}{cc|cccc} 
   \shline
   If update & Update by & MAD $\downarrow$  & MSE $\downarrow$  & Grad $\downarrow$ & dtSSD $\downarrow$ \\ 
   \hline
    - & - & 6.15 & 1.53 & 10.55 & 2.13  \\ 
   \checkmark & Alpha  &  5.65 & 0.89 & 8.27 & 1.68  \\ 
   \checkmark  & Mask  & {4.61} & {0.46} & {6.06} & {1.47}  \\ 
   \shline
   \end{tabular}\\

\caption{Effect of Fg/Bg update mechanism.} %
   \label{tab:update}

\end{table}

\vspace{6pt}
\noindent\textbf{Performance of different segmenters and backbones.}
We use Mark R-CNN \cite{he2017mask} to obtain an introductory mask for initiating Value in transformer. We replace Mask R-CNN with a recent method, Transfiner \cite{ke2022mask}. Table \ref{tab:segmenter} shows the results are similar, suggesting that the segmenter does not have a direct influence. Our method generates subsequent masks by itself and performs self-updates to mitigate the sensitivities associated with handling an ill-initialized mask. The results support our hypothesis that the impact of the segmenter is modest.
Further, we replace MobileNet \cite{sandler2018mobilenetv2} backbone with ResNet50 \cite{he2016deep}. As expected, the model with ResNet50 is able to learn rich feature representations and further enhances performance. However, it comes at a cost of higher computational complexity.

\begin{table}[!h]
\centering
\tablestyle{4pt}{1.1} 
   \begin{tabular}{cc|cccc} 
   \shline
    & Model & MAD $\downarrow$  & MSE $\downarrow$  & Grad $\downarrow$ & dtSSD $\downarrow$ \\ 
   \hline
   \multirow{2}{*}{Segmenter} &Mask R-CNN & {4.61} & {0.46} & {6.06} & {1.47}  \\ 
   & Transfiner  & 4.58 & 0.45 & 6.01 & 1.47  \\ 
   \hline
   \multirow{2}{*}{Backbone} & MobileNetV2 & {4.61} & {0.46} & {6.06} & {1.47}  \\
   & ResNet50 & 4.53 & 0.42 & 4.99 & 1.34 \\
   \shline
   \end{tabular}\\
\caption{Performance of variants with different segmenters and backbones.} %
   \label{tab:segmenter}
\end{table}

\subsection{\noindent\textbf{Speed}} The purpose of our method is to provide an alternative for reliable video matting in complex real-world scenarios. We manage to enhance the accuracy over previous attempts and still maintain real-time capability. For example, the inference time of our model for a HD video on an Nvidia GTX 2080Ti GPU is 42 fps.  
Comparatively, our method does not achieve the same speed as MODNet \cite{ke2022modnet} and RVM \cite{lin2022robust}, but it is capable of attaining high accuracy in challenging real-world videos. 
A few simple ways of enhancing model throughput are to perform mixed precision inferences or to integrate TensorRT \cite{tensorrt} with an ONNX \cite{onnx} backend. 
While beyond the scope of this study, we expect many of such speed improvement techniques to be applicable to our work.

\section{Conclusion}
We have introduced AdaM for reliable human matting in diverse and unstructured videos, where extensive foreground and background movements appear and affect the way by which many heretofore existing methods perform. 
The competitiveness of our method on the composited and real datasets suggests that explicit modeling of Fg/Bg appearance to help guide the matting task can be a vital contributing factor to video matting success. Comprehensive ablation studies indicate several key factors and advantages of the proposed model, including its design, temporal coherence, and generalizability to a wide range of real videos.

{\small
\bibliographystyle{ieee_fullname}
\bibliography{6_reference.bib}
}

\clearpage
\appendix
\pdfoutput=1
\twocolumn[{%
\renewcommand\twocolumn[1][]{#1}%
\begin{center}
\textbf{\Large Appendix}
\end{center}
\vspace{2mm}
}]

\vspace{6pt}
\section{Implementation Details} \label{sec:implem}
\vspace{6pt}

We describe the detailed network architecture of AdaM in this section.
Our model consists of an off-the-self segmentor $S$, an encoder $\Phi$, a Fg/Bg structuring network $\boldsymbol{\Xi}$, and a decoder $\Psi$. 
We denote that $I^t$ is the frame at time t; $\mathcal{F}^t$ is the feature set of $I^t$, where $\mathcal{F}^t = \{f^t_{1/2}, f^t_{1/4}, f^t_{1/8}, f^t_{1/16}\}$;  $m^0_s$ is the initial mask estimated by the segmenter $S$; $K^t$ and $V^t$ are Key and Value at time $t$; $\boldsymbol{\hat{K}}^{t}$ and $\boldsymbol{\hat{V}}^{t}_{f/b}$ are stored Key and Fg/Bg embedded Value; $m^t$, $\alpha^{c,t}$, and $\alpha^{f,t}$ are the mask, coarse alpha matte, fine-grained alpha matte outputs of the decoder; $m^t_d$ is the downsized mask of $m^t$; $E_f$ and $E_b$ are learnable foreground and background embeddings. We summarize the basic steps of AdaM in Algorithm \ref{alg1}.

\begin{algorithm}
\small 
\caption{AdaM}
\begin{algorithmic}
\label{alg1}
\STATE \textbf{Input:} Sequence of frames $I^0$, $I^1$, $I^2$, ..., $I^{T-1}$
\STATE \textbf{Initialize:} Extract features $\mathcal{F}^0=\Phi(I^0)$ and obtain initial segmentation mask $m^0_s=S(I^0)$;  Given $f^0_{1/16}$ and $m^0_s$, initialize $\boldsymbol{\hat{K}}^{0}=[K^0 ]$, and $\boldsymbol{\hat{V}}^{0}_{f/b}=[V^{0}_{f/b}]$ by Eq. 1 and 6 (main paper) 

\FOR {$t = 0,1, 2, ..., T-1$}
\STATE $\bullet$ Extract features: $\mathcal{F}^t=\Phi(I^t)$; $Z^t=f^t_{1/16}$ 
\STATE $\bullet$ Obtain $Q^t$, $K^t$, and $V^t$ by Eq. 1 in the main paper  
\STATE $\bullet$ Propagate Fg/Bg: $Z'^t=\boldsymbol{\Xi}(Q^t, \boldsymbol{\hat{K}}^{t}_{}, \boldsymbol{\hat{V}}^{t}_{f/b})$
\STATE $\bullet$ Decode outputs: $m^t, \alpha^{c,t}, \alpha^{f,t}=\Psi(Z'^t, f^t_{1/8}, f^t_{1/4}, f^t_{1/2})$
\STATE $\bullet$ Embed Fg/Bg: ${V}^{t}_{f/b} = V^{t} + (m^{t}_d  E_f + (1-m^{t}_d) E_b)$
\STATE $\bullet$ Update $\boldsymbol{\hat{K}}^{t+1}$:  $\boldsymbol{\hat{K}}^{t+1}=concat(\boldsymbol{\hat{K}}^{t}, K^t)$
\STATE $\bullet$ Update $\boldsymbol{\hat{V}}^{t+1}_{f/b}$:  $\boldsymbol{\hat{V}}^{t+1}_{f/b}=concat(\boldsymbol{\hat{V}}^{t}_{f/b}, V^t_{f/b})$
\STATE $\bullet$ Discard oldest $K^{t^{'}}$ and ${V}_{f/b}^{t^{'}}$ w.r.t the limitation criteria 
\ENDFOR
\end{algorithmic}
\end{algorithm}

\noindent\textbf{Encoder.}
MobileNetV2 \cite{sandler2018mobilenetv2} is used as our backbone. 
Following \cite{chen2017rethinking, chen2018encoder, howard2019searching, lin2022robust}, we modify the last block of our backbone using convolutions with a dilation rate of 2 and a stride of 1 to increase feature resolution.
For each video frame, the encoder network extracts features at 1/2, 1/4, 1/8, and 1/16 scales with different levels of abstraction ($F^{1/2}, F^{1/4}, F^{1/8}, F^{1/16}$ with feature channels of 16, 24, 32, and 1280, respectively.)
The low-resolution feature $F^{1/16}$ is fed into the transformer to model the foreground and background more efficiently. %
To reduce the computational complexity of the transformer network, the channel of the low-resolution feature is reduced from 1280 to 256 using a 1x1 convolution layer. The final feature channel of $F^{1/16}$ is 256. 

\vspace{6pt}
\noindent\textbf{Fg/Bg Structuring Network.}
The transformer in Fg/Bg Structuring Network consists of three layers with a hidden size of 256D. 
The transformer encoder consists of alternating layers of multiheaded self-attention (MSA) and MLP blocks. Residual connections and layernorm (LN) are applied after every block. The MLP contains two fully-connected layers with a GeLU non-linearity. %

\vspace{6pt}
\noindent\textbf{Decoder.}
Our decoder contains four upscaling blocks and two output blocks. 
In each upscaling block, the output features from the previous upscaling block and the corresponding features from the skip connection are concatenated. The concatenated features are passed through a layer of 3×3 convolution, Batch Normalization and ReLU activation. Finally, a layer of 3×3 convolution and a bilinear 2x upsampling layer are applied to generate output features. 
The feature channels at 1/16, 1/8, 1/4, 1/2 scales are 256, 128, 128, and 32, respectively.

We employ a two-stage refinement to refine alpha mattes progressively. 
The model first produces a Fg/Bg mask and a coarse alpha matte at 1/4 scale of the original resolution at the first output block, then predicts a finer alpha matte at full resolution at the second output block. 
In specific, after the first two upscaling blocks, the first output block takes the feature at 1/4 of the original resolution from the second upscaling block, and passes them through two parallel layers of 3×3 convolutions, producing a 2-channel Fg/Bg prediction $m_p$ and a 1-channel coarse alpha matte prediction $\alpha^c_p$. 
The second output block produces the final alpha matte at the original resolution. 
After the fourth upscaling block, the second output block takes the final feature map at the original resolution and passes it through two additional layers of 3×3 convolution, Batch Normalization and ReLU to obtain the final alpha matte at the original resolution.

\vspace{2pt}
\section{Additional Experiments} \label{sec:ablat}

\subsection{Bi-directional inference for offline scenarios} 
AdaM processes video frames in a causal manner in order to preserve real-time capabilities. 
Here, we examine an offline matting scenario in which we store the Fg/Bg feature information gathered during the forward pass and then process the sequence in reverse through time.
Without additional training, we observe that bi-directional inference leads to performance improvement as shown in Table \ref{tab:bidirection}. 
The advantage of bi-directional inference is that it allows comprehensive observation of video data through both forwards and backwards passes. 
\vspace{3pt}
\begin{table}[h]
\begin{center}
\footnotesize
\begin{tabular}{ c| c c c c}
\toprule
    VM 1920$\times$1080 & MAD $\downarrow$  & MSE $\downarrow$  & Grad $\downarrow$ & dtSSD $\downarrow$ \\ 
   \hline
   Online (forward) & {4.61} & {0.46} & {6.06} & {1.47}  \\
   Offline (bi-direction) & 4.57 & 0.43 & 5.92 & 1.43 \\
\bottomrule
\end{tabular}
\vspace{6pt}
\caption{Bi-directional inference  (MobileNetV2 model w/o HD).}
\label{tab:bidirection}
\end{center}
\end{table}
\begin{table*}[!h]
\centering
\tablestyle{10pt}{1.1} 
   \begin{tabular}{c|ccc|c|cccc} 
   \shline
   Model & Stage 1 & Stage 2 & Stage 3 & Alternating training in Stage 2 & MAD $\downarrow$  & MSE $\downarrow$  & Grad $\downarrow$ & dtSSD $\downarrow$ \\ 
   \hline
   (a) & & \checkmark &  &  & 4.85 & 0.57 & 7.88 & 1.60  \\ 
   (b) & \checkmark & \checkmark &  &  & 4.63 & 0.43 & 6.15 & 1.45  \\ 
   (c) & \checkmark & \checkmark &  & \checkmark  & {4.61} & {0.46} & {6.06} & {1.47}  \\ 
   (d) & \checkmark & \checkmark & \checkmark & \checkmark  & {4.42} & {0.39} & {5.12} & {1.39}  \\ 
   \shline
   \end{tabular}\\
\vspace{10pt}
\caption{Effect of different training strategies. All models are evaluated on VM 1920×1080 set.} %
   \label{tab:training}
\vspace{12pt}
\end{table*}
\begin{table*}[!h]
\centering
\tablestyle{10pt}{1.1} 
   \begin{tabular}{c|ccc|c|cccc} 
   \shline
   Model & Stage 1 & Stage 2 & Stage 3 & Alternating training in Stage 2 & MAD $\downarrow$  & MSE $\downarrow$  & Grad $\downarrow$ & dtSSD $\downarrow$ \\ 
   \hline
   (b) & \checkmark & \checkmark &  &  &  8.07 & 4.17 & 22.30 & 6.48  \\ 
   (c) & \checkmark & \checkmark &  & \checkmark  & {7.13} & {3.05} & {17.96} & {5.76}  \\ 
   (d) & \checkmark & \checkmark & \checkmark & \checkmark  & {5.94} & {2.79} & {16.61} & {5.45}  \\ 
   \shline
   \end{tabular}\\
\vspace{10pt}
\caption{Influence of different training strategies. The models are evaluated on CRGNN-R 1920×1080 set.} %
   \label{tab:crtraining}
\vspace{12pt}
\end{table*}
\begin{table*}[!h]
\centering
\tablestyle{10pt}{1.1} 
   \begin{tabular}{c|ccc|cccc} 
   \shline
   Model & $m_p$ & $\alpha^c_p$ & $\alpha^f_p$ & MAD $\downarrow$  & MSE $\downarrow$  & Grad $\downarrow$ & dtSSD $\downarrow$ \\ 
   \hline

  (e) & &  & \checkmark  & 5.65 & 0.89 & 8.27 & 1.68  \\ 
  (f) & \checkmark &  & \checkmark  & 4.74 & 0.48 &  6.68 & 1.51  \\ 
  (g) & \checkmark & \checkmark & \checkmark & {4.61} & {0.46} & {6.06} & {1.47}  \\ 
   \shline
   \end{tabular}\\
\vspace{10pt}

\caption{Effect of different model outputs. The models are evaluated on VM 1920×1080 set.} %
   \label{tab:loss}
\vspace{12pt}
\end{table*}
\subsection{Influence of training strategies}

As discussed in the implementation details (Section  4.2 of the main paper and Section 3.2 in the supplementary), our training pipeline includes three stages. 
In the second stage, we train our model alternately on VM SD data (odd iterations) and video segmentation data (even iterations) to prevent overfitting. 

In this experiment, we evaluate the influences of different training stages and the alternating choice. The results are reported in Table \ref{tab:training}.
To evaluate the effectiveness of different training strategies, we ablate the other stages and training choice step by step for a controlled evaluation within our framework. We draw several conclusions from the results.

First, we train our integrated model on the segmentation data at stage 1 to obtain better pre-trained weights for initiating the matting task.
As shown in the Table \ref{tab:training} (a) and (b), the model with both stages 1 and 2 performs better than the model with stage 2 only. When enabling alternating training in Model (c), we observe there is no significant performance difference in VM dataset.  
The objective of alternating training is to prevent the network from overfitting to synthetic data, so that the network will be able to generalize more effectively.
To validate its impact, we further test Model (b) to (d) on CRGNN-R dataset. As shown in Table \ref{tab:crtraining}, the model with alternating training (Model (c)) performs better on CRGNN-R dataset. For example, the MAD drops from 8.07 to 7.13. It shows that the alternating training strategy actually helps with model performance and generalization. 
As shown in Table \ref{tab:training} and \ref{tab:crtraining}, the last row (Model (d)) supports that our choice of integrating all three stages and alternating training improves results and achieves better performance.

\subsection{Impact of model outputs}

Section 3.4 of the main manuscript presents our loss functions, summarized below.
Our model predicts a Fg/Bg mask $m_p$, a coarse alpha matte $\alpha^c_p$, and a fine-grained alpha matte $\alpha^f_p$. The first output block in the decoder generates $m_p$ and $\alpha^c_p$, while the second output block produces $\alpha^c_p$.
The overall loss function $\mathcal{L}$ is: 
\vspace{6pt}
\begin{align}\label{eq:loss_all}
    \mathcal{L} =  \omega_{m} \mathcal{L}^{m}_{}(m_p) + & \omega^c_{\alpha}(\mathcal{L}^{\alpha}_{l1}(\alpha^c_p) + \mathcal{L}^{\alpha}_{lap} (\alpha^c_p)) \nonumber \\
    + &\omega^f_{\alpha}(\mathcal{L}^{\alpha}_{l1}(\alpha^f_p) + \mathcal{L}^{\alpha}_{lap} (\alpha^f_p)), 
\end{align} 
\vspace{1pt}

where $\omega_{m}$, $\omega^c_{\alpha}$, and $\omega^f_{\alpha}$ are loss weights. 

Here, we consider three variants to study the effects associated with the generation of Fg/Bg mask $m_p$, coarse alpha matte $\alpha^c_p$, and fine-grained alpha matte $\alpha^f_p$.

\vspace{6pt}
Variant I: the intermediate output of the first output block is disabled. In this case, the overall loss function becomes: 
\vspace{6pt}
\begin{align}
    \mathcal{L} = &\omega^f_{\alpha}(\mathcal{L}^{\alpha}_{l1}(\alpha^f_p) + \mathcal{L}^{\alpha}_{lap} (\alpha^f_p)), 
\end{align}%
\vspace{1pt}

Variant II: the network predicts a Fg/Bg mask $m_p$ as the sole intermediate output of the first output block. The second block output is unchanged.This results in the following overall loss function:
\vspace{6pt}
\begin{align}
    \mathcal{L} =  \omega_{m} \mathcal{L}^{m}_{}(m_p)
    + &\omega^f_{\alpha}(\mathcal{L}^{\alpha}_{l1}(\alpha^f_p) + \mathcal{L}^{\alpha}_{lap} (\alpha^f_p)), 
\end{align}
\vspace{1pt}

Variant III: the first and second outputs remain the same. The overall loss function is represented by Equation \ref{eq:loss_all}, which is the same as Equation (13) in the main manuscript.
\vspace{1pt}

We evaluate the performance of these three model variants and report the results in Table \ref{tab:loss}. 
Model (e) shows the result of disabling the first output block. In this case, the network could convert the final alpha matte prediction into a Fg/Bg mask and update Fg/Bg information in the Fg/Bg structuring network with the mask. As discussed in the main manuscript, an alpha matte is represented by a float value between 0 and 1. In the case of an incorrect alpha matte prediction, a direct conversion will produce Fg/Bg mask errors, which will propagate across frames. A notable performance loss can be observed when comparing Model (e) and (g).

In Variant II, no coarse alpha matte prediction is produced in the first output block. The second output block directly produces the alpha matte prediction. In Variant III, by producing an intermediate output of a coarse mask, we can constrain the coarse alpha matte prediction in the first output block and refine it gradually in the second block rather than predicting an alpha matte directly. This would allow the second output block to focus on refining the details after the intermediate alpha matte had provided the coarsely defined areas. Progressive refinement is also to prevent the model from performing unexpectedly when it takes a complex video input during inference.
As shown in Table \ref{tab:loss}, Model (g) yields the best performance.
With the supervision of the total intermediate output, the model benefits from better localized foreground areas and more discriminative features derived from different stages, thus realizing its full potential.

\subsection{Qualitative evaluation} \label{sec:more_exp}

In this section, we present additional qualitative results:
\begin{enumerate}[noitemsep]
  \item[1.] Figure \ref{fig:crg} illustrates the comparison results on the CRGNN real-video datasets \cite{wang2021video}. The results show the proposed method is able to produce more accurate foreground matting results. 
  \begin{figure*}[]
\captionsetup[subfigure]{labelformat=empty}
\begin{center}
    \scriptsize

\begin{minipage}[]{.99\textwidth}
    \centering
    \footnotesize
    \hspace{0.1em}\begin{subfigure}[b]{0.23\textwidth}
        \caption{Input}
        \includegraphics[width=\textwidth]{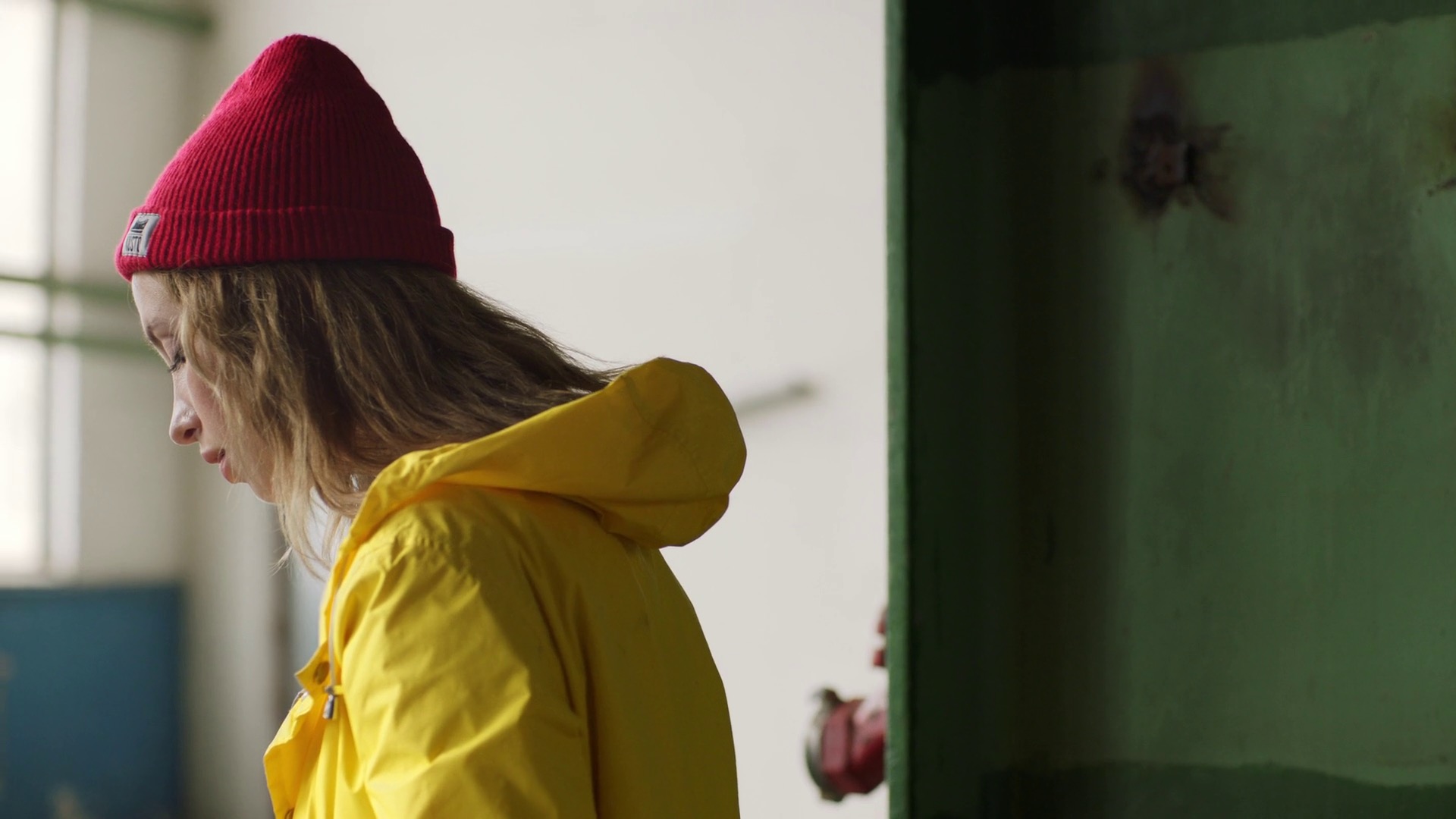}
    \end{subfigure}\hspace{0em}
    \begin{subfigure}[b]{0.23\textwidth}
        \caption{MODNet}
        \includegraphics[width=\textwidth]{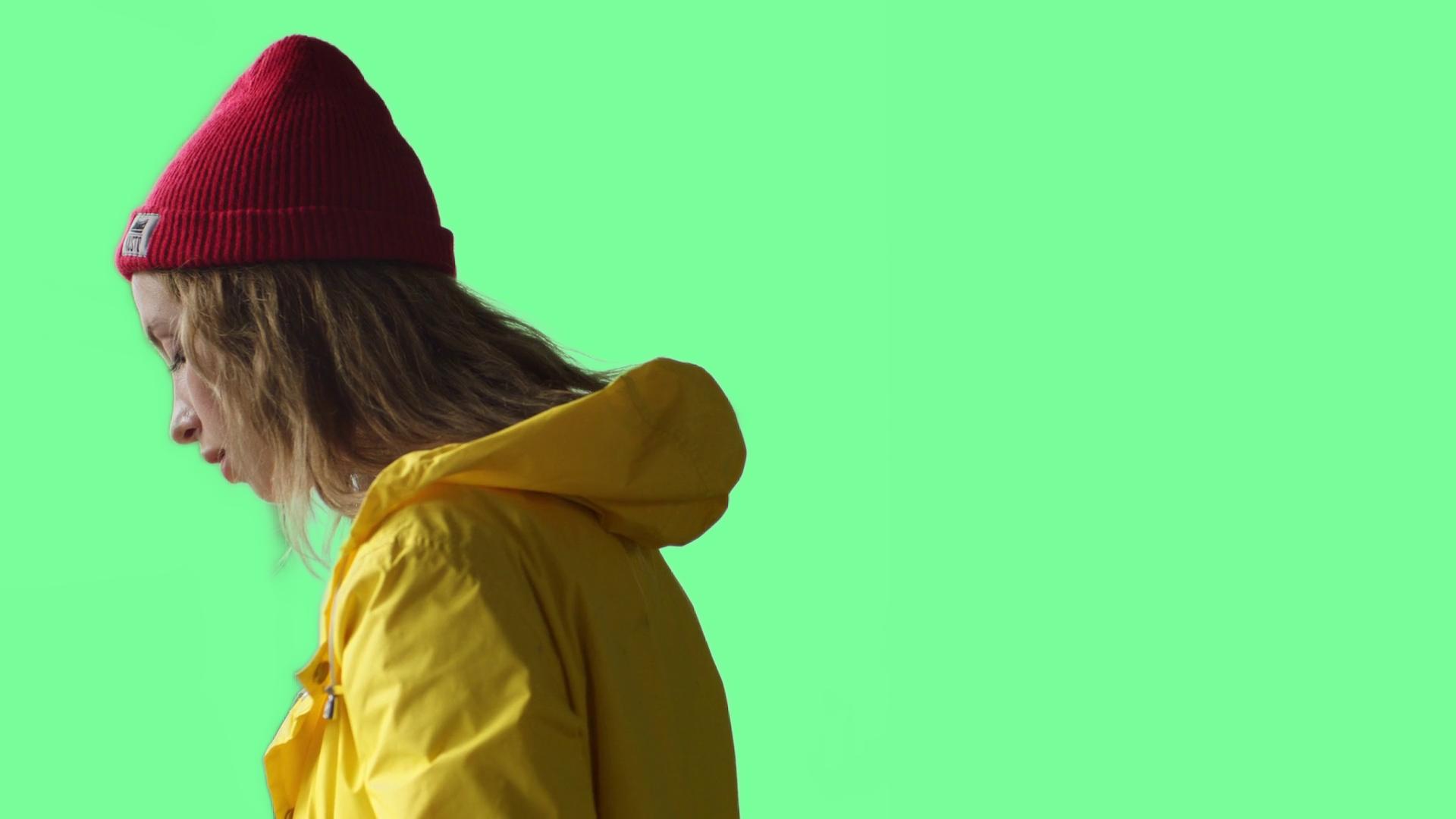}
    \end{subfigure}\hspace{0em}
    \begin{subfigure}[b]{0.23\textwidth}
        \caption{RVM}
        \includegraphics[width=\textwidth]{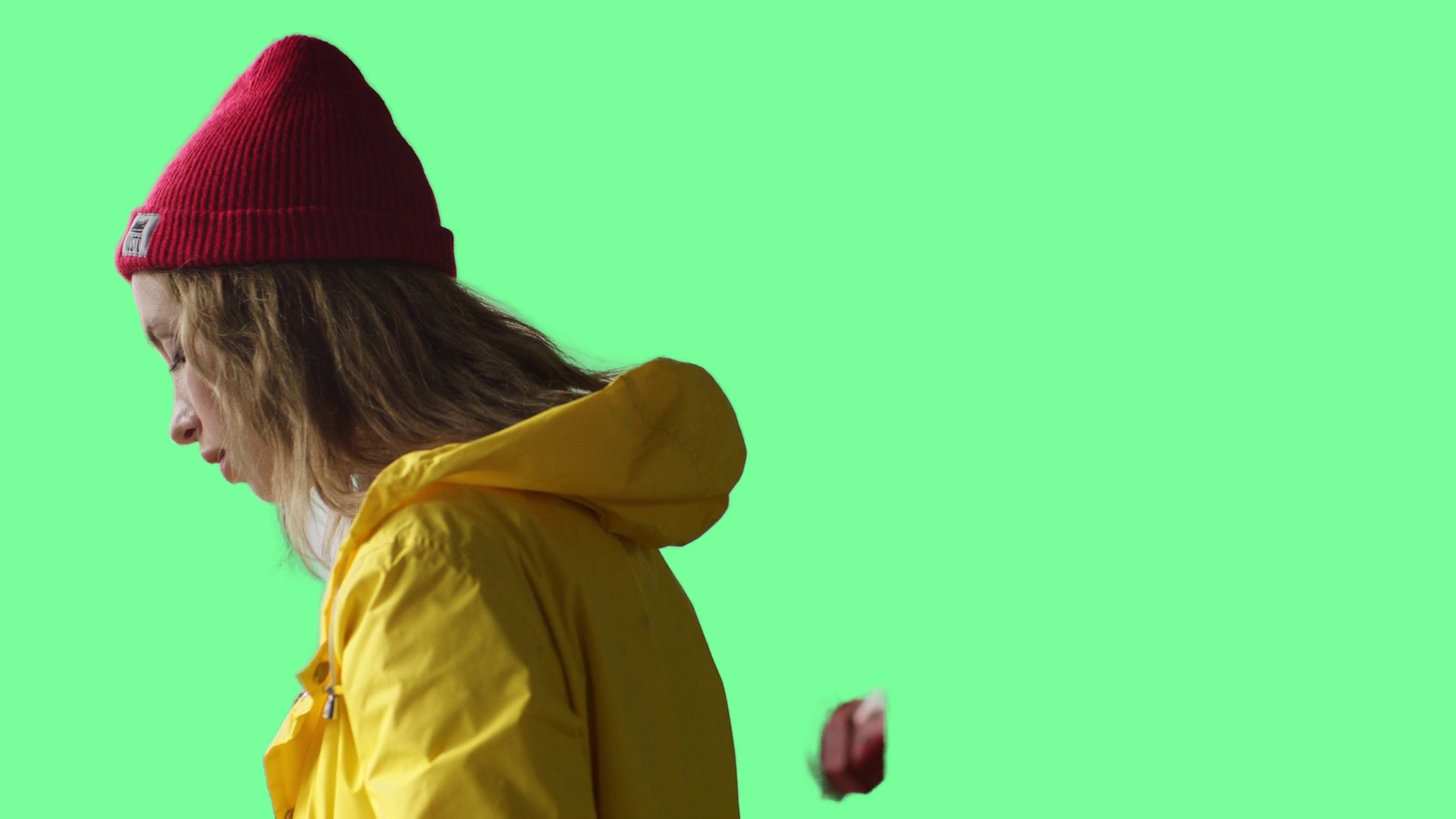}
    \end{subfigure}\hspace{0em}
    \begin{subfigure}[b]{0.23\textwidth}
        \caption{AdaM}
        \includegraphics[width=\textwidth]{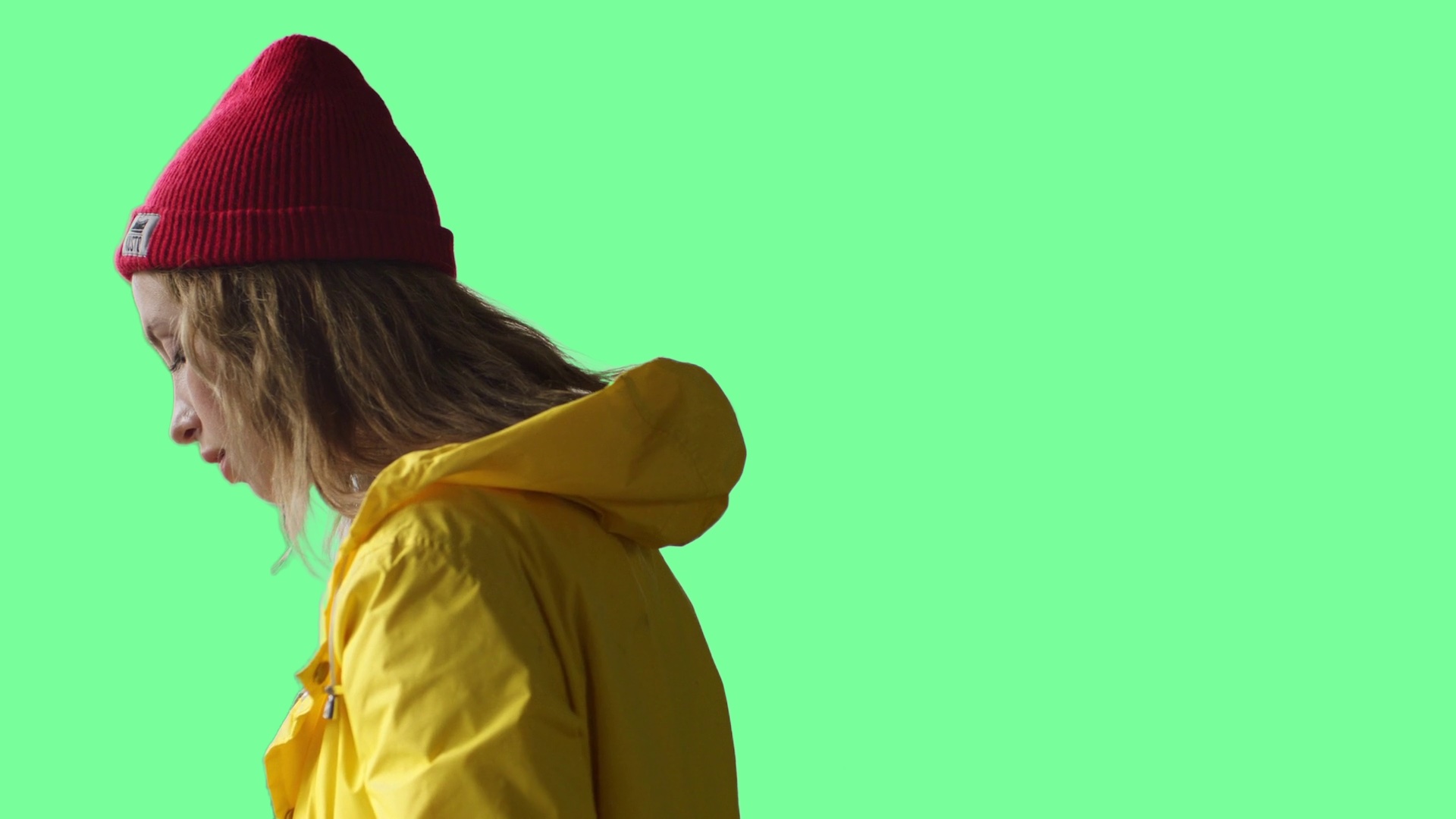}
    \end{subfigure}\hspace{0em}
   \vspace{1pt}
    \end{minipage}
\begin{minipage}[]{.99\textwidth}
        \centering
        \footnotesize
    \includegraphics[trim=0 0 0 0, clip,width=0.23\textwidth]{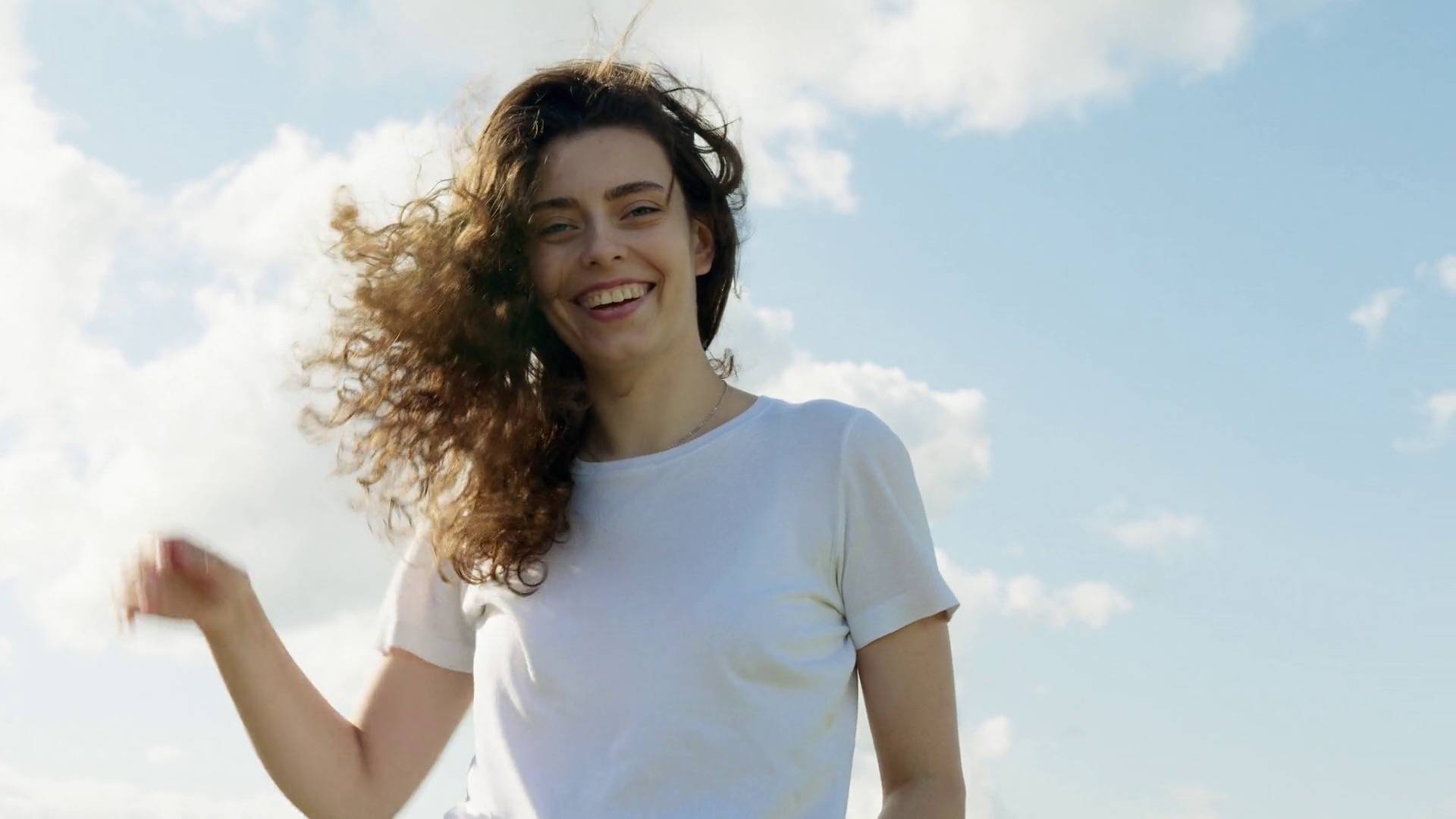}\hspace{0em}
    \includegraphics[trim=0 0 0 0, clip,width=0.23\textwidth]{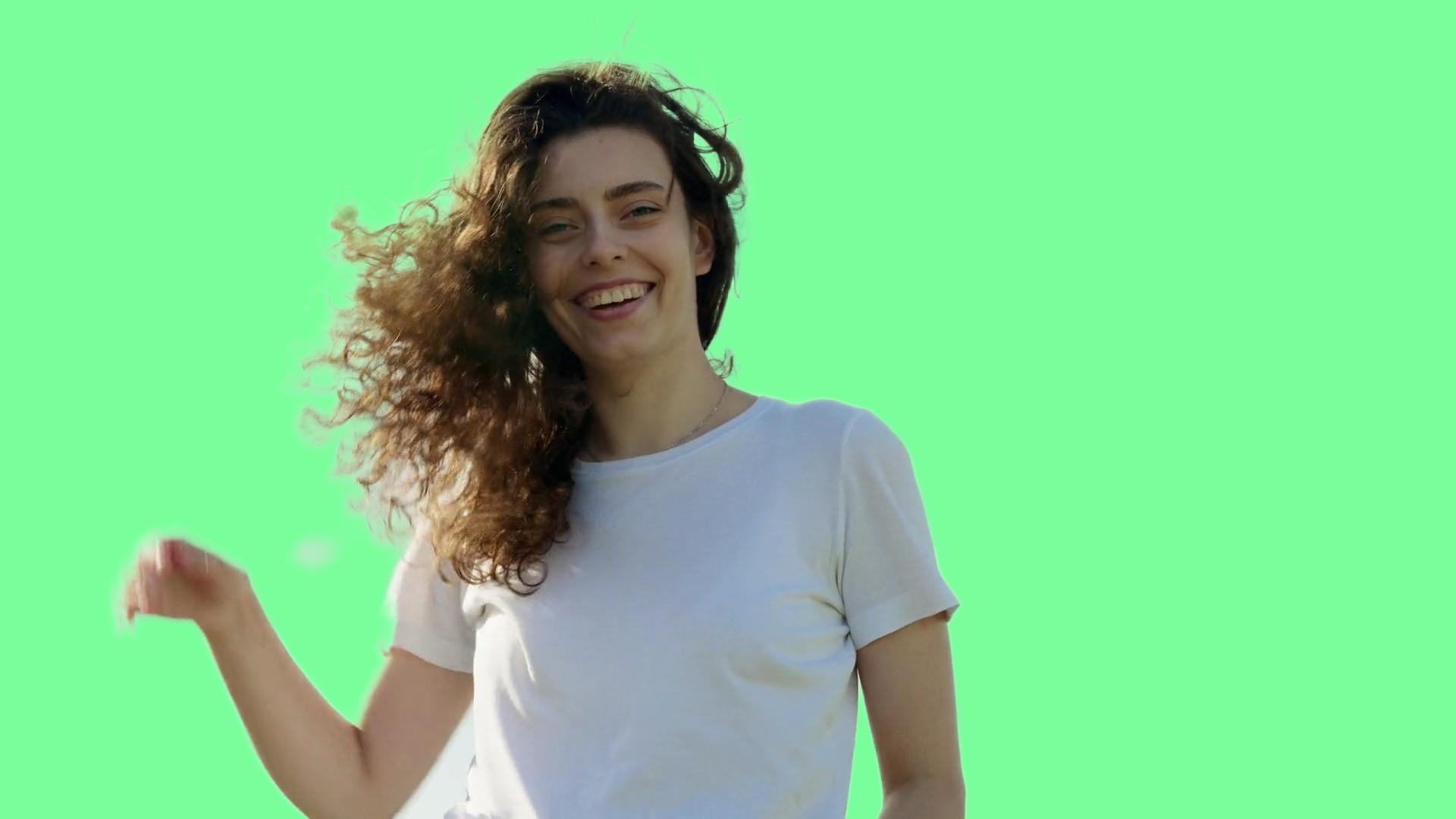}\hspace{0em}
    \includegraphics[trim=0 0 0 0, clip,width=0.23\textwidth]{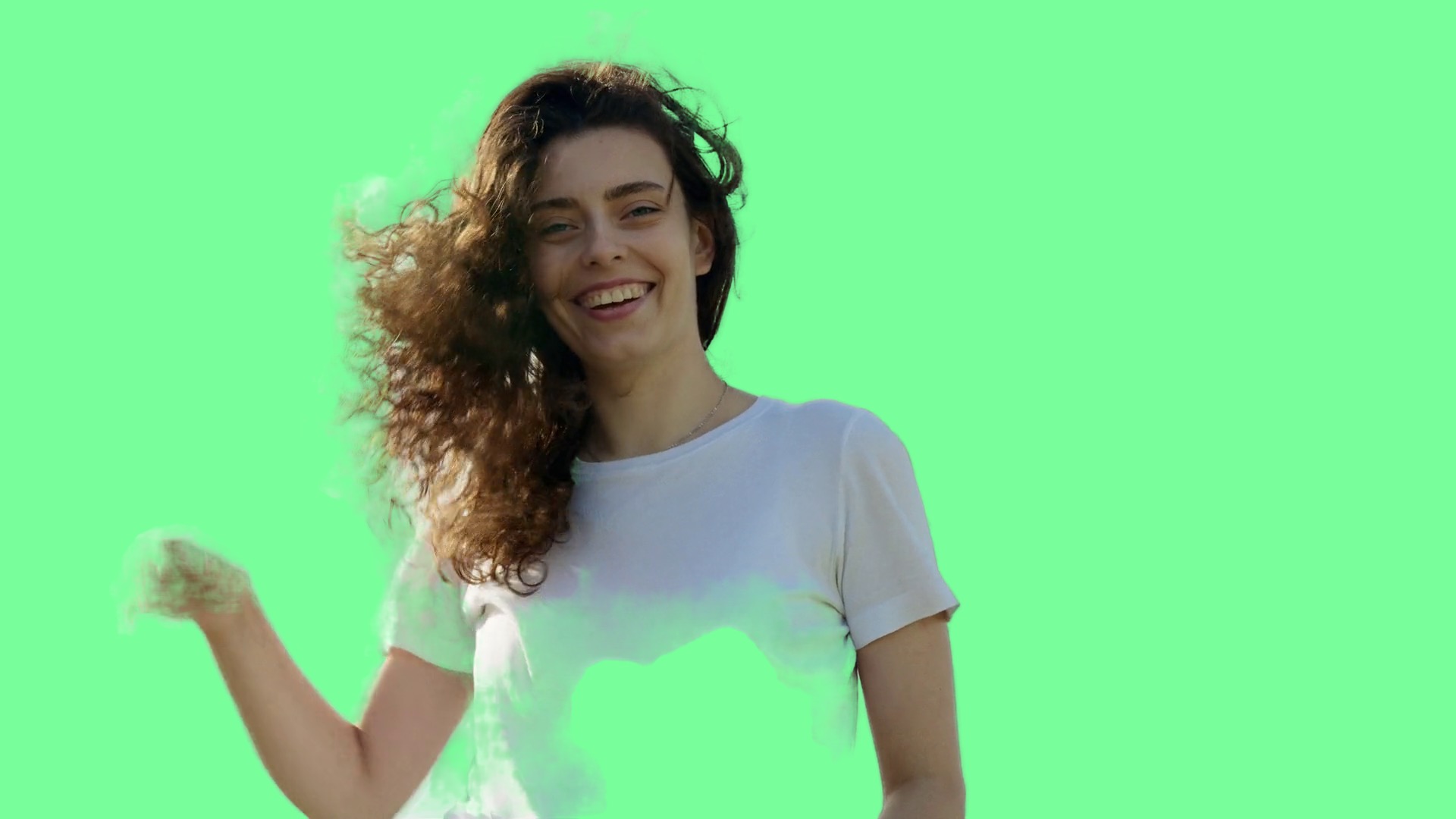}\hspace{0em}
    \includegraphics[trim=0 0 0 0, clip,width=0.23\textwidth]{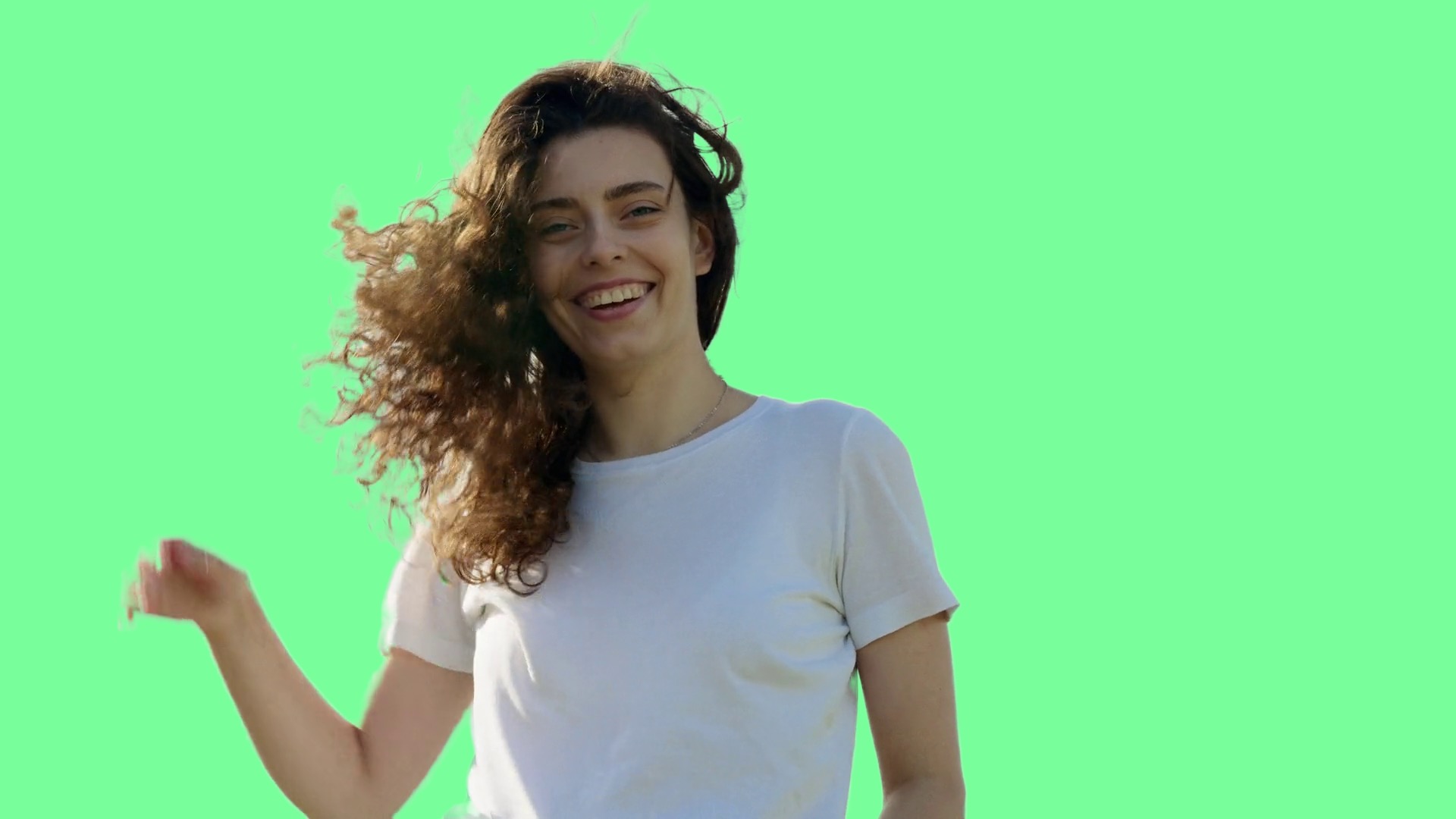}\hspace{0em}
    \end{minipage}

  \label{fig:crgnn}

  \end{center}

    \centering
\vspace{-5pt}
   \caption{\small Comparisons of our model to MODNet \cite{ke2022modnet} and RVM \cite{lin2022robust} on CRGNN-R test set.}
\vspace{5pt}   
\label{fig:crg}
\end{figure*}
  
  \item[2.] Figure \ref{fig:footage} shows three comparison samples of the challenging video footage released in RVM's GitHub \cite{lin2022robust}. The comparisons show AdaM's strength in real-world environments. Our method is able to yield reliable matting results with fewer artifacts.
  \begin{figure*}[]
\captionsetup[subfigure]{labelformat=empty}
\begin{center}
    \scriptsize

\begin{minipage}[]{.99\textwidth}
    \centering
    \footnotesize
    \begin{subfigure}[b]{0.23\textwidth}
        \caption{Input}
        \includegraphics[width=\textwidth]{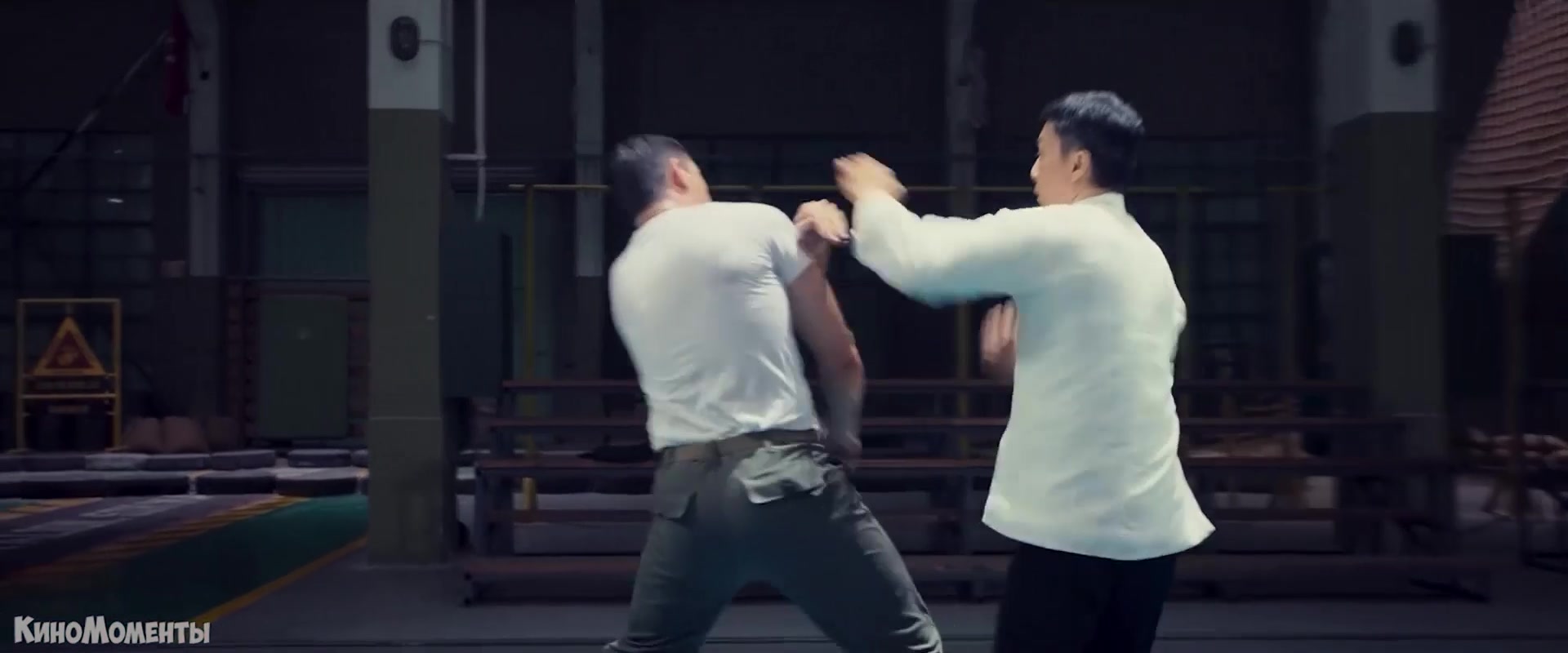}
    \end{subfigure}\hspace{0em}
    \begin{subfigure}[b]{0.23\textwidth}
        \caption{MODNet}
        \includegraphics[width=\textwidth]{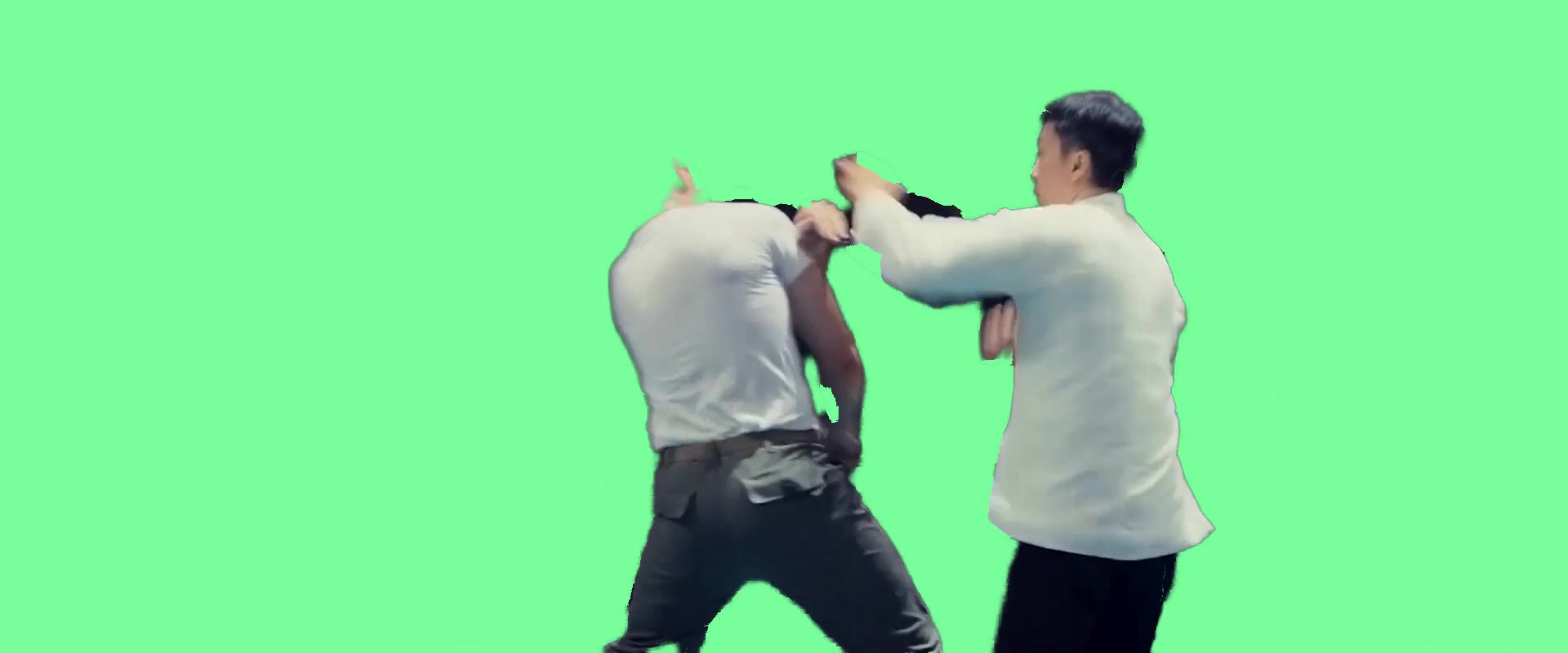}
    \end{subfigure}\hspace{0em}
    \begin{subfigure}[b]{0.23\textwidth}
        \caption{RVM}
        \includegraphics[width=\textwidth]{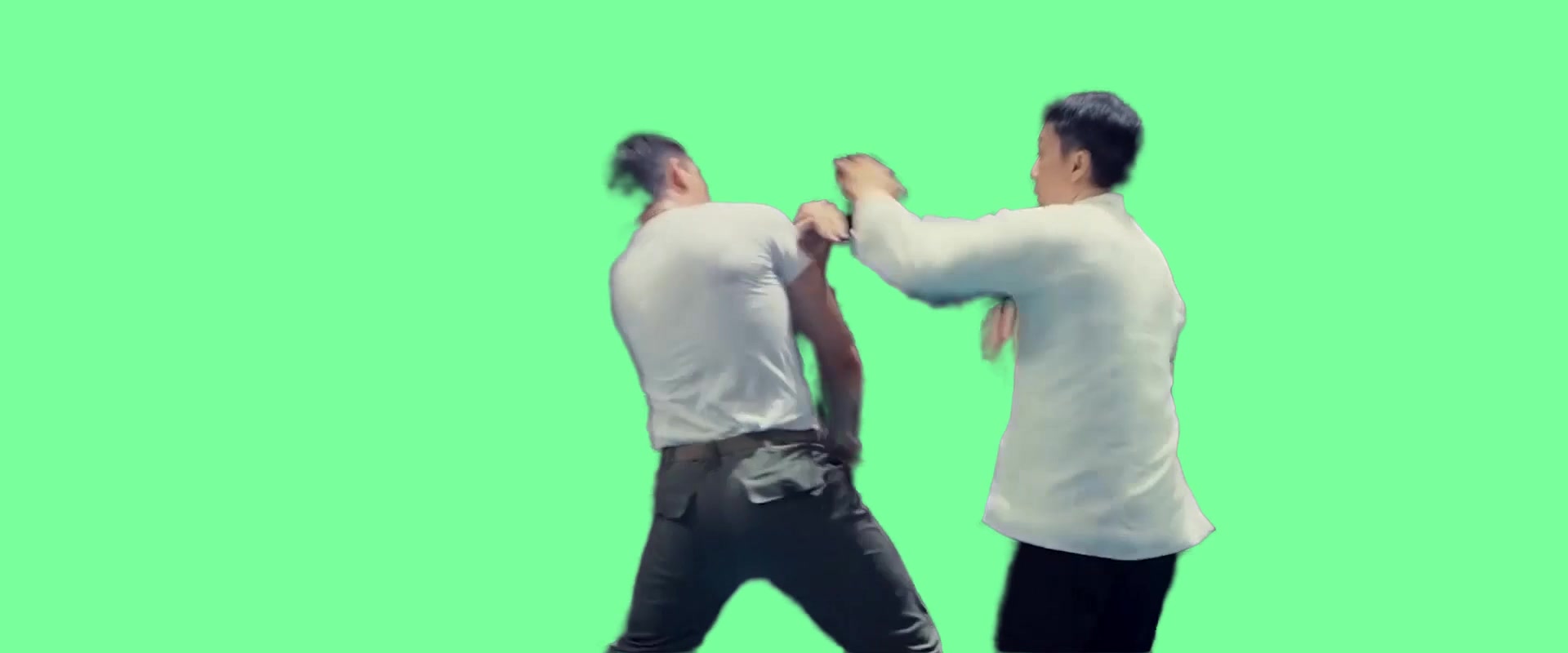}
    \end{subfigure}\hspace{0em}
    \begin{subfigure}[b]{0.23\textwidth}
        \caption{AdaM}
        \includegraphics[width=\textwidth]{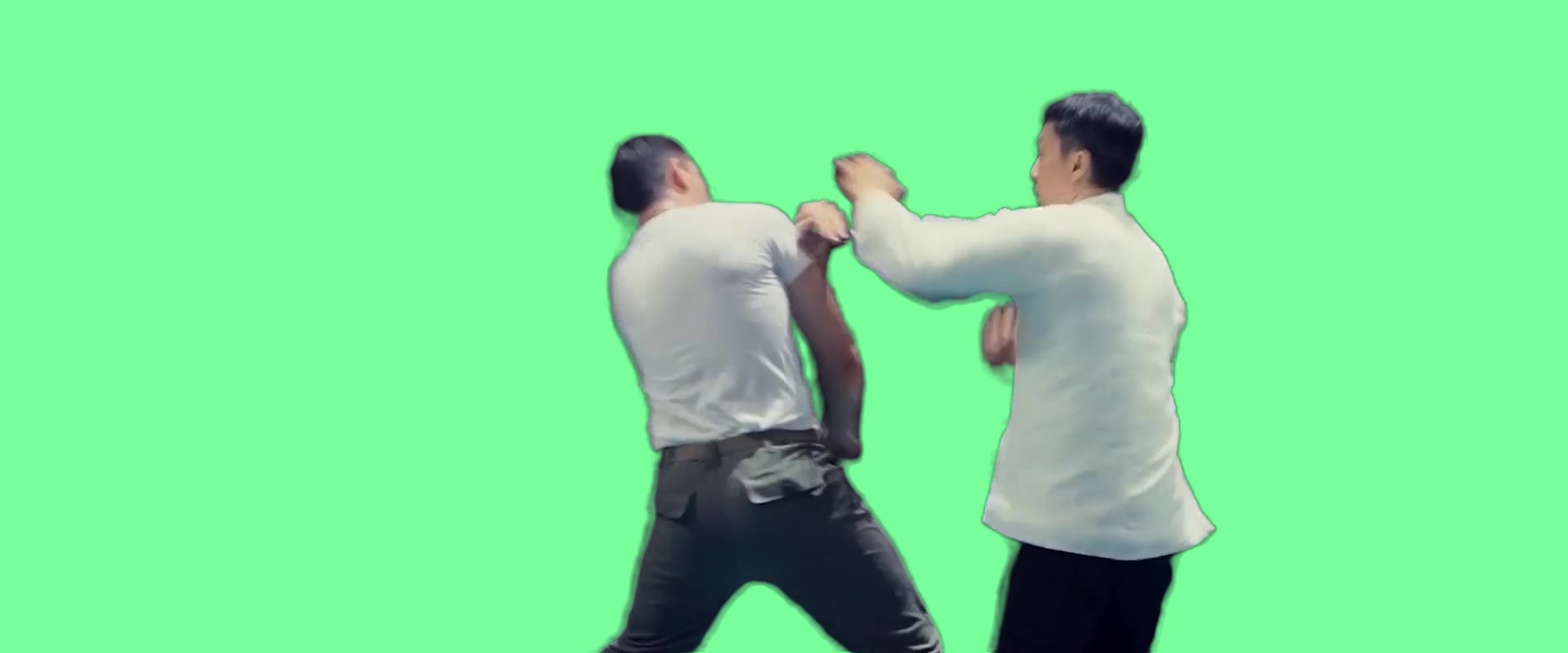}
    \end{subfigure}\hspace{0em}
   \vspace{1pt}
    \end{minipage}
\begin{minipage}[]{.99\textwidth}
        \centering
        \footnotesize
    \includegraphics[trim=0 0 0 0, clip,width=0.23\textwidth]{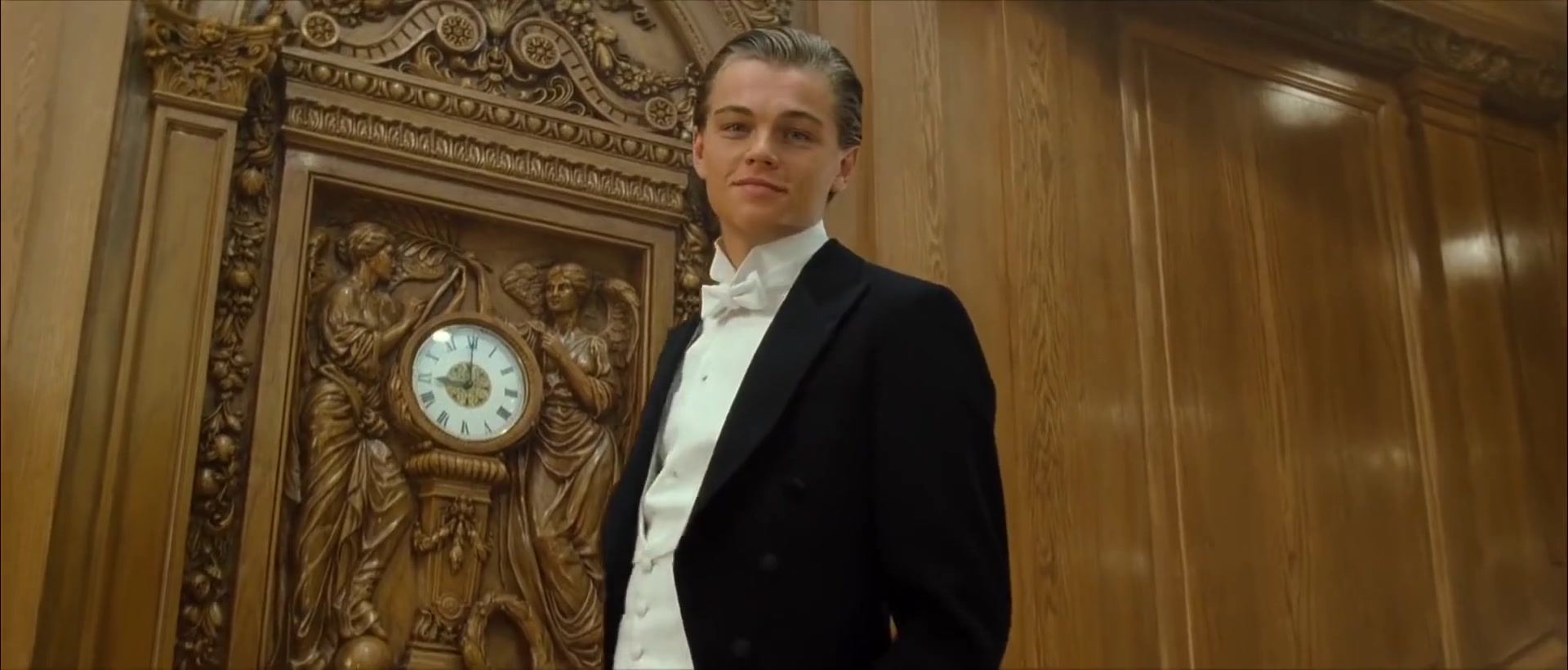}\hspace{0em}
    \includegraphics[trim=0 0 0 0, clip,width=0.23\textwidth]{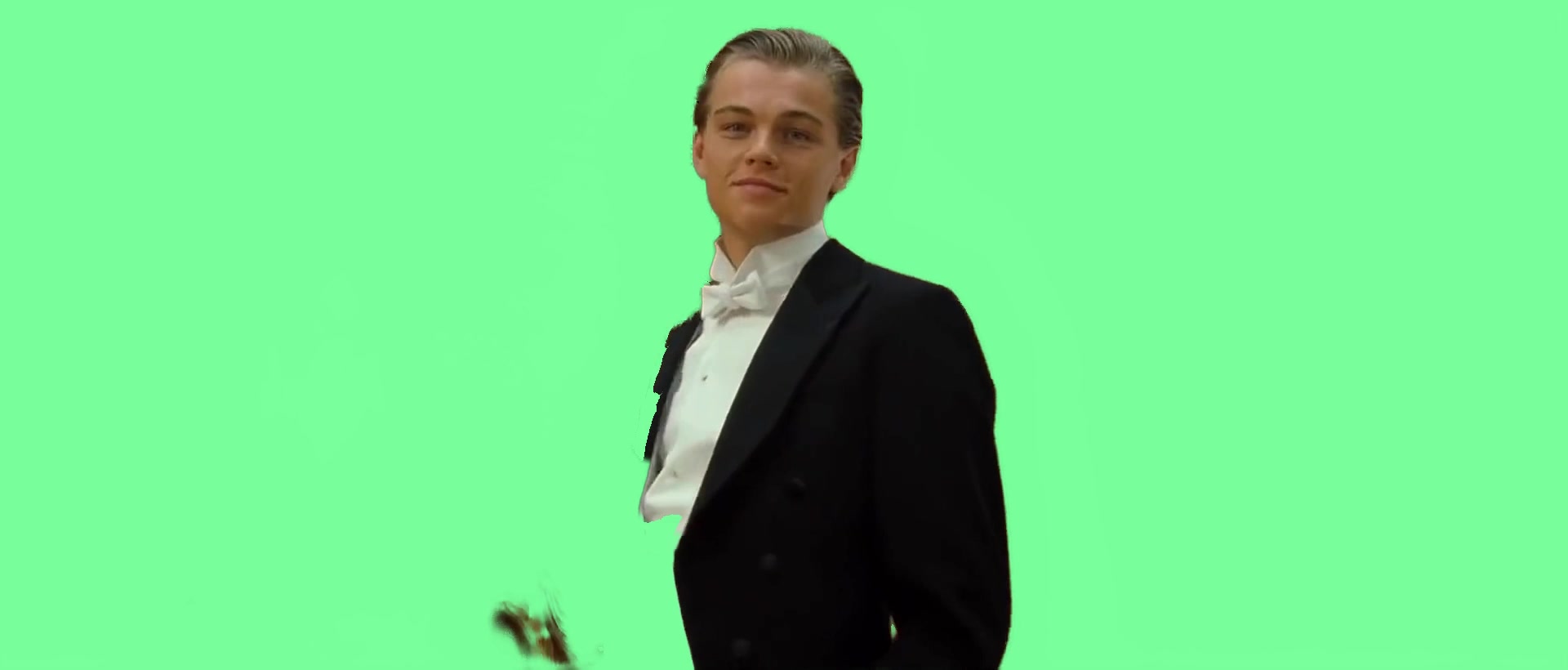}\hspace{0em}
    \includegraphics[trim=0 0 0 0, clip,width=0.23\textwidth]{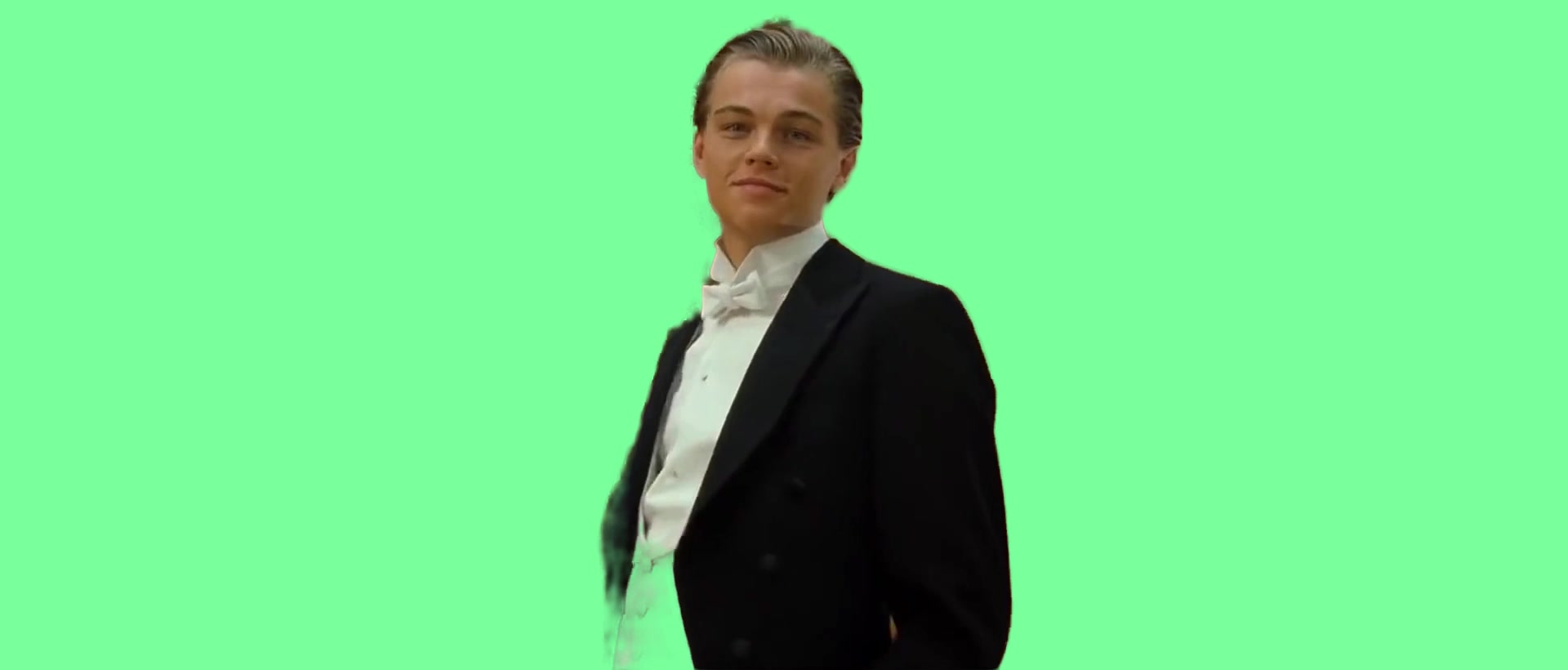}\hspace{0em}
    \includegraphics[trim=0 0 0 0, clip,width=0.23\textwidth]{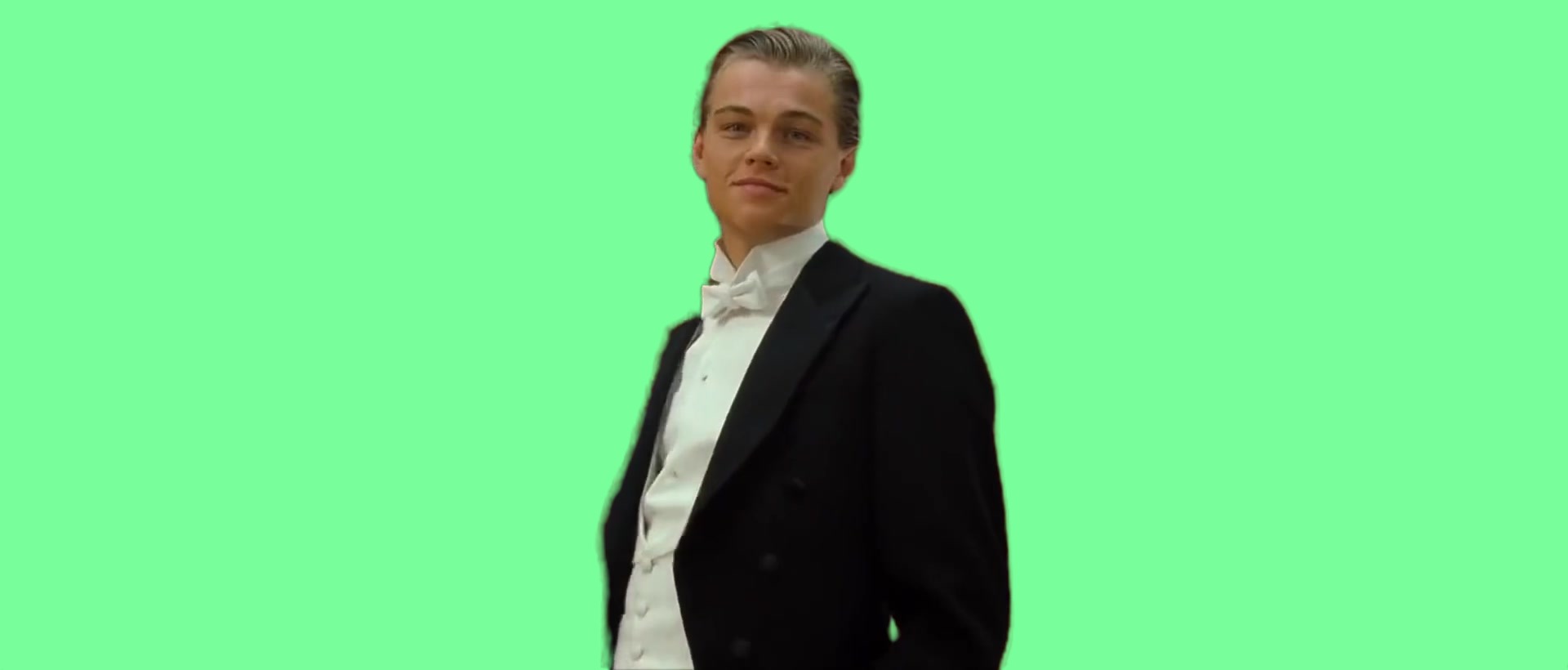}\hspace{0em}
    \vspace{1pt}
    \end{minipage}
\begin{minipage}[]{.99\textwidth}
        \centering
        \footnotesize
    \includegraphics[trim=0 0 0 0, clip,width=0.23\textwidth]{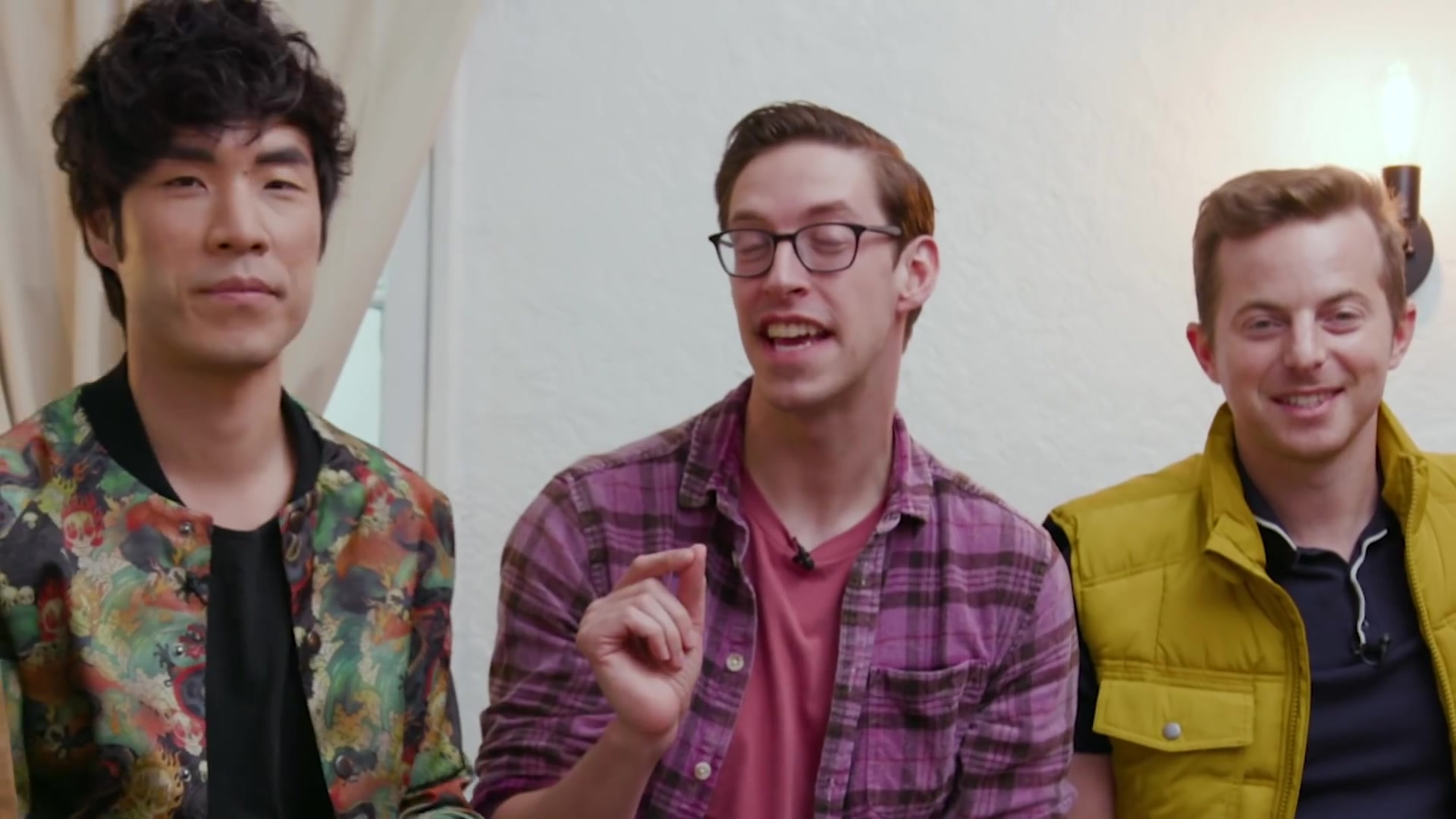}\hspace{0em}
    \includegraphics[trim=0 0 0 0, clip,width=0.23\textwidth]{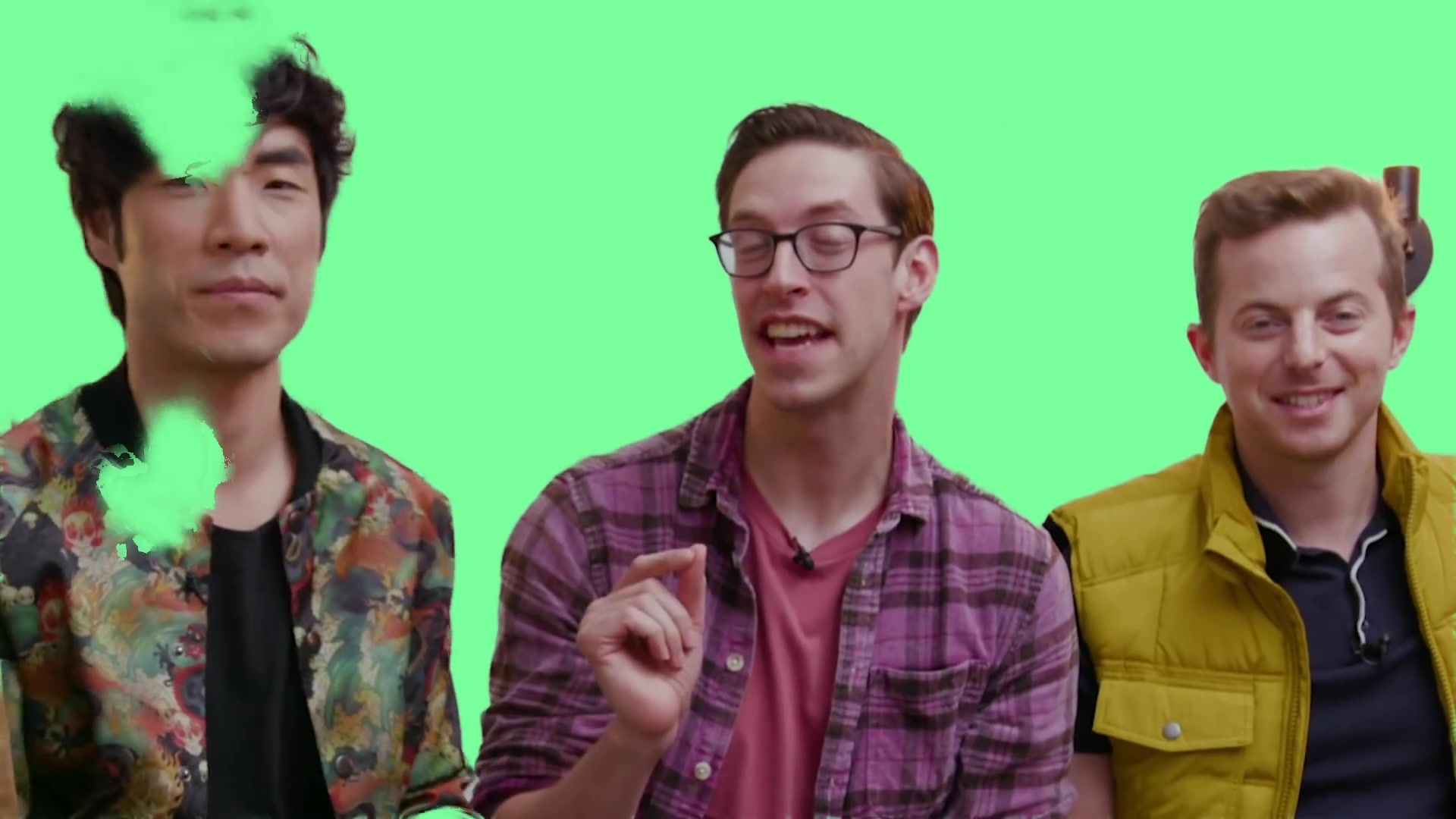}\hspace{0em}
    \includegraphics[trim=0 0 0 0, clip,width=0.23\textwidth]{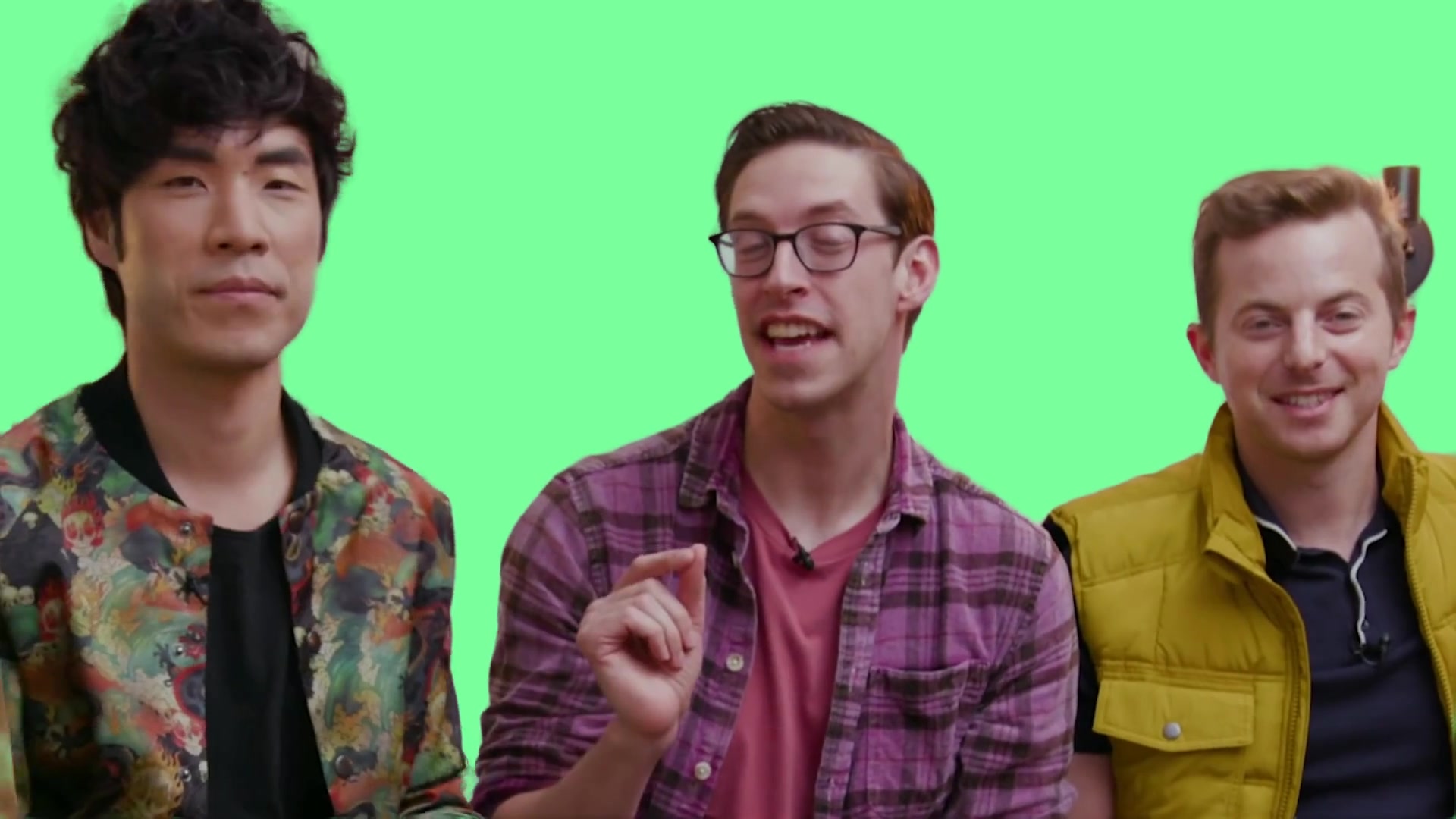}\hspace{0em}
    \includegraphics[trim=0 0 0 0, clip,width=0.23\textwidth]{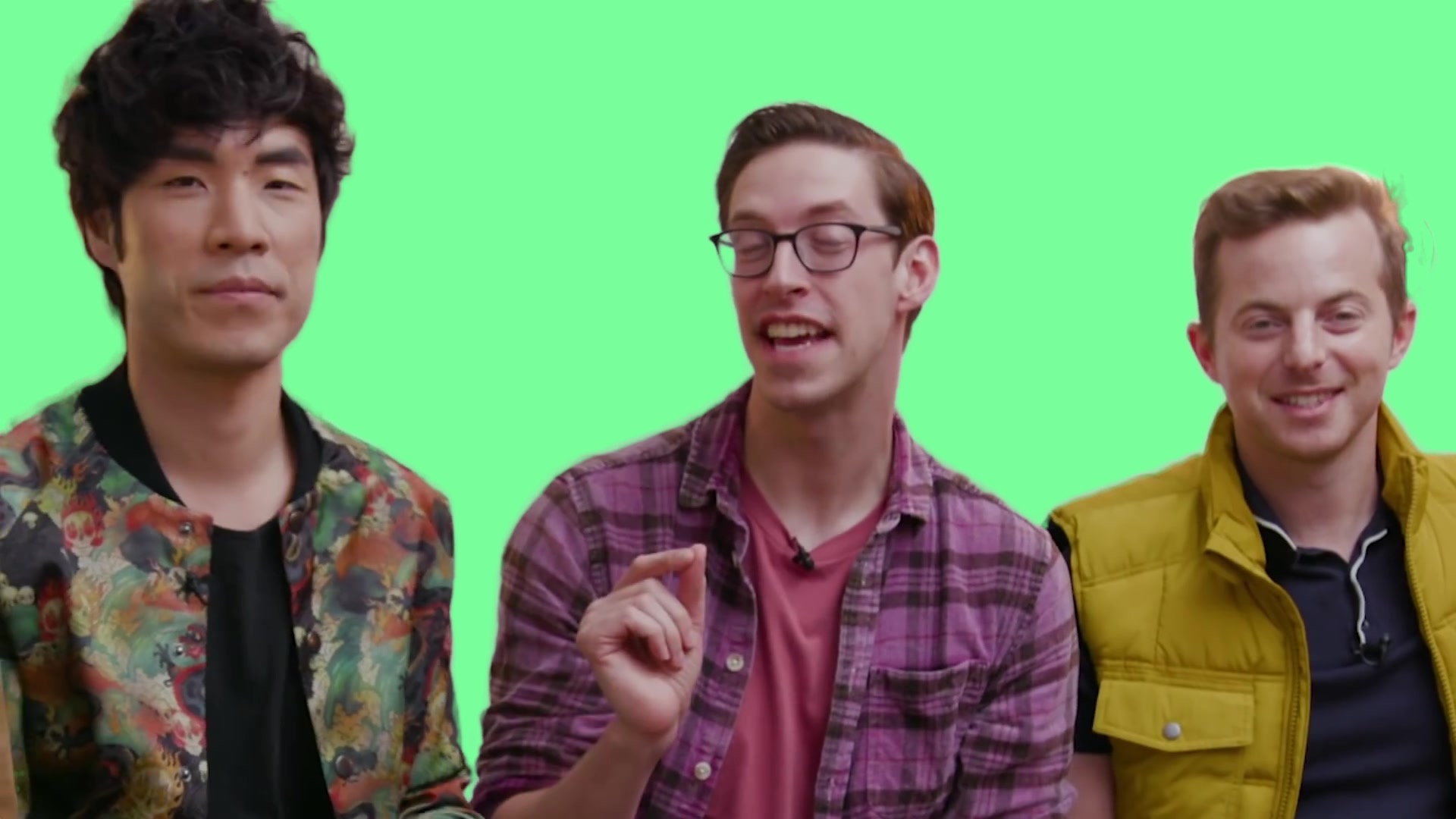}\hspace{0em}
   \vspace{-10pt}
    \vspace{1pt}
    \end{minipage}

  \label{fig:crgnn}

  \end{center}

    \centering
\vspace{5pt}   
   \caption{\small Comparisons of our model to MODNet \cite{ke2022modnet} and RVM \cite{lin2022robust} on Footage clips released on RVM's GitHub \cite{lin2022robust}}
\vspace{5pt}   
\label{fig:footage}
\end{figure*}
  
  \item[3.] Figure \ref{fig:temporal} presents another example of temporal consistency comparison. Over time, our model produces more consistent and coherent results.
\begin{figure*}[]
\captionsetup[subfigure]{labelformat=empty}
\begin{center}
    \scriptsize

\begin{minipage}[]{.99\textwidth}
    \centering
    \footnotesize
    \begin{subfigure}[b]{0.192\textwidth}
        \caption{$t$}
        \includegraphics[width=\textwidth]{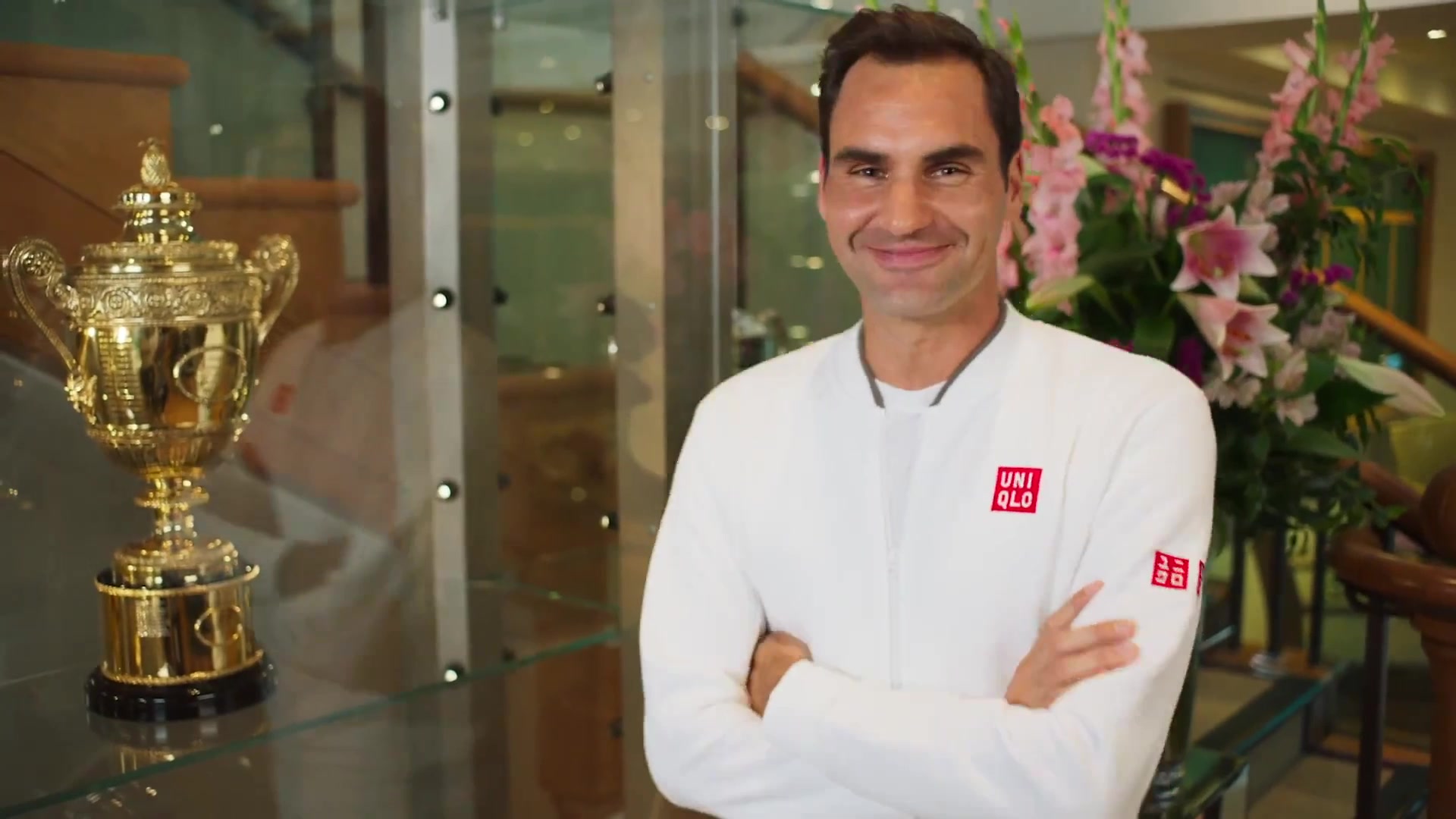}
    \end{subfigure}\hspace{0em}
    \begin{subfigure}[b]{0.192\textwidth}
        \caption{$t+5$}
        \includegraphics[width=\textwidth]{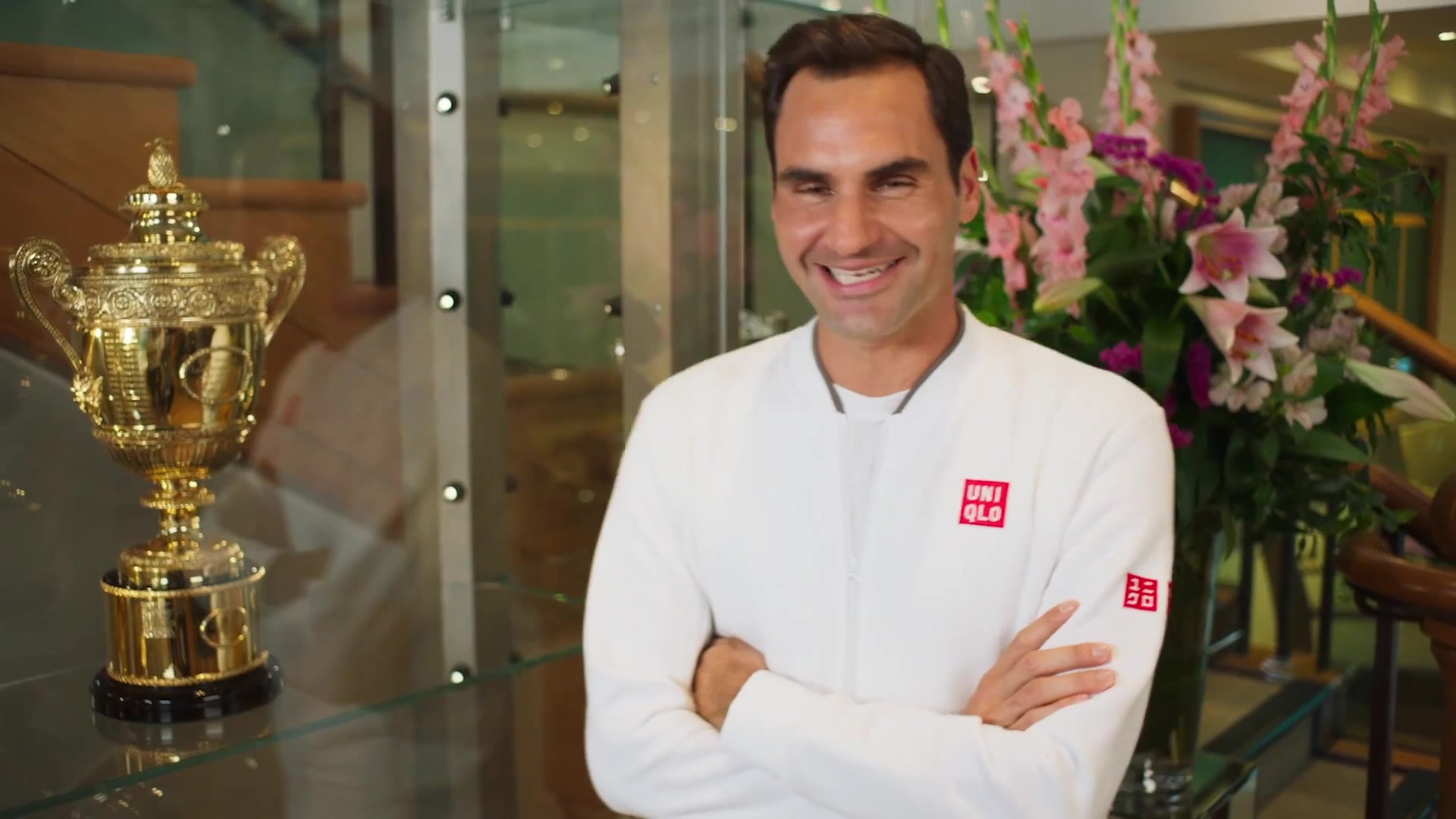}
    \end{subfigure}\hspace{0em}
    \begin{subfigure}[b]{0.192\textwidth}
        \caption{$t+10$}
        \includegraphics[width=\textwidth]{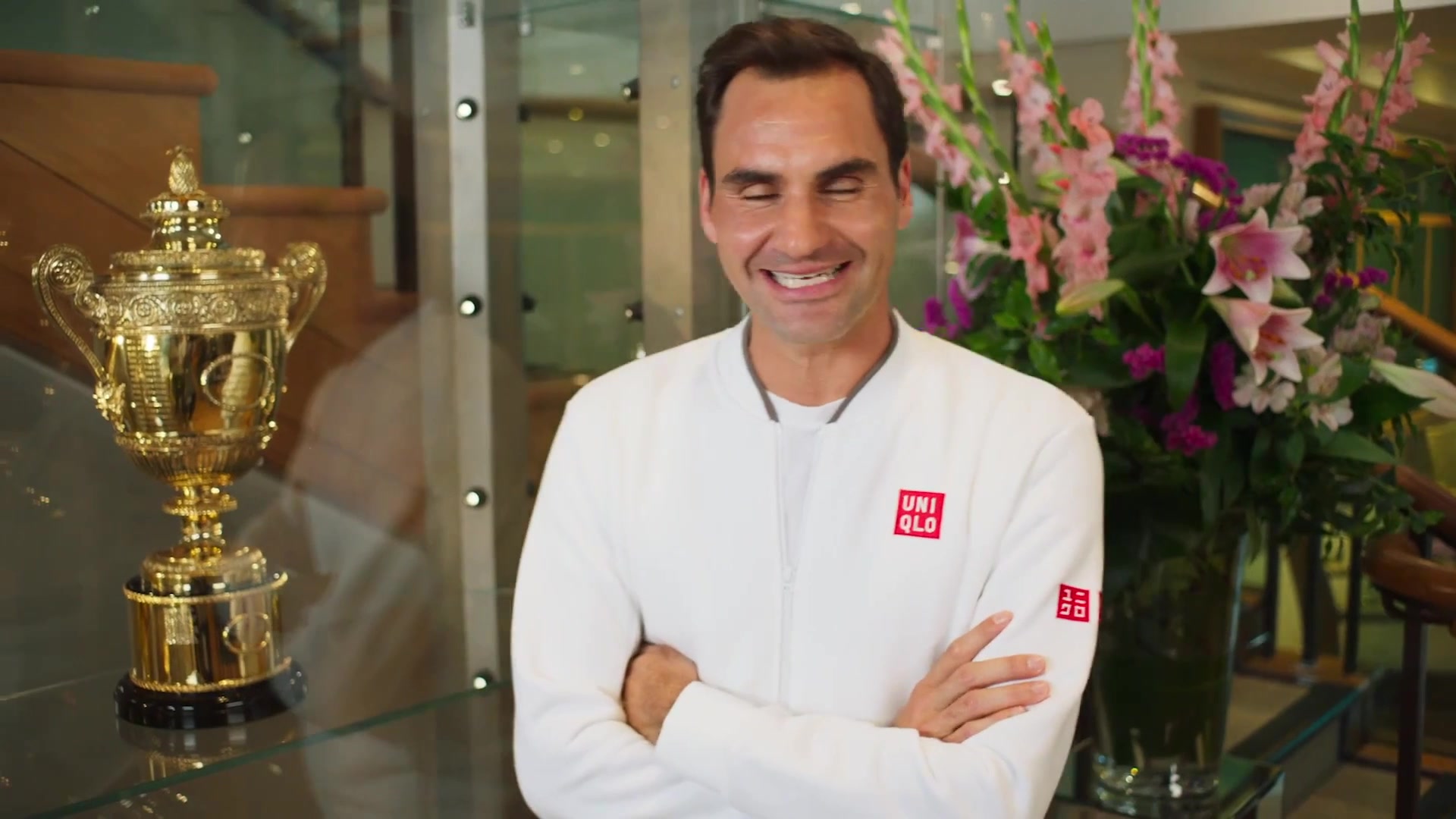}
    \end{subfigure}\hspace{0em}
    \begin{subfigure}[b]{0.192\textwidth}
        \caption{$t+15$}
        \includegraphics[width=\textwidth]{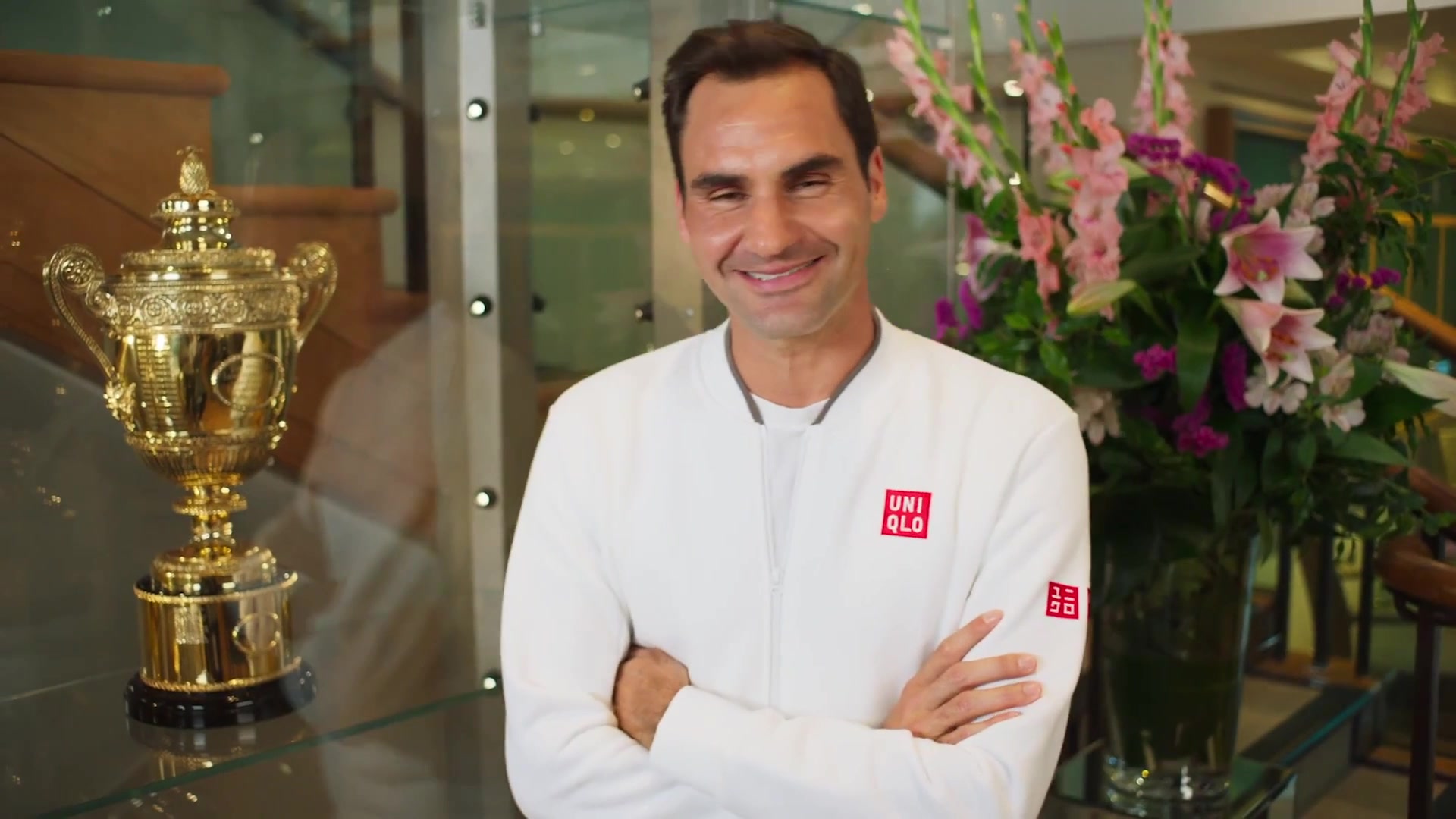}
    \end{subfigure}\hspace{0em}
    \begin{subfigure}[b]{0.192\textwidth}
        \caption{$t+20$}
        \includegraphics[width=\textwidth]{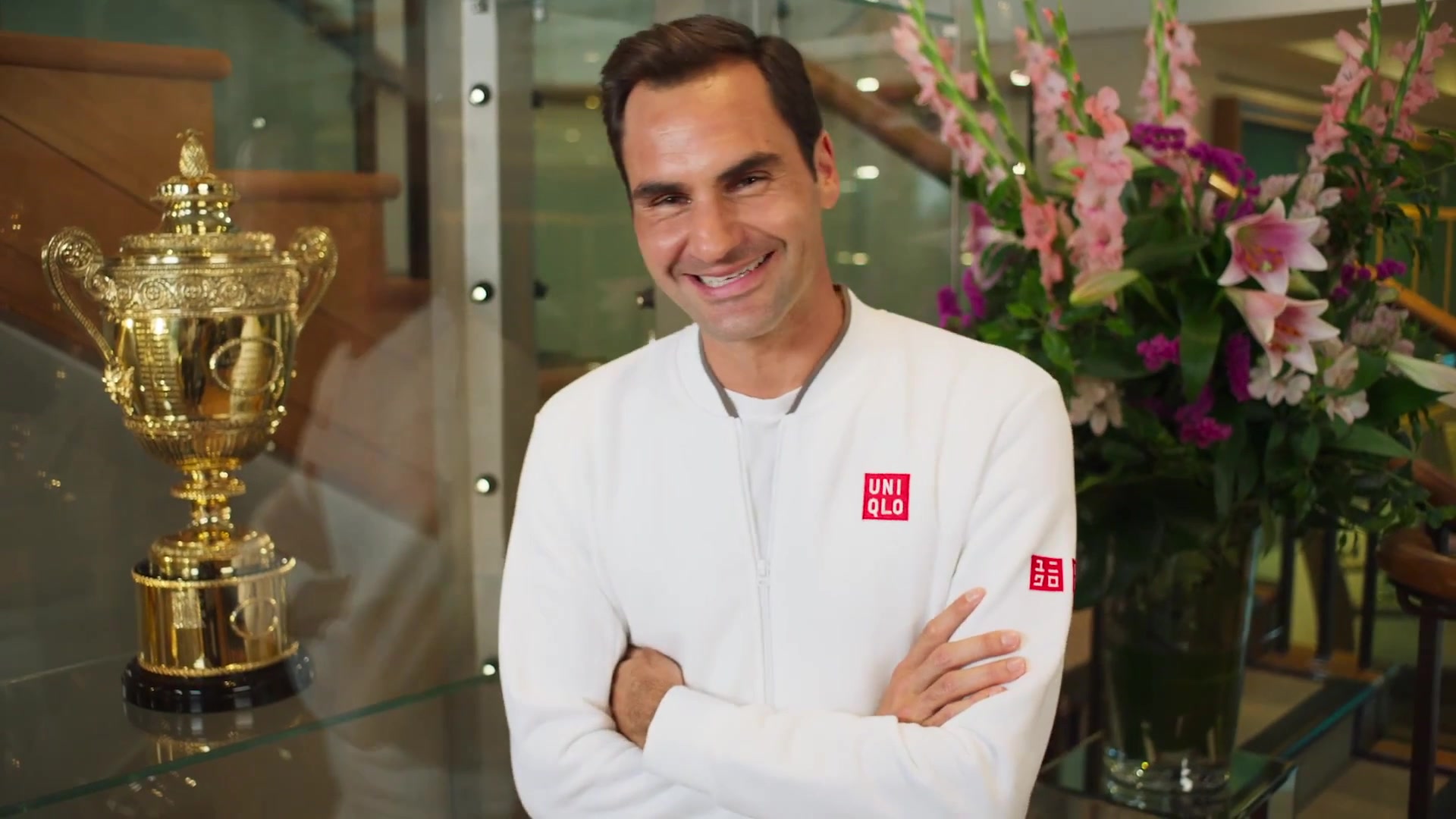}
    \end{subfigure}\hspace{0em}
    \subcaption{(a) Input.}
   \vspace{10pt}
    \end{minipage}

\begin{minipage}[]{.99\textwidth}
        \centering
        \footnotesize
    \includegraphics[trim=0 0 0 0, clip,width=0.192\textwidth]{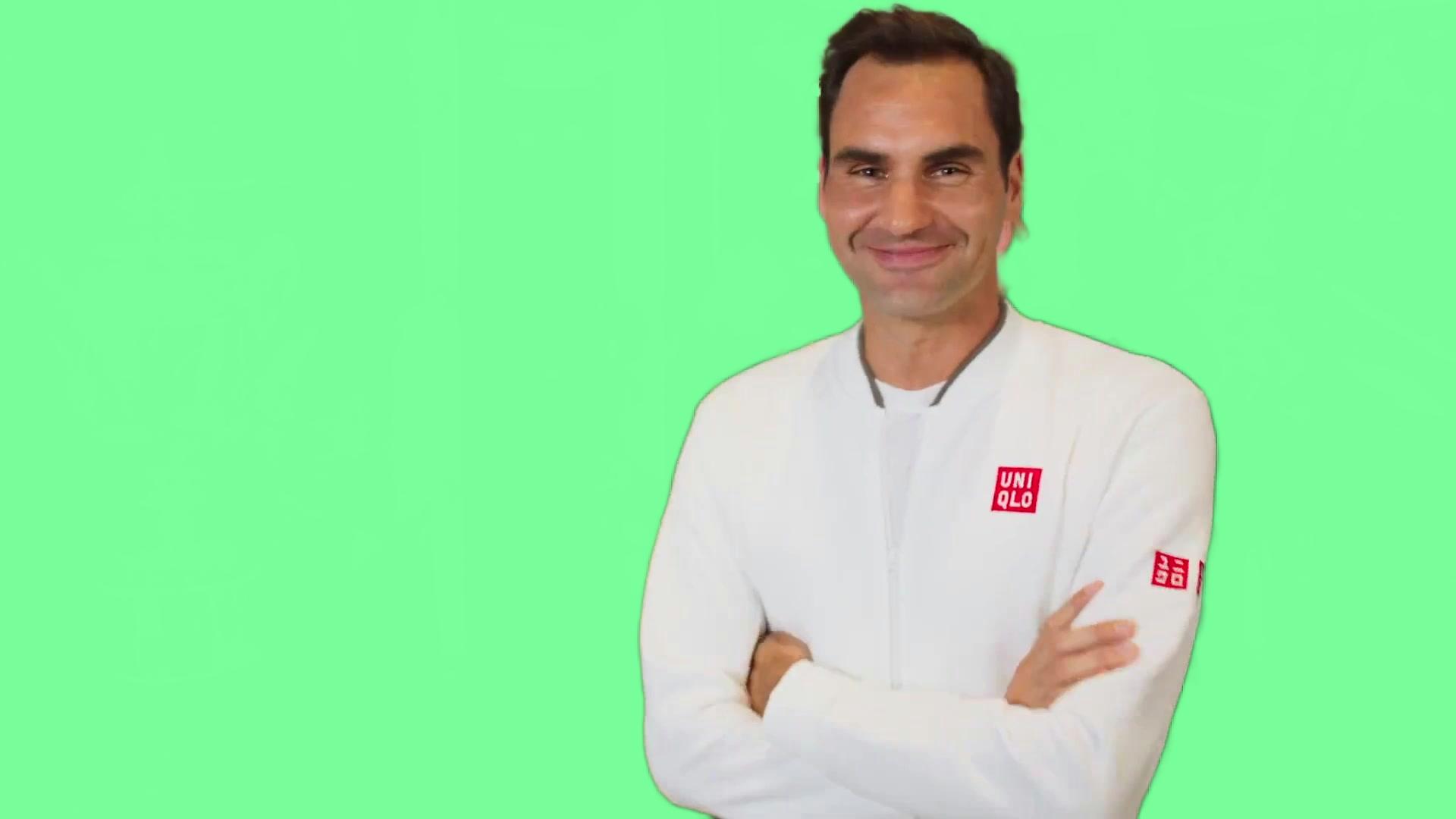}\hspace{0em}
    \includegraphics[trim=0 0 0 0, clip,width=0.192\textwidth]{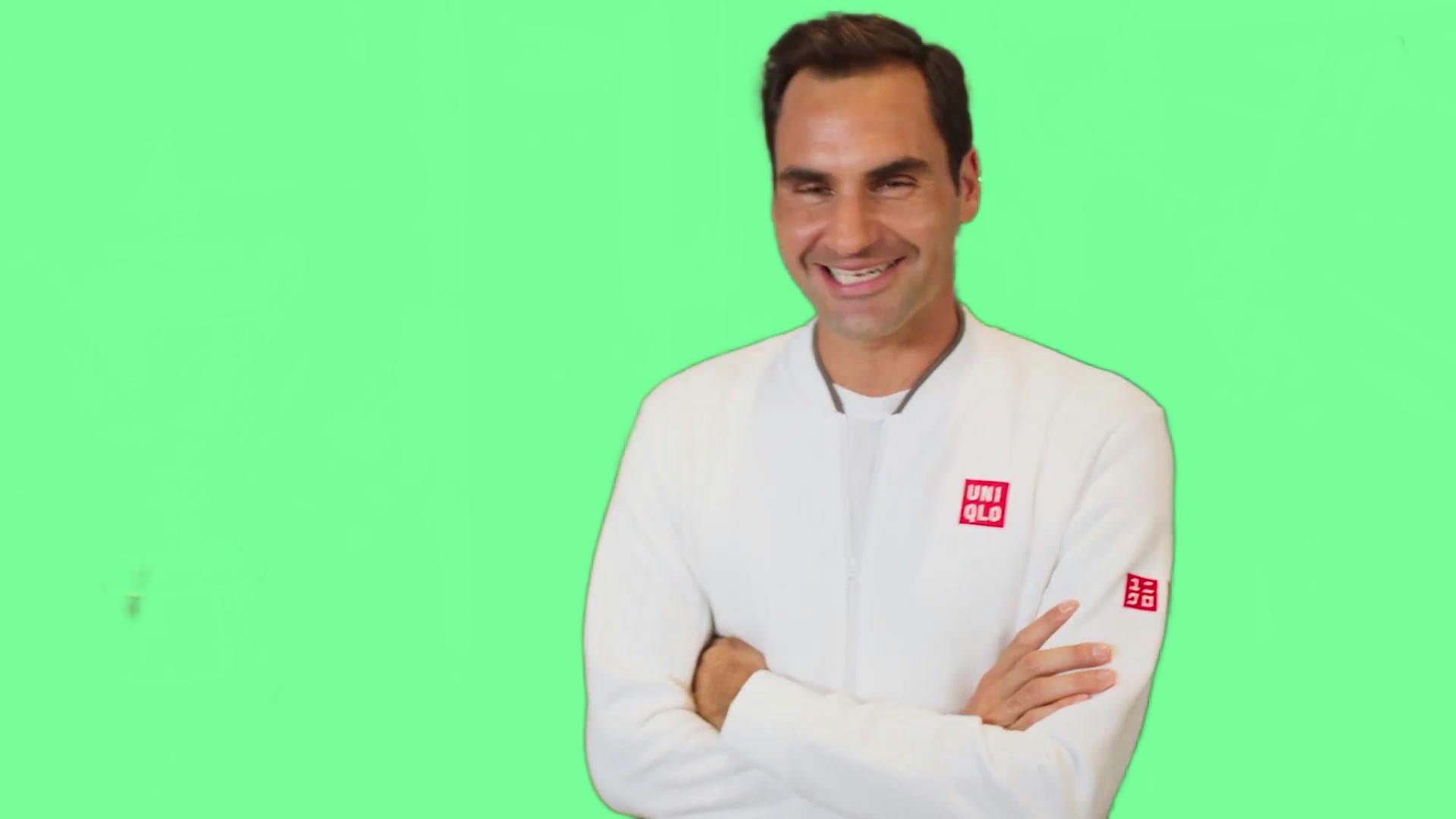}\hspace{0em}
    \includegraphics[trim=0 0 0 0, clip,width=0.192\textwidth]{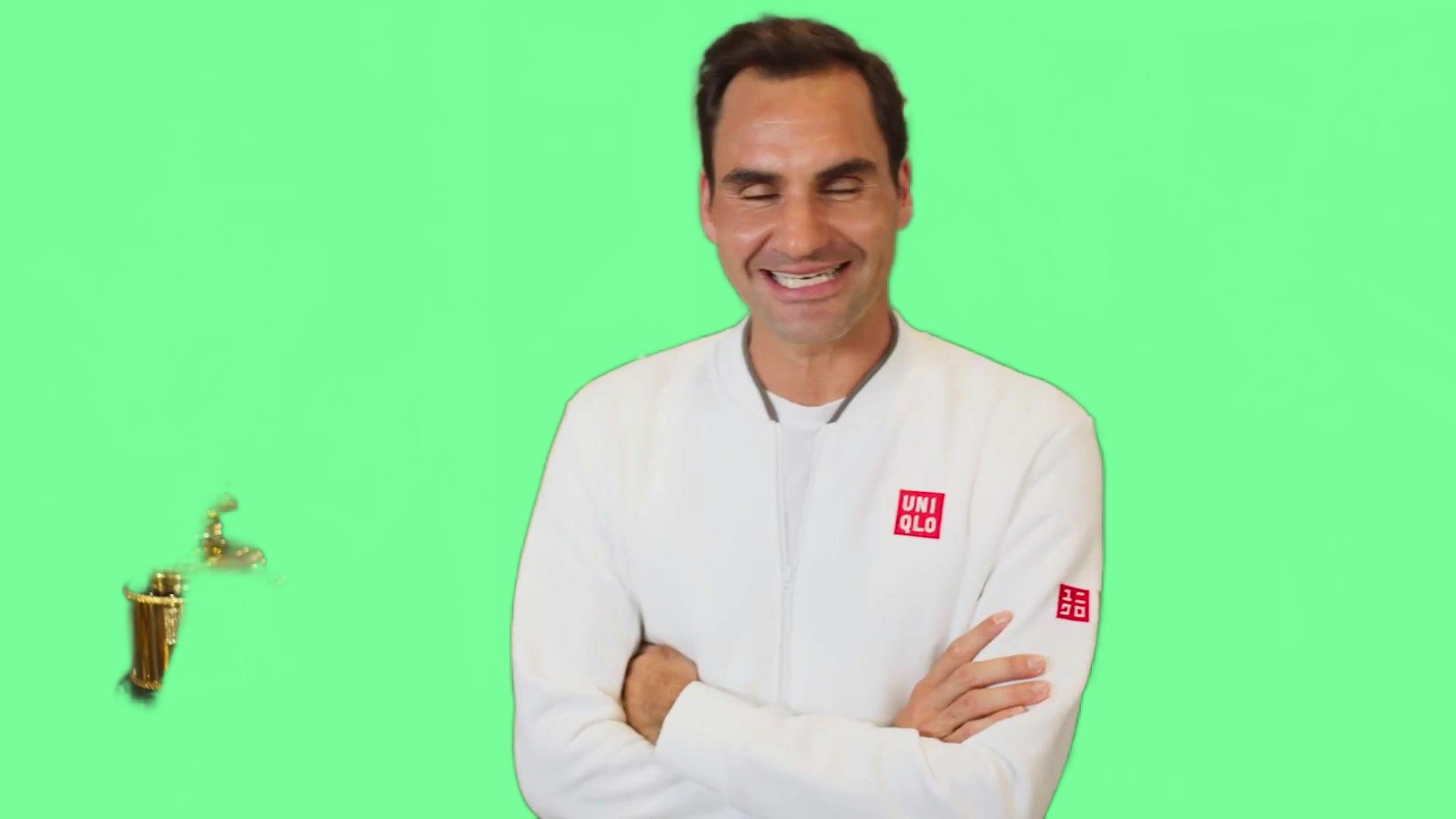}\hspace{0em}
    \includegraphics[trim=0 0 0 0, clip,width=0.192\textwidth]{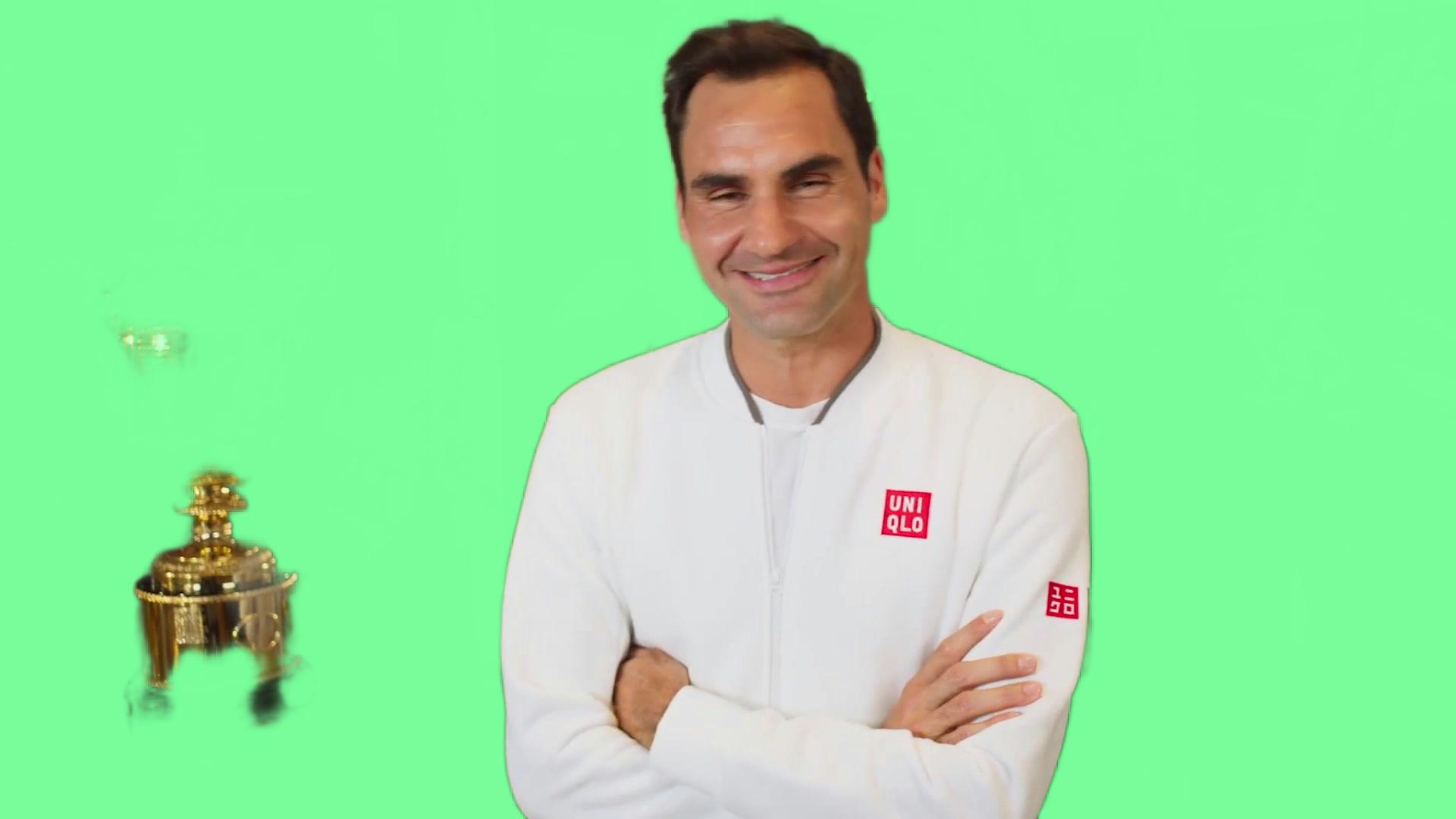}\hspace{0em}
    \includegraphics[trim=0 0 0 0, clip,width=0.192\textwidth]{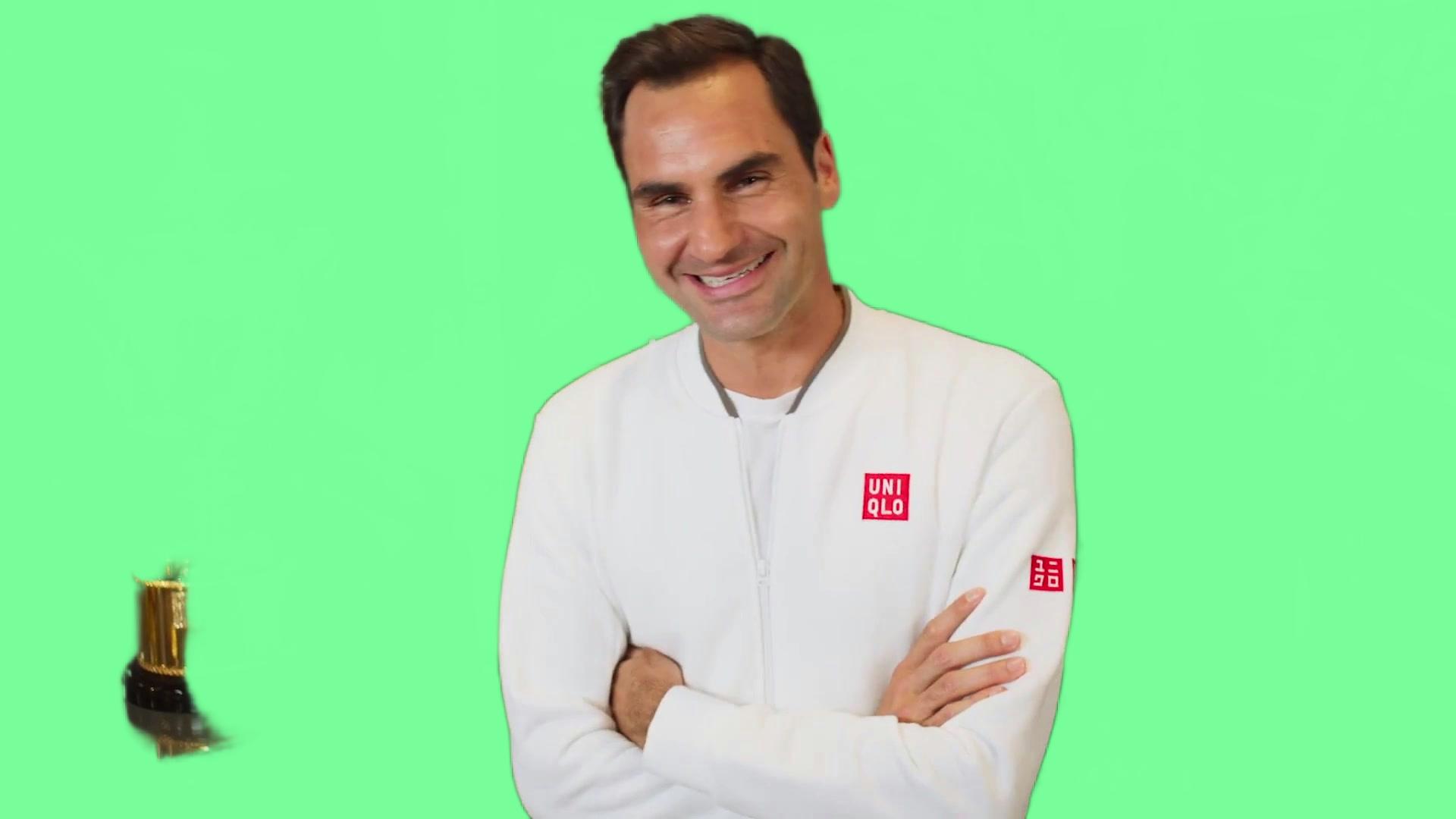}\hspace{0em}

    \subcaption{(b) MODNet.}
    \vspace{10pt}
    \end{minipage}
    
\begin{minipage}[]{.99\textwidth}
    \centering
    \footnotesize
    \includegraphics[trim=0 0 0 0, clip,width=0.192\textwidth]{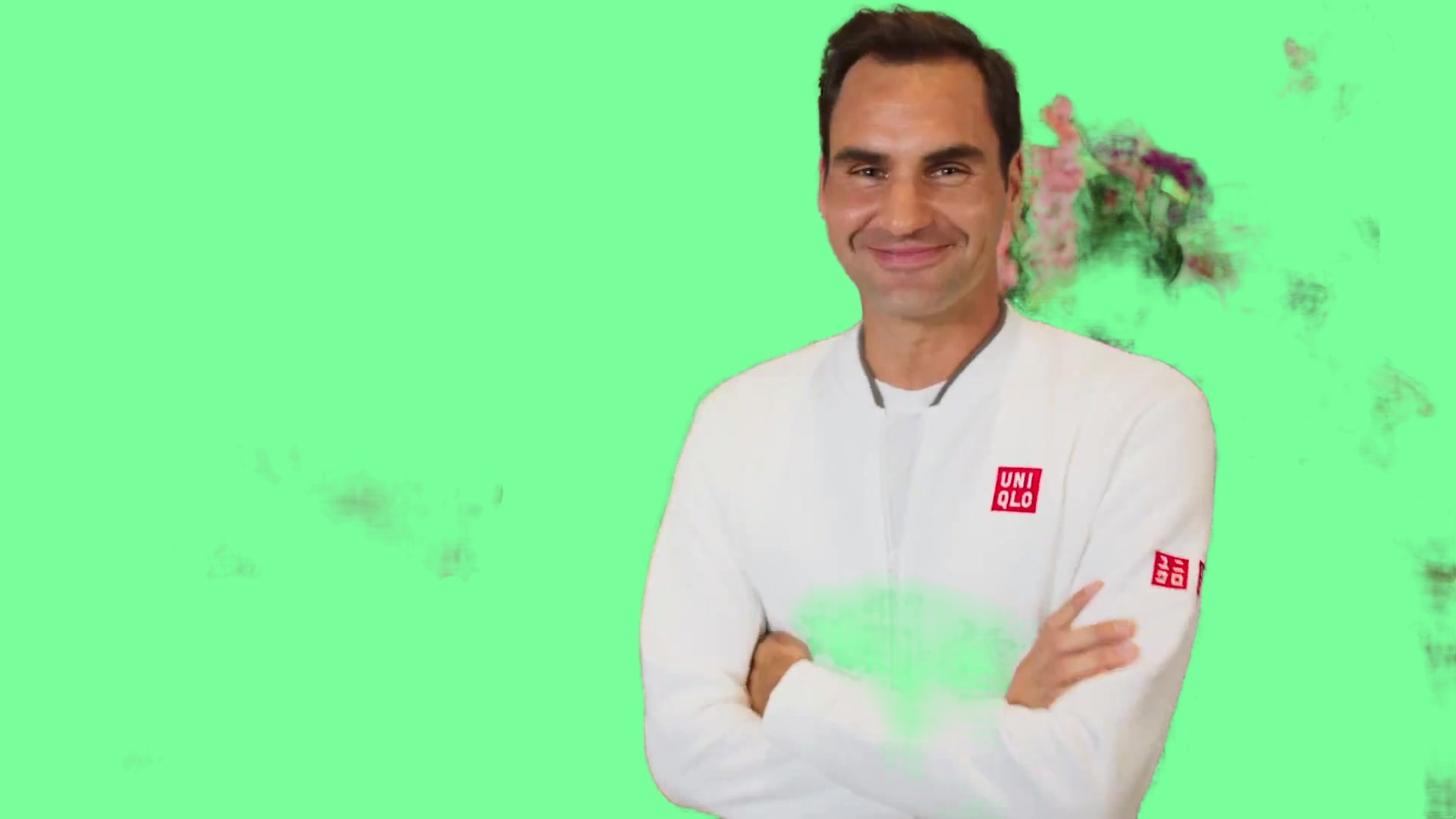}\hspace{0em}
    \includegraphics[trim=0 0 0 0, clip,width=0.192\textwidth]{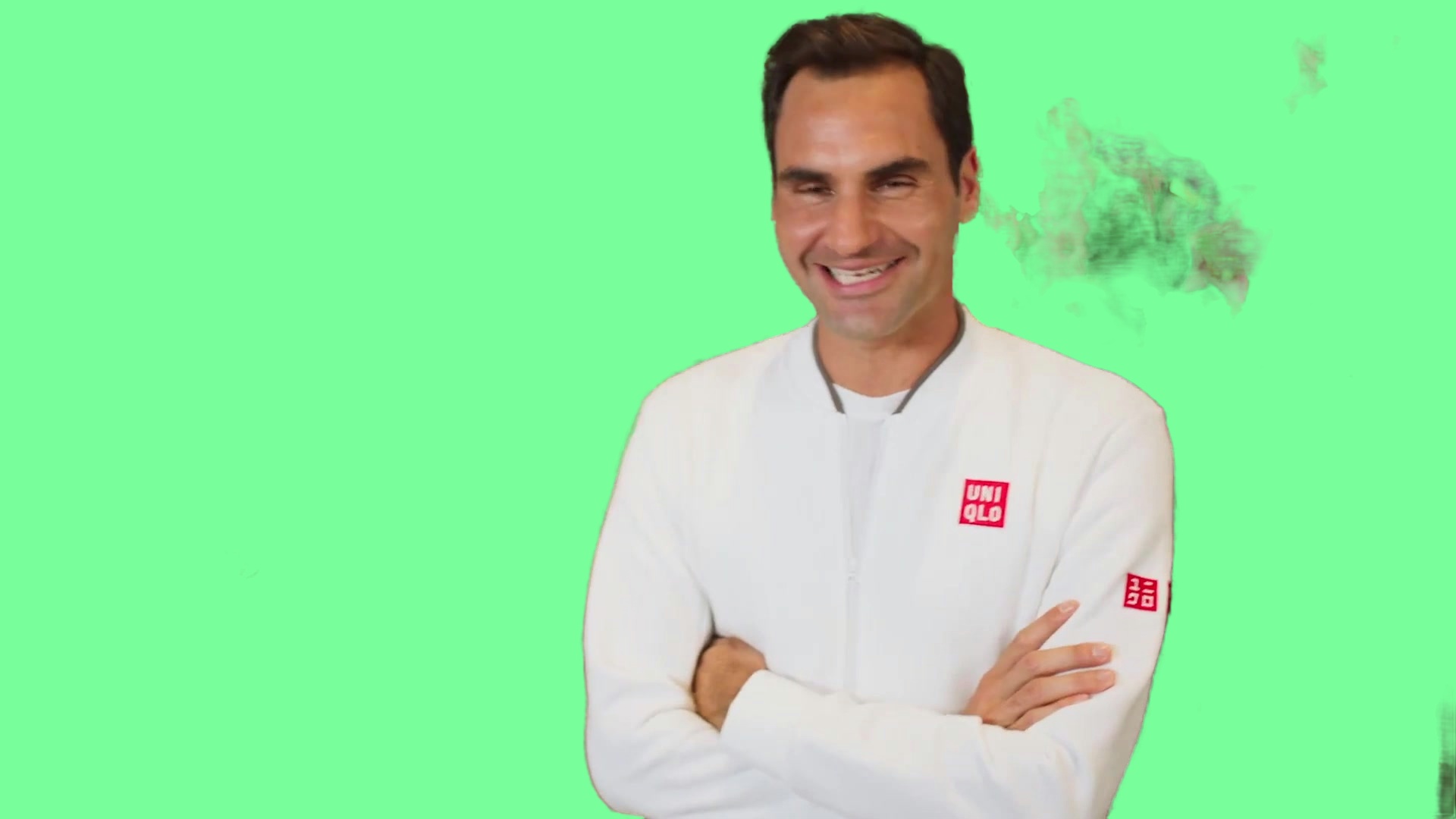}\hspace{0em}
    \includegraphics[trim=0 0 0 0, clip,width=0.192\textwidth]{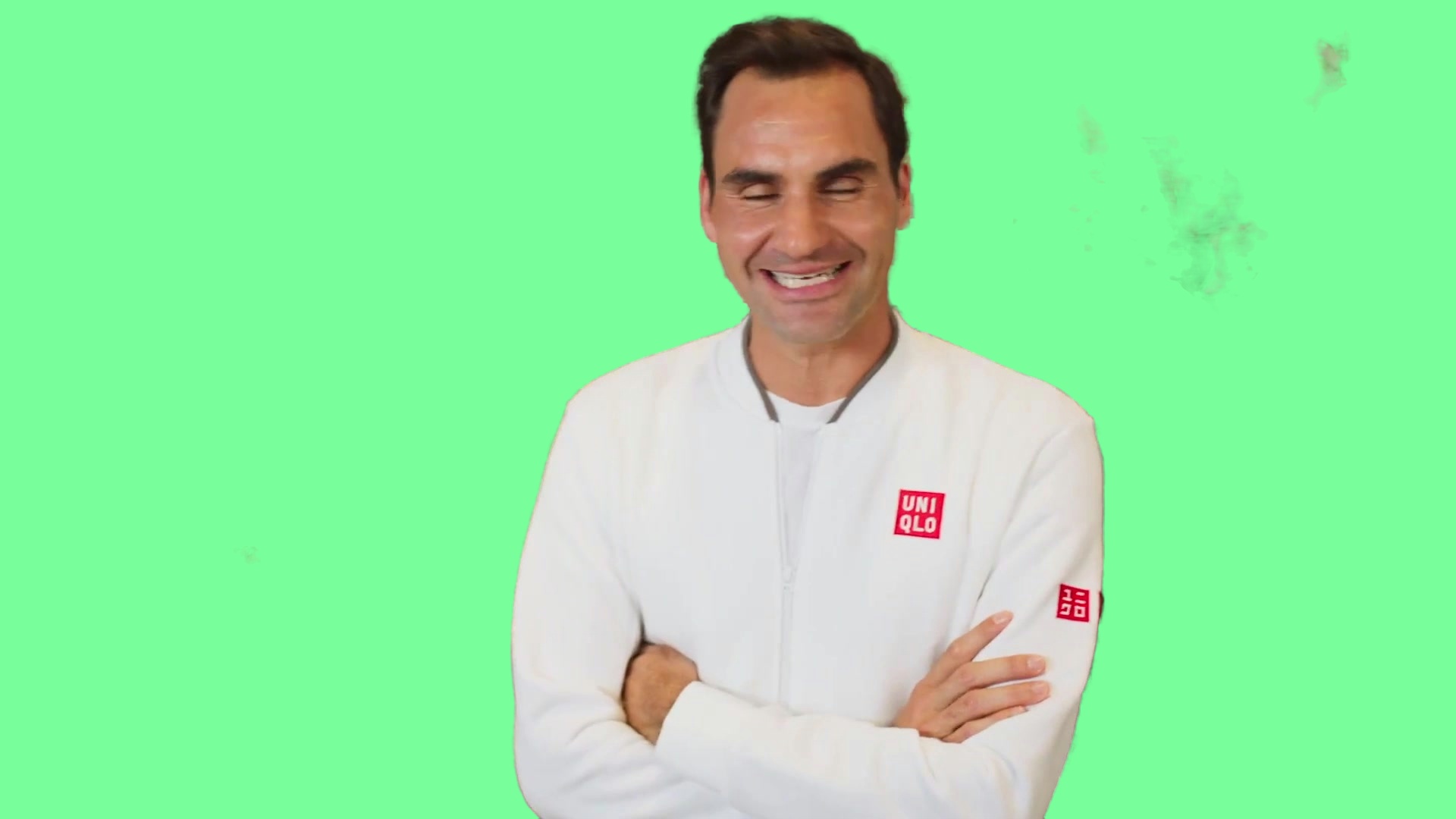}\hspace{0em}
    \includegraphics[trim=0 0 0 0, clip,width=0.192\textwidth]{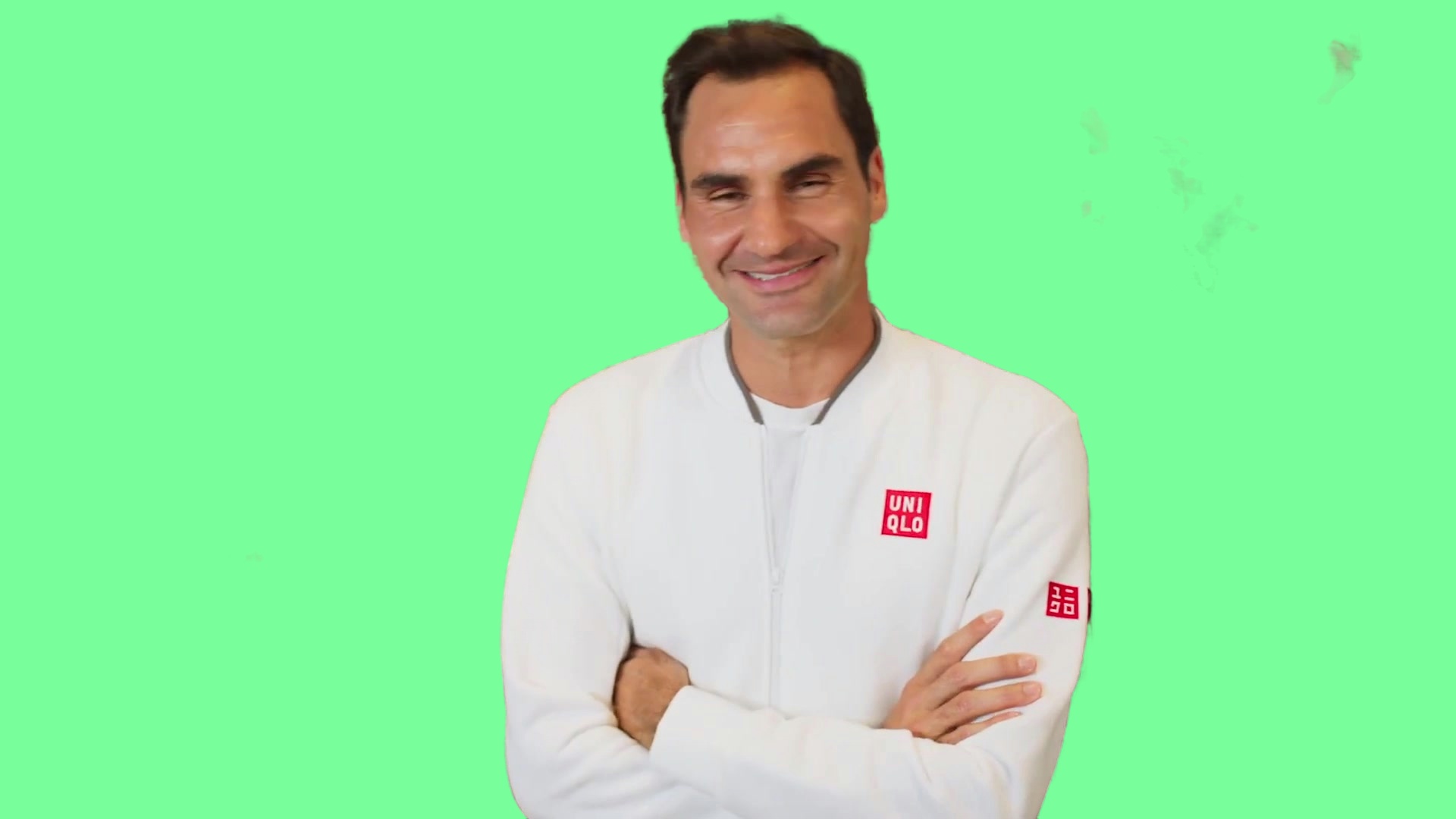}\hspace{0em}
    \includegraphics[trim=0 0 0 0, clip,width=0.192\textwidth]{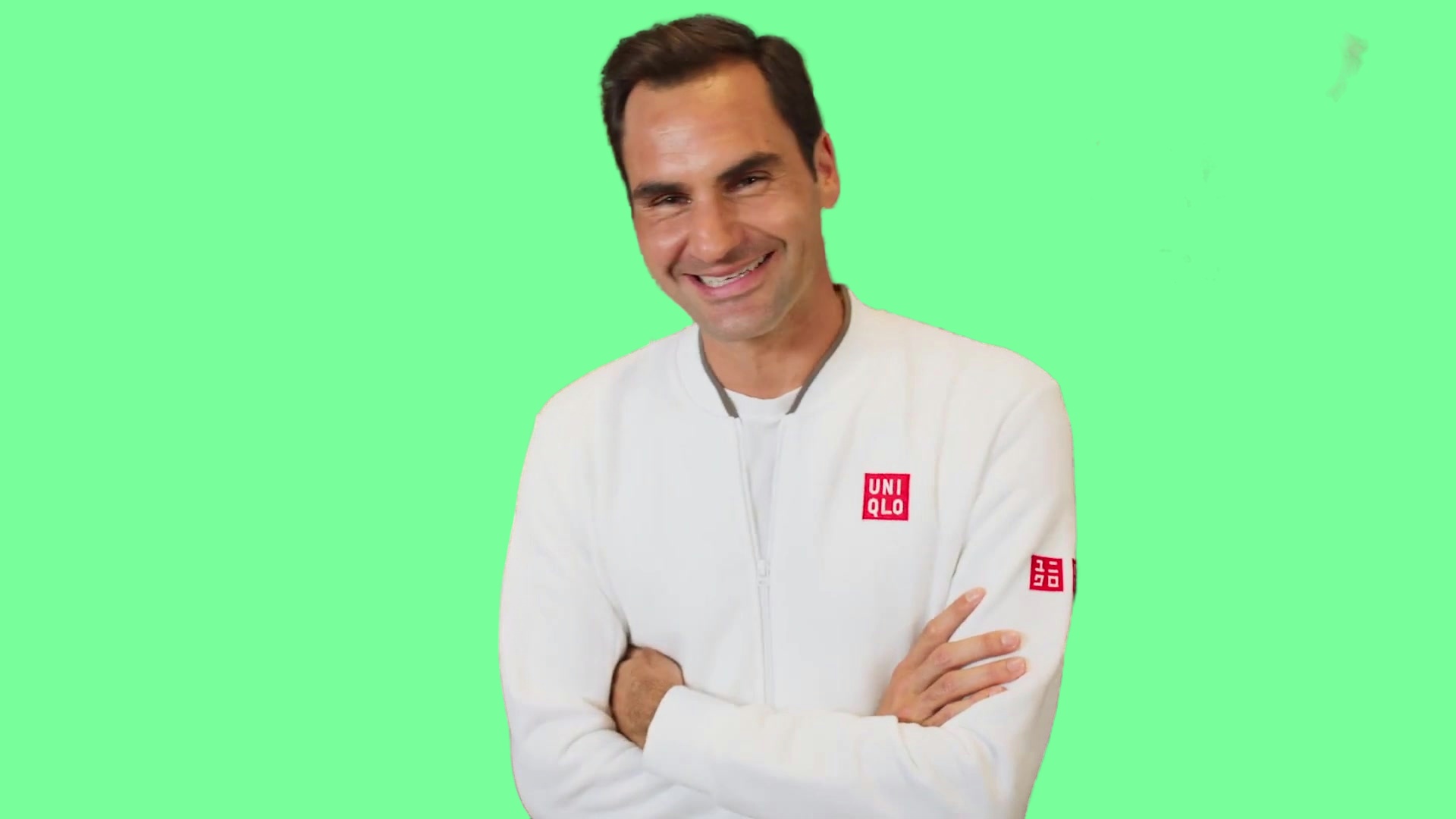}\hspace{0em}

    \subcaption{(c) RVM.}
    \vspace{10pt}
    \end{minipage}

\begin{minipage}[]{.99\textwidth}
        \centering
        \footnotesize
    \includegraphics[trim=0 0 0 0, clip,width=0.192\textwidth]{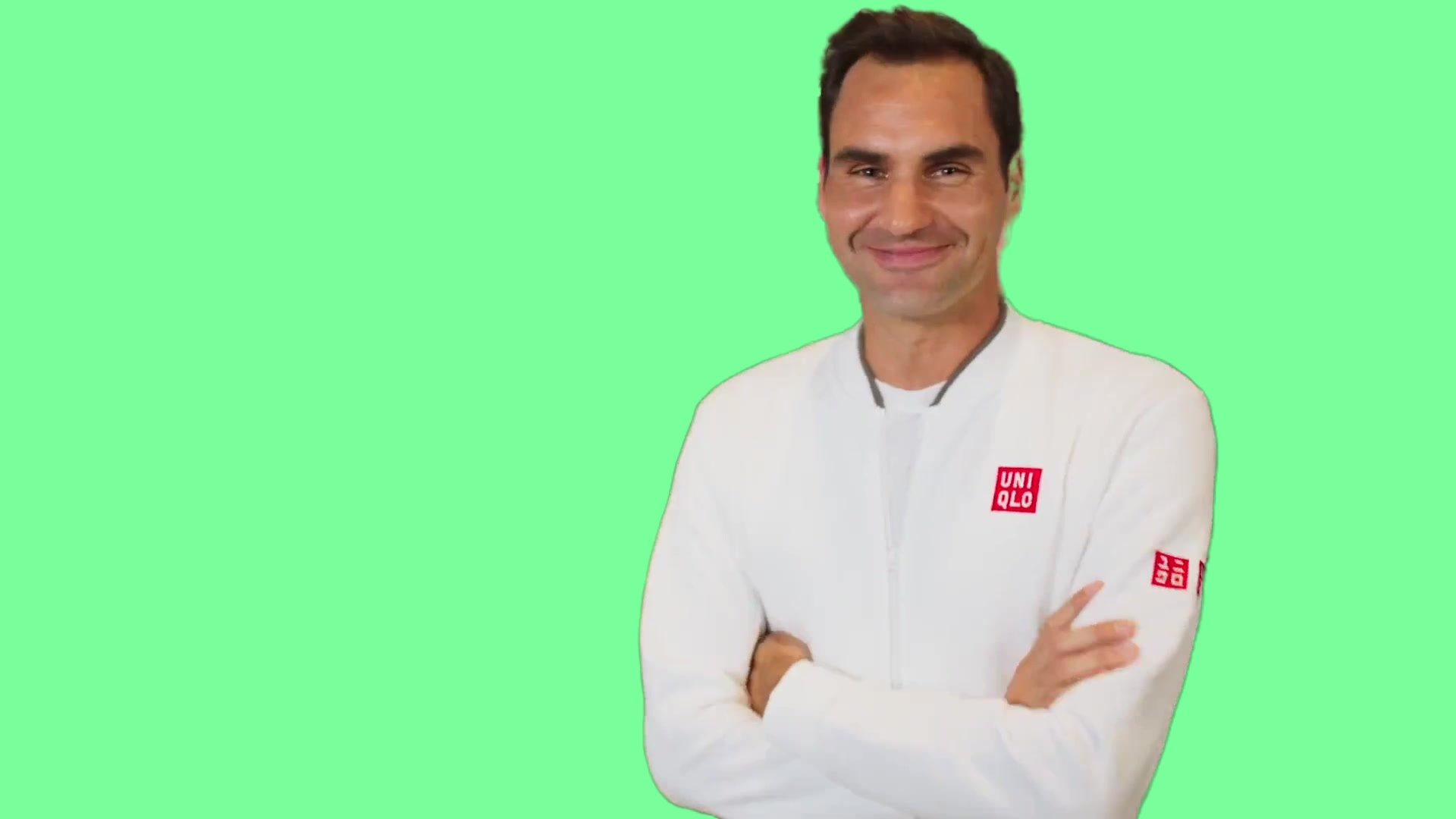}\hspace{0em}
    \includegraphics[trim=0 0 0 0, clip,width=0.192\textwidth]{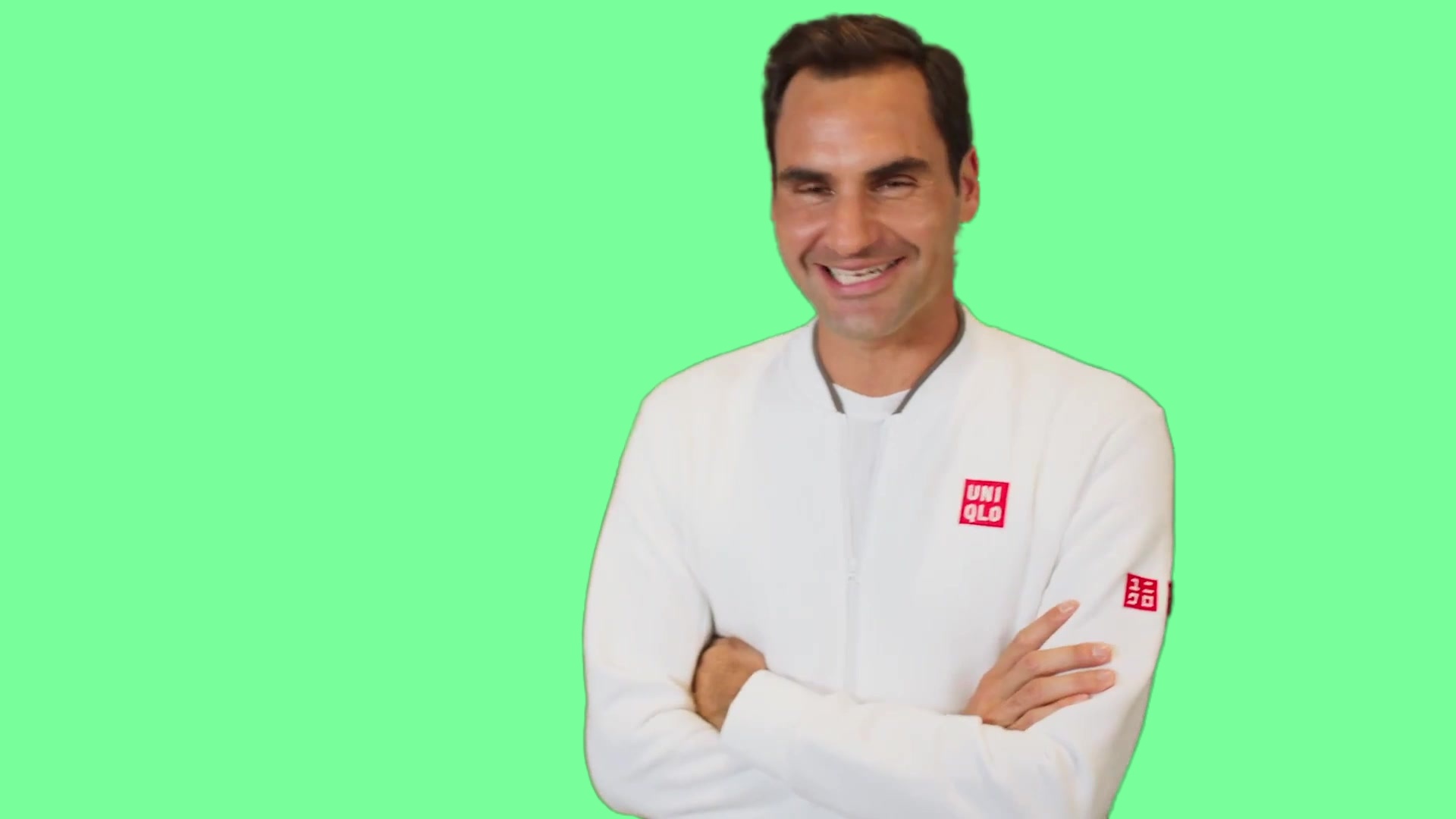}\hspace{0em}
    \includegraphics[trim=0 0 0 0, clip,width=0.192\textwidth]{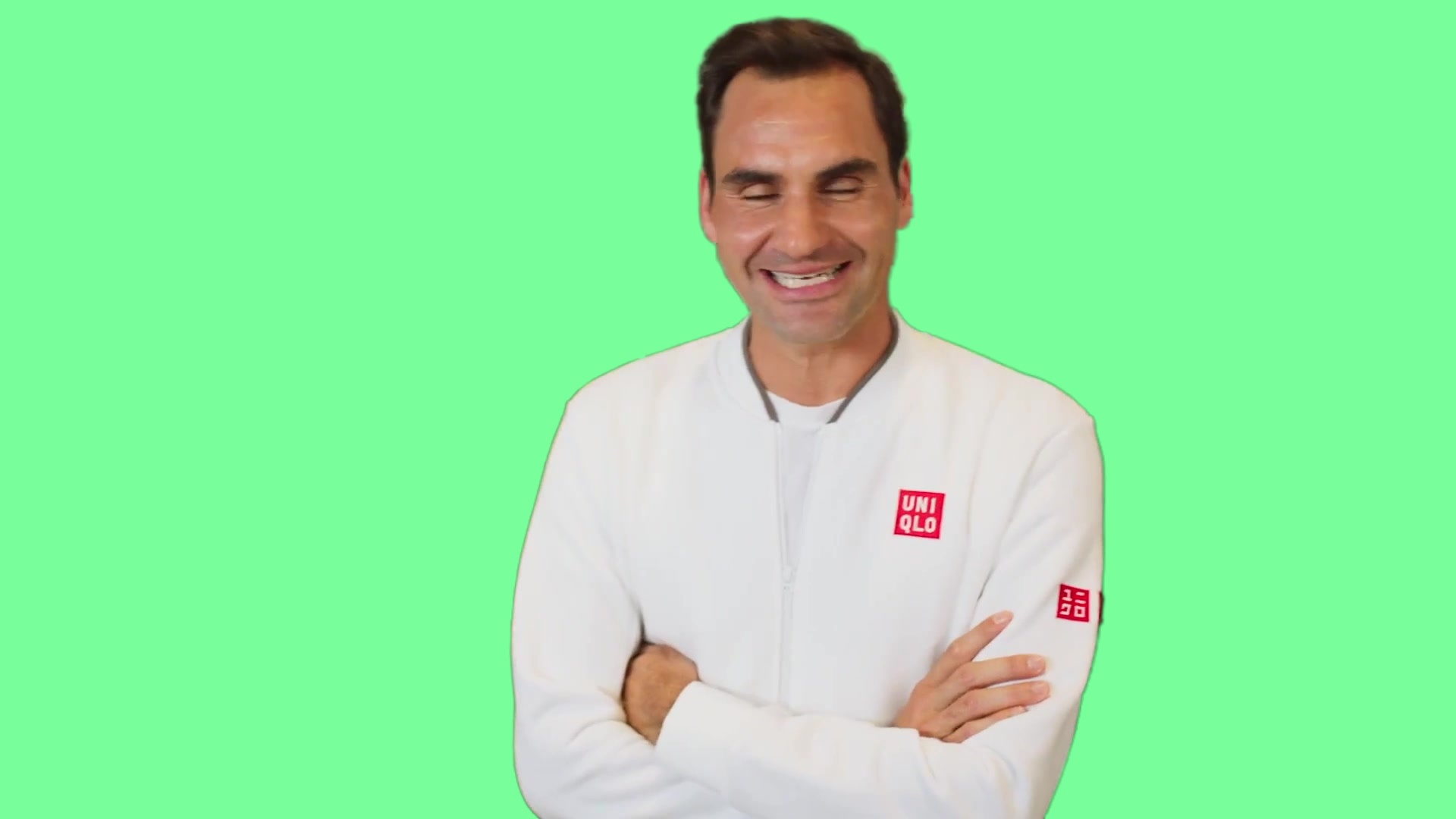}\hspace{0em}
    \includegraphics[trim=0 0 0 0, clip,width=0.192\textwidth]{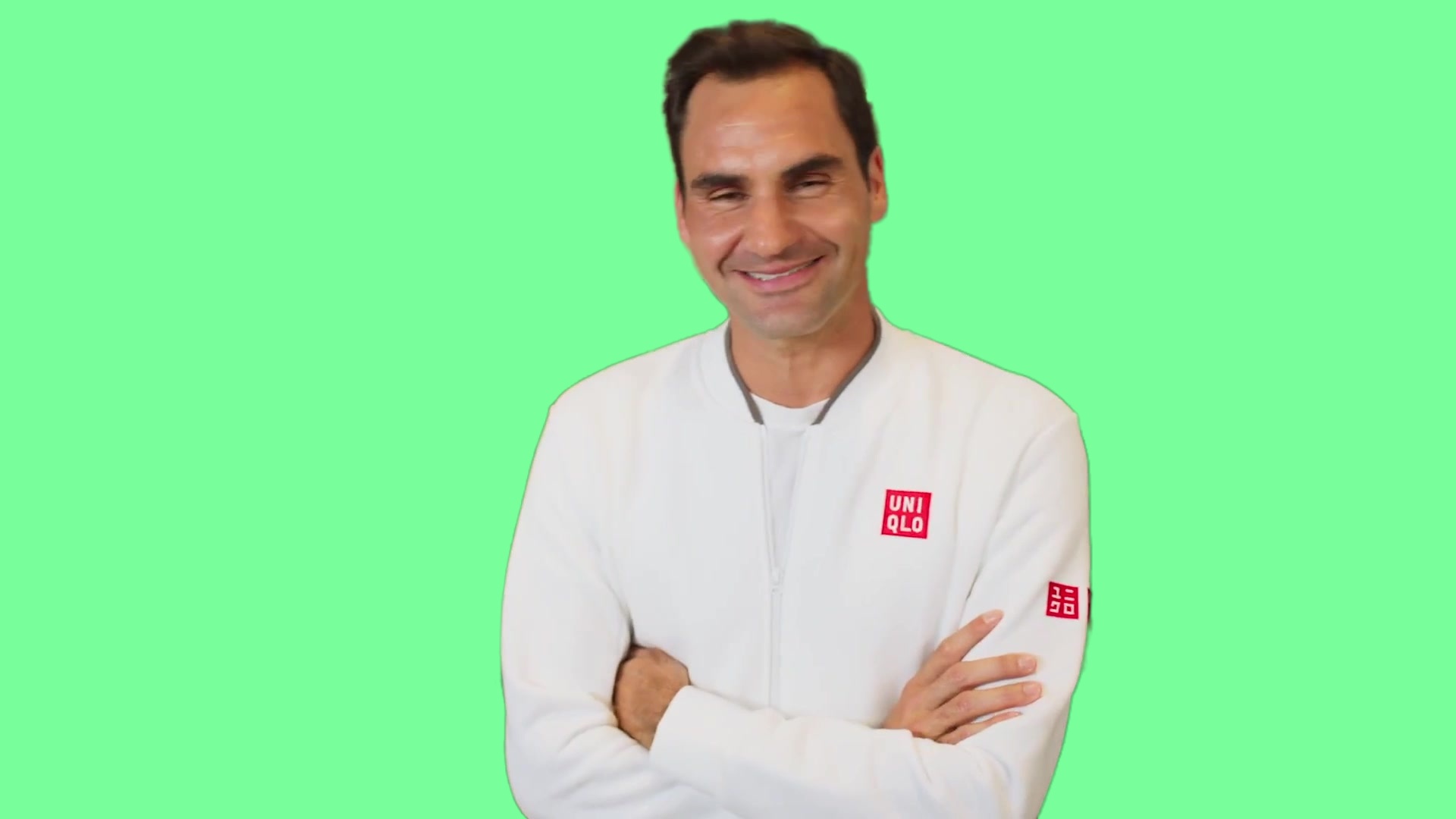}\hspace{0em}
    \includegraphics[trim=0 0 0 0, clip,width=0.192\textwidth]{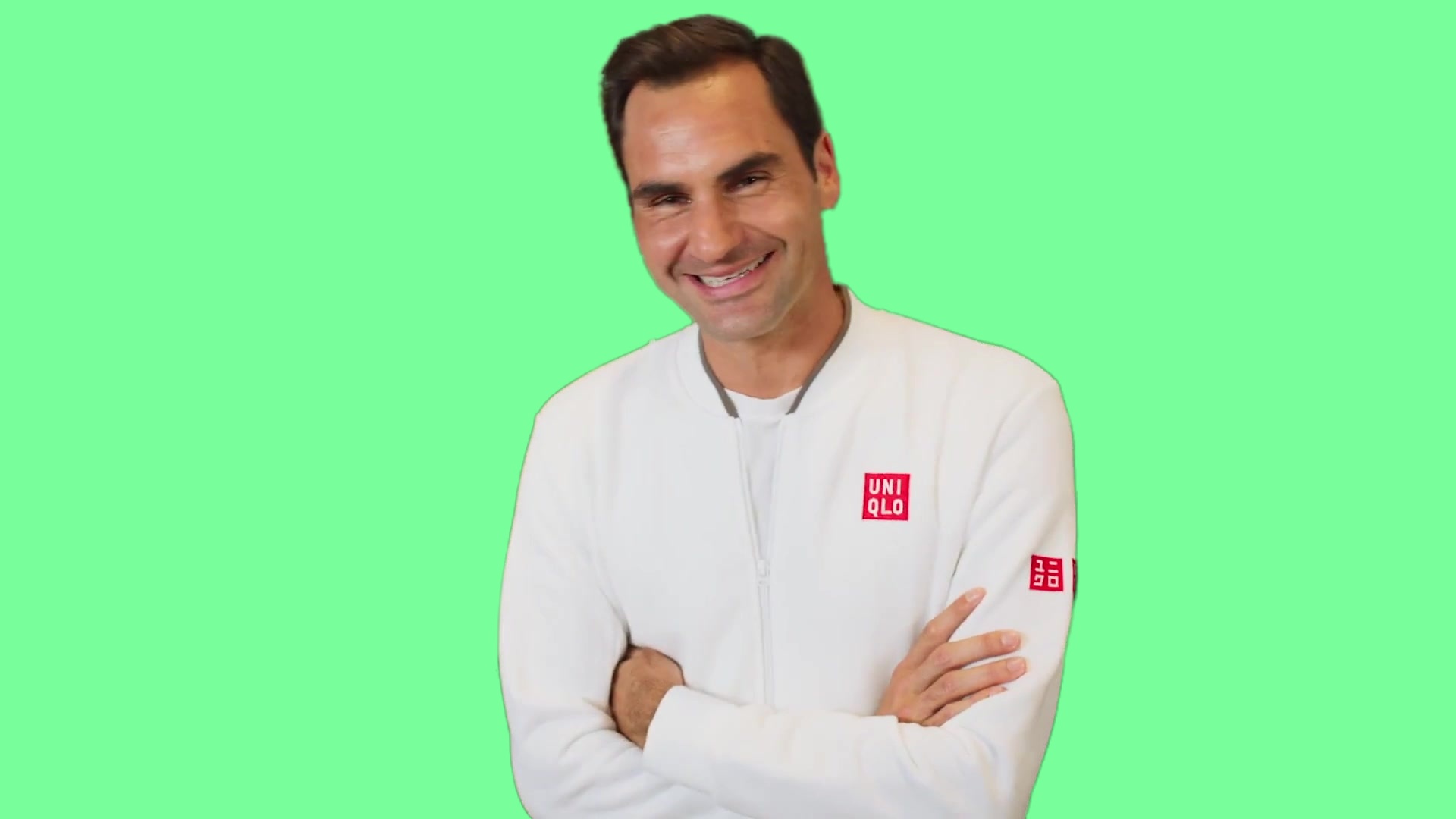}\hspace{0em}

    \subcaption{(d) AdaM.}
    \vspace{1pt}
    \end{minipage}

  \label{fig:crgnn}

  \end{center}

    \centering
   \caption{\small Temporal consistency comparisons on a real-world video crawled from Youtube.}
\label{fig:temporal}
\end{figure*}

  \item[4.] Figure \ref{fig:youtube} illustrates a series of test experiments conducted on real-world YouTube videos. The video clips are used to evaluate the robustness of the proposed algorithm in real-world scenarios. They demonstrate a variety of challenging scenarios, such as fast human motion, camera zooming in and out, rapid camera motion, low light conditions, and cluttered backgrounds. These examples demonstrate AdaM's competitiveness, suggesting it can provide reliable video matting and achieve great generalizability.

 \begin{figure*}[]
\captionsetup[subfigure]{labelformat=empty}
\begin{center}
    \scriptsize

\begin{minipage}[]{.99\textwidth}
    \centering
    \footnotesize
    \begin{subfigure}[b]{0.23\textwidth}
        \caption{Input}
        \includegraphics[width=\textwidth]{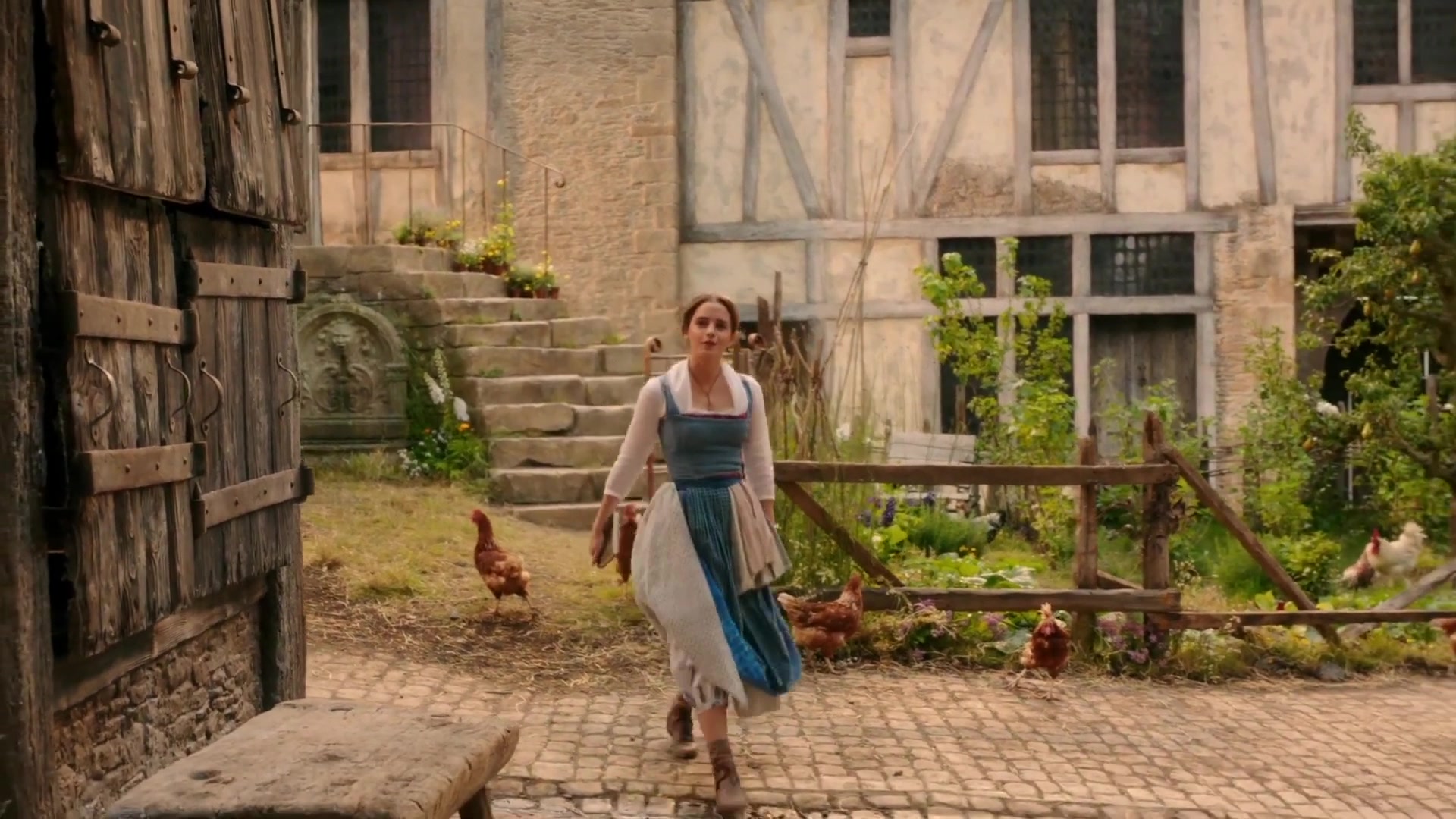}
    \end{subfigure}\hspace{0.2em}    
    \begin{subfigure}[b]{0.23\textwidth}
        \caption{MODNet}
        \includegraphics[width=\textwidth]{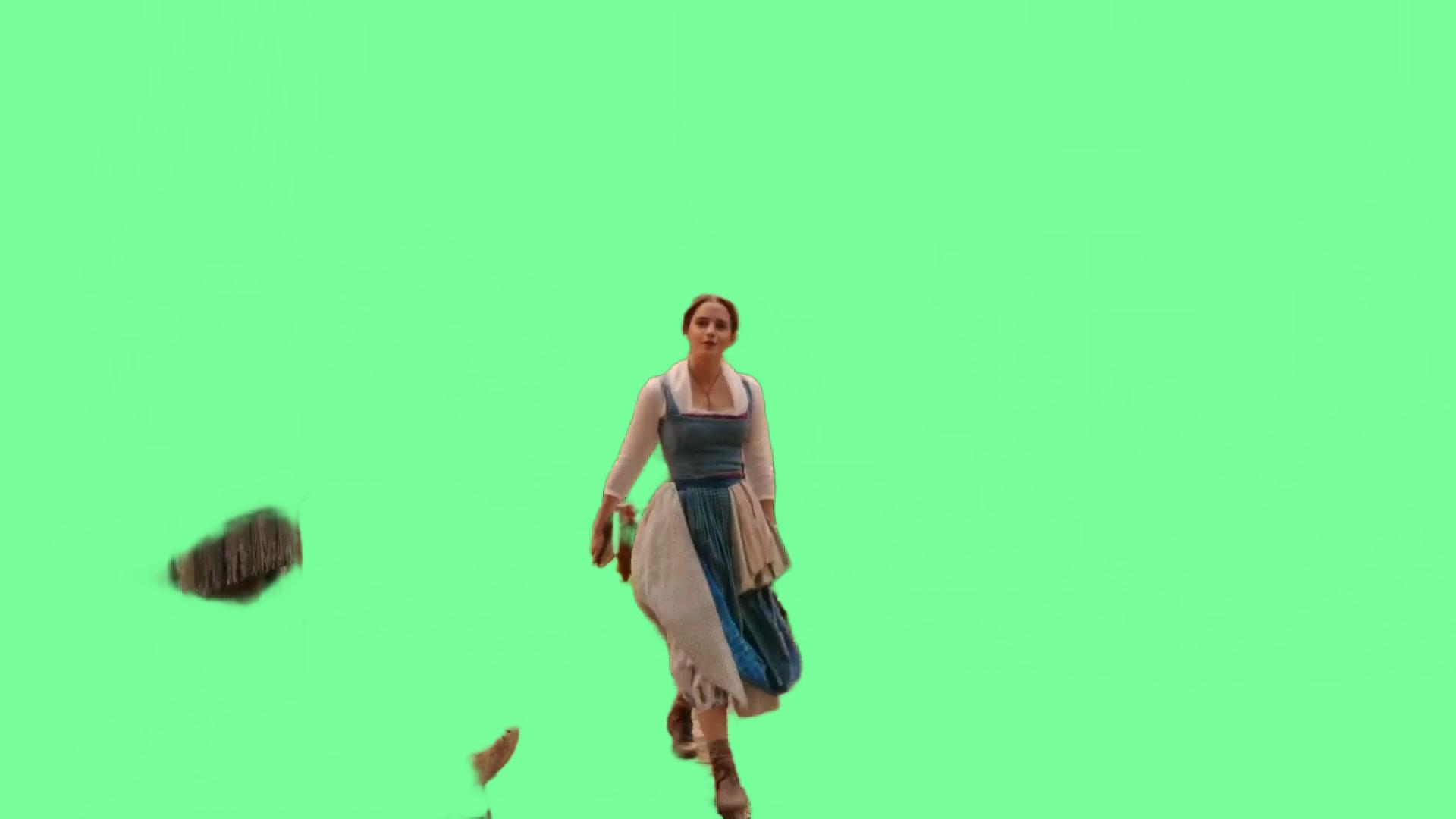}
    \end{subfigure}\hspace{0.2em}    
    \begin{subfigure}[b]{0.23\textwidth}
        \caption{RVM}
        \includegraphics[width=\textwidth]{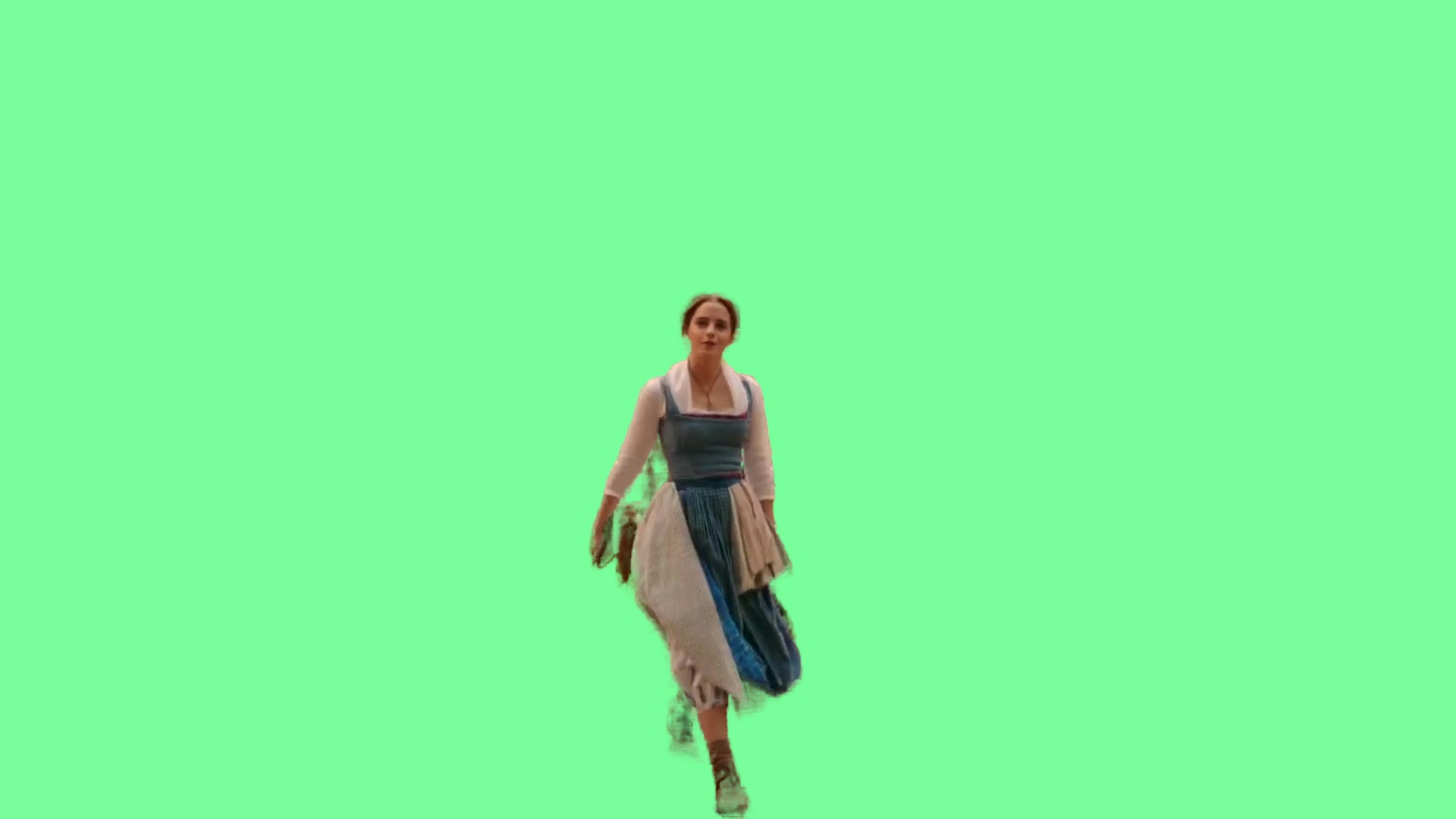}
    \end{subfigure}\hspace{0.2em}    
    \begin{subfigure}[b]{0.23\textwidth}
        \caption{AdaM}
        \includegraphics[width=\textwidth]{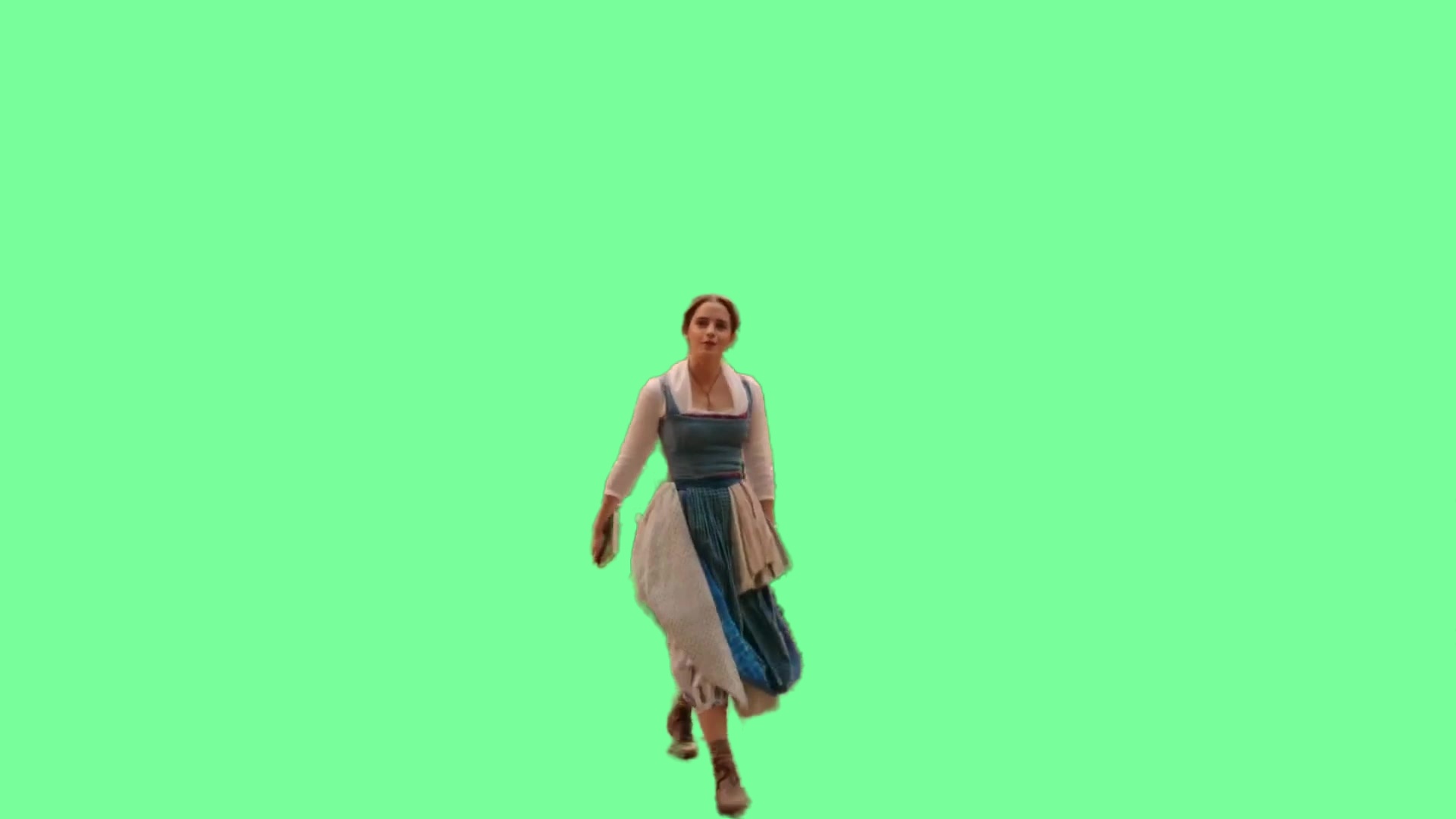}
    \end{subfigure}\hspace{0.2em}    
   \vspace{2pt}
    \end{minipage}
\begin{minipage}[]{.99\textwidth}
        \centering
        \footnotesize
    \includegraphics[trim=0 0 0 0, clip,width=0.23\textwidth]{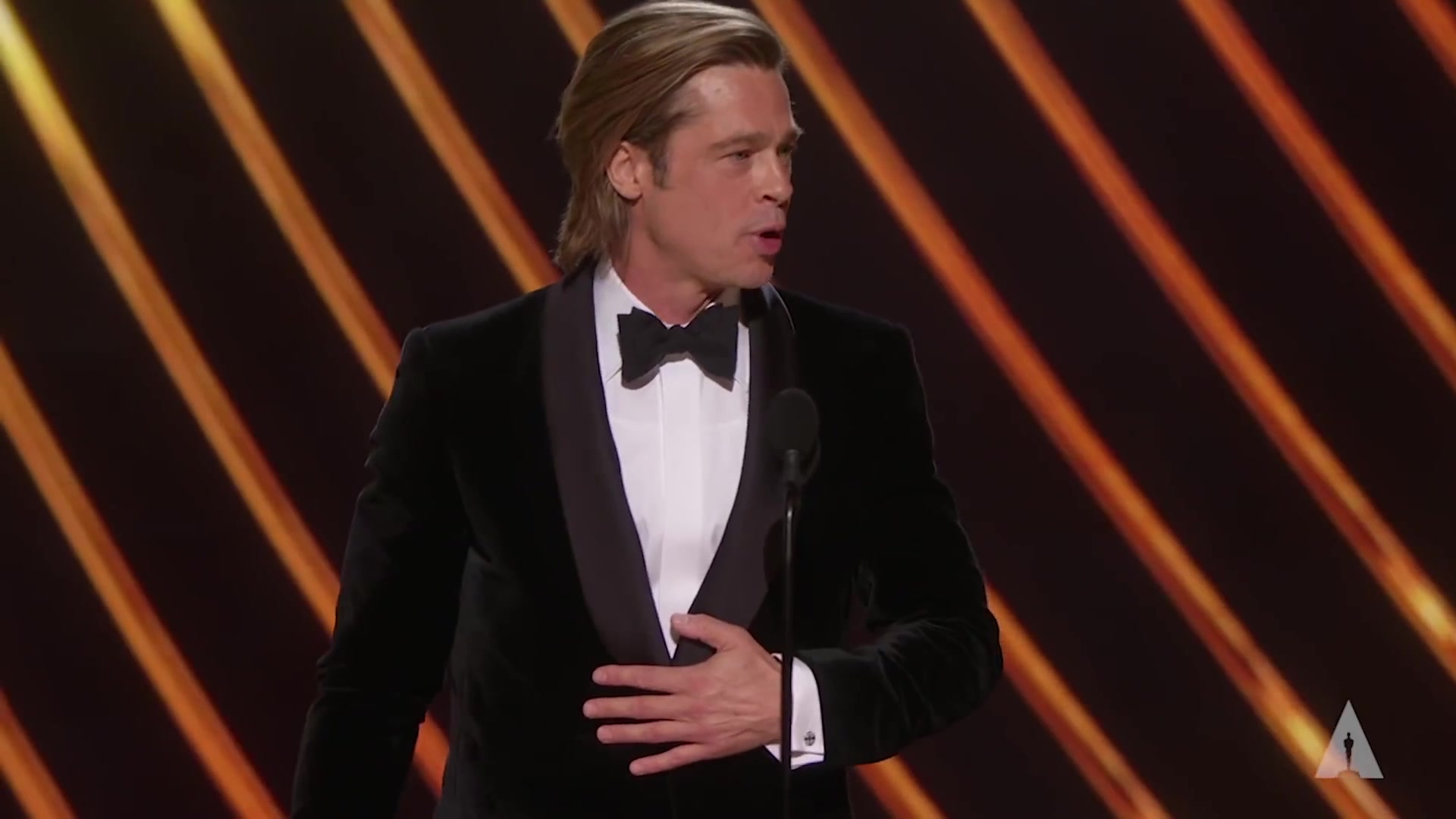}\hspace{0.2em}
    \includegraphics[trim=0 0 0 0, clip,width=0.23\textwidth]{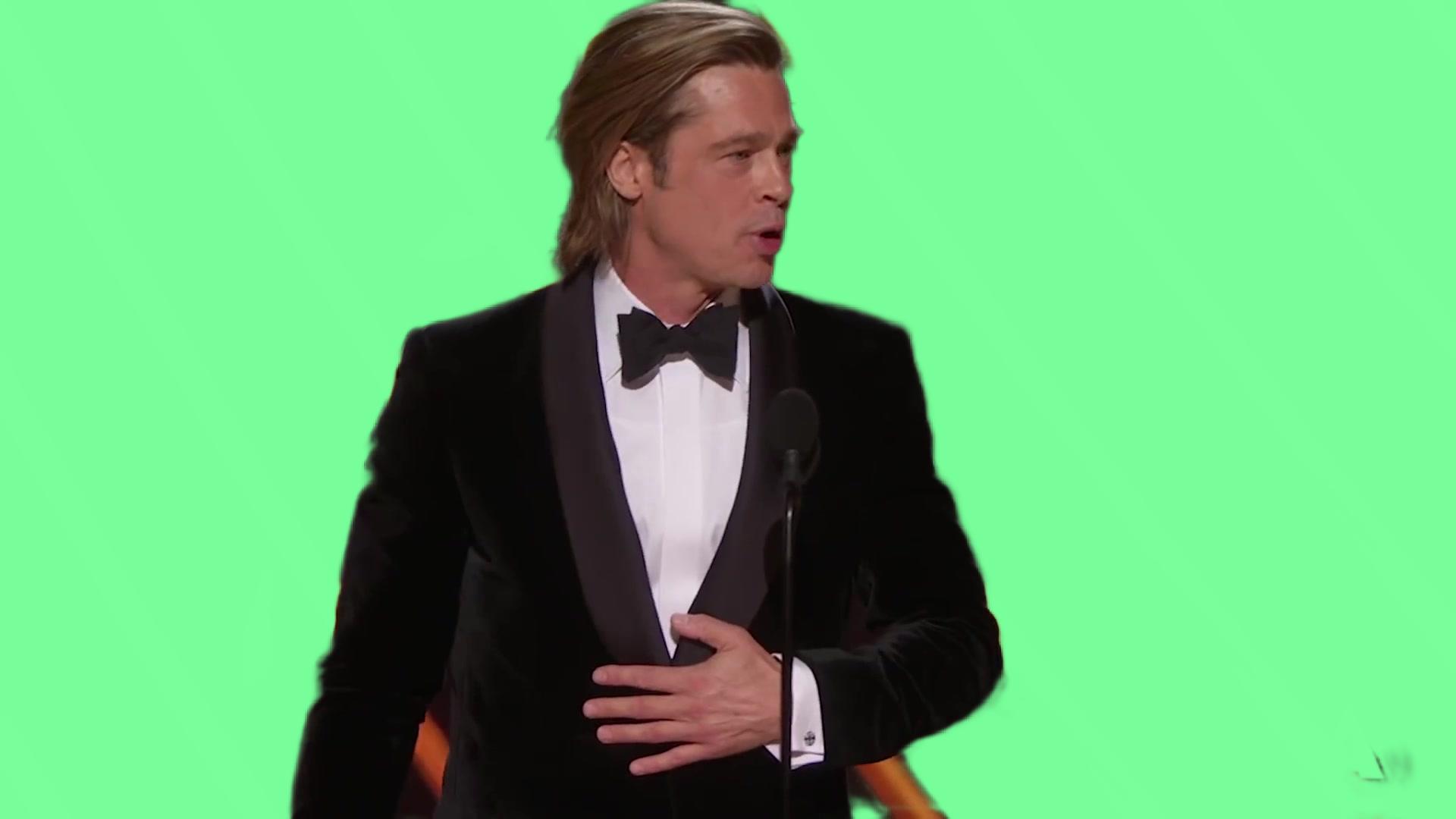}\hspace{0.2em}
    \includegraphics[trim=0 0 0 0, clip,width=0.23\textwidth]{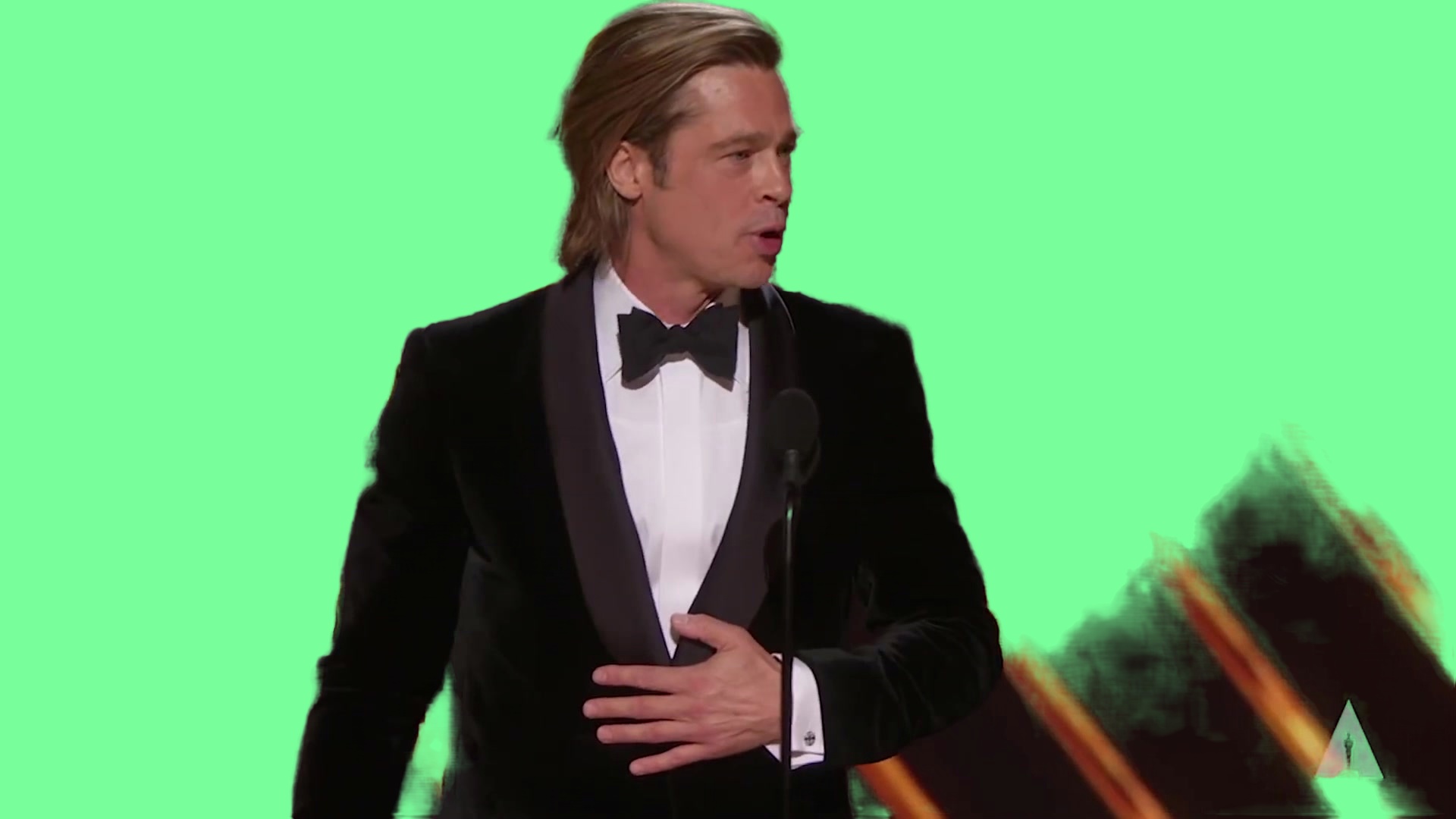}\hspace{0.2em}
    \includegraphics[trim=0 0 0 0, clip,width=0.23\textwidth]{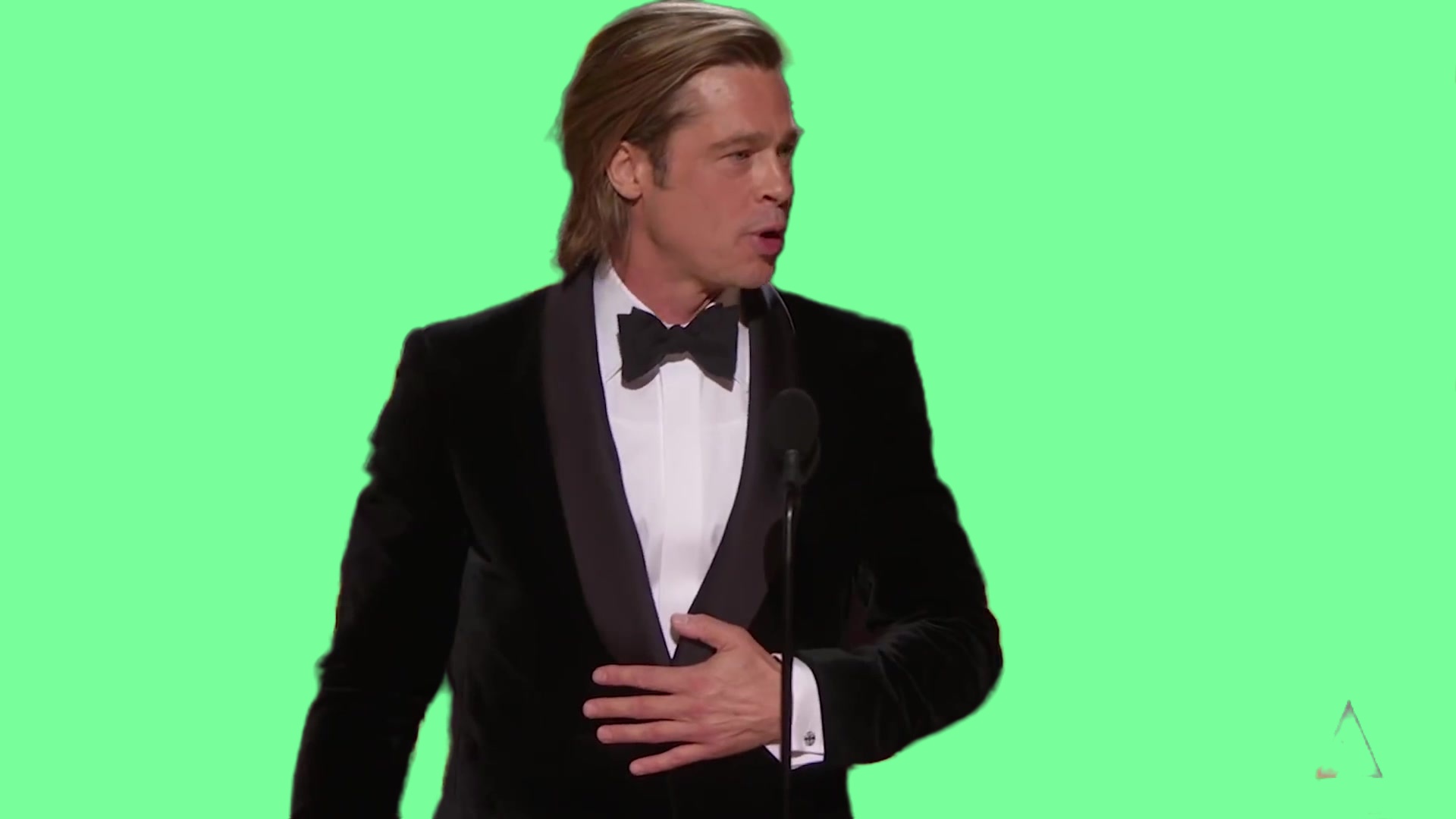}\hspace{0.2em}
    \vspace{2pt}
    \end{minipage}
\begin{minipage}[]{.99\textwidth}
        \centering
        \footnotesize
    \includegraphics[trim=0 0 0 0, clip,width=0.23\textwidth]{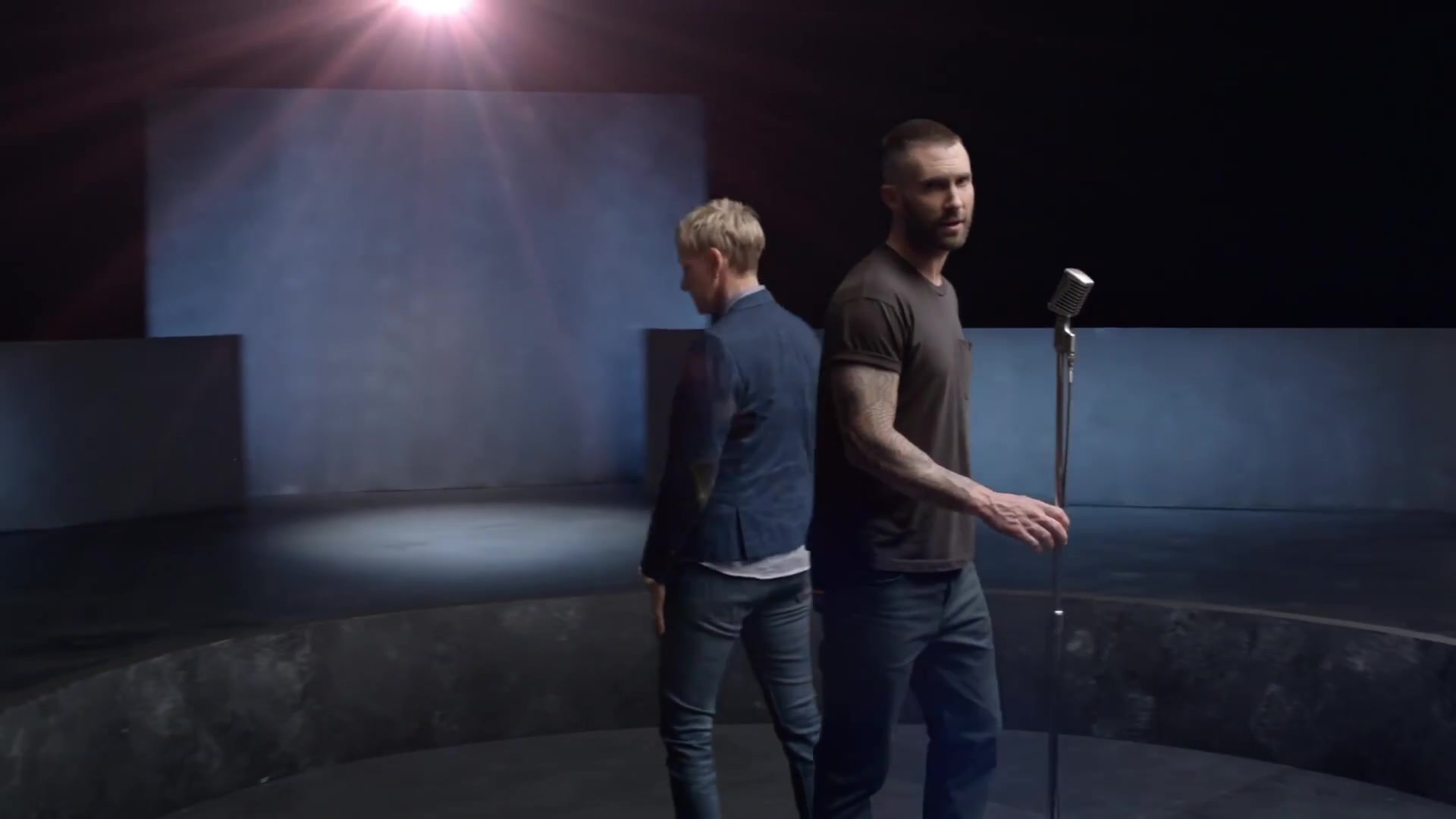}\hspace{0.2em}
    \includegraphics[trim=0 0 0 0, clip,width=0.23\textwidth]{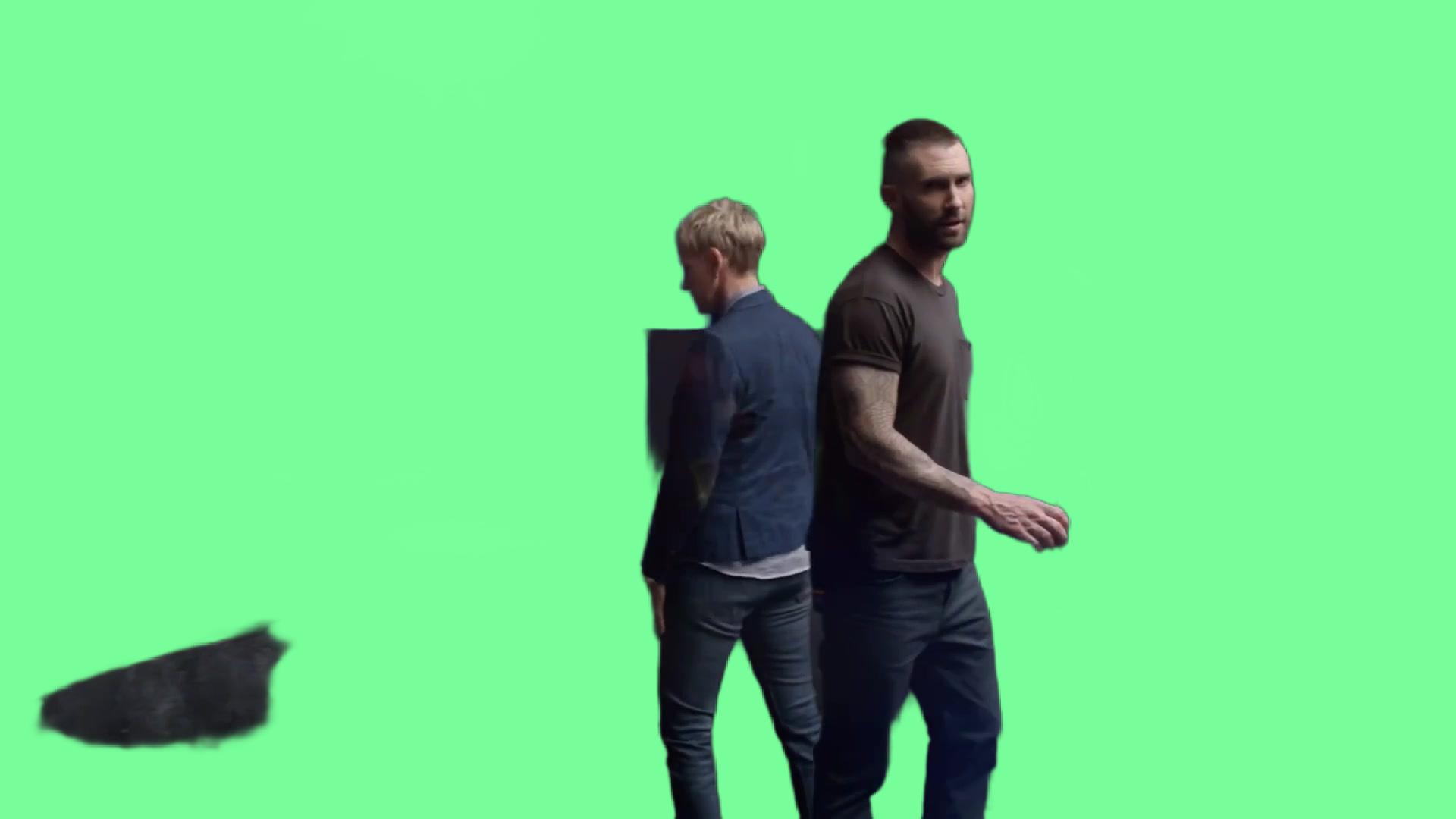}\hspace{0.2em}
    \includegraphics[trim=0 0 0 0, clip,width=0.23\textwidth]{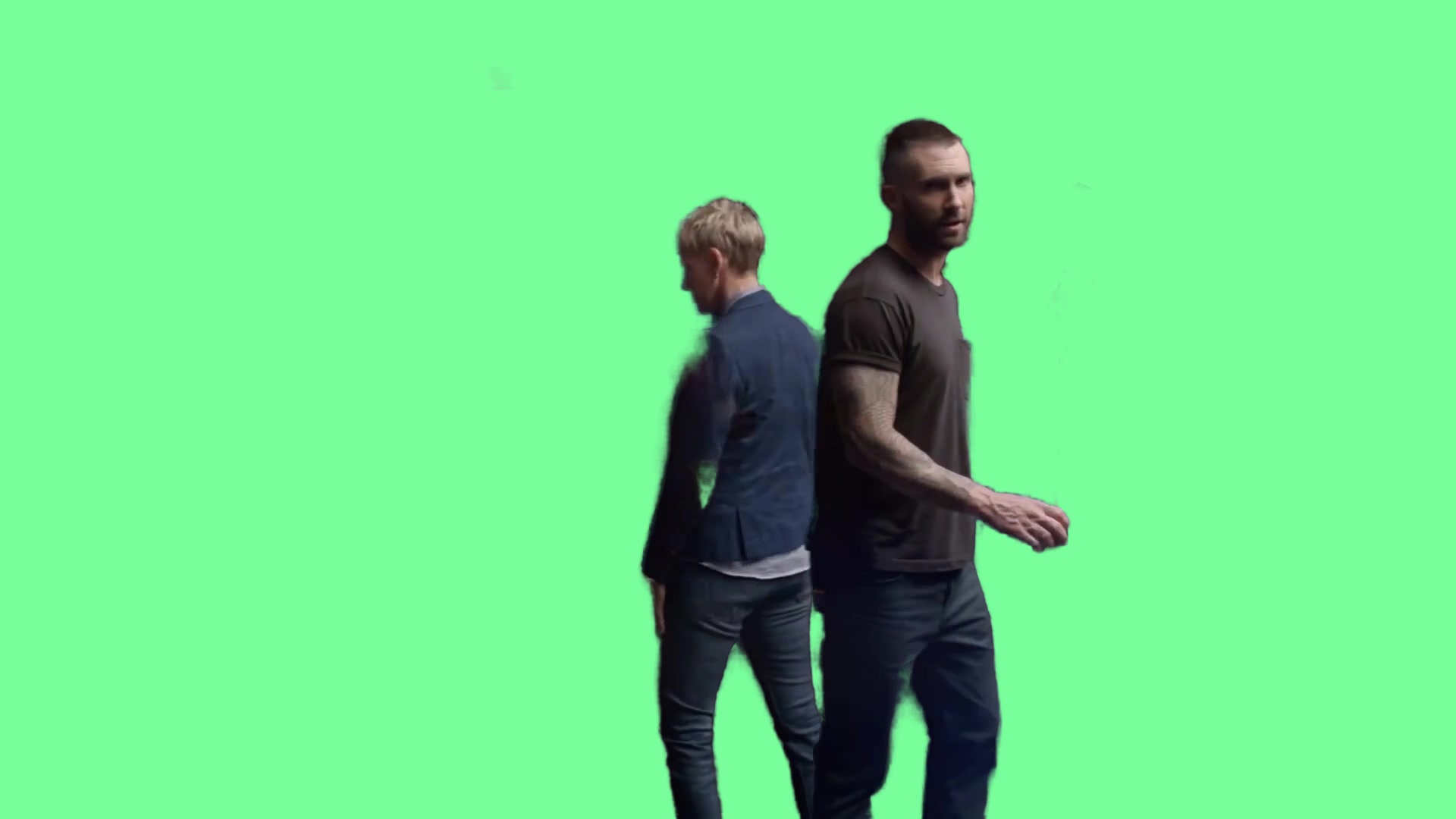}\hspace{0.2em}
    \includegraphics[trim=0 0 0 0, clip,width=0.23\textwidth]{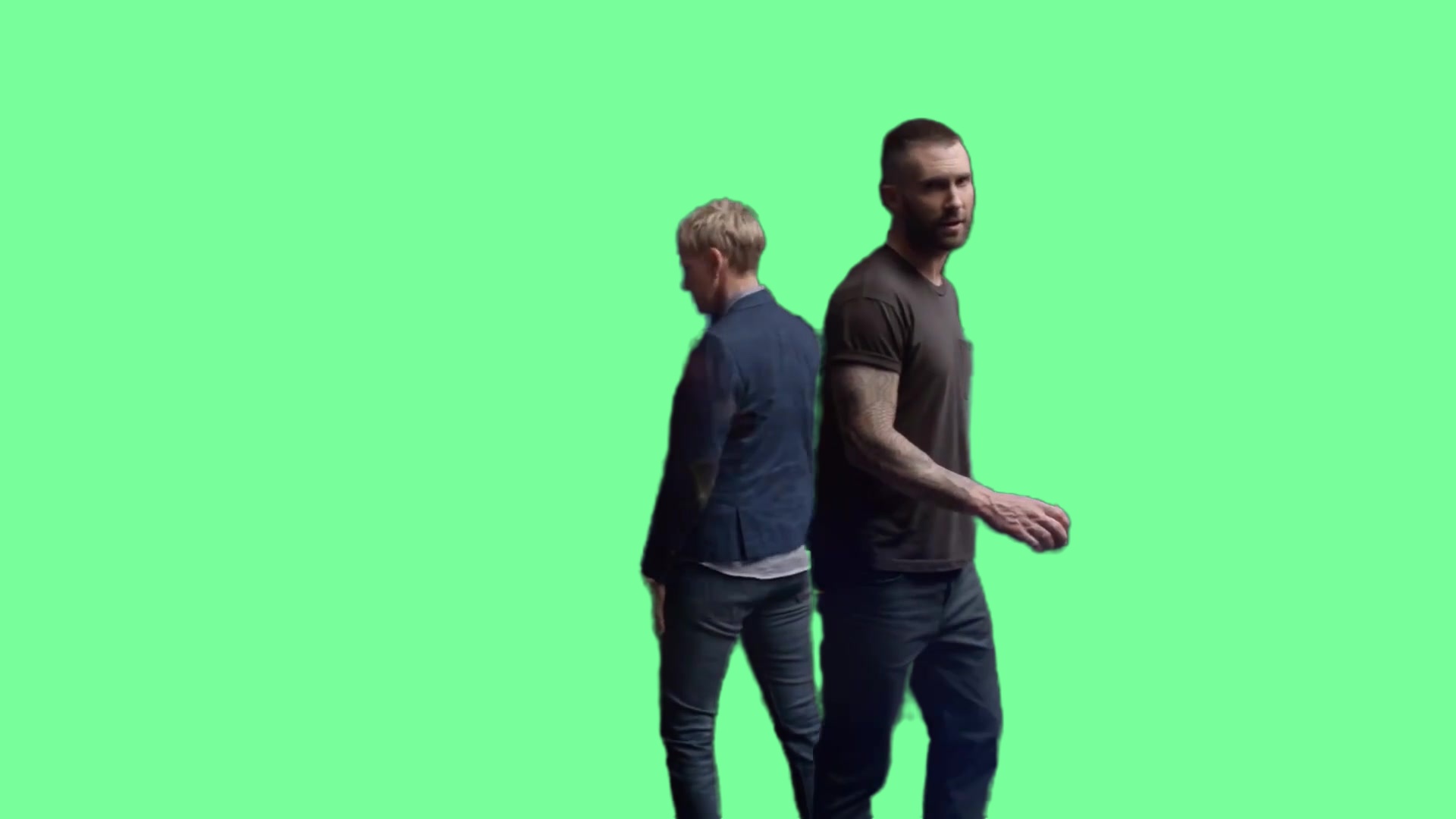}\hspace{0.2em}
    \vspace{2pt}
    \end{minipage}
\begin{minipage}[]{.99\textwidth}
        \centering
        \footnotesize
    \includegraphics[trim=0 0 0 0, clip,width=0.23\textwidth]{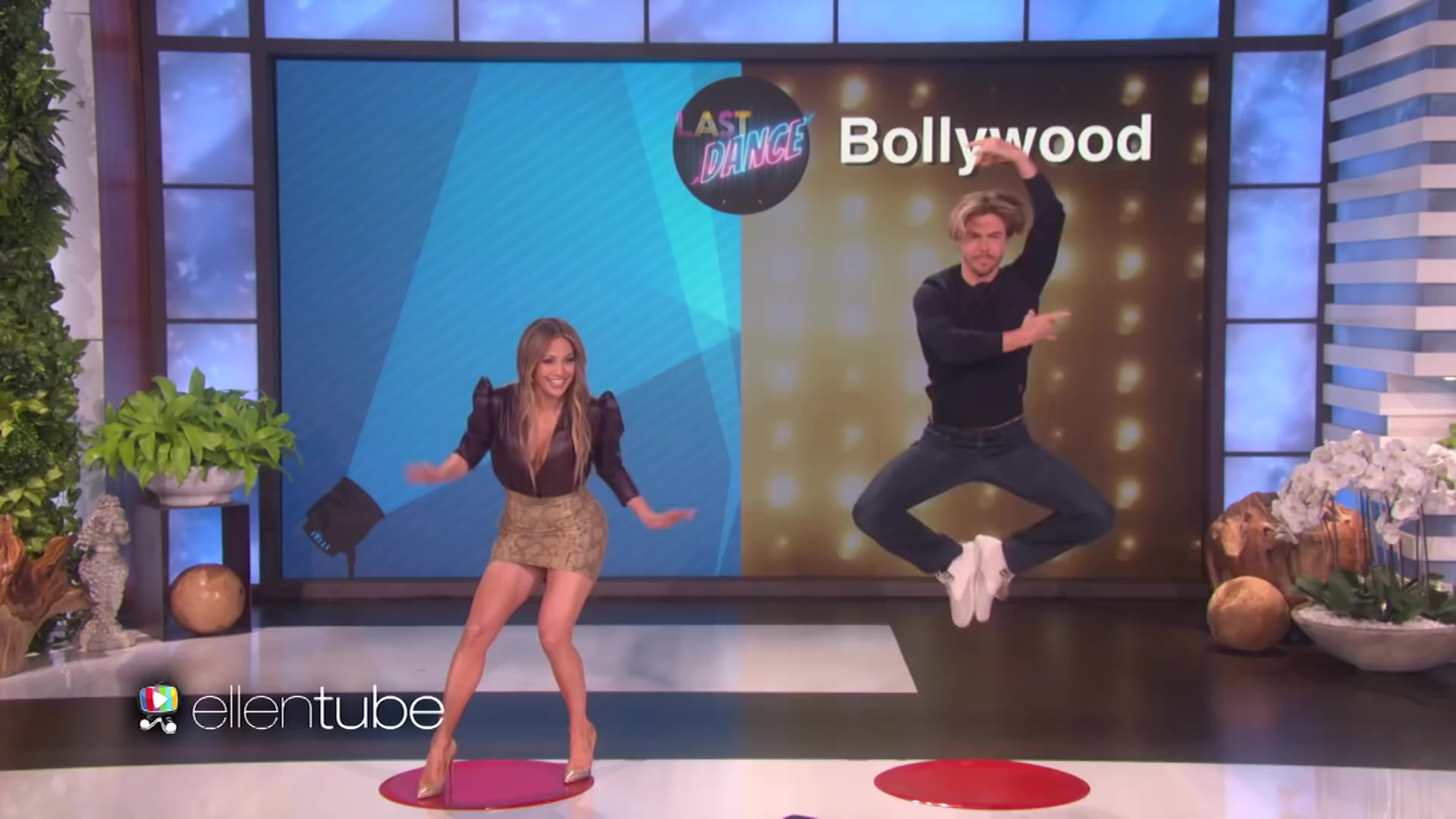}\hspace{0.2em}
    \includegraphics[trim=0 0 0 0, clip,width=0.23\textwidth]{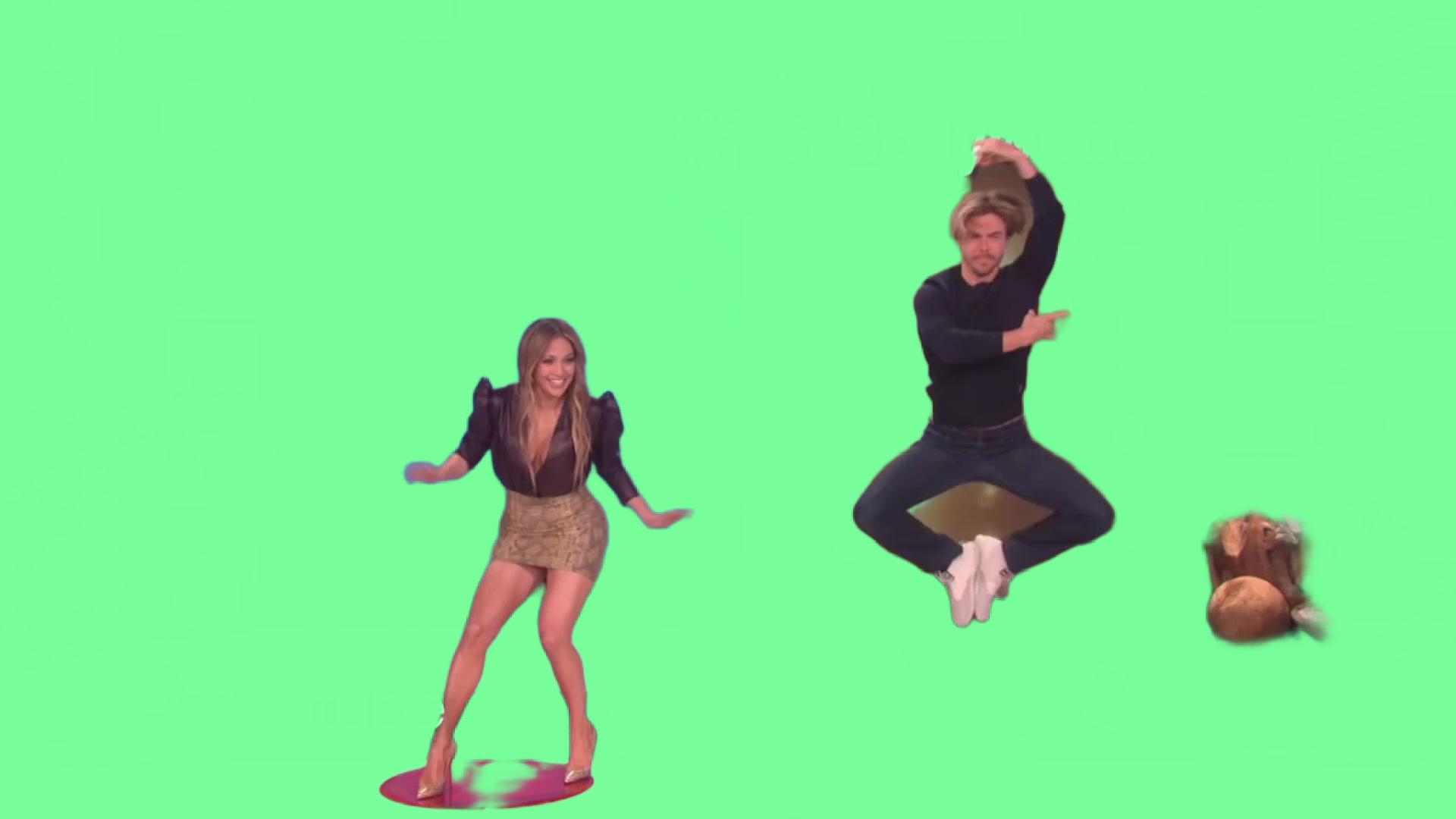}\hspace{0.2em}
    \includegraphics[trim=0 0 0 0, clip,width=0.23\textwidth]{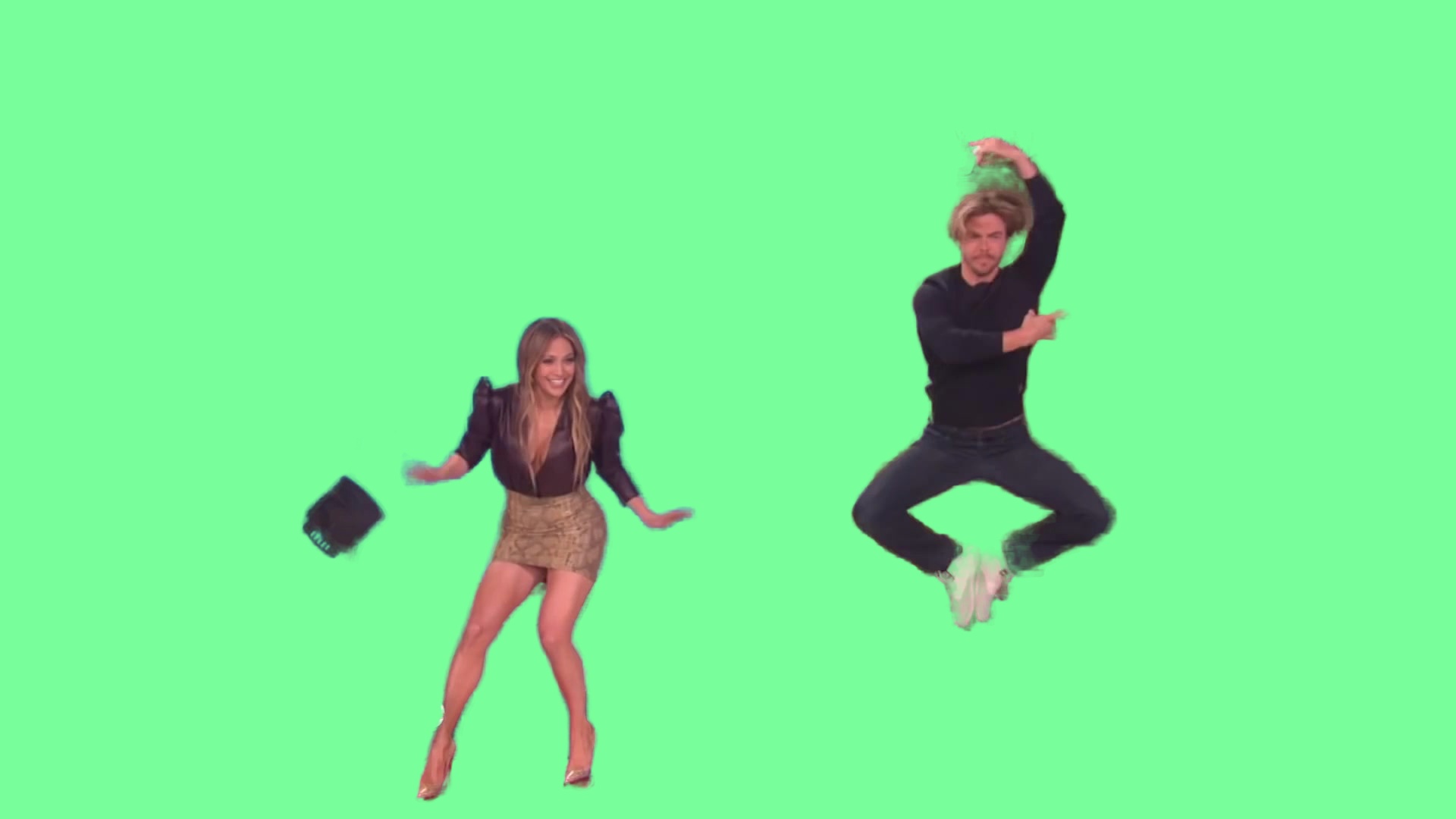}\hspace{0.2em}
    \includegraphics[trim=0 0 0 0, clip,width=0.23\textwidth]{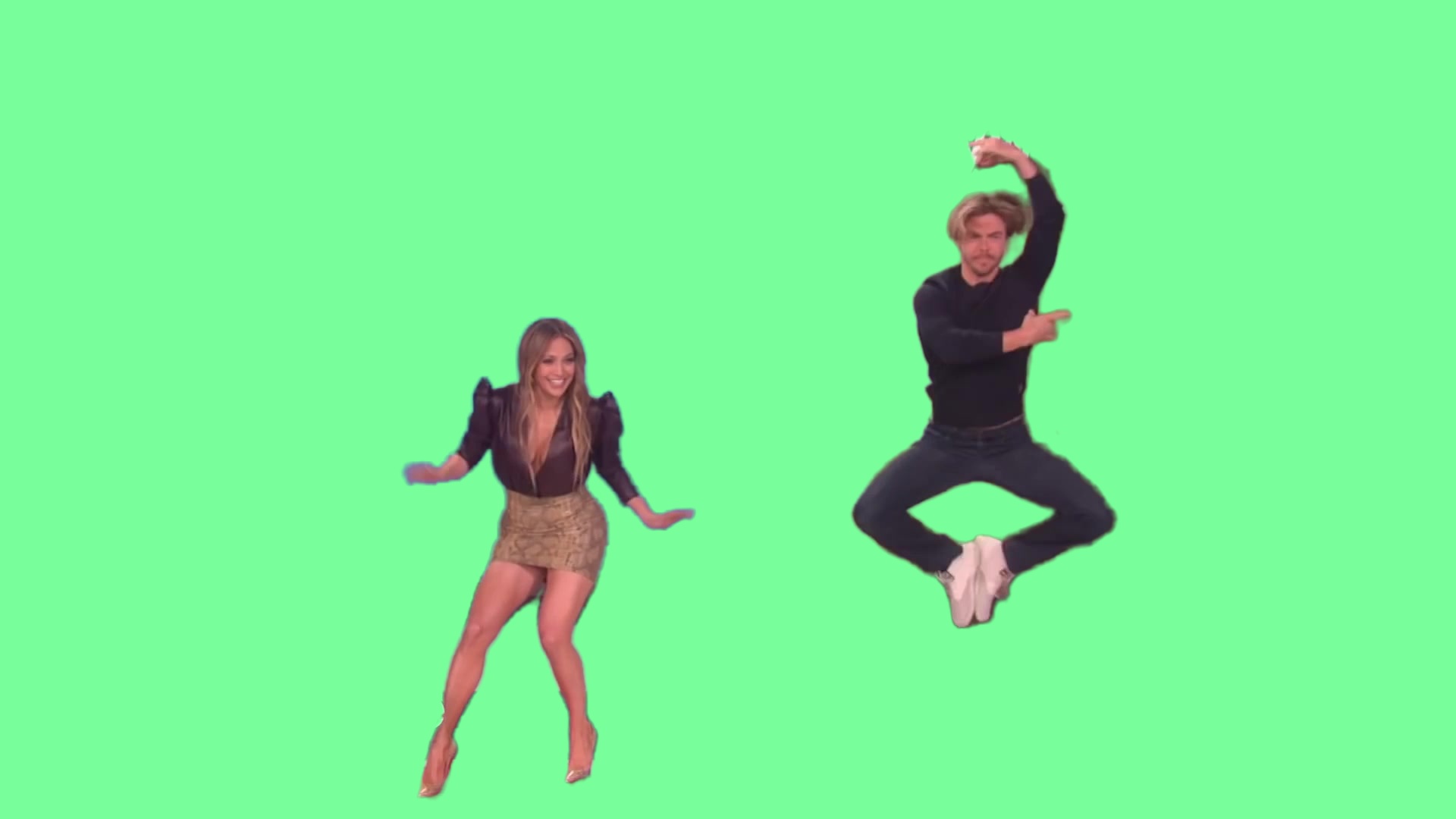}\hspace{0.2em}
    \vspace{2pt}
    \end{minipage} 
\begin{minipage}[]{.99\textwidth}
        \centering
        \footnotesize
    \includegraphics[trim=0 0 0 0, clip,width=0.23\textwidth]{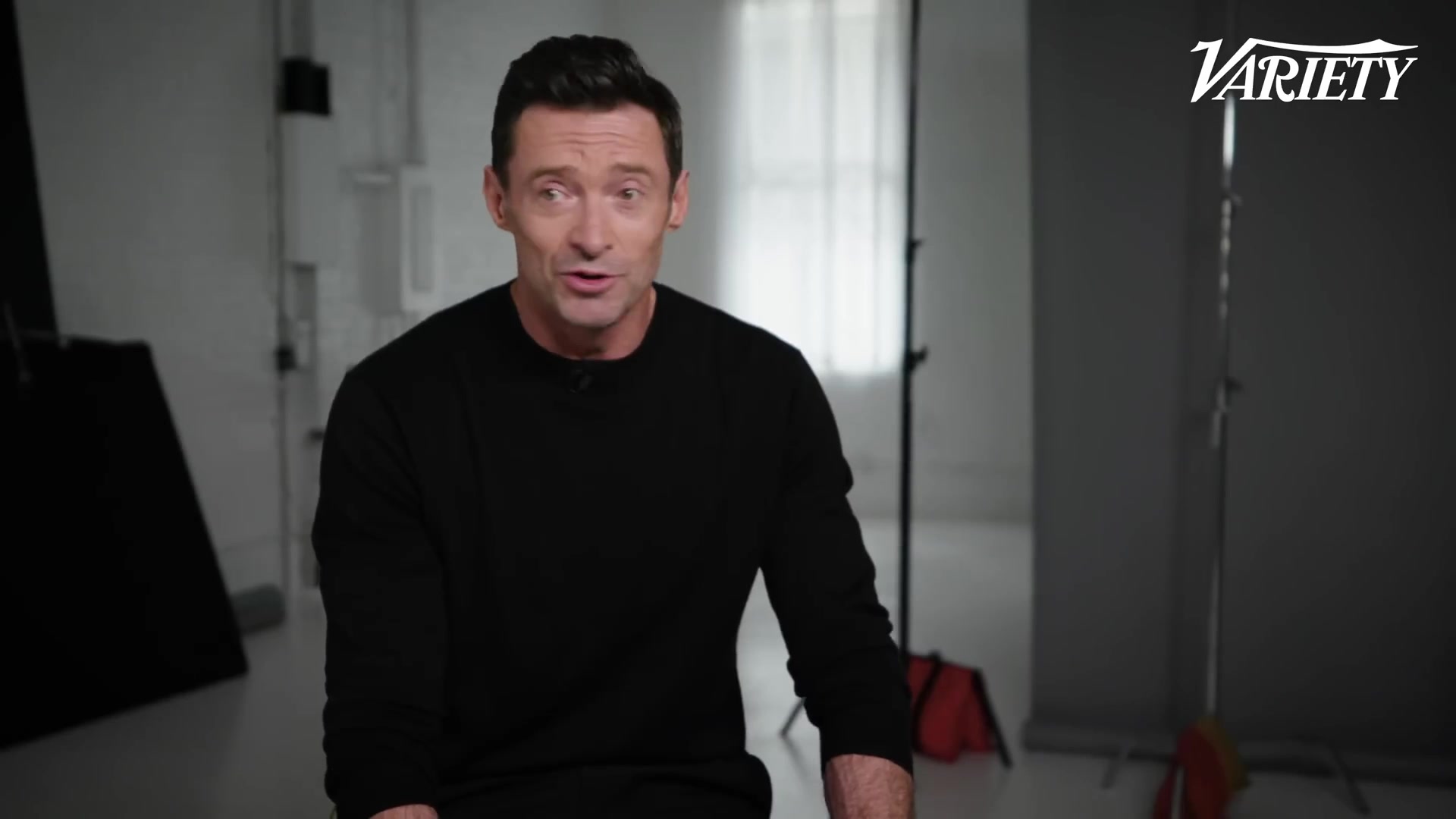}\hspace{0.2em}
    \includegraphics[trim=0 0 0 0, clip,width=0.23\textwidth]{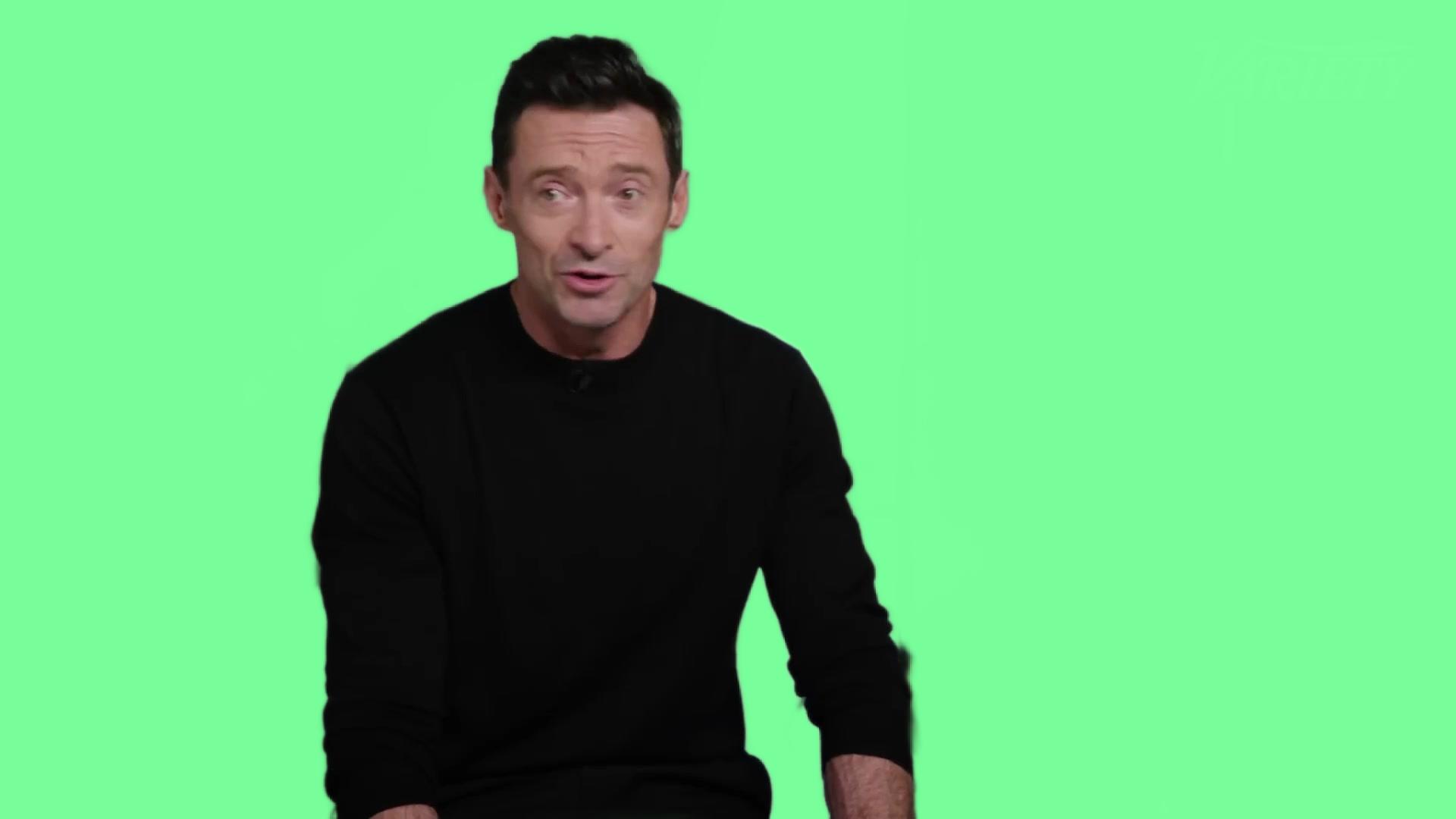}\hspace{0.2em}
    \includegraphics[trim=0 0 0 0, clip,width=0.23\textwidth]{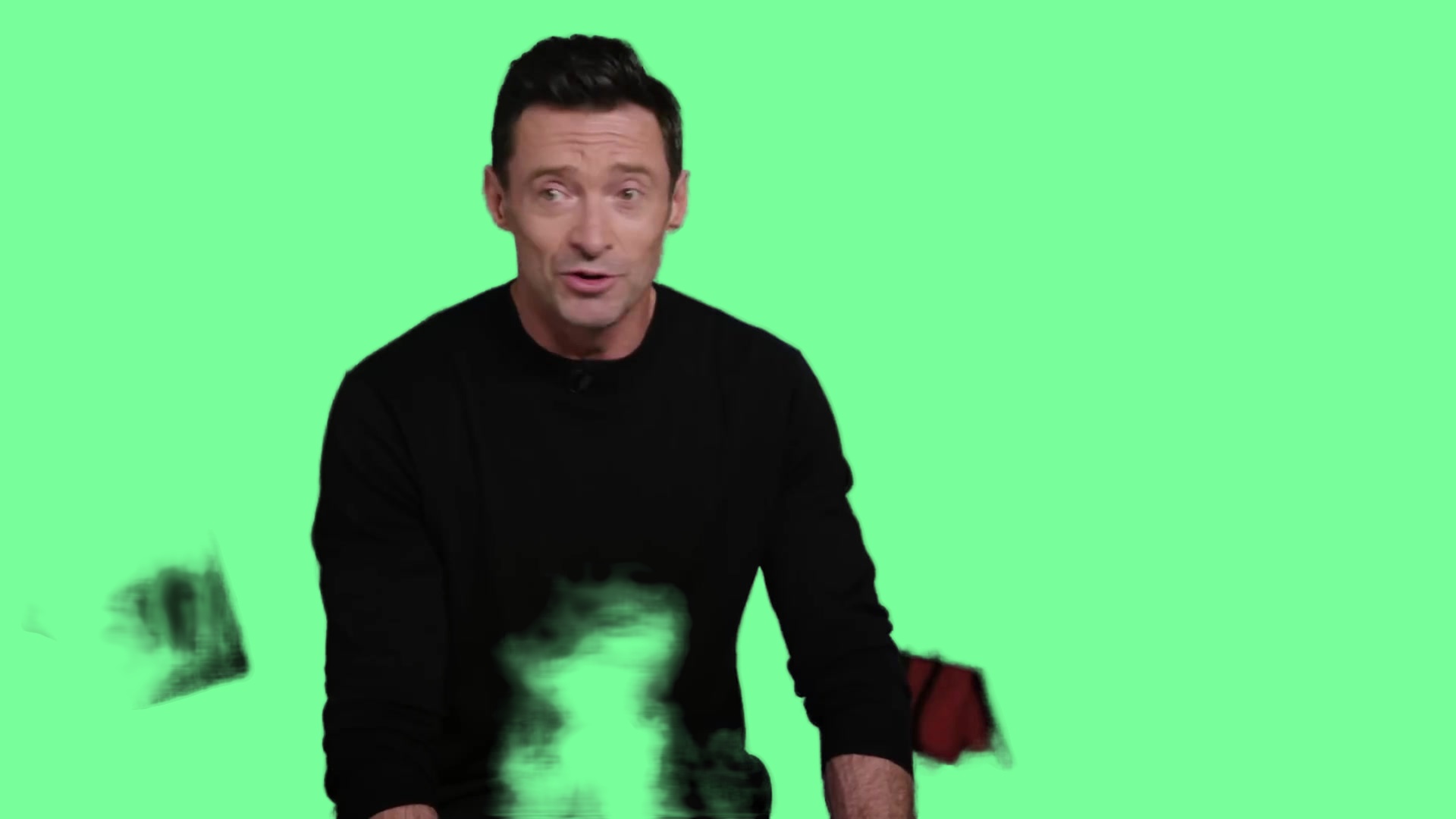}\hspace{0.2em}
    \includegraphics[trim=0 0 0 0, clip,width=0.23\textwidth]{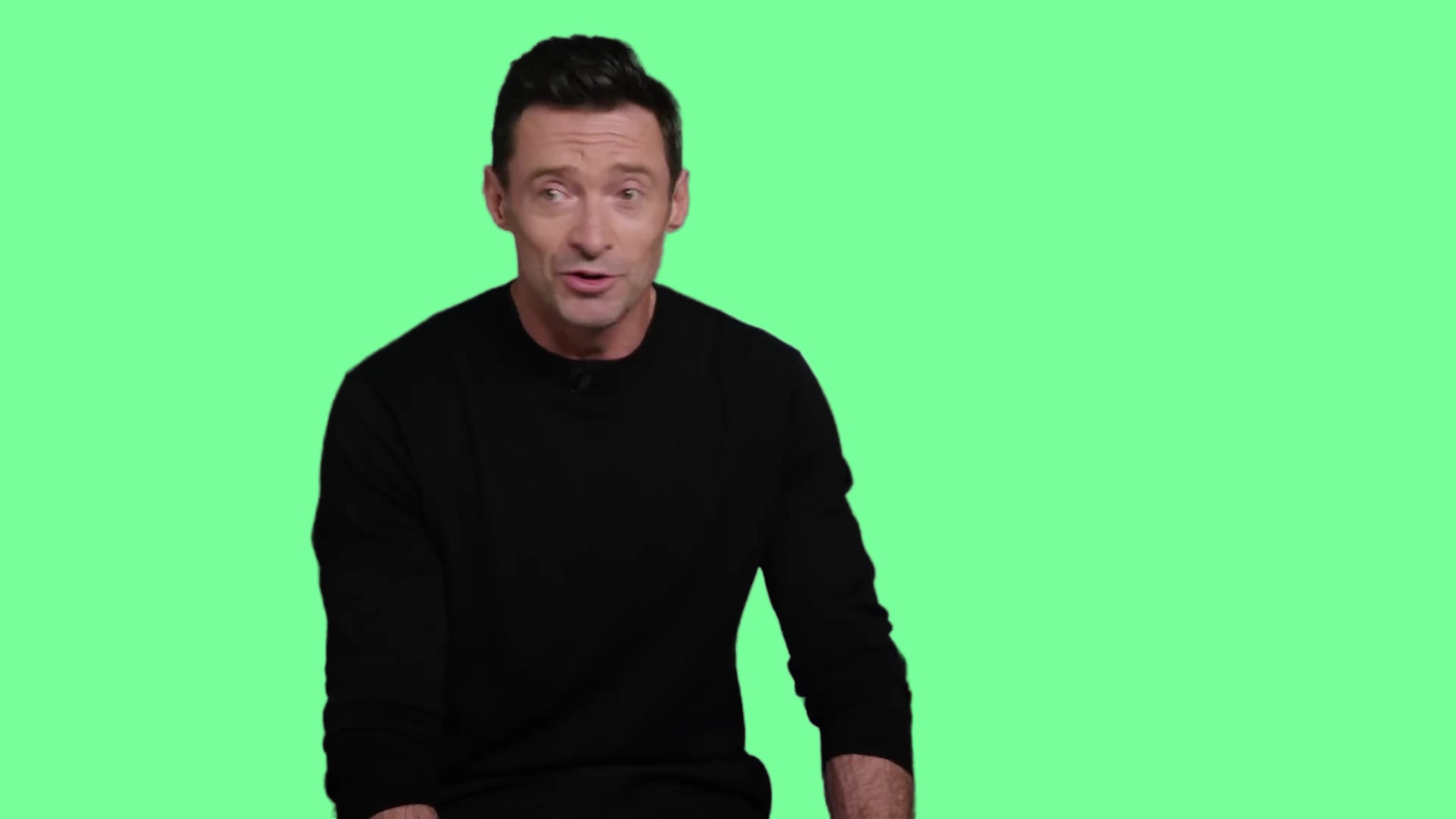}\hspace{0.2em}
    \vspace{2pt}
    \end{minipage} 
\begin{minipage}[]{.99\textwidth}
        \centering
        \footnotesize
    \includegraphics[trim=0 0 0 0, clip,width=0.23\textwidth]{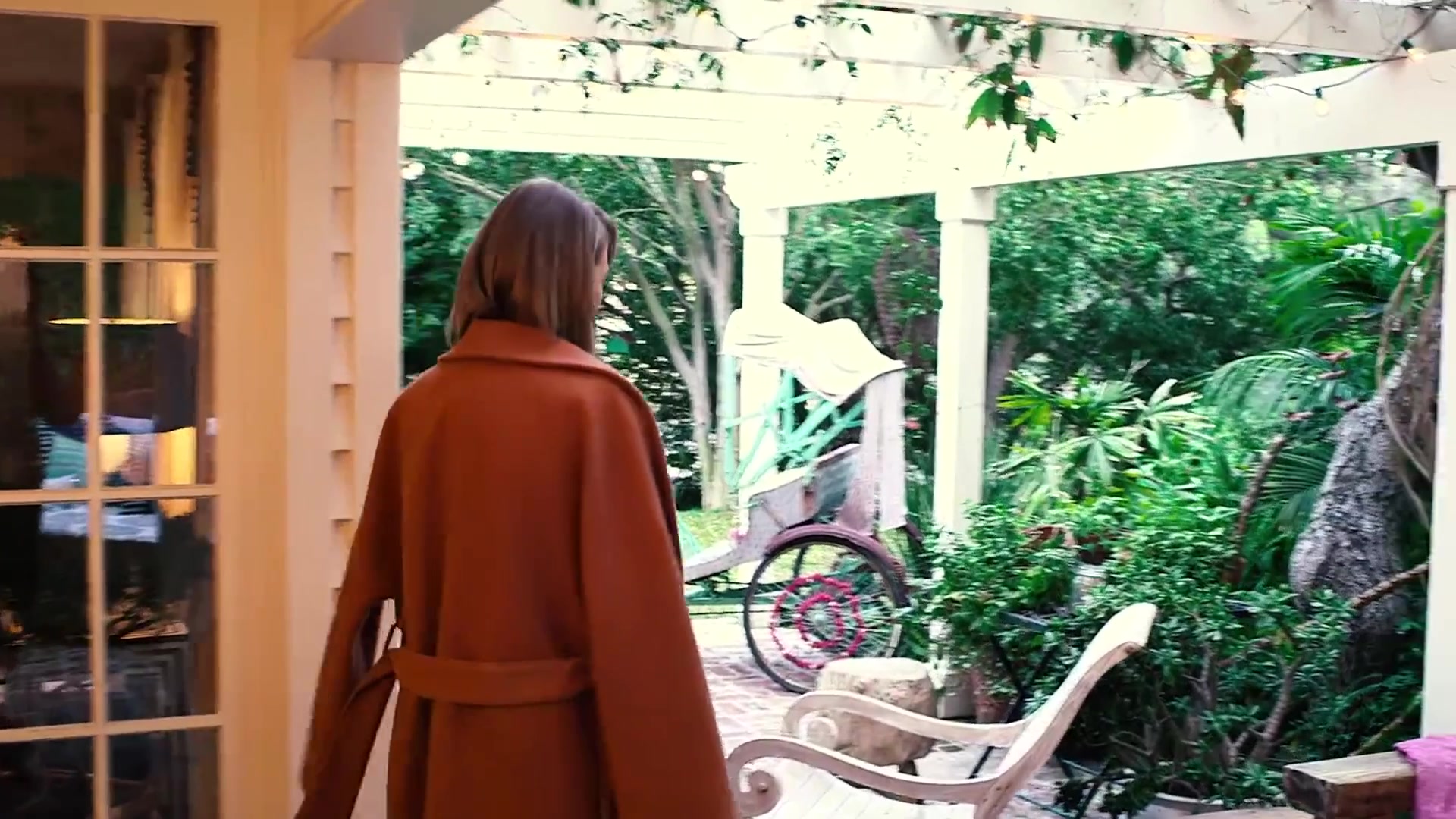}\hspace{0.2em}
    \includegraphics[trim=0 0 0 0, clip,width=0.23\textwidth]{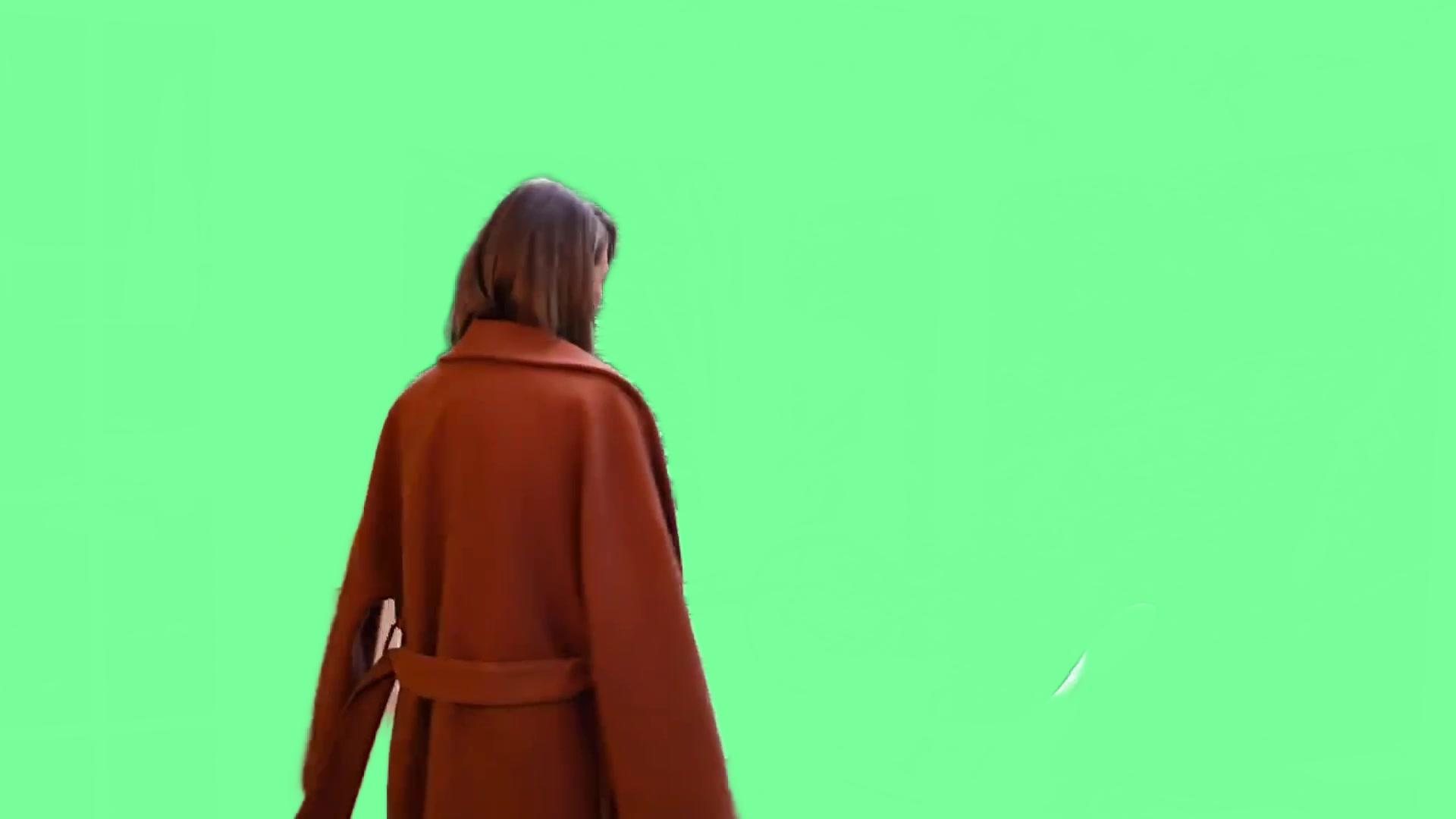}\hspace{0.2em}
    \includegraphics[trim=0 0 0 0, clip,width=0.23\textwidth]{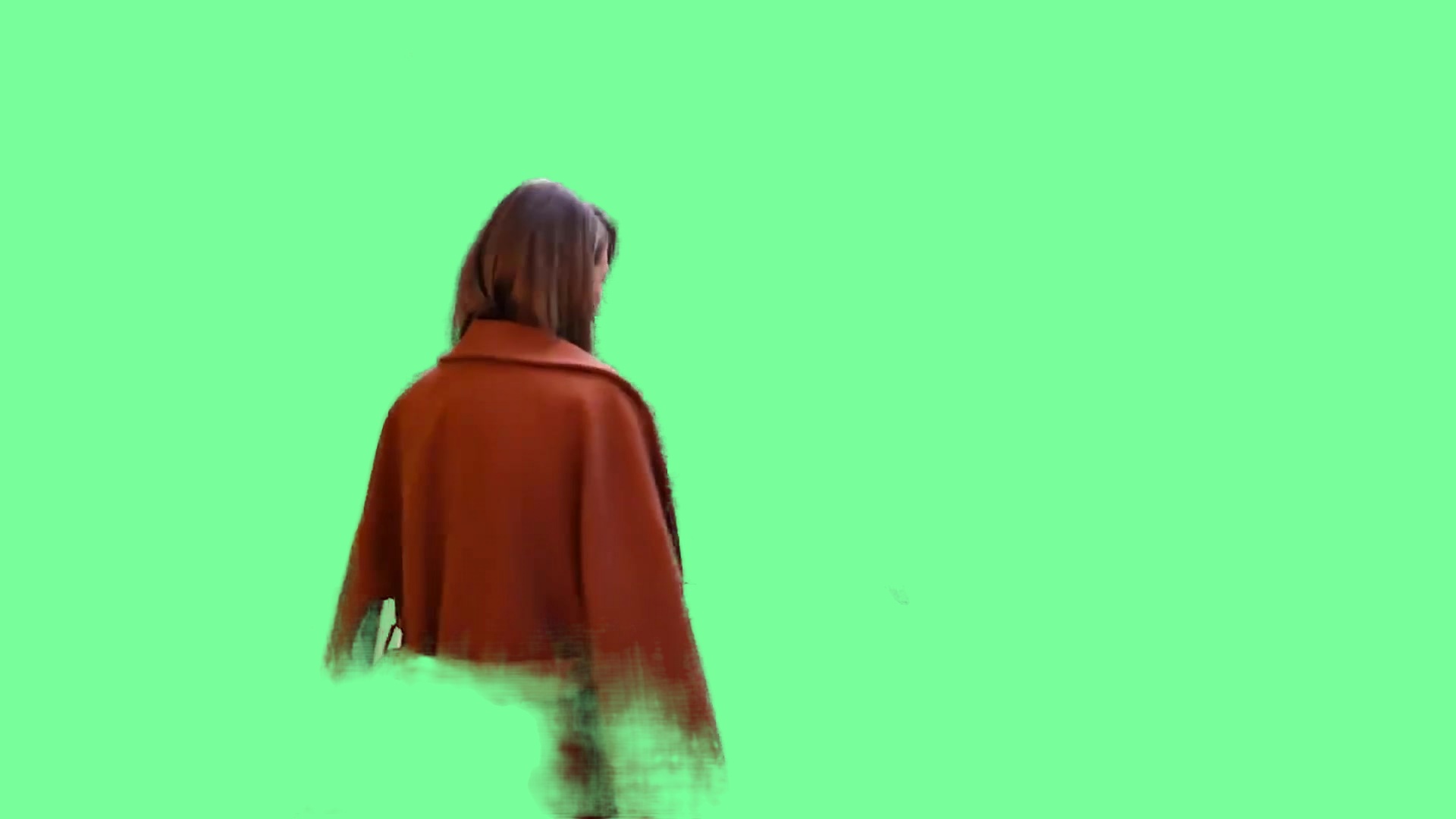}\hspace{0.2em}
    \includegraphics[trim=0 0 0 0, clip,width=0.23\textwidth]{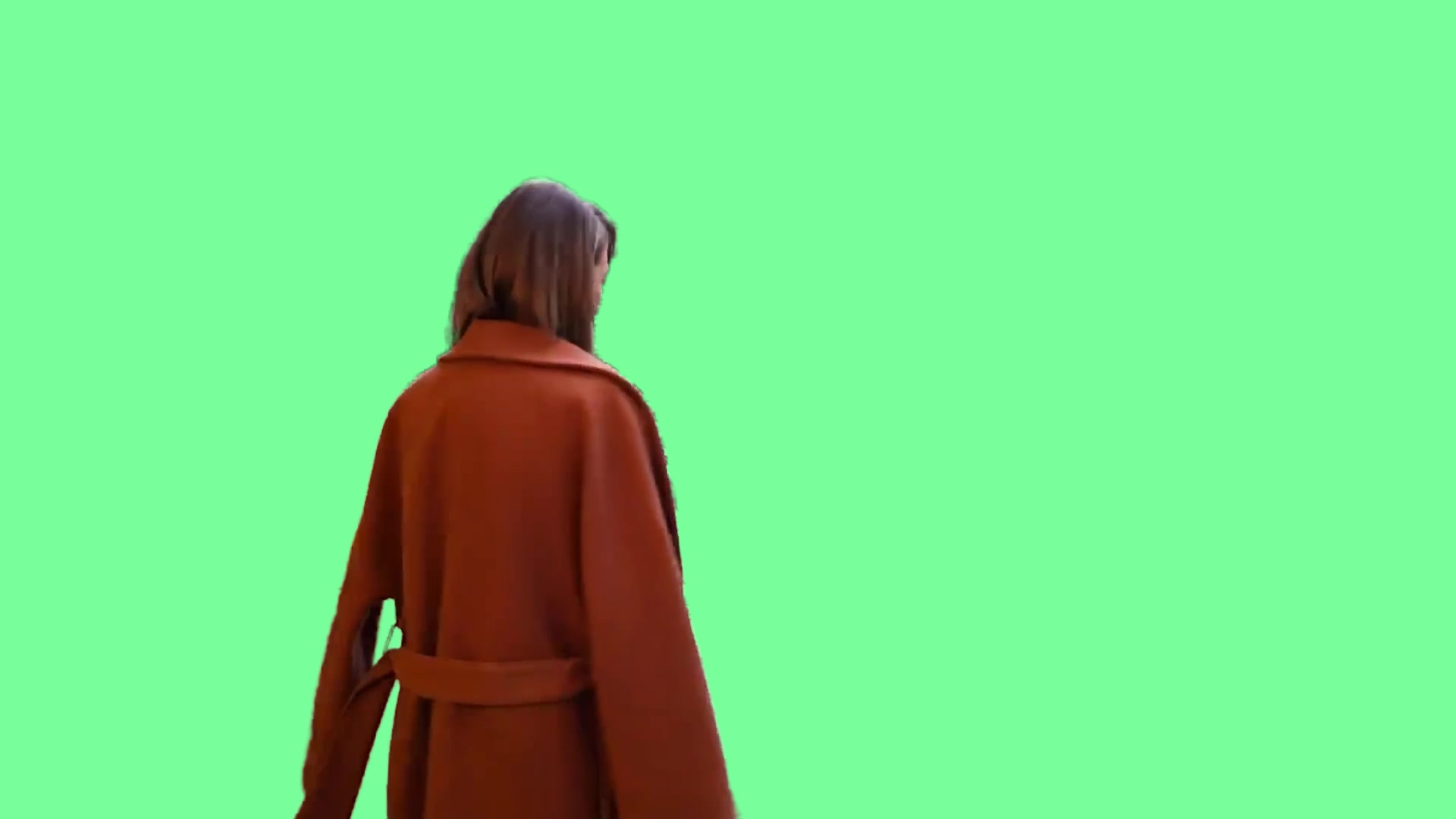}\hspace{0.2em}
    \vspace{2pt}
    \end{minipage} 
\begin{minipage}[]{.99\textwidth}
        \centering
        \footnotesize
    \includegraphics[trim=0 0 0 0, clip,width=0.23\textwidth]{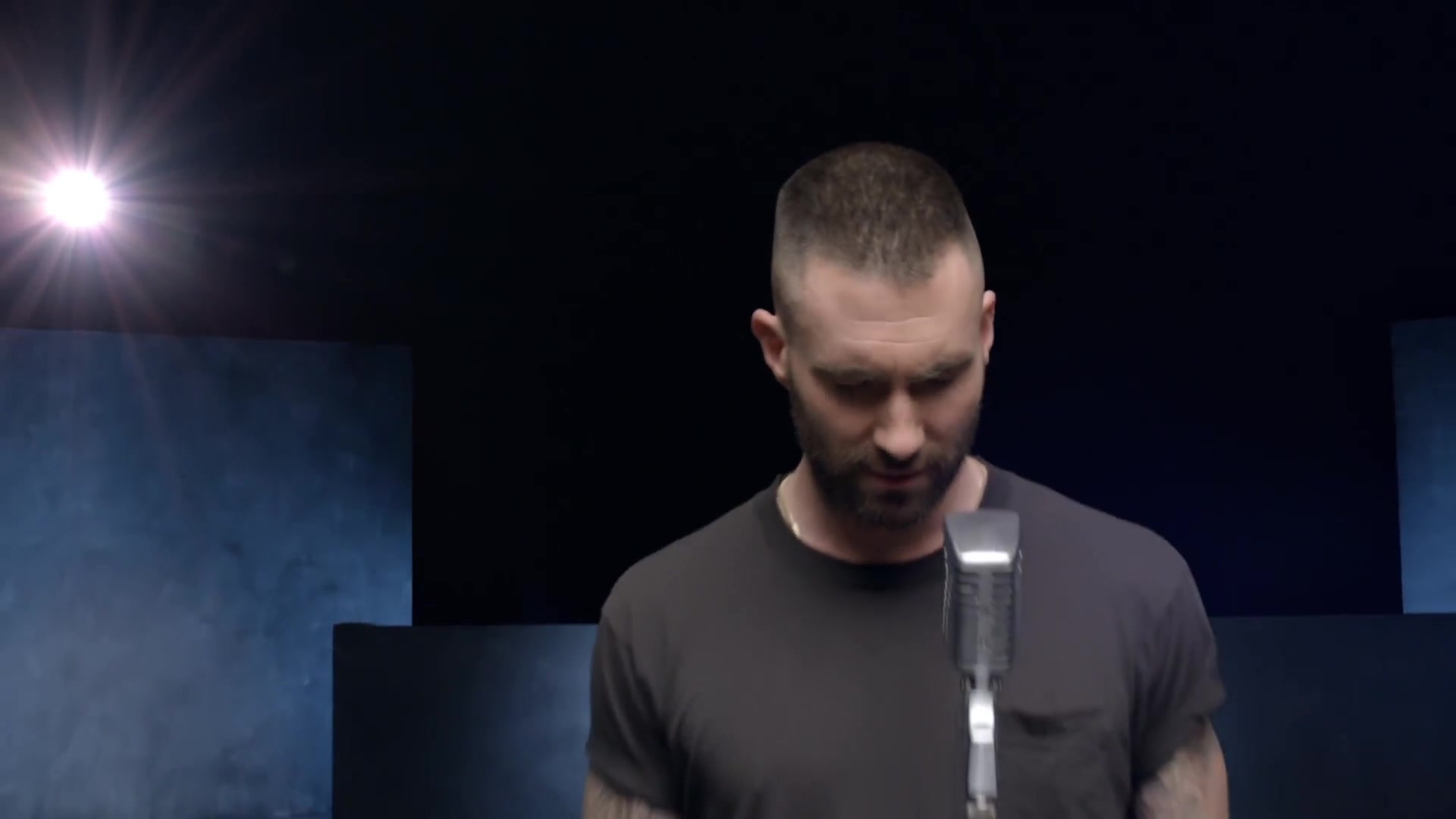}\hspace{0.2em}
    \includegraphics[trim=0 0 0 0, clip,width=0.23\textwidth]{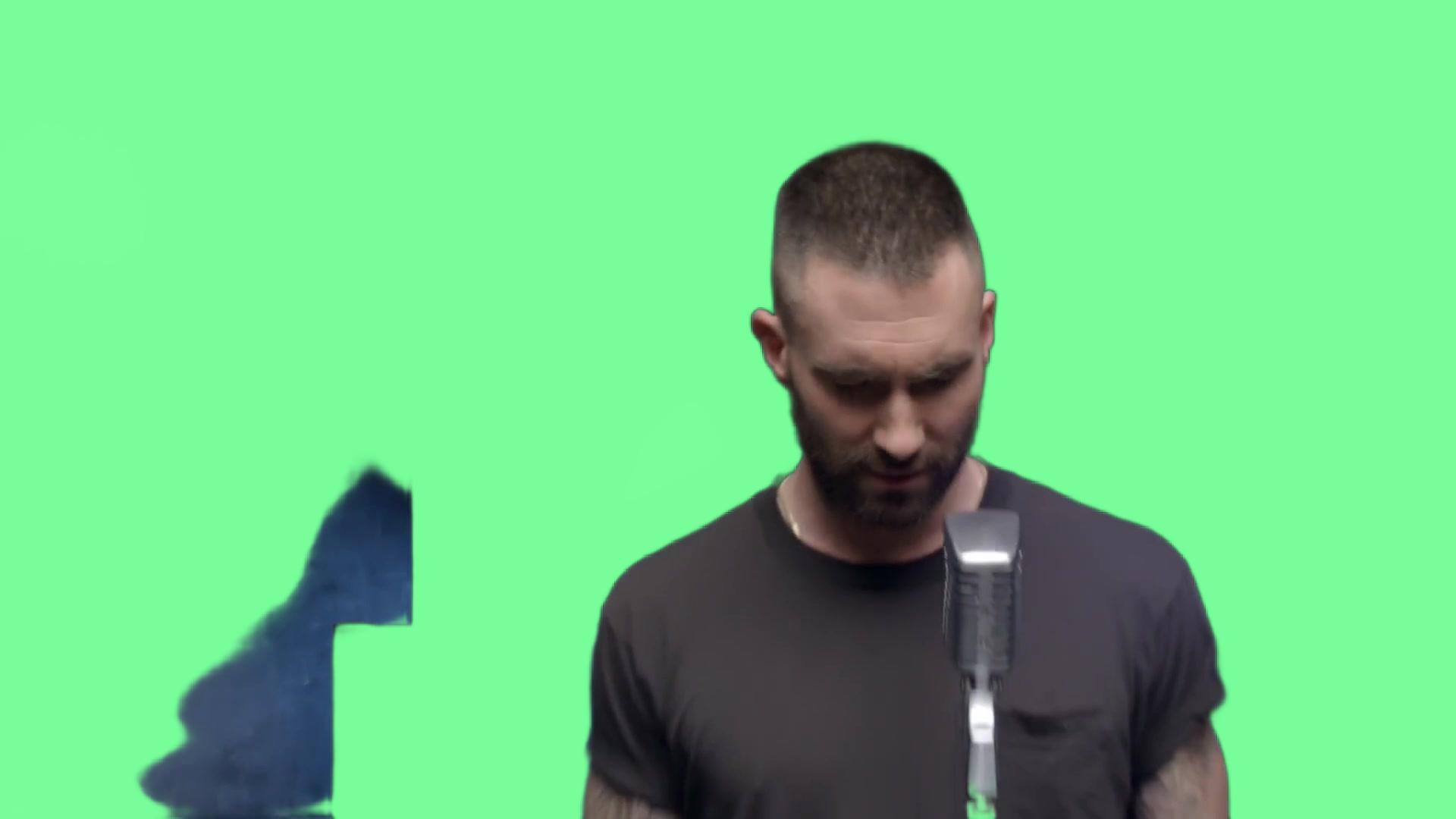}\hspace{0.2em}
    \includegraphics[trim=0 0 0 0, clip,width=0.23\textwidth]{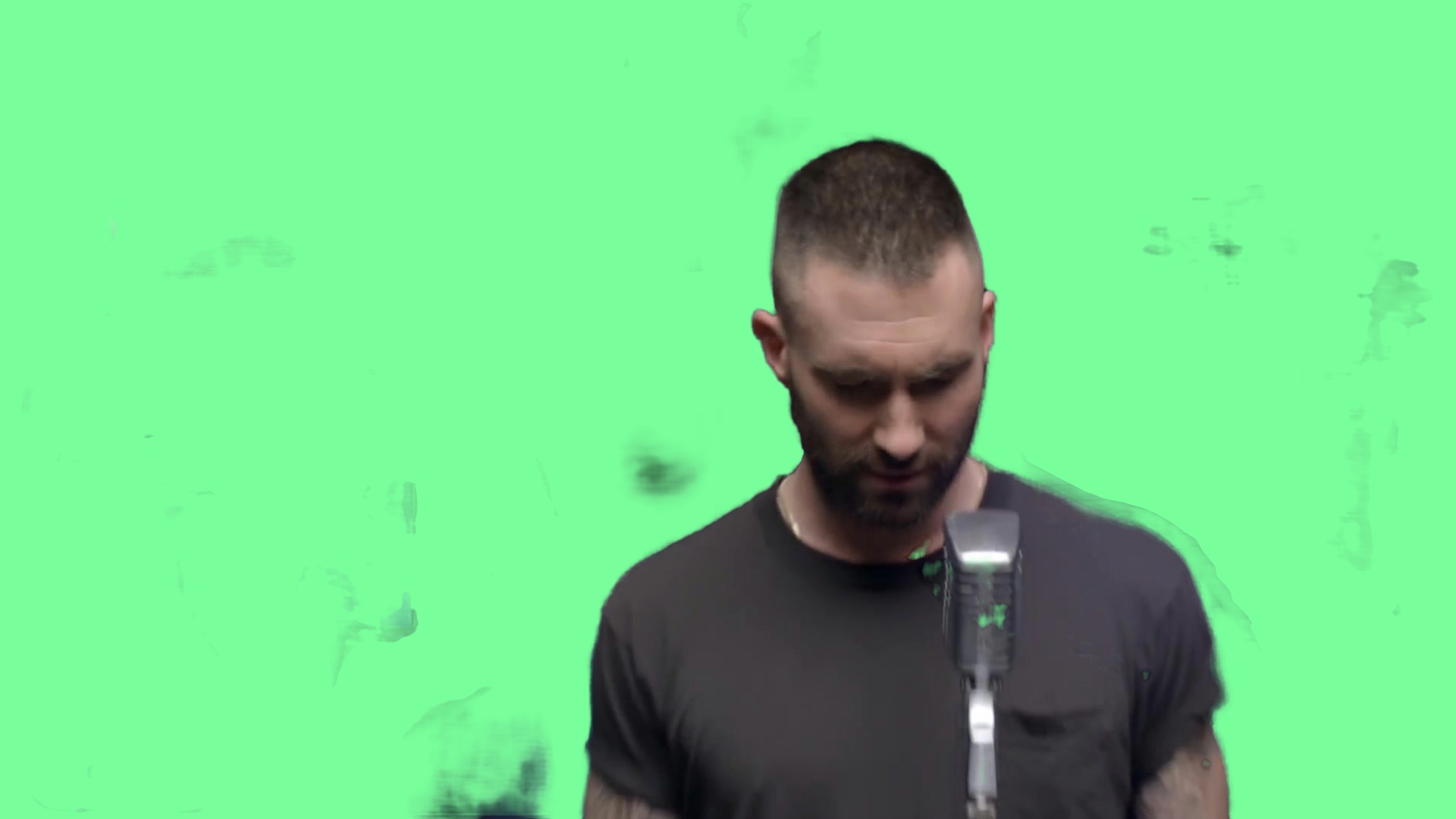}\hspace{0.2em}
    \includegraphics[trim=0 0 0 0, clip,width=0.23\textwidth]{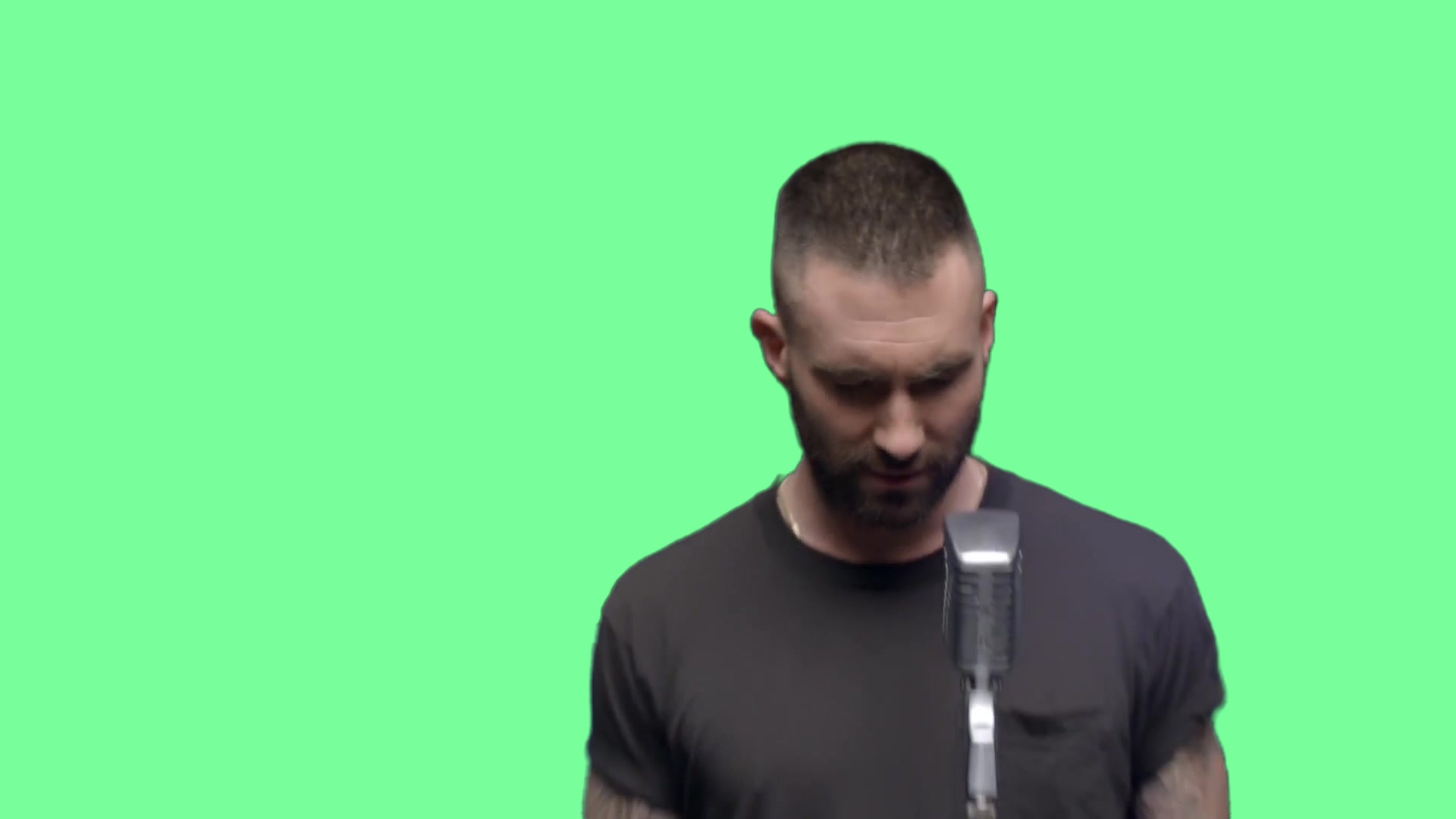}\hspace{0.2em}
    \vspace{2pt}
    \end{minipage}

\begin{minipage}[]{.99\textwidth}
        \centering
        \footnotesize
    \includegraphics[trim=0 0 0 0, clip,width=0.23\textwidth]{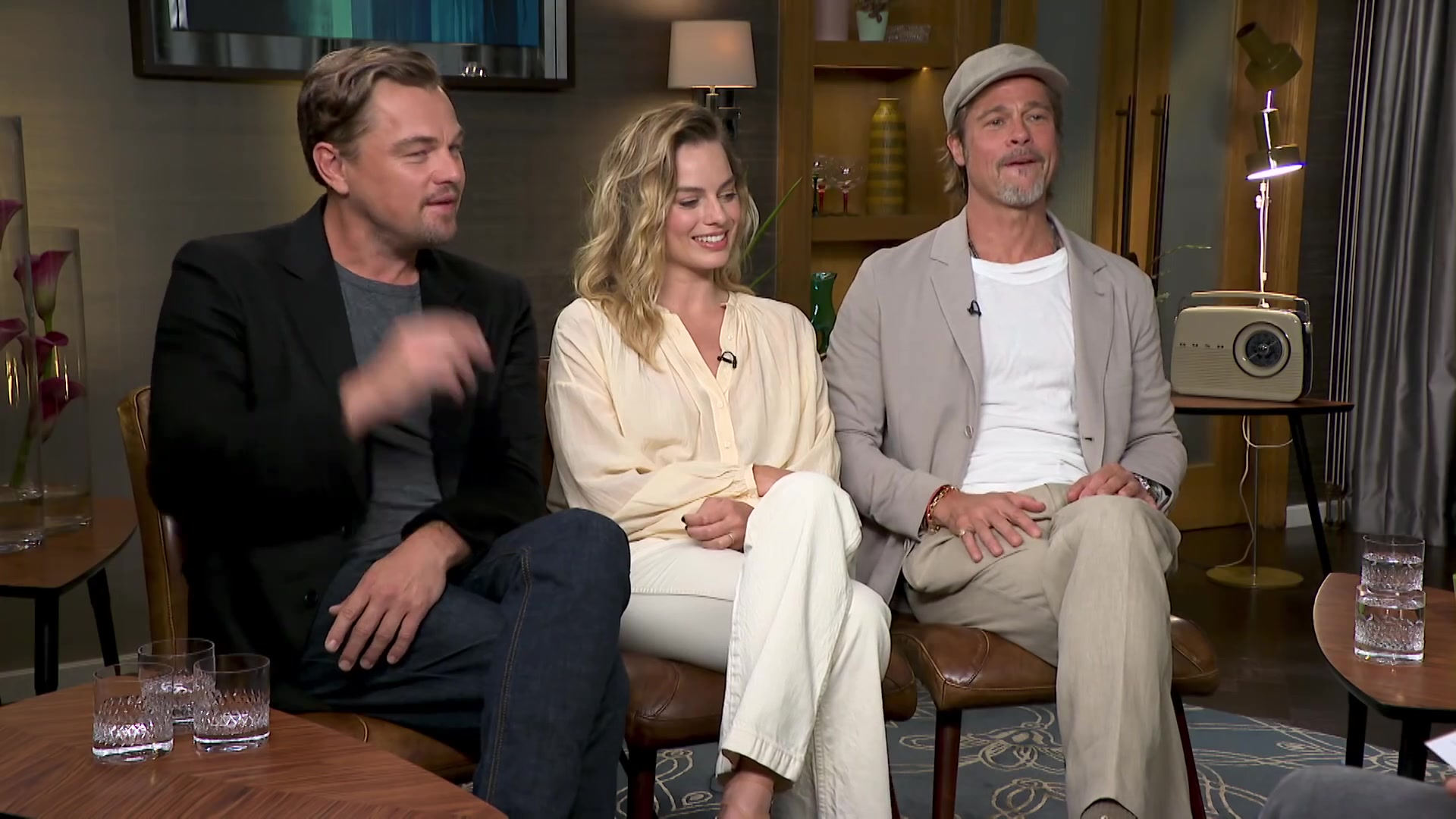}\hspace{0.2em}
    \includegraphics[trim=0 0 0 0, clip,width=0.23\textwidth]{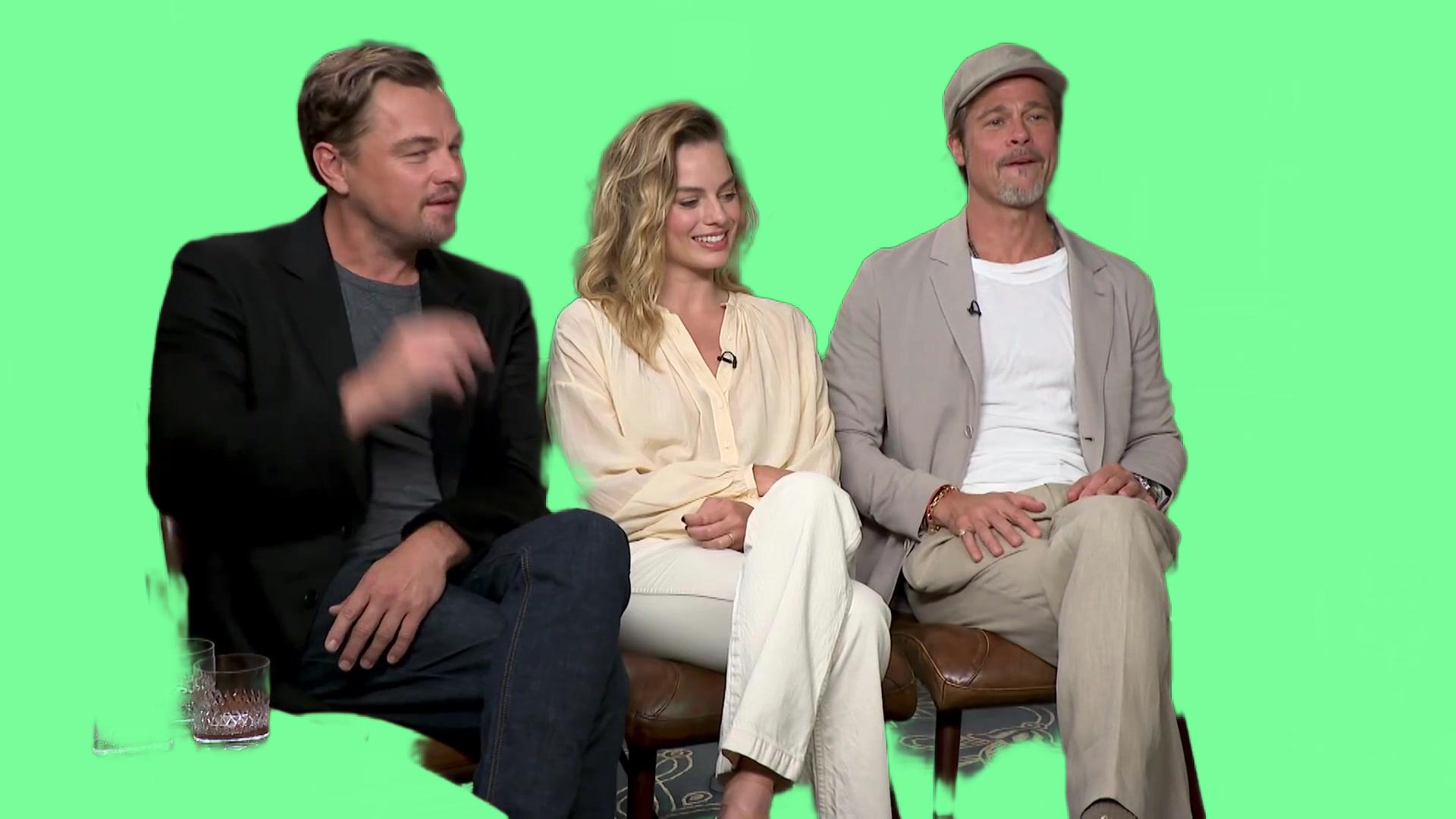}\hspace{0.2em}
    \includegraphics[trim=0 0 0 0, clip,width=0.23\textwidth]{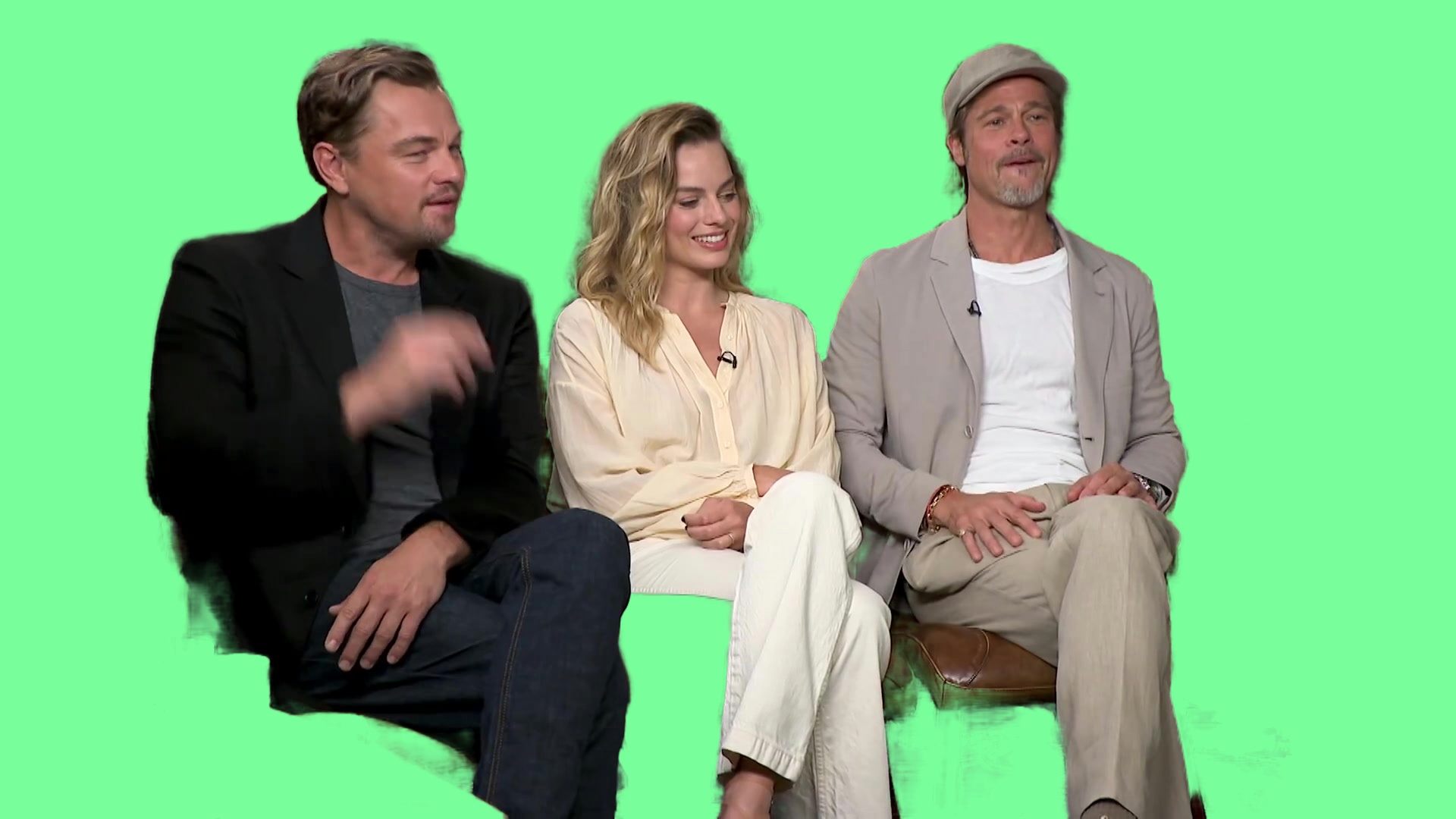}\hspace{0.2em}
    \includegraphics[trim=0 0 0 0, clip,width=0.23\textwidth]{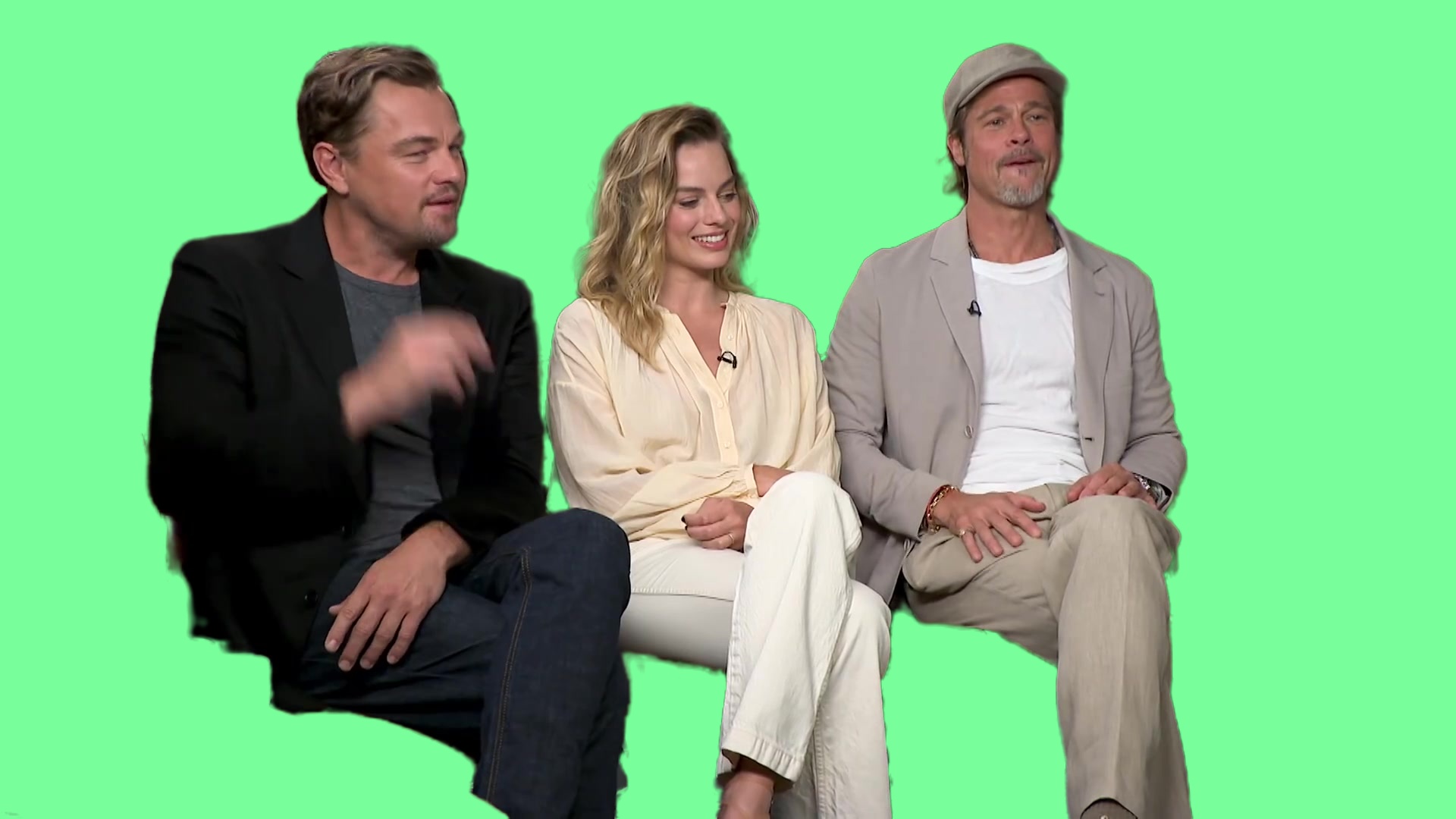}\hspace{0.2em}
    \vspace{2pt}
    \end{minipage} 
\begin{minipage}[]{.99\textwidth}
        \centering
        \footnotesize
    \includegraphics[trim=0 0 0 0, clip,width=0.23\textwidth]{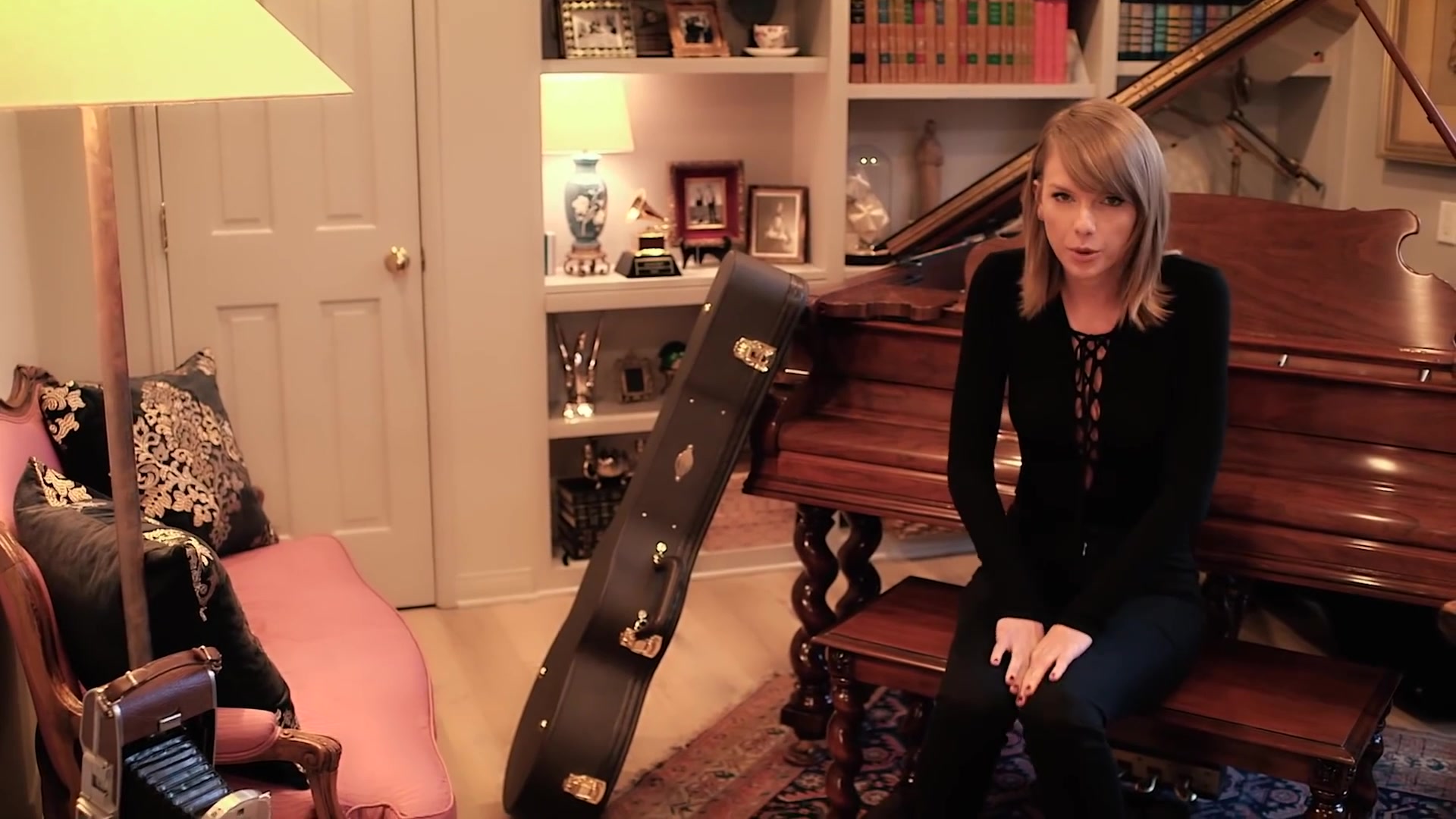}\hspace{0.2em}
    \includegraphics[trim=0 0 0 0, clip,width=0.23\textwidth]{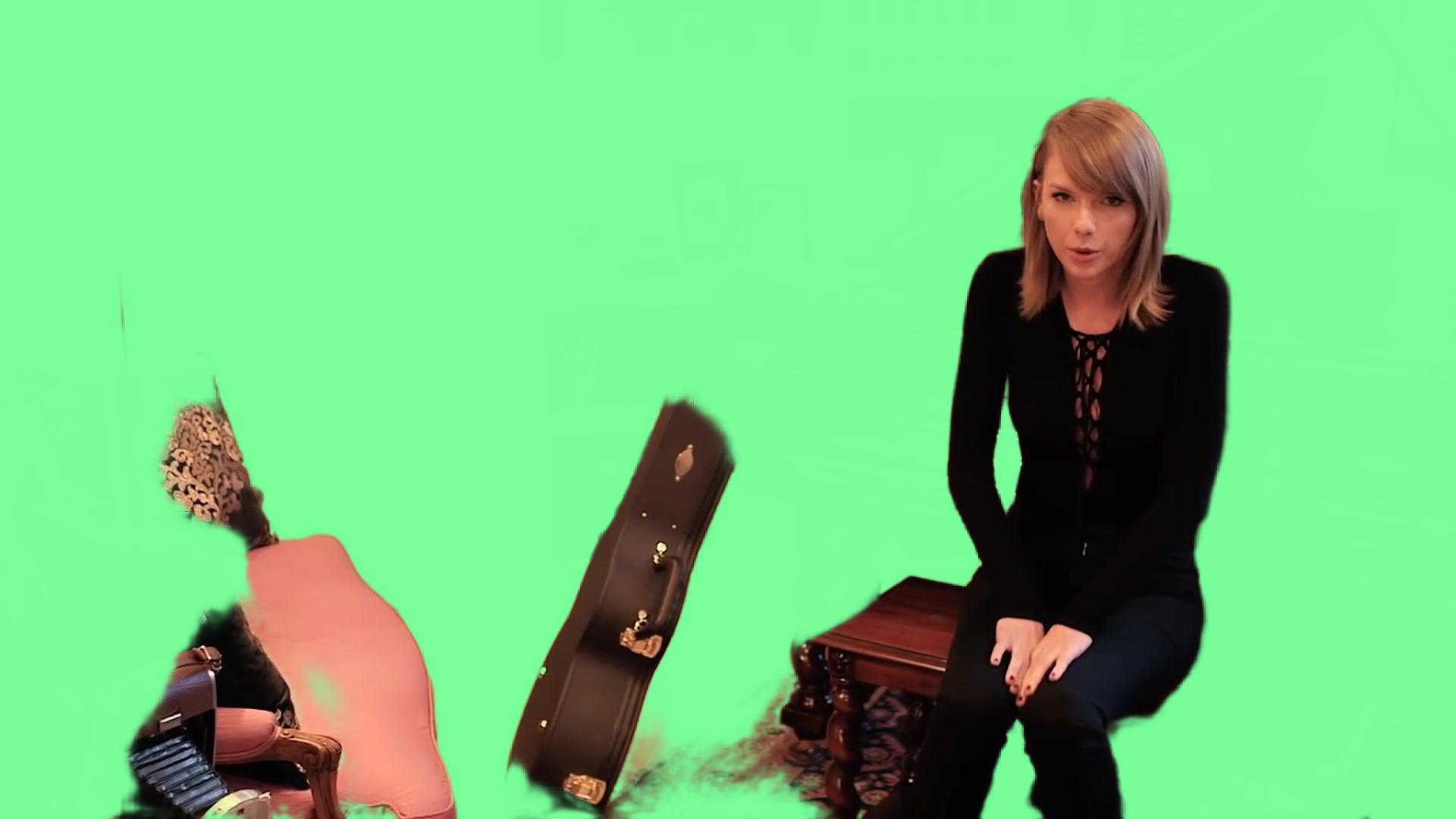}\hspace{0.2em}
    \includegraphics[trim=0 0 0 0, clip,width=0.23\textwidth]{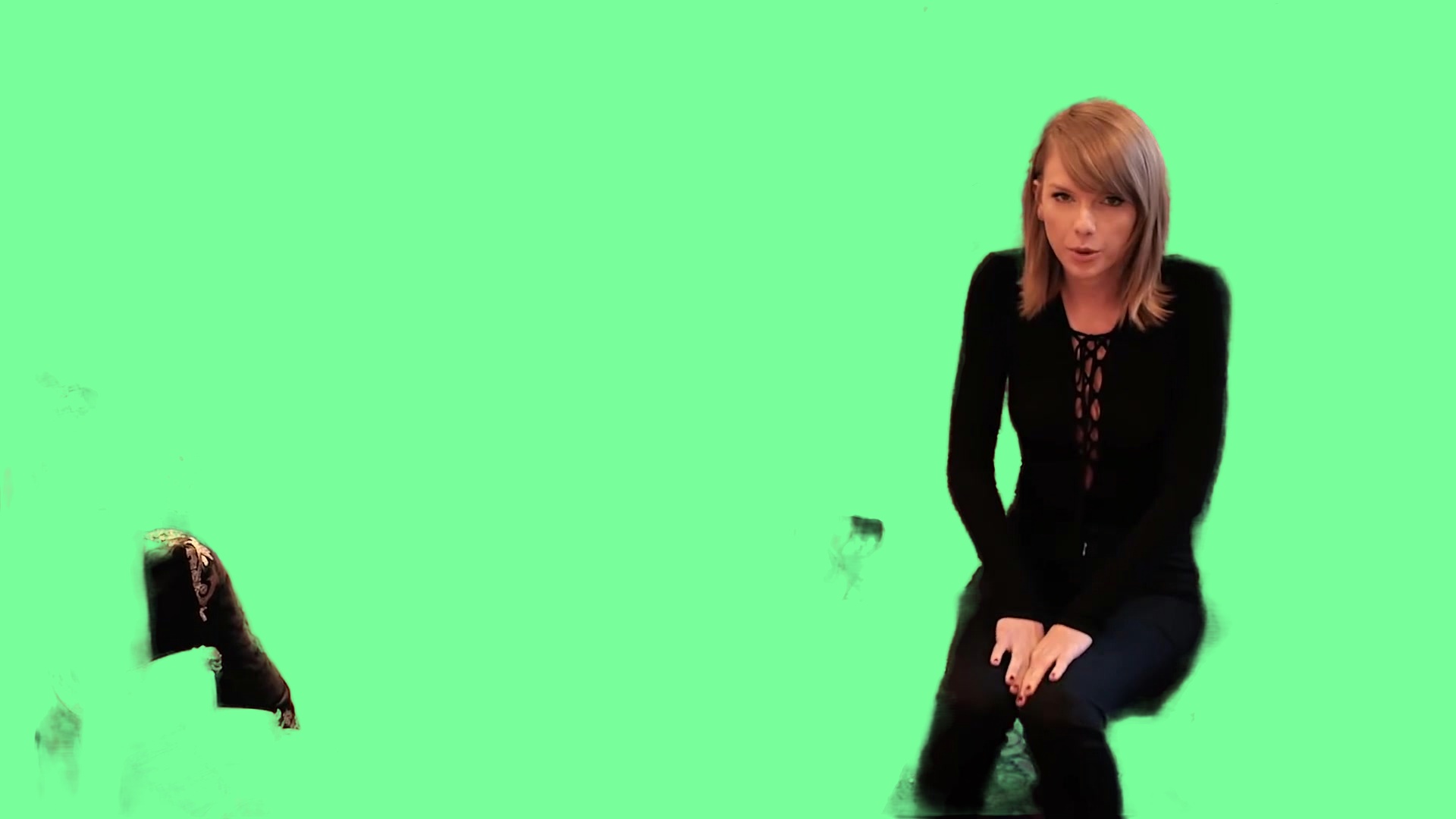}\hspace{0.2em}
    \includegraphics[trim=0 0 0 0, clip,width=0.23\textwidth]{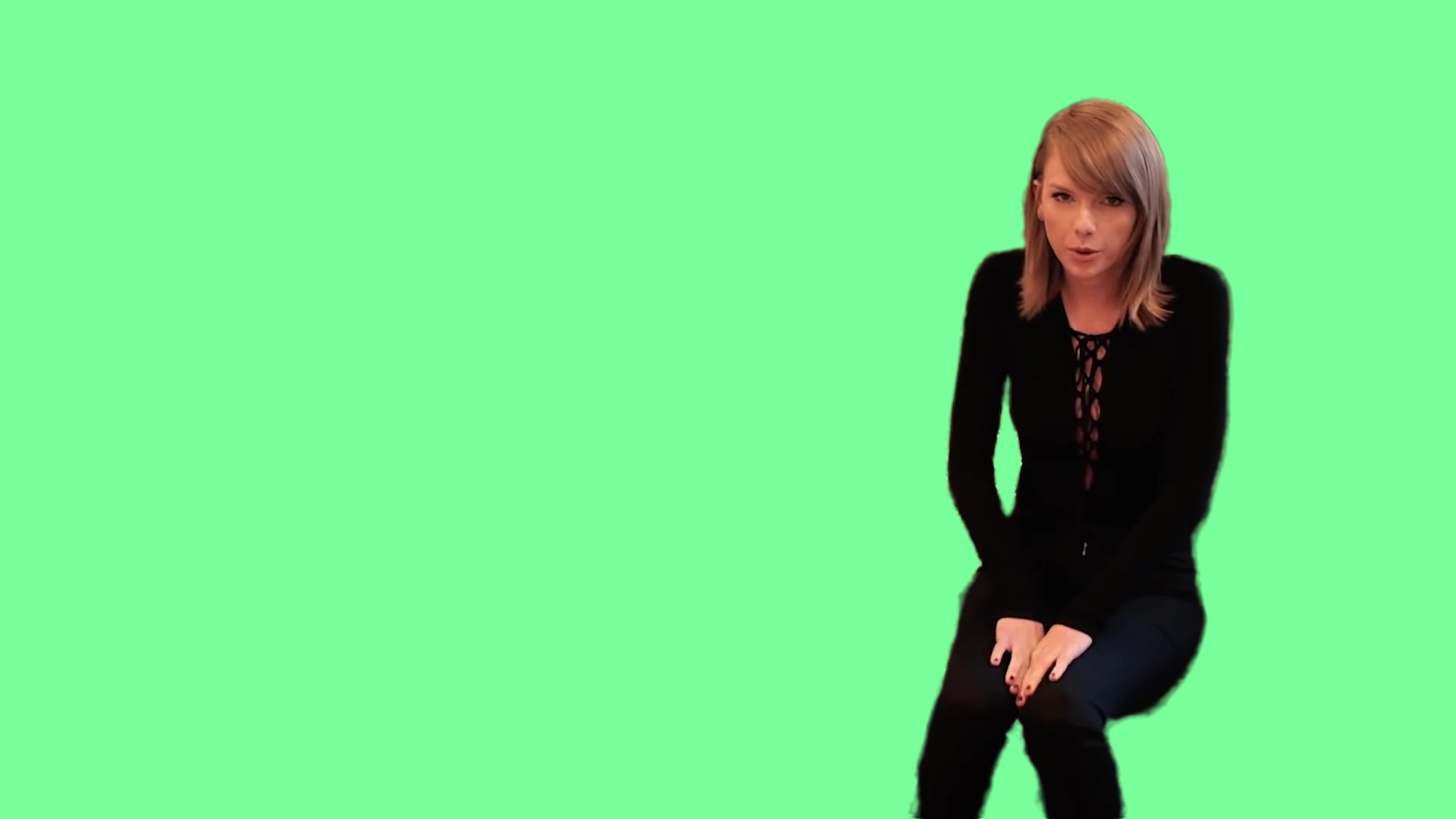}\hspace{0.2em}
    \vspace{2pt}
    \end{minipage}

  \end{center}

    \centering
\vspace{5pt}   
   \caption{\small Comparisons of our model to MODNet \cite{ke2022modnet} and RVM \cite{lin2022robust} on challenging real-world videos from YouTube.}
\vspace{5pt}   
\label{fig:youtube}
\end{figure*}
 
\end{enumerate}

\end{document}